\documentclass[MAL,examplefnt,biber]{nowfnt} % creates the journal version, needs biber version
\usepackage[utf8]{inputenc}

\title{Probabilistic Graphical Models: \\ A Concise Tutorial}

\subtitle{A Concise Tutorial}

\usepackage{amsmath}
\usepackage{amssymb}
\usepackage{mathtools}
\usepackage{amsthm}
\newcommand{\ind}{\perp\!\!\!\perp} 
\newcommand{\nind}{\not\!\perp\!\!\!\perp} 
\DeclareMathOperator*{\argmin}{argmin}
\DeclareMathOperator*{\argmax}{argmax}

\usepackage{booktabs}
\usepackage[most]{tcolorbox}
\newenvironment{properties}
    {\begin{tcolorbox}[ enhanced,breakable,
    attach boxed title to top center={yshift=-3mm,yshifttext=-1mm},
    colback=white,colframe=black!50,colbacktitle=white,boxrule=0.5pt,coltext=black,
    title=Properties,fonttitle=\color{black}\itshape,
    boxed title style={size=small,colframe=black,boxrule=0.75pt} ]
    \vspace{2mm}
    }
    { 
    \end{tcolorbox}
    }
\usepackage{booktabs}
\usepackage[most]{tcolorbox}
\newenvironment{reading}
    {\begin{tcolorbox}[ enhanced,breakable,
    attach boxed title to top center={yshift=-3mm,yshifttext=-1mm},
    colback=white,colframe=black!50,colbacktitle=white,boxrule=0.5pt,coltext=black,
    title=Further Reading,fonttitle=\color{black}\itshape,
    boxed title style={size=small,colframe=black,boxrule=0.75pt} ]
    \vspace{2mm}
    }
    { 
    \end{tcolorbox}
    }

\newcommand{\convexpath}[2]{
[   
    create hullnodes/.code={
        \global\edef\namelist{#1}
        \foreach [count=\counter] \nodename in \namelist {
            \global\edef\numberofnodes{\counter}
            \node at (\nodename) [draw=none,name=hullnode\counter] {};
        }
        \node at (hullnode\numberofnodes) [name=hullnode0,draw=none] {};
        \pgfmathtruncatemacro\lastnumber{\numberofnodes+1}
        \node at (hullnode1) [name=hullnode\lastnumber,draw=none] {};
    },
    create hullnodes
]
($(hullnode1)!#2!-90:(hullnode0)$)
\foreach [
    evaluate=\currentnode as \previousnode using \currentnode-1,
    evaluate=\currentnode as \nextnode using \currentnode+1
    ] \currentnode in {1,...,\numberofnodes} {
  let
    \p1 = ($(hullnode\currentnode)!#2!-90:(hullnode\previousnode)$),
    \p2 = ($(hullnode\currentnode)!#2!90:(hullnode\nextnode)$),
    \p3 = ($(\p1) - (hullnode\currentnode)$),
    \n1 = {atan2(\y3,\x3)},
    \p4 = ($(\p2) - (hullnode\currentnode)$),
    \n2 = {atan2(\y4,\x4)},
    \n{delta} = {-Mod(\n1-\n2,360)}
  in 
    {-- (\p1) arc[start angle=\n1, delta angle=\n{delta}, radius=#2] -- (\p2)}
}
-- cycle
}

\makeatletter
\tikzset{
    nomorepostaction/.code=\makeatletter\let\tikz@postactions\pgfutil@empty,
    my axis/.style={
        postaction={
            decoration={
                markings,
                mark=at position 1 with {
                    \arrow[thick]{>}
                }
            },
            decorate,
            nomorepostaction
        },
        thin,
        -, % switch off other arrow tips
        every path/.append style=my axis % this is necessary
    }
}
\makeatother

\RequirePackage{algorithm}
\RequirePackage[noend]{algorithmic}
\newcommand{\algorithmicinput}{\textbf{Input:}}
\newcommand{\algorithmicoutput}{\textbf{Output:}}
\newcommand{\INPUT}{\item[\algorithmicinput]}
\newcommand{\OUTPUT}{\item[\algorithmicoutput]}

\usepackage{enumitem}

\usepackage{tikz}
\usepackage{adjustbox}
\usepackage{pgfplots}
\usetikzlibrary {arrows.meta,arrows,patterns,decorations.pathreplacing,chains,positioning,shapes.geometric,shapes.callouts,calc,backgrounds,trees,pgfplots.fillbetween,quotes,calligraphy}
\usepackage{pgfplots,pgfmath,pgffor}

\usepackage{pifont}% http://ctan.org/pkg/pifont % for checkmarks, etc
\maintitleauthorlist{
Jacqueline Maasch \\
Department of Computer Science \\
Cornell Tech \\
New York, NY, USA 
\and
Willie Neiswanger \\
Department of Computer Science \\
University of Southern California\\
Los Angeles, CA, USA
\and
Stefano Ermon \\
Department of Computer Science \\
Stanford University\\
Stanford, CA, USA 
\and
Volodymyr Kuleshov \\
Department of Computer Science \\
Cornell Tech\\
New York, NY, USA
}

\issuesetup
{%
 copyrightowner={J.~Maasch and W.~Neiswanger and S.~Ermon and V.~Kuleshov},
 volume        = xx,
 issue         = xx,
 pubyear       = 2025,
 isbn          = xxx-x-xxxxx-xxx-x,
 eisbn         = xxx-x-xxxxx-xxx-x,
 doi           = 10.1561/XXXXXXXXX,
 firstpage     = 1, %Explain
 lastpage      = 197 %This line needed to avoid compile errors
}

\usepackage{mwe}

\author[1]{Maasch,Jacqueline}
\author[2]{Willie Neiswanger}
\author[3]{Ermon,Stefano}
\author[1]{Kuleshov,Volodymyr}

\affil[1]{Cornell Tech, New York, NY, USA}
\affil[2]{University of Southern California, Los Angeles, CA, USA}
\affil[3]{Stanford University, Stanford, CA, USA}

\begin{document}

\makeabstracttitle

\begin{abstract}
Probabilistic graphical modeling is a branch of machine learning that uses probability distributions to describe the world, make predictions, and support decision-making under uncertainty. Underlying this modeling framework is an elegant body of theory that bridges two mathematical traditions: \textit{probability} and \textit{graph theory}. This framework provides compact yet expressive representations of joint probability distributions, %over complex domains, 
yielding powerful generative models for probabilistic reasoning.

This tutorial provides a concise introduction to the formalisms, methods, and applications of this modeling framework. After a  review of basic probability and graph theory, we explore three dominant themes: (1) the \textit{representation} of multivariate distributions in the intuitive visual language of graphs, (2) algorithms for \textit{learning} model parameters and graphical structures from data, and (3) algorithms for \textit{inference}, both exact and approximate. %Challenges in time complexity, sample efficiency, and modeling assumptions are explored. Discussions are contextualized with real-world applications in computer vision, natural language processing, and other domains.
\end{abstract}

% JM: added for arxiv.
\clearpage\tableofcontents

%%%%%%%%%%%%%%%%%%%%%%%%
%% Introduction
%%%%%%%%%%%%%%%%%%%%%%%%

\chapter{Introduction}
\label{sec:intro}

\begin{figure}[!t]
    \centering
\begin{tikzpicture}[scale=0.8, 
    >=Stealth,   
    background rectangle/.style={fill=SpringGreen!15},
    show background rectangle
]

% Labels
\node (label) at (0,0) {\small $p(\mathbf{x}) = p(a) p(b \mid a) p(c) p(d \mid b,e) p(e \mid b,c)$};

% Curly braces
\draw[very thick, black, decorate, decoration={calligraphic brace, amplitude=1ex, raise=1ex,mirror}] (4, -1) -- (4, 1) node[pos=.5, right=3ex,text width=4cm] 
{\footnotesize\textsc{joint distribution over}  $\mathbf{X} = \{A,B,C,D,E\}$};

\end{tikzpicture}
%%%%%%%%%%%%%%%%%%%%%%%%%%%%%%%%%%%
%%%%%%%%%%%%%%%%%%%%%%%%%%%%%%%%%%%
\begin{tikzpicture}[scale=0.8, 
    >=Stealth,   
    background rectangle/.style={fill=SeaGreen!15},
    show background rectangle
]

\draw[thick,SeaGreen!15] (-4,0) rectangle (4,2.7);

% Nodes
\node[draw,circle,thick,scale=0.8] (A) at (-2,2) {$A$};
\node[draw,circle,thick,scale=0.8] (B) at (0,2) {$B$};
\node[draw,circle,thick,scale=0.8] (C) at (2,2) {$C$};
\node[draw,circle,thick,scale=0.8] (D) at (-2,0) {$D$};
\node[draw,circle,thick,scale=0.8] (E) at (2,0) {$E$};

% Edges
\draw[thick,->] (A) -- (B);
\draw[thick,->] (B) -- (D);
\draw[thick,->] (C) -- (E);
\draw[thick,->] (B) -- (E);
\draw[thick,->] (E) -- (D);

% Curly braces
\draw[very thick, black, decorate, decoration={calligraphic brace, amplitude=1ex, raise=1ex,mirror}]
(4, -0.5) -- (4, 2.5) node[pos=.5, right=3ex,text width=4cm] {\footnotesize\textsc{directed acyclic graph of} $\mathbf{X}$};

\end{tikzpicture}

    \caption{This tutorial demonstrates how probability and graph theory provide complementary languages for representing, learning, and querying models of real-world phenomena. The reader will gain intuition for reasoning over complex domains, both in terms of probability distributions and in terms of their graphical representations. For example, this tutorial will explain when and why the factorization of joint distribution $p(\mathbf{x})$ implies the directed graph of $\mathbf{X}$ depicted above.}
    \label{fig:intro_dag}
\end{figure}
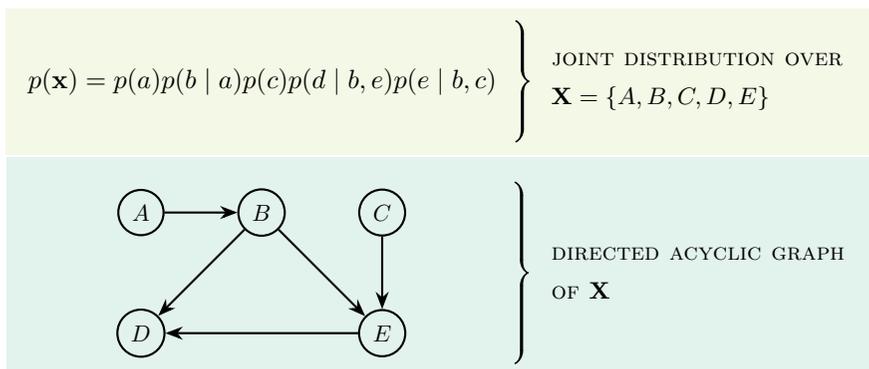

Probabilistic graphical modeling is a branch of machine learning that uses probability distributions to describe the world and to make useful predictions about it. Underlying this modeling framework is an elegant body of theory that bridges two mathematical traditions: \textit{probability} and \textit{graph theory} (Figure \ref{fig:intro_dag}). Probabilistic graphical modeling also intersects with philosophy in intriguing ways, especially through questions of \textit{causality}. This tutorial aims to provide a concise introduction to the formalisms, methods, and applications of this powerful framework.

Probabilistic graphical modeling has had far-reaching impacts on theoretical and applied research in computing and beyond. It has been used to solve problems across diverse domains, including medicine, biology, chemistry, physics, electrical engineering, natural language processing, and computer vision. This combination of elegant theory and practical applications has made probabilistic graphical modeling one of the most fascinating topics in modern computer science and artificial intelligence. 

Testament to the impact of this research area, the Association for Computing Machinery awarded the 2011 Turing Award\footnote{The Turing Award is widely regarded as the ``Nobel Prize of Computing.''} to Judea Pearl for establishing the foundations of probabilistic graphical modeling and introducing a calculus for causal reasoning \citep{gotlieb2022turing}. %acm_turing_pearl
These contributions ``revolutionized the field of artificial intelligence'' and precipitated vast advancements across engineering, the natural sciences, and the social sciences \citep{acm_turing_pearl}.

In the remainder of this brief introduction, we provide a high-level overview of what to expect from this tutorial.

\section{A Framework for Reasoning Under Uncertainty}

When solving a real-world problem using mathematics, it is very common to define a mathematical model of the world in the form of an equation.
Perhaps the simplest model would be a linear equation of the form
\[
y = \boldsymbol{\beta}^T \mathbf{x},
\]
where $y$ is an outcome variable that we want to predict and $\mathbf{x}$ are observed variables that affect this outcome. For example, $y$ may be the price of a house, while $\mathbf{x}$ denotes a series of features that affect this price (e.g., location, number of bedrooms, age of the house, etc.). We assume that $y$ is a linear function of this input, parameterized by $\boldsymbol{\beta}$.

Often, the real world that we are trying to model is very complicated. In particular, it often involves a significant amount of \textit{uncertainty} (e.g., a house's price $y$ can fluctuate based on the inherent variability in buyers’ subjective preferences about its features $\mathbf{x}$). It is therefore very natural to deal with this uncertainty by modeling the world in the form of a probability distribution,
\[
p(\mathbf{x},y).
\]
Given such a model, we could ask various questions. For example: \textit{Given that the house costs \$100,000, what is the probability that it has three bedrooms?}

The probabilistic aspect of modeling is very important. Typically, we cannot perfectly predict the future. We rarely have sufficient knowledge about the world, and many real-world processes are stochastic in nature. Additionally, we need to assess the confidence of our predictions. Often, predicting a single value is not enough: we need the system to output its \textit{beliefs} about the world, and to what extent these are uncertain.

In this tutorial, we present principled ways of \textit{reasoning about uncertainty}. We draw from both probability and graph theory to derive efficient machine learning algorithms for this task, answering such questions as the following.
\begin{itemize}
    \item \textit{What are the tradeoffs between computational complexity and the richness of a probabilistic model?}
    \item \textit{What is the best model for inferring facts about the future, given a fixed dataset and computational budget?}
    \item \textit{How does one combine prior knowledge with observed evidence in a principled way to make predictions?}
    %\item \textit{How does one rigorously infer whether $A$ is the cause of $B$, or vice versa? }
\end{itemize}
We will contextualize our discussions with applications to real-world problems, such as image and language analysis.

\subsubsection{Challenges in Probabilistic Modeling}

To get a first taste of the challenges that lie ahead, consider a simple application of probabilistic modeling: spam classification.

Suppose we have a model $p(y, x_1, \dotsc, x_n)$ of word occurrences in spam and non-spam mail. The binary random variables $\{X_i\}_{i=1}^n$ encode whether the $i^{th}$ English word is present in the email. The binary random variable $Y$ indicates whether the email is spam. In order to classify a new email, we can look at the probability $ p(y=1 \mid x_1, \dotsc, x_n) $.

What is the ``size'' of the function $p$ that we just defined? Our model defines a probability in $[0,1]$ for each combination of inputs $\{y, x_1, \dotsc, x_n\}$. Specifying each of these probabilities requires the calculation of a staggering $2^{n+1}$ different values, one for each assignment to our $n+1$ binary variables. Since $n$ is the size of the English vocabulary, this is clearly impractical from both a computational standpoint (i.e., \textit{how do we store a data object of this size?}) and from a statistical perspective (i.e., \textit{how do we efficiently estimate the parameters from limited data?}). More generally, this example illustrates one of the main challenges that this tutorial will deal with: probability distributions are inherently exponentially-sized objects, and so to manipulate them we must make simplifying assumptions about their structure.

The main simplifying assumption that we will make in this tutorial is that of \textit{conditional independence} among variables. For example, suppose that the English words in our spam example are all conditionally independent given $y$. That is, the event that a given word occurs in the email is independent of whether a different word occurs, given that the email is spam. This is clearly an oversimplification, as some words are likely to co-occur in English text regardless of $y$ (e.g., \textit{ibuprofen} and \textit{pain}). However, these events will indeed be independent for most words (e.g., \textit{penguin} and \textit{muffin} do not have an obvious relationship to each other in the English language). We can therefore justify that this assumption will not significantly degrade the accuracy of the model.

We refer to this particular choice of independencies as the \textit{Naive Bayes assumption}. Given this assumption, we can write the model probability as a product of \textit{factors},
\[
p(y, x_1, \dotsc, x_n) = p(y) \prod_{i=1}^n p(x_i \mid y).
\]
Each factor $p(x_i \mid y)$ can be completely described by a small number of parameters (four parameters with two degrees of freedom to be exact). The entire distribution is parameterized by $O(n)$ parameters, which we can tractably estimate from data and make predictions over.

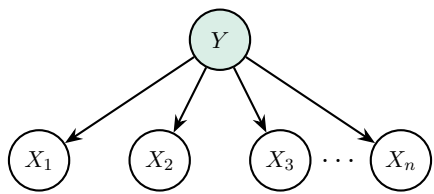
\begin{figure}
    \centering
\begin{tikzpicture}[scale=0.8, >=Stealth, node distance=2cm]
%>={Stealth[width=5pt,length=5pt]}

% Nodes
\node[draw, thick, circle, minimum size=1cm,fill=SeaGreen!20, scale=0.8] (Y) at (0, 0) {$Y$};
\node[draw, thick, circle, minimum size=1cm,scale=0.8] (X1) at (-3, -2) {$X_1$};
\node[draw, thick, circle, minimum size=1cm,scale=0.8] (X2) at (-1, -2) {$X_2$};
\node[draw, thick, circle, minimum size=1cm,scale=0.8] (X3) at (1, -2) {$X_3$};
\node[draw, thick, circle, minimum size=1cm,scale=0.8] (Xn) at (3, -2) {$X_n$};
\node at (2, -2) {$\dots$}; % Dots in between

% Edges
\draw[->, thick] (Y) -- (X1);
\draw[->, thick] (Y) -- (X2);
\draw[->, thick] (Y) -- (X3);
\draw[->, thick] (Y) -- (Xn);

\end{tikzpicture}

    \caption{Graphical representation of the Naive Bayes spam classification model. We can interpret the directed graph as conveying a story about how the data was generated. First, a spam/non-spam label was chosen at random ($Y = y$). Then, a subset of $n$ possible English words were sampled independently and at random ($\{X_i\}_{i=1}^n = \{x_i\}_{i=1}^n$). Thus, the graphical representation of this data generating process displays directed edges from $Y$ to each $X_i$ (denoting dependence), but no edges among the individual $X_i$ (denoting independence).}
    \label{fig:spam}
\end{figure}

\subsubsection{Describing Probabilities with Graphs}

The previous independence assumption can be conveniently represented in the form of a \textit{directed graph}, where directed edges $Y \to X_i$ visually represent the relationships that are also captured in the joint factorization $p(y) \prod_{i=1}^n p(x_i \mid y)$ (Figure \ref{fig:spam}). This graphical representation has the immediate advantage of being easy to understand. It can be interpreted as telling us a story: an email was generated by first choosing at random whether the email is spam or not (indicated by $Y = y$), and then by sampling words one at a time. Conversely, if we have a story for how our dataset was generated, we can naturally express it as a graph with an associated probability distribution.

More importantly, we want to submit various queries to our model (e.g., what is the probability of spam given that I see the word \textit{congratulations}?). Answering these questions will require specialized inference algorithms that are most naturally defined using graph-theoretic concepts. We will also use graph theory to analyze the speed of learning algorithms and to quantify the computational complexity (e.g., NP-hardness) of different learning tasks. In essence, there is an intimate connection between probability distributions and graphs that will be exploited throughout this tutorial for the purposes of defining, learning, and querying probabilistic models.

\section{A Bird's Eye View of This Tutorial}

\subsubsection{Tutorial Structure}

After briefly reviewing the fundamentals of probability and graph theory (Chapter \ref{sec:preliminaries}), our discussion will be divided into three major themes: {representation} (Chapter \ref{sec:representation}), {inference} (Chapters \ref{sec:inference} and \ref{sec:approximate_inference}), and {learning} (Chapter \ref{sec:learning}). %We explore interrelations among these themes in our concluding discussion of the variational autoencoder (Chapter \ref{sec:discussion_vae}).
\begin{enumerate}
    \item \textbf{Representation} (i.e., \textit{how to specify a model}). How do we express a probability distribution that models some real-world phenomenon? This is not a trivial problem: we have seen that a naive model for classifying spam messages with $n$ possible words generally requires the specification of $O(2^n)$ parameters. We will address this difficulty via general techniques for constructing tractable models. These recipes will make heavy use of graph theory. %Given a probability distribution, we will see how the properties of its graphical representation can impact the efficiency of downstream learning and inference. 
%%%
    \item \textbf{Inference} (i.e., \textit{how to ask the model questions}). Given a probabilistic model, how do we obtain answers to relevant questions about the world? Such questions often reduce to querying the marginal or conditional probabilities of certain events of interest. More concretely, we will typically be interested in asking two types of questions:

\begin{enumerate}
    \item \textit{Marginal inference.} What is the probability of a given variable in our model after we sum everything else out? Such queries generally take the form
        \[
        p(x_1) = \sum_{x_2} \sum_{x_3} \cdots \sum_{x_n} p(x_1, x_2, \dotsc, x_n). 
        \]
    \item \textit{Maximum a posteriori (MAP) inference.} Here, we ask for the most likely assignment of variables. For example, we can determine the most likely spam message under our model. Such queries generally take the form
        \[
        \underset{x_1, \dots, x_n}{\operatorname{argmax}} \; p(x_1,\dotsc,x_n, y=1).
        \]
\end{enumerate}
Queries often involve \textit{evidence}, in which case we fix the assignment of a subset of the variables (e.g., $y = 1$ above).

It turns out that inference is a very challenging task. For many probabilistic models of interest, it will be NP-hard to answer the kinds of questions that we have described. Crucially, whether inference is tractable will depend on the structure of the graph that describes the underlying probability distribution. We will show that even when \textit{exact inference} is intractable, we can still obtain useful answers via \textit{approximate inference} methods. %Interestingly, many of the algorithms described in this tutorial will be heavily based on work done in the statistical physics community in the mid-20th century.  
%%%%
    \item \textbf{Learning} (i.e., \textit{how to fit a model to real-world data}). Our last key task refers to fitting a model to a dataset (e.g., a large number of labeled examples of spam). By inspecting the data, we can infer useful patterns (e.g., which words are found more frequently in spam emails). We can then leverage our knowledge of these patterns to make predictions about the future. However, we will see that learning and inference are also inherently linked in a more subtle way: inference is a key subroutine that is called repeatedly within learning algorithms. The topic of learning also shares important connections with two prominent fields:
    \begin{itemize}
        \item \textit{Computational learning theory}, which deals with questions such as generalization from limited data and overfitting; and
        \item \textit{Bayesian statistics}, which offers an elegant mathematical language for combining prior knowledge and observed evidence in a principled way.
    \end{itemize}
\end{enumerate}

The three themes of representation, inference, and learning are closely linked. As we have alluded, the derivation of efficient inference and learning algorithms requires an adequately represented model. Furthermore, learning will require inference as a subroutine. Thus, it is best to always keep these three tasks in mind rather than focusing on them in isolation. In Chapter \ref{sec:discussion_vae}, we highlight these interconnections with a concluding discussion on a powerful deep probabilistic model: the \textit{variational autoencoder} \citep{kingma_introduction_2019}, whose widespread influence was recognized with the Test of Time Award at the 2024 International Conference on Learning Representations \citep{vae_iclr_tot_blog, vae_iclr_tot}.

%\williex{Could include comparison to prior works (but in a more-friendly way); emphasize the goals of this book in relation to prior texts (e.g., compared to Koller-Friedman, is more concise, different register/tone, connects better to (prerequisite for) modern deep generative modeling.}

\subsubsection{Relation to Prior Works} 

This tutorial is intended to be a  concise and accessible overview of the state of the field. Throughout, we offer recommendations for exceptional resources that provide additional depth. We encourage the reader to explore the works of Judea Pearl for many authoritative resources on probabilistic and causal inference through a graphical lens \citep{pearl1988probabilistic,geffner2022probabilistic}.  \citet{koller2009probabilistic} provide an extensive textbook that makes a strong accompaniment to this tutorial. Many seminal machine learning texts dedicate chapters to related topics, including \citet{bishop2006pattern}, \citet{murphy2012machine}, and \citet{murphy2023advanced}. This tutorial  discusses several topics that have been previously described in \textit{Foundations and Trends}, including graphical models and variational inference \citep{wainwright_graphical_2008}, conditional random fields \citep{sutton_introduction_2012}, Monte Carlo methods \citep{lindsten_backward_2013, naesseth_elements_2019}, Bayesian inference \citep{angelino_patterns_2016}, and variational autoencoders \citep{kingma_introduction_2019}. This tutorial provides a compact survey of the core theory and methods in this large literature, highlighting their interrelationships and drawing connections to modern deep generative modeling.

%On the general topic of probabilistic graphical modeling, we highly recommend the following texts as accompaniments to this tutorial.

%\vspace{5mm}

%\begin{reading}
%    \begin{itemize}[leftmargin=*]
%        \item \fullcite{pearl1988probabilistic}.
        %\item \fullcite{bishop2006pattern}.
%        \item \fullcite{koller2009probabilistic}.
%    \end{itemize}
%\end{reading}

%%%%%%%%%%%%%%%%%%%%%%%%
%% Preliminaries
%%%%%%%%%%%%%%%%%%%%%%%%

\chapter{Preliminaries}
\label{sec:preliminaries}

%%%%%%
%% Review of probability theory
%%%%%%

The foundations of probabilistic graphical modeling lie centrally in \textit{probability theory} and \textit{graph theory}.\footnote{As for most machine learning topics, the reader may also benefit from a review of statistics, linear algebra, calculus, and information theory. See \citet{cover2006elements,murphy2022probabilistic, strang2023introduction} for useful treatments of these topics.} This chapter provides a non-exhaustive review of both. Sections \ref{sec:probability} and \ref{sec:random_variables} cover the basics of probability and random variables, including  probability spaces, notions of variance and expectation, and joint, marginal, and conditional probability distributions. Section \ref{sec:graph_theory}  provides an overview of basic graph theoretic concepts, such as directed and undirected graphs, their properties and common substructures, and special categories of graphs that are useful for inference and learning (e.g., tree-structured graphs).

\paragraph{A Note on Notation}

Moving forward, we will adhere to the following notation for clarity. A random variable will be denoted by a capital letter (e.g., $X$), while the specific values it takes on will be in lowercase (e.g., $X = x$). Sets and multivariate random variables (i.e., random vectors or sets of variables) will be denoted by boldface capitals (e.g., $\mathbf{X}$), with vector values in bold lowercase (e.g., $\mathbf{X} = \mathbf{x}$).
Graphs will be denoted by calligraphic letters (e.g., $\mathcal{G}$). As the nodes in graph objects represent random variables, these will generally be uppercase (e.g., $X \leftarrow Z \rightarrow Y$). Probability density functions associated with distributions will often
be expressed with lowercase letters (e.g., $p(\mathbf{y} \mid \mathbf{x})$). Point estimates, such as maximum likelihood or maximum a posteriori (MAP) estimates, will be denoted using a hat (e.g., $\hat{\mathbf{x}}$). Model parameters will be denoted by Greek letters (e.g., $\theta$), whether scalar or vector-valued.

%\section{Review of Probability Theory}

%We provide a review of probability concepts. Some of the review materials have been adapted from [CS229 Probability Notes](http://cs229.stanford.edu/section/cs229-prob.pdf) and [STATS310 Probability Theory Notes](https://web.stanford.edu/class/stats310a/lnotes.pdf).

\section{Elements of Probability}
\label{sec:probability}

The concept of probability has been explored through various lenses. In one sense, probability can be interpreted as a \textit{frequency of occurrence}. For example, we might think of probability as the percentage of ``successes'' in a series of repeated trials that can succeed or fail. This is often referred to as the \textit{frequentist} interpretation. In another sense, probability can be conceptualized as a \textit{measure of uncertainty}: a degree of \textit{subjective belief} or \textit{reasonable expectation} informed by prior knowledge \citep{bertsekas2008introduction,murphy2022probabilistic}. This is often referred to as the \textit{Bayesian}\footnote{In reference to the 18$^{th}$ century mathematician Thomas Bayes and Bayes' rule (Definition \ref{def:bayes_rule}), which we will discuss in this chapter.} interpretation.

Each of these viewpoints can be useful, though some cases do not accommodate every interpetation. For example, if we wish to model the uncertainty of an event that does not have a long-run frequency (e.g., an event that can strictly occur once at most), the Bayesian interpretation of probability will be more natural than the frequentist perspective. Throughout this tutorial, we will draw from both the Bayesian and frequentist traditions. We will occasionally take a formal Bayesian stance, as in Section \ref{sec:learning_bayesian_models} on learning Bayesian models.

\subsection{Set Theory}

We begin with the basic elements of probability to establish the definition of probabilities on \textit{sets}. In one view of probability theory, we can treat probability as a \textit{measure of the size of a set} \citep{chan2021introduction}. We will briefly review the basic language of set theory, as this will crop up throughout this tutorial.

\begin{definition}[Set]
    A set is a collection of \textit{objects} or \textit{elements}, e.g.,
    \[
    \mathbf{A} \coloneqq \{A_i\}_{i=1}^n = \{A_1,\dots,A_n\}.
    \]
\end{definition}

For each element $A_i$ belonging to $\mathbf{A}$, we say that $A_i \in \mathbf{A}$. Note that elements themselves can be sets (yielding a set of sets). We can define operations on sets, such as intersection ($\cap$), union ($\cup$), and difference ($\setminus$) (Figure \ref{fig:set_operations}). For a more extensive review of set theory, see \citet{kunen1980set,jech2002set}.

%The joy of sets: fundamentals of contemporary set theory 
%Using the language of sets and set operations.

%\input{tables/table_set_notation}

\tikzset{
B/.style = {decorate,
            decoration={calligraphic brace, amplitude=4pt,
            raise=1pt, mirror},% for mirroring of brace
            very thick,
            pen colour=black},
dot/.style = {circle, fill, inner sep=2pt, outer sep=0pt}
        }

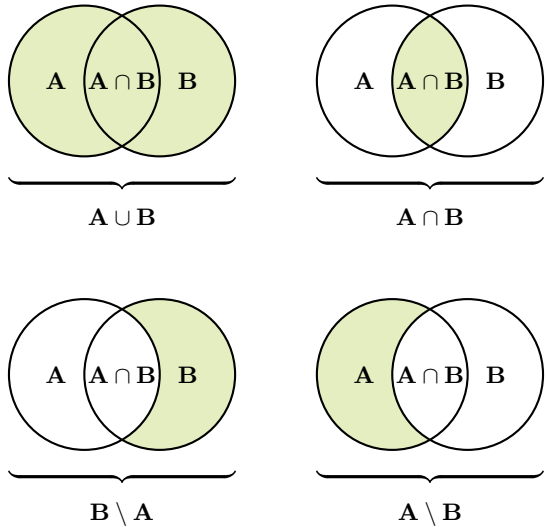
\begin{figure}[!t]
    \centering

\begin{tikzpicture}[scale=0.5]

    % Circles for sets A and B
    \draw[SpringGreen,fill=SpringGreen!40] (1,0) circle(2);
    \draw[SpringGreen,fill=SpringGreen!40] (-1,0) circle(2);

    \draw[thick,black] (1,0) circle(2);
    \draw[thick,black] (-1,0) circle(2);

    % Labels
    \node[] at (1.75,0) {\footnotesize$\mathbf{B}$};
    \node[] at (-1.75,0) {\footnotesize$\mathbf{A}$};
    \node[] at (0,0) {\footnotesize$\mathbf{A} \cap \mathbf{B}$};
    \draw[B] (-3,-2.5) -- node[below=3mm] {\footnotesize$\mathbf{A} \cup \mathbf{B}$} (3,-2.5);
\end{tikzpicture}
%%%%%%%%%%%%%%%%%%%%%%%%%%%%
\hspace{8mm}
%%%%%%%%%%%%%%%%%%%%%%%%%%%%
\begin{tikzpicture}[scale=0.5]

    % Circles for sets A and B
    \draw[white,fill=white] (1,0) circle(2);
    \draw[white,fill=white] (-1,0) circle(2);

    \begin{scope}
      \clip (-1,0) circle(2);
      \fill[SpringGreen!40] (1,0) circle(2);
    \end{scope}

    \draw[thick,black] (1,0) circle(2);
    \draw[thick,black] (-1,0) circle(2);

    % Labels
    \node[] at (1.75,0) {\footnotesize$\mathbf{B}$};
    \node[] at (-1.75,0) {\footnotesize$\mathbf{A}$};
    \node[] at (0,0) {\footnotesize$\mathbf{A} \cap \mathbf{B}$};
    \draw[B] (-3,-2.5) -- node[below=3mm] {\footnotesize$\mathbf{A} \cap \mathbf{B}$} (3,-2.5);
\end{tikzpicture} \\ 
%%%%%%%%%%%%%%%%%%%%%%%%%%%%
\vspace{8mm}
%%%%%%%%%%%%%%%%%%%%%%%%%%%%
\begin{tikzpicture}[scale=0.5]

    % Circles for sets A and B
    \draw[SpringGreen,fill=SpringGreen!40] (1,0) circle(2);
    \draw[white,fill=white] (-1,0) circle(2);

    \draw[thick,black] (1,0) circle(2);
    \draw[thick,black] (-1,0) circle(2);
    
    % Labels
    \node[] at (1.75,0) {\footnotesize$\mathbf{B}$};
    \node[] at (-1.75,0) {\footnotesize$\mathbf{A}$};
    \node[] at (0,0) {\footnotesize$\mathbf{A} \cap \mathbf{B}$};
    \draw[B] (-3,-2.5) -- node[below=3mm] {\footnotesize$ \mathbf{B} \setminus \mathbf{A}$} (3,-2.5);
\end{tikzpicture}
%%%%%%%%%%%%%%%%%%%%%%%%%%%%
\hspace{8mm}
%%%%%%%%%%%%%%%%%%%%%%%%%%%%
\begin{tikzpicture}[scale=0.5]

    % Circles for sets A and B
    \draw[SpringGreen!40,fill=SpringGreen!40] (-1,0) circle(2);
    \draw[white,fill=white] (1,0) circle(2);

    \draw[thick,black] (1,0) circle(2);
    \draw[thick,black] (-1,0) circle(2);
    
    % Labels
    \node[] at (1.75,0) {\footnotesize$\mathbf{B}$};
    \node[] at (-1.75,0) {\footnotesize$\mathbf{A}$};
    \node[] at (0,0) {\footnotesize$\mathbf{A} \cap \mathbf{B}$};
    \draw[B] (-3,-2.5) -- node[below=3mm] {\footnotesize$ \mathbf{A} \setminus \mathbf{B}$} (3,-2.5);
\end{tikzpicture}

    \caption{Set operations on two sets.}
    \label{fig:set_operations}
\end{figure}

\subsection{Probability Spaces}

The probability space ($\Omega, \mathcal{F}, \mathbb{P}$) is a fundamental framework for expressing a random process.
The sample space $\Omega$ is the set of all the outcomes of a random experiment. Here, each outcome $\omega \in \Omega$ can be thought of as a complete description of the state of the real world at the end of the experiment. The event space $\mathcal{F}$ is a subset of all possible sets of outcomes. It represents the collection of subsets of possible interest to us, where we denote elements of $\mathcal{F}$ as \textit{events}. The mapping $\mathbb{P}$ assigns probabilities to each event $A \in \mathcal{F}$. Although each $A \in \mathcal{F}$ is itself a set, we omit boldface here to align with standard probability notation and avoid clutter. For a probability space, $\mathcal{F}$ is furthermore a $\sigma$-\textit{algebra}, and $\mathbb{P}$ is a \textit{probability measure}, which we define as follows.

\begin{definition}[$\sigma$-algebra] \label{def:sigma_algebra}
    Let $2^{\Omega}$ denote the power set of $\Omega$. We call $\mathcal{F} \subseteq 2^{\Omega}$ a $\sigma$-algebra if the following holds.
\begin{itemize}
    \item $\Omega \in \mathcal{F}$.
    \item \textit{Closed under complement.} If $A \in\mathcal{F}$, then $A^C \in\mathcal{F}$ (where $A^C \coloneqq \Omega \backslash A$ is the complement of $A$).
    \item \textit{Closed under countable unions.} If $A_i \in \mathcal{F}$ for $i = 1, 2, 3, \dots, n$, then $\bigcup_{i=1}^n A_i \in \mathcal{F}$.
\end{itemize}
\end{definition}

\noindent A pair $(\Omega, \mathcal{F})$ where $\mathcal{F}$ is a $\sigma$-algebra is called a \textit{measurable space}.

\begin{definition}[Probability measure] \label{def:probability_measure}
     Given a measurable space $(\Omega, \mathcal{F})$, a measure $\mu$ is any set function $\mu: \mathcal{F} \rightarrow [0, \infty]$ that satisfies the following properties.
\begin{itemize}
    \item $\mu(A) \ge \mu(\emptyset) \coloneqq 0$ for all $A \in \mathcal{F}$.
    \item $\mu\big(\bigcup_i A_i \big) = \sum_i \mu(A_i)$ for any countable collection of disjoint sets $A_i \in \mathcal{F}$.
\end{itemize}
When the above properties are satisfied and $\mu(\Omega) = 1$, we call $\mu$ a \textit{probability measure} and denote it as $\mathbb{P}$.
\end{definition}

A probability measure $\mathbb{P}$ boasts many useful properties. We enumerate some of these properties below.

\begin{definition}[Axioms of probability, \citealt{kolmogoroff1933grundbegriffe}] \label{def:axioms_of_probability}
We define the following three axioms on probability measure $\mathbb{P}$.
    \begin{enumerate}
        \item \textit{Nonnegativity}. For any event $A$, $0 \leq \mathbb{P}(A) \leq 1$.
        \item \textit{Normalization}. $\mathbb{P}(\Omega) = 1$.
        \item \textit{Countable additivity.} For pairwise disjoint sets $A_n \in \mathcal{F}$,
            \[
            \mathbb{P} \left( \bigcup_{n=1}^\infty A_n \right) = \sum_{n=1}^\infty \mathbb{P} (A_n).
            \]
    \end{enumerate}
\end{definition}

Note that these axioms are consistent but not complete \citep{kolmogorov2018foundations}. Additionally, we remark the following useful properties.
\begin{properties}
    \begin{itemize}[leftmargin=*]
    \item \textit{Monotonicity.} $A \subseteq B \implies \mathbb{P}(A) \leq \mathbb{P}(B)$.
    \item \textit{Intersection.} $\mathbb{P}(A \cap B) \leq \min(\mathbb{P}(A), \mathbb{P}(B))$.
    \item \textit{Union Bound.} $\mathbb{P}(A \cup B) \leq \mathbb{P}(A) + \mathbb{P}(B)$.
    \item \textit{Complement.} $\mathbb{P}(\Omega - A) = 1 - \mathbb{P}(A)$.
    \item \textit{Law of Total Probability.} If $A_1, \dotsc, A_k$ are a set of disjoint events such that $\bigcup^k_{i=1} A_i = \Omega$, then $\sum^k_{i=1} \mathbb{P}(A_i) = 1$.
\end{itemize}
\end{properties}

\vspace{3mm}

\begin{example}
    Consider tossing a six-sided die. The sample space is $\Omega = \{1, 2, 3, 4, 5, 6\}$. We can define different event spaces and probability measures on this sample space. For example, the simplest event space is the trivial event space $\mathcal{F} = \{\emptyset, \Omega\}$. Note that this $\mathcal{F}$ is a $\sigma$-algebra, as $\emptyset$ and $\Omega$ are complements of each other. The unique probability measure for this $\mathcal{F}$ satisfying the requirements above is given by $\mathbb{P}(\emptyset) = 0$, $\mathbb{P}(\Omega) = 1$. Another event space is the set of all subsets of $\Omega$. We can construct a valid probability measure for this $\mathcal{F}$ by assigning the probability of each set in the event space to be $\frac{i}{6}$ where $i$ is the number of elements of that set; for example, $\mathbb{P}(\{1, 2, 3, 4\}) = \frac{4}{6}$ and $\mathbb{P}(\{1, 2, 3\}) = \frac{3}{6}$. Intuitively, this probability measure could correspond to the probability that a random fair die roll belongs to a given subset of $\Omega$.
\end{example}

\subsection{Independence of Events}

Now that we understand the notion of an \textit{event}, we can make statements on how events relate to each other. In particular, the concepts of \textit{dependence} and \textit{independence} are fundamental in probability and statistics.

\begin{definition}[Independence of events] \label{def:independence_events}
    Let $A$ and $B$ be events. $A$ and $B$ are independent (denoted $A \ind B$) if and only if the following statements are true.
    \begin{align*}
        A \ind B \iff \mathbb{P}(A \cap B) &= \mathbb{P}(A)\mathbb{P}(B) \\
        A \ind B \iff \mathbb{P}(A \mid B) &= \mathbb{P}(A).
    \end{align*}
\end{definition}
In words, $A$ and $B$ are independent if and only if their joint probability is equal to the product of the probabilities of each individual event. Alternatively, we can say that the conditional probability of $A$ given that we known $B$ is equal to the probability of $A$ alone. Intuitively, if $A$ and $B$ are independent, then observing $B$ does not have any effect on the probability of $A$. 

Independence generalizes beyond the two-event case. A finite set of events is \textit{pairwise independent} if every pair in the set is independent. A finite set of events is \textit{mutually independent} if every event is independent of the intersection of any subset of events. Additionally, we have the fundamental notion of \textit{conditional independence}: events $A$ and $B$ are conditionally independent if they are independent given another finite set of events (i.e., once we have already observed the other set of events, observing $B$ does not have any effect on the probability of $A$). Later, we will see how the notion of conditional independence of random variables plays a pivotal role in graphical modeling and structure learning.

\subsection{Conditional Probability}

Conditional probability is a measure of the probability of an event given that another event has already occurred. Conditional probabilities show up frequently throughout probability theory and probabilistic graphical modeling, in such laws as Baye's rule (Definition \ref{def:bayes_rule}) and the chain rule (Definitions \ref{def:chain_rule}, \ref{def:chain_rule_variables}).

\begin{definition}[Conditional probability of events] \label{def:conditional_probability_events}
    Let $B$ be an event with non-zero probability. The conditional probability of any event $A$ given $B$ is defined as
    \[
    \mathbb{P}(A \mid B) = \frac {\mathbb{P}(A \cap B)}{\mathbb{P}(B)}.
    \]
    In words, $\mathbb{P}(A \mid B)$ is the probability measure of the event $A$ after observing the occurrence of event $B$.
\end{definition}

\begin{definition}[Chain rule of probability] \label{def:chain_rule}
    Let $A_1, \dotsc, A_k$ be events, $\mathbb{P}(A_i) >0$. Then the chain rule states that
    \begin{align*}
    & \mathbb{P}(A_1 \cap A_2 \cap \dotsb \cap A_k) \\
    &= \mathbb{P}(A_1) \mathbb{P}(A_2 | A_1) \mathbb{P}(A_3 | A_2 \cap A_1 ) \dotsb \mathbb{P}(A_k | A_1 \cap A_2 \cap \dotsb \cap A_{k-1}).
    \end{align*}
\end{definition}
Note that for $k=2$ events, this is just the definition of conditional probability:
\[
\mathbb{P}(A_1 \cap A_2) = \mathbb{P}(A_1) \mathbb{P}(A_2 | A_1).
\]
In general, the chain rule is derived by applying the definition of conditional probability multiple times, as in the following example.

\begin{example}[Chain rule applied to four events] Given events $A_1, A_2, A_3, A_4$:
    \begin{align*}
& \mathbb{P}(A_1 \cap A_2 \cap A_3 \cap A_4) \\
&= \mathbb{P}(A_1 \cap A_2 \cap A_3) \mathbb{P}(A_4 \mid A_1 \cap A_2 \cap A_3) \\
&= \mathbb{P}(A_1 \cap A_2) \mathbb{P}(A_3 \mid A_1 \cap A_2) \mathbb{P}(A_4 \mid A_1 \cap A_2 \cap A_3) \\
&= \mathbb{P}(A_1) \mathbb{P}(A_2 \mid A_1) \mathbb{P}(A_3 \mid A_1 \cap A_2) \mathbb{P}(A_4 \mid A_1 \cap A_2 \cap A_3).
\end{align*}
\end{example}

\section{Random Variables}
\label{sec:random_variables}

Oftentimes, we do not care to know the probability of a particular event. Instead, we want to know probabilities over some function of these events. For example, consider an experiment in which we flip 10 coins. Here, the elements of the sample space $\Omega$ are length-10 sequences of heads and tails, and the event space $\mathcal{F}$ is all subsets of $\Omega$. We observe sequences of coin flips; for instance, $\omega_0 = \langle H, H, T, H, T, H, H, T, T, T \rangle \in \Omega$. However, in practice we may not care to directly know the particular probability $\mathbb{P}(\omega_0)$ of a sequence or even the probability over a set of sequences in $\mathcal{F}$. Instead, we might want to know the number of coins that come up heads or the length of the longest run of tails. These quantities are functions of $\omega \in \Omega$, which we refer to as \textit{random variables}.

More formally, define a mapping $X : \Omega \to E$ between two measurable spaces $(\Omega, \mathcal{F})$ and $(E, \mathcal{E})$, where $\mathcal{E}$ is a $\sigma$-algebra on $E$. Then, $X$ is a random variable if $X^{-1}(B) := \{\omega: X(\omega) \in B \} \in \mathcal{F}$ for all $B \in \mathcal{E}$. Intuitively, this means that every set $B$ is associated with a set of outcomes that belongs to $\mathcal{F}$ and has a well-defined probability. Typically, we denote random variables using upper case letters $X(\omega)$ or more simply $X$ (where the dependence on the random outcome $\omega$ is implied). We denote the value that a random variable may take on using lower case letters $x$. Thus, $X = x$ denotes the event that the random variable $X$ takes on the value $x \in E$.

\begin{example}
    In our experiment above, suppose that $X(\omega)$ is the number of heads which occur in the sequence of tosses $\omega$. Given that only 10 coins are tossed, $X(\omega)$ can take only a finite number of values (0 through 10), so it is known as a discrete random variable. Here, the probability of the set associated with a random variable $X$ taking on some specific value $k$ is $\mathbb{P}(X = k) := \mathbb{P}(\{\omega : X(\omega) = k\}) = \mathbb{P}(\omega \in \text{all sequences with k heads})$. Note that the set of all sequences with $k$ heads is an element of $\mathcal{F}$, given that $\mathcal{F}$ consists of all subsets of $\Omega$.
\end{example}

\begin{example}
    Suppose that $X(\omega)$ is a random variable indicating the amount of time it takes for a radioactive particle to decay ($\omega$ for this example could be some underlying characterization of the particle that changes as it decays). In this case, $X(\omega)$ takes on an infinite number of possible values, so it is called a continuous random variable. We denote the probability that $X$ takes on a value between two real constants $a$ and $b$ (where $a < b$) as $\mathbb{P}(a \leq X \leq b) := \mathbb{P}(\{\omega : a \leq X(\omega) \leq b\})$.
\end{example}

When describing the event that a random variable takes on a certain value, we often use the \textit{indicator function} $\mathbf{1}\{A\}$ which takes value 1 when event $A$ happens and 0 otherwise. For example, for a random variable $X$,
\[
\mathbf{1}\{X > 3\} = \begin{cases}
    1, & \text{if }X > 3 \\
    0, & \text{otherwise}
    \end{cases}
\]

In order to specify the probability measures used when dealing with random variables, it is often convenient to specify alternative functions from which the probability measure governing an experiment immediately follows. In the following three sections, we describe these functions: the cumulative distribution function (CDF), the probability mass function (PMF) for discrete random variables, and the probability density function (PDF) for continuous random variables. For the rest of this section, we suppose that $X$ takes on real values, i.e., $E = \mathbb{R}$. 

\subsection{Cumulative Distribution Functions}

\begin{definition}[Cumulative distribution function (CDF)] \label{def:cdf}
    A CDF is a function $F_X : \mathbb{R} \to [0, 1]$ which specifies a probability measure as
    \[
    F_X(x) \coloneqq \mathbb{P}(X \leq x).
    \]
    By using this function, one can calculate the probability that $X$ takes on a value between any two real constants $a$ and $b$ (where $a < b$).
\end{definition}

\par \vspace{2mm} \begin{properties}
\begin{itemize}[leftmargin=*]
    \item $0 \leq F_X(x) \leq 1$. This follows from the definition of the probability measure.
    \item $\lim_{x \to -\infty} F_X(x) = 0$. As $x$ approaches $-\infty$, the corresponding set of $\omega$ where $X(\Omega) \le x$ approaches $\emptyset$, for which $\mathbb{P}(\emptyset) = 0$. 
    \item $\lim_{x \to +\infty} F_X(x) = 1$. As $x$ approaches $\infty$, the corresponding set of $\omega$ where $X(\omega) \le x$ approaches $\Omega$, for which $\mathbb{P}(\Omega) = 1$.
    \item $x \leq y \implies F_X(x) \leq F_X(y)$. This follows from the fact that the event that $X \le x$ is a subset of $X \le y$ for $x \leq y$.
    \end{itemize}
\end{properties}

\subsection{Probability Mass Functions}

Suppose a random variable $X$ takes on a finite set of possible values (i.e., $X$ is a discrete random variable). A simpler way to represent the probability measure associated with a random variable is to directly specify the probability of each value that the random variable can assume. Let $\text{Val}(X)$ refer to the set of possible values that the random variable $X$ may assume. For example, if $X(\omega)$ is a random variable indicating the number of heads out of ten tosses of coin, then $\text{Val}(X) = \{0, 1, 2, \dotsc, 10\}$. We can then define a probability mass function with respect to $X$. 

\begin{definition}[Probability mass function (PMF)] \label{def:pmf}
    A PMF is a function $p_X : \text{Val}(X) \to [0, 1]$ such that $p_X(x) = \mathbb{P}(X = x)$.
\end{definition} 

\par \vspace{2mm} \begin{properties}
    \begin{itemize}[leftmargin=*]
        \item $0 \leq p_X(x) \leq 1$.
        \item $\sum_{x \in A} p_X(x) = \mathbb{P}(X \in A)$, as probability measures apply over countable unions of disjoint sets. 
        \item $\sum_{x \in \text{Val}(X)} p_X(x) = 1$. Applying the previous property, we have that $\sum_{x \in \text{Val}(X)} p_X(x) = \mathbb{P}(X \in \text{Val}(X)) = \mathbb{P}(\Omega) = 1$.
    \end{itemize}
\end{properties}

\subsection{Probability Density Functions}

For some continuous random variables, the cumulative distribution function $F_X(x)$ is differentiable everywhere. In these cases, we define the \textit{probability density function} (PDF) as the derivative of the CDF.
\begin{definition}[Probability density function (PDF)]
    \[
    f_X(x) \coloneqq \frac{dF_X(x)}{dx}.
    \]
\end{definition}
Note that the PDF for a continuous random variable may not always exist (i.e., if $F_X(x)$ is not differentiable everywhere). According to the properties of differentiation, for very small $\delta x$,
\[
\mathbb{P}(x \leq X \leq x + \delta x) \approx f_X(x) \delta x.
\]
Both CDFs and PDFs (when they exist) can be used for calculating the probabilities of different events. But it should be emphasized that the value of PDF at any given point $x$ is not the probability of that event, i.e., $f_X(x) \neq \mathbb{P}(X = x)$. Because $X$ can take on infinitely many values, it holds that $\mathbb{P}(X = x) = 0$. On the other hand, $f_X(x)$ can take on values larger than one (but the integral of $f_X(x)$ over any subset of $\mathbb{R}$ will be at most one).

\par \vspace{2mm} \begin{properties}
    \begin{itemize}[leftmargin=*]
    \item $f_X(x) \geq 0$.
    \item $\int^{\infty}_{-\infty} f_X(x) dx = 1$.
    \item $\int_{x \in A} f_X(x) dx = \mathbb{P}(X \in A)$.
    \end{itemize}
\end{properties}

\subsection{Expectation}

The \textit{expectation} or \textit{expected value} of a random variable tells us something about what outcome to expect in the average case. In the discrete setting, the expectation is a generalization of the weighted average: it is the arithmetic mean of the possible outcomes of a random variable weighted by the probabilities of observing these outcomes. Given this arithmetic operation, the expectation can take on a value that itself is never observed in reality. In the continuous case, summation is replaced by integration. We can take expected values of functions of random variables, or of random variables themselves.

\begin{definition}[Expected value] \label{def:expectation}
Let $X$ denote a discrete random variable with PMF $p_X(x)$ and $g : \mathbb{R} \to \mathbb{R}$ denote an arbitrary function. In this case, $g(X)$ can be considered a random variable, with an associated \textit{expectation} or \textit{expected value}. We define the expected value of $g(X)$ as
    \[
    \mathbb{E}[g(X)] \coloneqq \sum_{x \in \text{Val}(X)} g(x)p_X(x).
    \]
If $X$ is a continuous random variable with PDF $f_X(x)$, then the expected value of g(X) is defined as
\[
\mathbb{E}[g(X)] \coloneqq \int^{\infty}_{-\infty} g(x)f_X(x)dx.
\]
\end{definition}

Intuitively, the expectation of $g(X)$ can be thought of as a weighted average of the values that $g(x)$ can take on, where the weights are given by $p_X(x)$ or $f_X(x)$ which add up to $1$ over all $x$. As a special case of the above, note that the expectation, $\mathbb{E}[X]$ of a random variable itself is found by letting $g(x) = x$; this is also known as the mean of the random variable $X$.

\par \vspace{2mm} \begin{properties}
    \begin{itemize}[leftmargin=*]
    \item $\mathbb{E}[a] = a$ for any constant $a \in \mathbb{R}$.
    \item $\mathbb{E}[af(X)] = a\mathbb{E}[f(X)]$ for any constant $a \in \mathbb{R}$.
    \item \textit{Linearity of Expectation.}
        \[
        \mathbb{E}[f(X) + g(X)] = \mathbb{E}[f(X)] + \mathbb{E}[g(X)].
        \]
    \item For a discrete random variable $X$,
        \[
        \mathbb{E}[\mathbf{1}\{X = k\}] = \mathbb{P}(X = k).
        \]
    \end{itemize}
\end{properties}

\subsection{Variance}

\begin{definition}[Variance]
    The variance of a random variable $X$ is a measure of how concentrated the distribution of a random variable $X$ is around its mean. Formally, the variance of a random variable $X$ is defined
    \[
    \text{Var}[X] \coloneqq \mathbb{E}[(X - \mathbb{E}[X])^2].
    \]
\end{definition}

Using the properties in the previous section, we can derive an alternate expression for variance:
\begin{align*}
& \mathbb{E}[(X - \mathbb{E}[X])^2] \\
&= \mathbb{E}[X^2 - 2\mathbb{E}[X]X + \mathbb{E}[X]^2] \\
&= \mathbb{E}[X^2] - 2\mathbb{E}[X]\mathbb{E}[X] + \mathbb{E}[X]^2 \\
&= \mathbb{E}[X^2] - \mathbb{E}[X]^2,
\end{align*}
where the second equality follows from the linearity of expectation and the fact that $\mathbb{E}[X]$ is actually a constant with respect to the outer expectation.

\par \vspace{2mm} \begin{properties}
    \begin{itemize}[leftmargin=*]
    \item $\text{Var}[a] = 0$ for any constant $a \in \mathbb{R}$.
    \item $\text{Var}[af(X)] = a^2 \, \text{Var}[f(X)]$ for any constant $a \in \mathbb{R}$.
    \end{itemize}
\end{properties}

 \vspace{2mm} 

\begin{example}
    Calculate the mean and the variance of the uniform random variable $X$ with PDF $f_X(x) = 1, \forall x \in [0, 1], 0$ elsewhere.
    \begin{align*}
    \mathbb{E}[X] &= \int^{\infty}_{-\infty} x f_X(x) dx = \int^1_0 x dx = \frac{1}{2} \\
    \mathbb{E}[X^2] &= \int^{\infty}_{-\infty} x^2 f_X(x)dx = \int^1_0 x^2 dx = \frac{1}{3} \\
    \text{Var}[X] &= \mathbb{E}[X^2] - \mathbb{E}[X]^2 = \frac{1}{3} - \frac{1}{4} = \frac{1}{12}
    \end{align*}
\end{example}

\begin{example}
    Suppose that $g(x) = \mathbf{1}\{x \in A\}$ for some subset $A \subseteq \Omega$. What is $\mathbb{E}[g(X)]$? \\

    \noindent \textit{Discrete case.}
    \[
    \mathbb{E}[g(X)] = \sum_{x \in \text{Val}(X)} \mathbf{1}\{x \in A \} P_X(x) = \sum_{x \in A} P_X(x) = \mathbb{P}(X \in A).
    \]
    \textit{Continuous case.}
    \[
    \mathbb{E}[g(X)] = \int_{-\infty}^\infty \mathbf{1}\{x \in A \} f_X(x) dx = \int_{x\in A} f_X(x) dx = \mathbb{P}(X \in A).
    \]
\end{example}

\subsection{Common Random Variables}

Here, we define several useful random variables that are commonly encountered in probabilistic modeling. See \citet{ross2010introduction, murphy2012machine} for additional examples and elaboration on their properties.

\renewcommand{\arraystretch}{1.25}
\begin{table}[!t]
    \centering
    \begin{tabular}{l l l c c}
        \toprule[1pt]
        & \multicolumn{2}{c}{\textsc{Distribution}} & \textsc{Mean} & \textsc{Variance} \\
        \midrule[1pt]
        \textit{Discrete} & Bernoulli & Ber($p$) & $p$ & $p(1-p)$\\
        & Binomial & Bin($n,p$) & $np$ & $np(1-p)$ \\
        & Geometric & Geo($p$) & $\frac{1}{p}$ & $\frac{1-p}{p^2}$ \\
        & Poisson & Poi($\lambda$) & $\lambda$ & $\lambda$ \\
        \midrule
        \textit{Continuous} & Uniform & U(\textit{a,b}) & $\frac{a+b}{2}$ & $\frac{(b-a)^2}{12}$ \\
        & Gaussian & $\mathcal{N}(\mu,\sigma^2)$ & $\mu$ & $\sigma^2$ \\
        & Exponential & Expon($\lambda$)  & $\frac{1}{\lambda}$ & $\frac{1}{\lambda^2}$ \\
        \bottomrule[1pt]
    \end{tabular}
    \caption{Means and variances for common probability distributions. %Adapted from \citet{maleki_2020}.
    }
    \label{tab:rv_discrete_means_variances}
\end{table}
\renewcommand{\arraystretch}{1}
\subsubsection{Discrete Random Variables}

\begin{definition}[Bernoulli random variables]
     The Bernoulli distribution models the outcome of an experiment (or \textit{Bernoulli trial}) with potential values 0 and 1, which can be construed as 1 = \textit{success} and 0 = \textit{failure}. A random variable $X$ is said to be Bernoulli if its PMF $p_X(x)$ is given by the following form.
    \begin{align*}
        X &\sim \text{Ber}(p), \quad 0 \leq p \leq 1 \\
        p_X(x) &= \begin{cases}
            p, & \text{if }x = 1 \\
            1-p, & \text{if }x = 0.
        \end{cases}
    \end{align*}
    For example, a Bernoulli random variable can represent the outcome of a coin flip ($H=1, T=0$) that comes up heads with probability $p$.
\end{definition}

\begin{definition}[Binomial random variables]
    The binomial distribution models the number of successes in $n$ successive independent Bernoulli trials. A random variable $X$ is said to be binomial if its PMF $p_X(x)$ takes the following form.
    \begin{align*}
        X &\sim \text{Bin}(n, p), \quad 0 \leq p \leq 1 \\
        p_X(x) &= \binom{n}{x} \cdot p^x (1-p)^{n-x}.
    \end{align*}
    For example, a binomial random variable can represent the number of heads in $n$ independent flips of a coin with heads probability $p$. The Bernoulli distribution can also be viewed as a binomial distribution with $n = 1$.
\end{definition}

\begin{definition}[Geometric random variables]
    Imagine that successive independent Bernoulli trials are performed until we reach the first success, where each trial has a probability $p$ of success. A geometric random variable $X$ represents the number of trials required until the first success, with PMF $p_X(x)$ expressed as follows.
    \begin{align*}
        X &\sim \text{Geo}(p), \quad 0 \leq p \leq 1 \\
        p_X(x) &= p(1 - p)^{x-1}.
    \end{align*}
    Continuing with the coin flip example, a geometric random variable can represent the number of flips of a coin until the first heads, for a coin that comes up heads with probability $p$.
\end{definition}

\begin{definition}[Poisson  random variables]
    The Poisson distribution is a probability distribution over the nonnegative integers $\{0,1,2,\dots\}$ that models the frequency of rare events. A Poisson random variable with PMF $p_X(x)$ is given by
    \begin{align*}
        X &\sim \text{Poi}(\lambda), \quad \lambda > 0 \\
        p_X(x) &= e^{-\lambda} \frac{\lambda^x}{x!}
    \end{align*}
    where $x$ is the number of times the event of interest occurs, $e$ is Euler's number, and parameter $\lambda$ is the expected number of events. The term $e^{-\lambda}$ is a normalization constant that guarantees that the distribution sums to 1.
\end{definition}

\begin{figure}[!t]
    \centering
\definecolor{dodgerblue}{rgb}{0.12, 0.56, 1.0}
\definecolor{dogwoodrose}{rgb}{0.84, 0.09, 0.41}
\definecolor{mediumaquamarine}{rgb}{0.4, 0.8, 0.67}
%\begin{figure}
%    \centering
\begin{tikzpicture}[scale=0.7]
  \begin{axis}[
    no markers, domain=-5:5, samples=100,
    xlabel=$x$, ylabel={$f(x)$},
    height=6cm, width=10cm,
    legend style={at={(0.95,0.95)},anchor=north east},
    grid = major,
    ]

    % Gaussian with variance 1
    \addplot[dodgerblue, ultra thick] {1/sqrt(2*pi)*exp(-0.5*x^2)};
    \addlegendentry{$\sigma^2=1$}

    % Gaussian with variance 0.5
    \addplot[dogwoodrose, ultra thick, dotted] {1/sqrt(2*pi*0.5)*exp(-0.5*x^2/0.5)};
    \addlegendentry{$\sigma^2=0.5$}

    % Gaussian with variance 2
    \addplot[mediumaquamarine, ultra thick, dashed] {1/sqrt(2*pi*2)*exp(-0.5*x^2/2)};
    \addlegendentry{$\sigma^2=2$}

  \end{axis}
\end{tikzpicture}
%    \caption{Caption}
%    \label{fig:gaussian_variances}
%\end{figure}
 \\
\vspace{5mm}
\definecolor{dodgerblue}{rgb}{0.12, 0.56, 1.0}
\definecolor{dogwoodrose}{rgb}{0.84, 0.09, 0.41}
\definecolor{mediumaquamarine}{rgb}{0.4, 0.8, 0.67}
%\begin{figure}
%    \centering
\begin{tikzpicture}[scale=0.7]
  \begin{axis}[
    xlabel={$k$}, ylabel={$p(k)$}, %ylabel={Probability},
    xtick=data, legend style={at={(0.95,0.95)},anchor=north east},
    grid=major, width=10cm, height=6cm,
    ]

    % Poisson with lambda = 2
    \addplot[
      dodgerblue, mark=*,
      smooth, very thick, domain=0:10, samples at={0,1,...,10}
    ] {exp(-2) * 2^x / factorial(x)};
    \addlegendentry{$\lambda=2$}

    % Poisson with lambda = 4
    \addplot[
      dogwoodrose, mark=square*,
      smooth, very thick, domain=0:10, samples at={0,1,...,10}
    ] {exp(-4) * 4^x / factorial(x)};
    \addlegendentry{$\lambda=4$}

    % Poisson with lambda = 6
    \addplot[
      mediumaquamarine, mark=triangle*,
      smooth, very thick, domain=0:10, samples at={0,1,...,10}
    ] {exp(-6) * 6^x / factorial(x)};
    \addlegendentry{$\lambda=6$}

  \end{axis}
\end{tikzpicture}
    %\caption{Poisson PMFs with different values of $\lambda$.}
    %\label{fig:poisson_pmf_bars}
%\end{figure}
    \caption{Gaussian PDF (top) and Poisson PMF (bottom) with varying parameters. Note that the Poisson PMF is defined over integer values, and smoothed curves are visual guides only.}
    \label{fig:gaussian_poisson}
\end{figure}
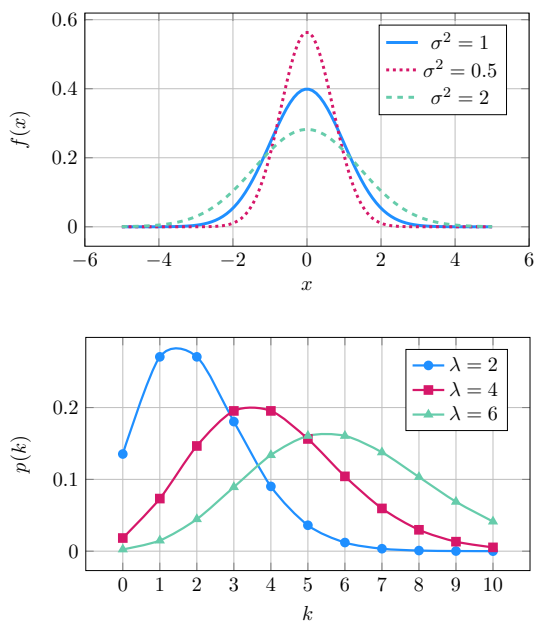

\subsubsection{Continuous Random Variables}

\begin{definition}[Uniform random variables]
    A uniformly distributed random variable with PDF $f_X(x)$ assigns equal probability density to every value between $a$ and $b$ on the real line.
    \begin{align*}
        X &\sim \text{U}(a, b), \quad a < b \\
        f_X(x) &= \begin{cases}
        \frac{1}{b-a}, & \text{if }a \leq x \leq b \\
        0, & \text{otherwise}.
    \end{cases}
    \end{align*}
\end{definition}

\begin{definition}[Gaussian random variables]
    The Gaussian or \textit{normal} distribution is parameterized by mean $\mu$ and variance $\sigma^2$. Its PDF $f_X(x)$ takes the form of a bell-shaped curve that is symmetric around $\mu$.
    \begin{align*}
        X &\sim \mathcal{N}(\mu, \sigma^2) \\
        f_X(x) &= \frac{1}{\sqrt{2\pi}\sigma}e^{-\frac{(x-\mu)^2}{2\sigma^2}}.
    \end{align*}
    The properties of univariate and multivariate Gaussian distributions offer many mathematical conveniences. Historically, Gaussian distributions have been commonly assumed to model continuous random variables of unknown form in the natural and social sciences.
\end{definition}

\begin{figure}[!t]
    \centering
\definecolor{royalblue}{rgb}{0.25, 0.41, 0.88}
\begin{tikzpicture}[scale=0.6]
  \begin{axis}[
      width=9.7cm, height=8cm,
      samples=100,
      domain=0:5,
      axis lines=middle,
      axis line style ={line width=1pt, draw=black},
      xtick={0,1,2,3,4,5},
      ytick={0,0.2,0.4,0.6,0.8},
      ymax=1,
      grid=both,
      major grid style={line width=.2pt, draw=gray!50},
      minor grid style={line width=.2pt, draw=gray!50},
      xlabel={$x$},
      ylabel={$F(x)$},
      xlabel style={at={(axis description cs:0.5,-0.1)}, anchor=north},
      ylabel style={at={(axis description cs:-0.07,0.95)}, anchor=south},
    ]
    \addplot[
      very thick,
      royalblue,
    ]
    {1 - exp(-x)};
    \node[font=\Large] at (axis cs:1,0.5) [anchor=north west] {$F(x) = 1 - e^{-\lambda x}, \; \; \lambda=1$};
  \end{axis}
\end{tikzpicture}
\definecolor{crimson}{rgb}{0.86, 0.08, 0.24}
\begin{tikzpicture}[scale=0.6]
  \begin{axis}[
      width=9.7cm, height=8cm,
      samples=100,
      domain=0:5,
      axis lines=middle,
      axis line style ={line width=1pt, draw=black},
      xtick={0,1,2,3,4,5},
      ytick={0,0.2,0.4,0.6,0.8},
      grid=both,
      major grid style={line width=.2pt, draw=gray!50},
      minor grid style={line width=.2pt, draw=gray!50},
      xlabel={$x$},
      ylabel={$f(x)$},
      xlabel style={at={(axis description cs:0.5,-0.1)}, anchor=north},
      ylabel style={at={(axis description cs:-0.07,0.95)}, anchor=south},
    ]
    \addplot[
      very thick,
      crimson,
    ]
    {exp(-x)};
    \node[font=\Large] at (axis cs:1.5,0.5) [anchor=north west] {$f(x) = \lambda e^{-\lambda x}, \; \; \lambda=1$};
  \end{axis}
\end{tikzpicture}
    \caption{The CDF (left) and PDF (right) of an exponential random variable with $\lambda = 1$.}
    \label{fig:exponential_cdf_pdf}
\end{figure}
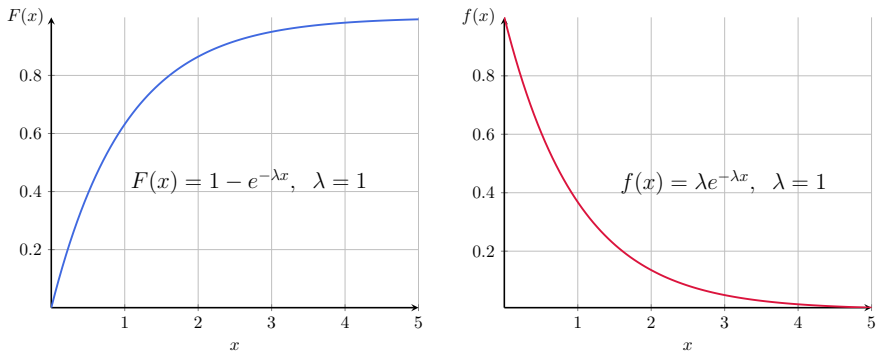

\begin{definition}[Exponential random variables]
    The exponential distribution models the distance (or time) between events in a Poisson process with a constant average rate of change $\lambda$, wherein events occur continuously and independently. It is a special case of the gamma distribution and the continuous analog to the geometric distribution. Its PDF $f_X(x)$ is a decaying probability density over the nonnegative reals.
    \begin{align*}
        X &\sim \text{Expon}(\lambda), \quad \lambda > 0 \\
        f_X(x) &= \begin{cases}
        \lambda e^{-\lambda x}, & \text{if }x \geq 0 \\
        0, & \text{otherwise}.
    \end{cases}
    \end{align*}
\end{definition}

\subsection{Joint and Marginal Cumulative Distribution Functions}

Thus far, we have considered single random variables. In many situations, however, there may be more than one quantity that we are interested in knowing during a random experiment. For instance, in an experiment where we flip a coin ten times, we may care about both the number of heads that come up (denoted $X(\omega)$) as well as the length of the longest run of consecutive heads (denoted $Y(\omega)$).

For ease of exposition, the remainder of this chapter will consider the setting of two random variables. However, the general multivariate case can consider arbitrarily many variables. We first discuss joint and marginal CDFs, then joint and marginal PMFs and PDFs.

Suppose that we have two random variables, $X$ and $Y$. One way to work with these two random variables is to consider each of them separately. If we do that we will need the CDFs $F_X(x)$ and $F_Y (y)$. But if we want to know about the values that $X$ and $Y$ assume simultaneously during outcomes of a random experiment, we require a more complicated structure known as the \textit{joint CDF} of $X$ and $Y$.
\begin{definition}[Joint cumulative distribution function of two random variables] The joint CDF is given by
    \begin{align*}
        F_{XY} (x, y) = \mathbb{P}(X \leq x, Y \leq y).
    \end{align*}
\end{definition}

Given the joint CDF, the probability of any event involving $X$ and $Y$ can be calculated. The joint CDF $F_{XY} (x, y)$ and the CDFs $F_X(x)$ and $F_Y (y)$ of each variable separately are related as follows.
\begin{definition}[Marginal cumulative distribution function] The marginal CDFs $F_X(x)$ and $F_Y(y)$ of $F_{XY} (x, y)$ are given by
    \begin{align*}
    F_X(x) &= \lim_{y \to \infty} F_{XY} (x, y) \\
    F_Y(y) &= \lim_{x \to \infty} F_{XY} (x, y).
    \end{align*}
\end{definition}

\par \vspace{2mm} \begin{properties}
    \begin{itemize}[leftmargin=*]
    \item $0 \leq F_{XY} (x, y) \leq 1$.
    \item $\lim_{x,y\to \infty} F_{XY} (x, y) = 1$.
    \item $\lim_{x,y\to -\infty} F_{XY} (x, y) = 0$.
    \end{itemize}
\end{properties}

\subsection{Joint and Marginal Probability Mass Functions}

Like the CDF, the notion of joint and marginal distributions can be applied to the probability mass functions of discrete random variables.

\begin{definition}[Joint probability mass function of two random variables]
    If $X$ and $Y$ are discrete random variables, then the joint PMF $p_{XY} : \text{Val}(X) \times \text{Val}(Y) \to [0, 1]$ is defined by
    \[
    p_{XY}(x, y) = \mathbb{P}(X = x, Y = y)
    \]
    where $0 \leq p_{XY}(x, y) \leq 1$ for all $x, y,$ and
    \[
    \sum_{x \in \text{Val}(X)} \sum_{y \in \text{Val}(Y)} p_{XY}(x, y) = 1.
    \]
\end{definition}

The joint PMF over two variables relates to the PMF for each variable separately as follows.
\begin{definition}[Marginal probability mass function]
    The marginal PMF $p_X(x)$ of $X$ is given by
    \[
    p_X(x) = \sum_y p_{XY} (x, y),
    \]
    and analogously for $p_Y(y)$ of $Y$.
\end{definition}
In statistics, the process of forming the marginal distribution with respect to one variable by summing out the other variable is often known as \textit{marginalization}.

\subsection{Joint and Marginal Probability Density Functions}

Let $X$ and $Y$ be two continuous random variables with joint CDF $F_{XY}$. In the case that $F_{XY}(x, y)$ is everywhere differentiable in both $x$ and $y$, then we can define the \textit{joint PDF}.
\begin{definition}[Joint probability density function of two random variables] The joint PDF for random variables $X$ and $Y$ takes the form
    \begin{align*}
        f_{XY}(x, y) = \frac{\partial^2F_{XY}(x, y)}{\partial x \partial y}.
    \end{align*}
\end{definition}

Like in the single-dimensional case, $f_{XY} (x, y) \neq \mathbb{P}(X = x, Y = y)$, but rather
\[
\int \int_{(x,y) \in A} f_{XY} (x, y) dx dy = \mathbb{P}((X, Y) \in A).
\]
Note that the values of the PDF $f_{XY}(x, y)$ are always nonnegative, but they may be greater than 1. Nonetheless, it must be the case that $\int^{\infty}_{-\infty} \int^{\infty}_{-\infty} f_{XY}(x,y) = 1$.

Analogous to the discrete case, we can define the \textit{marginal PDFs} or \textit{marginal densities} of $X$ and $Y$.
\begin{definition}[Marginal probability density function] The marginal PDF for $X$ is given as
    \begin{align*}
        f_X(x) = \int^{\infty}_{-\infty} f_{XY} (x, y)dy,
    \end{align*}
    and analogously for $f_Y (y)$ of $Y$.
\end{definition}

\subsection{Conditional Distributions}

Conditional distributions seek to answer the question, \textit{what is the probability distribution over $Y$, when we know that $X$ must take on a certain value $x$}?

\begin{definition}[Conditional probability mass function]
    In the discrete case, the conditional PMF of $Y$ given $X$ is simply
    \[
    p_{Y \mid X} (y \mid x) = \frac{p_{XY}(x, y)}{p_X(x)},
    \]
    assuming that $p_X(x) \neq 0$.
\end{definition}

In the continuous case, the situation is technically a little more complicated because the probability that a continuous random variable $X$ takes on a specific value $x$ is equal to zero. Ignoring this technical point, we simply define, by analogy to the discrete case, the \textit{conditional probability density} of $Y$ given $X = x$ as follows.

\begin{definition}[Conditional probability density function]
    For continuous random variables $X$ and $Y$, the conditional PDF of $Y$ given $X$ is simply
    \[
    f_{Y \mid X}(y \mid x) = \frac{f_{XY} (x, y)}{f_X(x)},
    \]
    provided $f_X(x) \neq 0$.
\end{definition}

\subsection{Chain Rule}

The chain rule that we previously derived for events (Definition \ref{def:chain_rule}) can also be applied to random variables.
\begin{definition}[Chain rule for random variables] \label{def:chain_rule_variables}
    Let $\mathbf{X} = \{X_i\}_{i=1}^n$ be set of random variables. By the chain rule of probability, the joint distribution over $\mathbf{X}$ factorizes as
    \begin{align*}
    %& p_{X_1, \cdots X_n} (x_1, \cdots, x_n) \\
    %&= p_{X_1} (x_1) p_{X_2 \mid X_1} (x_2 \mid x_1) \cdots p_{X_n \mid X_1, \cdots, X_{n-1}} (x_n \mid x_1, \cdots, x_{n-1}) \\
    %&= 
    p(\mathbf{x}) &= \prod_{i=1}^n p(x_i \mid x_1, \cdots , x_{i-1}) 
    = \prod_{i=1}^n p(x_i \mid \mathbf{x}_{<i}).
    \end{align*}
\end{definition}

\noindent Thus, for $\mathbf{X} = \{X_1, X_2, X_3, X_4\}$, we have the factorization
\[
p(\mathbf{x}) = p(x_1)p(x_2 \mid x_1)p(x_3 \mid x_1,x_2)p(x_4 \mid x_1,x_2,x_3).
\]

\subsection{Bayes' Rule}

A useful formula that often arises when deriving expressions for conditional probability is \textit{Bayes' rule}, also known as \textit{Bayes' theorem}. This formula arises from the chain rule.

\begin{definition}[Bayes' rule] \label{def:bayes_rule}
For discrete random variables $X$ and $Y$, Bayes' rule states
    \[
    P_{Y \mid X}(y \mid x) = \frac{P_{XY}(x, y)}{P_X(x)} = \frac{P_{X \mid Y} (x \mid y) P_Y(y)}{\sum_{y' \in \text{Val}(Y)} P_{X \mid Y} (x \mid y') P_Y(y')}.
    \]
    When $X$ and $Y$ are continuous, we have
    \[
    f_{Y \mid X}(y\mid x) = \frac{f_{XY}(x, y)}{f_X(x)} = \frac{f_{X \mid Y} (x \mid y) f_Y(y)}{\int^{\infty}_{- \infty} f_{X\mid Y} (x \mid y') f_Y (y') dy'}.
    \]
\end{definition}

\subsection{Independence of Random Variables}

The notion of independence among random variables is central to factorizing joint distributions, learning graphical structures, and many other fundamental tasks in probabilistic graphical modeling. Variables can be \textit{marginally independent} or \textit{conditionally independent} given a set of additional variables. We denote independence with the symbol $\ind$, and dependence with $\nind$.

\begin{definition}[Independence of random variables]
    Two random variables $X$ and $Y$ are independent if the following holds for all values of $x$ and $y$.
    \[
    F_{XY} (x, y) = F_X(x)F_Y(y) \Longrightarrow X \ind Y
    \]
    Equivalently,
\begin{itemize}
    \item For discrete random variables, $X \ind Y$ when
        \begin{itemize}
            \item $p_{XY} (x, y) = p_X(x)p_Y(y)$ for all $x \in \text{Val}(X)$, $y \in \text{Val}(Y)$.
            \item $p_{Y\mid X}(y \mid x) = p_Y(y)$ whenever $p_X(x) \neq 0$ for all $y \in \text{Val}(Y)$.
        \end{itemize}
    \item For continuous random variables, $X \ind Y$ when
        \begin{itemize}
            \item $f_{XY} (x, y) = f_X(x)f_Y(y)$ for all $x, y \in \mathbb{R}$.
            \item $f_{Y\mid X}(y \mid x) = f_Y(y)$ whenever $f_X(x) \neq 0$ for all $y \in \mathbb{R}$.
        \end{itemize}
\end{itemize}
\end{definition}

Informally, two random variables $X$ and $Y$ are independent if knowing the value of one variable will never have any effect on the conditional probability distribution of the other variable. That is, you know all the information about the pair $(X, Y)$ by just knowing $f(x)$ and $f(y)$. The following lemma formalizes this observation.

\begin{lemma}
    If $X$ and $Y$ are independent, then for any subsets $A, B \subseteq \mathbb{R}$, we have
        \[
        \mathbb{P}(X \in A, Y \in B) = \mathbb{P}(X \in A)\mathbb{P}(Y \in B).
        \]
\end{lemma}

By using the above lemma, one can prove that if $X$ is independent of $Y$ then any function of $X$ is independent of any function of $Y$.

\subsection{Expectation and Covariance}

Suppose that we have two random variables, $X$ and $Y$. Let $g : \mathbb{R}^2 \to \mathbb{R}$ denote a function of these two random variables.

\begin{definition}[Expected value of a function over two random variables]
    When $X$ and $Y$ are discrete, the expected value of $g$ is defined as
    \[
    \mathbb{E}[g(X,Y)] = \sum_{x \in \text{Val}(X)} \sum_{y \in \text{Val}(Y)} g(x, y)p_{XY}(x, y).
    \]
    For continuous random variables $X, Y$, the analogous expression is
    \[
    \mathbb{E}[g(X, Y)] = \int^{\infty}_{-\infty} \int^{\infty}_{-\infty} g(x, y)f_{XY}(x, y)dxdy.
    \]
\end{definition}

We can use the concept of expectation to study the relationship of two random variables with each other. In particular, we can examine their \textit{covariance}, a measure of their joint variability.

\begin{definition}[Covariance of two random variables]
    The covariance of two random variables $X$ and $Y$ is defined as
    \[
    \text{Cov}[X, Y] = \mathbb{E}[(X - \mathbb{E}[X])(Y - \mathbb{E}[Y])].
    \]
\end{definition}

Intuitively, the covariance of $X$ and $Y$ measures how often and by how much $X$ and $Y$ are both greater than or less than their respective means. If larger values of $X$ correspond with larger values of $Y$ and vice versa, then covariance is positive. If larger values of $X$ correspond with smaller values of $Y$ and vice versa, then covariance is negative. Using an argument similar to that for variance, we can rewrite this as
\begin{align*}
\text{Cov}[X, Y]
&= \mathbb{E}[(X - \mathbb{E}[X])(Y - \mathbb{E}[Y])] \\
&= \mathbb{E}[XY - X\mathbb{E}[Y] - Y \mathbb{E}[X] + \mathbb{E}[X]\mathbb{E}[Y]] \\
&= \mathbb{E}[XY] - \mathbb{E}[X]\mathbb{E}[Y] - \mathbb{E}[Y]\mathbb{E}[X] + \mathbb{E}[X]\mathbb{E}[Y] \\
&= \mathbb{E}[XY] - \mathbb{E}[X]\mathbb{E}[Y].
\end{align*}

Here, the key step in showing the equality of the two forms of covariance is in the third equality, where we use the fact that $\mathbb{E}[X]$ and $\mathbb{E}[Y]$ are actually constants which can be pulled out of the expectation. When $\text{Cov}[X, Y] = 0$, we say that $X$ and $Y$ are \textit{uncorrelated}. %\jm{Elaborate on usefulness of covariance, e.g., for statistical independence in jointly Gaussian distributions, structure learning, etc.}

\par \vspace{2mm} \begin{properties}
    \begin{itemize}[leftmargin=*]
    \item \textit{Linearity of expectation.}
        \[
        \mathbb{E}[f(X, Y) + g(X, Y)] = \mathbb{E}[f(X, Y)] + \mathbb{E}[g(X, Y)].
        \]
    \item $\text{Var}[X + Y] = \text{Var}[X] + \text{Var}[Y] + 2 \, \text{Cov}[X, Y]$.
    \item If $X$ and $Y$ are independent, then $\text{Cov}[X, Y] = 0$. However, if $\text{Cov}[X, Y] = 0$, it is not necessarily true that $X$ and $Y$ are independent. For example, let $X \sim \text{Uniform}(-1, 1)$ and let $Y = X^2$. Then, $\text{Cov}[X, Y] = \mathbb{E}[X^3] - \mathbb{E}[X]\mathbb{E}[X^2] = \mathbb{E}[X^3] - 0\cdot \mathbb{E}[X^2] = 0$ even though $X$ and $Y$ are not independent.
    \item If $X$ and $Y$ are independent, then $\mathbb{E}[f(X)g(Y)] = \mathbb{E}[f(X)]\mathbb{E}[g(Y)]$.
    \end{itemize}
\end{properties}

\vspace{5mm}

\begin{reading}
%\textit{Texts on probability theory}
    \begin{itemize}[leftmargin=*]
        \item \fullcite{grinstead1997introduction}.
        \item \fullcite{bertsekas2008introduction}.
        \item \fullcite{murphy2022probabilistic}.
    \end{itemize}
\end{reading}

\clearpage

%%%%%%%%%%%%%%%%%%%%%%%%%%
% Graph Theory
%%%%%%%%%%%%%%%%%%%%%%%%%%

\section{Basic Graph Theory}
\label{sec:graph_theory}

We will now review some basic graph theoretic concepts. Graph theory is the branch of mathematics that is broadly concerned with (1) the properties of graphical objects, which model the pairwise relationships among entities in a system; (2) the operations and transformations that can be performed on these objects; and (3) the design and analysis of algorithms that act on graph objects.

Fundamentally, a \textit{graph} is an abstract mathematical object defined by a set of \textit{nodes} and a set of \textit{edges} that link those nodes. Nodes denote entities in the system, while edges model their relationships.
\begin{definition}[Graph]
    Let $\mathcal{G} = (\mathbf{V},\mathbf{E})$ be a graph, where $\mathbf{V} = \{V_i\}_{i=1}^n$ is the  node set (also called vertices) and $\mathbf{E} = \{E_i\}_{i=1}^m$ is the edge set. %Let the cardinality of the node set be $|\mathbf{V}| = n$ and the cardinality of the edge set be $|\mathbf{E}| = m$.
\end{definition}

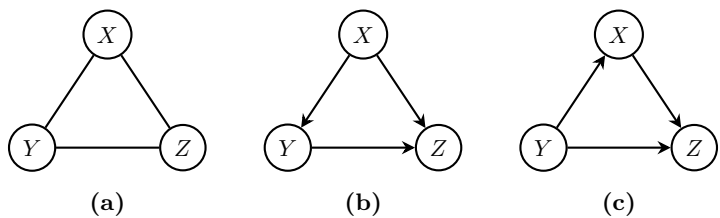
\begin{figure}
    \centering

\begin{tikzpicture}[every edge quotes/.style = {font=\footnotesize, fill=white,sloped}]
  \node[circle,black,thick,draw,scale=0.8] (x) at (1,1.5) {$X$};
  \node[circle,black,thick,draw,scale=0.8] (y) at (0, 0) {$Y$};
  \node[circle,black,thick,draw,scale=0.8] (z) at (2, 0) {$Z$};
  %%%
  \draw[thick,black,-]  (x) edge[] (y);
  \draw[thick,black,-]  (y) edge[] (z);
  \draw[thick,black,-]  (x) edge[] (z);
  %%%
  \node[] (label) at (1,-0.75) {\footnotesize\textbf{(a)}};
\end{tikzpicture}
%%%%%%%%%%%%%%%%
\hspace{5mm}
%%%%%%%%%%%%%%%%
\begin{tikzpicture}[every edge quotes/.style = {font=\footnotesize, fill=white,sloped}]
  \node[circle,black,thick,draw,scale=0.8] (x) at (1,1.5) {$X$};
  \node[circle,black,thick,draw,scale=0.8] (y) at (0, 0) {$Y$};
  \node[circle,black,thick,draw,scale=0.8] (z) at (2, 0) {$Z$};
  %%%
  \draw[thick,black,-{Stealth[width=5pt,length=5pt]}]  (x) edge[] (y);
  \draw[thick,black,-{Stealth[width=5pt,length=5pt]}]  (y) edge[] (z);
  \draw[thick,black,-{Stealth[width=5pt,length=5pt]}]  (x) edge[] (z);
  %%%
  \node[] (label) at (1,-0.75) {\footnotesize\textbf{(b)}};
\end{tikzpicture}
%%%%%%%%%%%%%%%%
\hspace{5mm}
%%%%%%%%%%%%%%%%
\begin{tikzpicture}[every edge quotes/.style = {font=\footnotesize, fill=white,sloped}]
  \node[circle,black,thick,draw,scale=0.8] (x) at (1,1.5) {$X$};
  \node[circle,black,thick,draw,scale=0.8] (y) at (0, 0) {$Y$};
  \node[circle,black,thick,draw,scale=0.8] (z) at (2, 0) {$Z$};
  %%%
  \draw[thick,black,-{Stealth[width=5pt,length=5pt]}]  (y) edge[] (x);
  \draw[thick,black,-{Stealth[width=5pt,length=5pt]}]  (y) edge[] (z);
  \draw[thick,black,-{Stealth[width=5pt,length=5pt]}]  (x) edge[] (z);
  %%%
  \node[] (label) at (1,-0.75) {\footnotesize\textbf{(c)}};
\end{tikzpicture}

    \caption{Undirected \textbf{(a)} and directed graphs \textbf{(b,c)} with the same adjacencies.}
    \label{fig:directed_vs_undirected}
\end{figure}

\paragraph{Undirected, Directed, and Mixed Graphs} Many kinds of graphs can arise. A basic dichotomy in graph theory is that of undirected graphs versus directed graphs. If the edges of $\mathcal{G}$ have no orientation or directionality (as signified by a lack of arrowheads), we say that $\mathcal{G}$ is an \textit{undirected graph} (Figure \ref{fig:directed_vs_undirected}a). If the edges of $\mathcal{G}$ do display a fixed orientation (e.g., $X \to Y$), we say that $\mathcal{G}$ is a \textit{directed graph} or \textit{digraph} (Figure \ref{fig:directed_vs_undirected}b). This is a false dichotomy, as there exist several noteworthy \textit{mixed graph} classes that contain both directed and undirected edge types. Mixed graphs can even contain \textit{bidirected edges} and edges with alternative endpoints denoting uncertain directionality. Such mixed graphs are famously useful for expressing uncertainty and unmeasured variables in \textit{causal graphical modeling}. For the remainder of this chapter, we will restrict our attention to directed and undirected graphs, but we will revisit mixed graphs in Section \ref{sec:structure_learning}.

\begin{figure}[!t]
    \centering

\begin{tikzpicture}[every edge quotes/.style = {font=\footnotesize, fill=white,sloped}]
  \node[circle,WildStrawberry,fill=WildStrawberry!10,thick,draw,scale=0.7] (u) at (0, 1) {\color{black}$U$};
  \node[circle,black,thick,draw,scale=0.7] (v) at (0, 0) {$V$};
  \node[] (label_pos) at (0,0.75) {}; 
  %%%
  \draw[thick,black,-{Stealth[width=5pt,length=5pt]}]  (u) edge[] (v);
  %%%
  \node[below=of label_pos,align=flush center,text width=3cm] (label) {\footnotesize\textbf{(a)} \textsc{parent}}; 
\end{tikzpicture}
%%%%%%%%%%%%%%%%%%%%%%
\hspace{0mm}
%%%%%%%%%%%%%%%%%%%%%%
\begin{tikzpicture}[every edge quotes/.style = {font=\footnotesize, fill=white,sloped}]
  \node[circle,black,thick,draw,scale=0.7] (u) at (0, 1) {$U$};
  \node[circle,WildStrawberry,fill=WildStrawberry!10,thick,draw,scale=0.7] (v) at (0, 0) {\color{black}$V$};
  \node[] (label_pos) at (0,0.75) {}; 
  %%%
  \draw[thick,black,-{Stealth[width=5pt,length=5pt]}]  (u) edge[] (v);
  %%%
  \node[below=of label_pos,align=flush center,text width=3cm] (label) {\footnotesize\textbf{(b)} \textsc{child}}; 
\end{tikzpicture} 
%%%%%%%%%%%%%%%%%%%%%%
\hspace{0mm}
%%%%%%%%%%%%%%%%%%%%%%
\begin{tikzpicture}[every edge quotes/.style = {font=\footnotesize, fill=white,sloped}]
  \node[circle,WildStrawberry,fill=WildStrawberry!10,thick,draw,scale=0.7] (u) at (0, 1) {\color{black}$U$};
  \node[circle,WildStrawberry,fill=WildStrawberry!10,thick,draw,scale=0.7] (v) at (0, 0) {\color{black}$V$};
  \node[] (label_pos) at (0,0.75) {}; 
  %%%
  \draw[thick,black,-]  (u) edge[] (v);
  %%%
  \node[below=of label_pos,align=flush center,text width=3cm] (label) {\footnotesize\textbf{(c)} \textsc{neighbors}}; 
\end{tikzpicture} \\
%%%%%%%%%%%%%%%%%%%%%%
\vspace{5mm}
%%%%%%%%%%%%%%%%%%%%%%
\begin{tikzpicture}[every edge quotes/.style = {font=\footnotesize, fill=white,sloped}]
  \node[circle,WildStrawberry,fill=WildStrawberry!10,thick,draw,scale=0.7] (u) at (0, 1) {\color{black}$U$};
  \node[circle,black,thick,draw,scale=0.7] (v) at (0, 0) {$V$};
  \node[circle,WildStrawberry,fill=WildStrawberry!10,thick,draw,scale=0.7] (w) at (0, -1) {\color{black}$W$};
  \node[] (label_pos) at (0,-0.25) {};  
  %%%
  \draw[thick,black,-{Stealth[width=5pt,length=5pt]}]  (u) edge[] (v);
  \draw[thick,black,-{Stealth[width=5pt,length=5pt]}]  (w) edge[] (v);
  %%%
  \node[below=of label_pos,align=flush center,text width=3.5cm] (label) {\footnotesize\textbf{(d)} \textsc{spouses} }; 
\end{tikzpicture} 
%%%%%%%%%%%%%%%%%%%%%%
\hspace{0mm}
%%%%%%%%%%%%%%%%%%%%%%
\begin{tikzpicture}[every edge quotes/.style = {font=\footnotesize, fill=white,sloped}]
  \node[circle,WildStrawberry,fill=WildStrawberry!10,thick,draw,scale=0.7] (u) at (0, 1) {\color{black}$U$};
  \node[circle,WildStrawberry,fill=WildStrawberry!10,thick,draw,scale=0.7] (v) at (0, 0) {\color{black}$V$};
  \node[circle,black,thick,draw,scale=0.7] (w) at (0, -1) {$W$};
  \node[] (label_pos) at (0,-0.25) {};  
  %%%
  \draw[thick,black,-{Stealth[width=5pt,length=5pt]}]  (u) edge[] (v);
  \draw[thick,black,-{Stealth[width=5pt,length=5pt]}]  (v) edge[] (w);
  %%%
  \node[below=of label_pos,align=flush center,text width=3.1cm] (label) {\footnotesize\textbf{(e)} \textsc{ancestors of w}}; 
\end{tikzpicture}
%%%%%%%%%%%%%%%%%%%%%%
\hspace{0mm}
%%%%%%%%%%%%%%%%%%%%%%
\begin{tikzpicture}[every edge quotes/.style = {font=\footnotesize, fill=white,sloped}]
  \node[circle,black,thick,draw,scale=0.7] (u) at (0, 1) {$U$};
  \node[circle,WildStrawberry,fill=WildStrawberry!10,thick,draw,scale=0.7] (v) at (0, 0) {\color{black}$V$};
  \node[circle,WildStrawberry,fill=WildStrawberry!10,thick,draw,scale=0.7] (w) at (0, -1) {\color{black}$W$};
  \node[] (label_pos) at (0,-0.25) {}; 
  %%%
  \draw[thick,black,-{Stealth[width=5pt,length=5pt]}]  (u) edge[] (v);
  \draw[thick,black,-{Stealth[width=5pt,length=5pt]}]  (v) edge[] (w);
  %%%
  \node[below=of label_pos,align=flush center,text width=3.5cm] (label) {\footnotesize\textbf{(f)} \textsc{descendants of u}}; 
\end{tikzpicture} \\
%%%%%%%%%%%%%%%%%%%%%%
\vspace{5mm}
%%%%%%%%%%%%%%%%%%%%%%
\begin{tikzpicture}[every edge quotes/.style = {font=\footnotesize, fill=white,sloped}]
  \node[circle,WildStrawberry,fill=WildStrawberry!10,thick,draw,scale=0.7] (u) at (0, 1) {\color{black}$U$};
  \node[circle,black,thick,draw,scale=0.7] (v) at (-1, 0) {$V$};
  \node[circle,black,thick,draw,scale=0.7] (w) at (1, 0) {$W$};
  \node[] (label_pos) at (0,-0.25) {}; 
  \node[circle,black,thick,draw,scale=0.7] (x) at (1, -1) {$X$};
  \node[] (label_pos) at (0,-0.25) {}; 
  %%%
  \draw[thick,black,-{Stealth[width=5pt,length=5pt]}]  (u) edge[] (v);
  \draw[thick,black,-{Stealth[width=5pt,length=5pt]}]  (u) edge[] (w);
  \draw[thick,black,-{Stealth[width=5pt,length=5pt]}]  (w) edge[] (x);
  %%%
  \node[below=of label_pos,align=flush center,text width=3.5cm] (label) {\footnotesize\textbf{(g)} \textsc{root}}; 
\end{tikzpicture}
%%%%%%%%%%%%%%%%%%%%%%
\hspace{0mm}
%%%%%%%%%%%%%%%%%%%%%%
\begin{tikzpicture}[every edge quotes/.style = {font=\footnotesize, fill=white,sloped}]
  \node[circle,black,thick,draw,scale=0.7] (u) at (0, 1) {$U$};
  \node[circle,WildStrawberry,fill=WildStrawberry!10,thick,draw,scale=0.7] (v) at (-1, 0) {\color{black}$V$};
  \node[circle,black,thick,draw,scale=0.7] (w) at (1, 0) {$W$};
  \node[] (label_pos) at (0,-0.25) {}; 
  \node[circle,WildStrawberry,fill=WildStrawberry!10,thick,draw,scale=0.7] (x) at (1, -1) {\color{black}$X$};
  \node[] (label_pos) at (0,-0.25) {}; 
  %%%
  \draw[thick,black,-{Stealth[width=5pt,length=5pt]}]  (u) edge[] (v);
  \draw[thick,black,-{Stealth[width=5pt,length=5pt]}]  (u) edge[] (w);
  \draw[thick,black,-{Stealth[width=5pt,length=5pt]}]  (w) edge[] (x);
  %%%
  \node[below=of label_pos,align=flush center,text width=3.5cm] (label) {\footnotesize\textbf{(h)} \textsc{leaves}}; 
\end{tikzpicture}
    \caption{Roles that a given (highlighted) node can play with respect to others.}
    \label{fig:node_roles}
\end{figure}
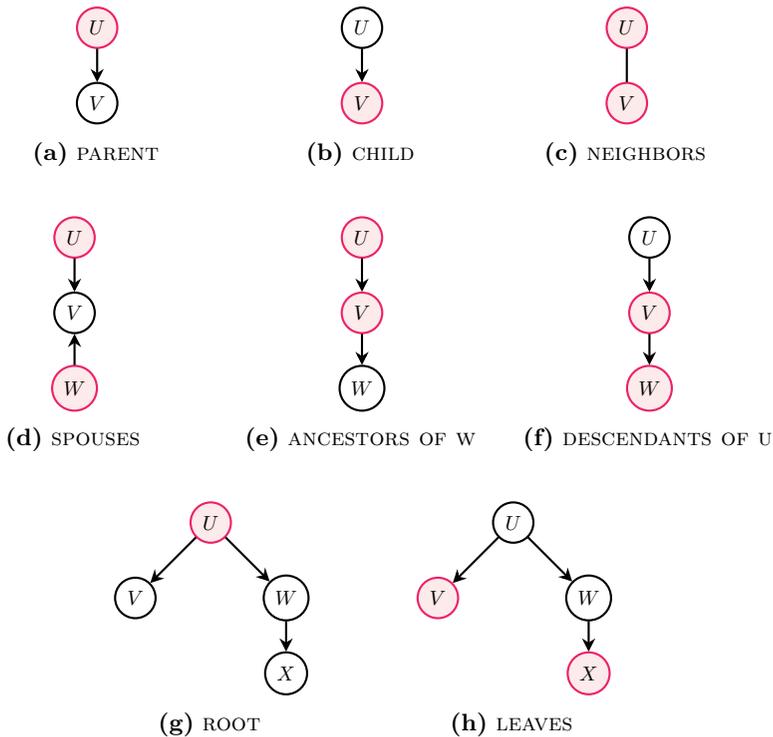

\paragraph{Properties and Substructures} An enormous range of graph, node, and edge properties can be enumerated, but here we will focus on basic vocabulary. Nodes that are connected by an edge are said to be \textit{adjacent} or \textit{neighbors}. The \textit{degree} of a node is the number of neighbors that it has. In a directed graph, the \textit{in-degree} of a node denotes the total number of neighbors with edges pointing into it (e.g., $\cdots \to Z$). Analogously, the \textit{out-degree} is equal to the number of edges pointing out of a node (e.g., $Z \to \cdots$). In Figure \ref{fig:directed_vs_undirected}b, the in-degree of $Z$ is 2 and its out-degree is 0, while the in-degree of $X$ is 0 and its out-degree is 2.

Sets of adjacent nodes can form various noteworthy substructures. A sequence of linked nodes can form a \textit{path}: $V_i - V_{i+1} - \cdots - V_{i+n}$ in the undirected case, or $V_i \to V_{i+1} \to \cdots \to V_{i+n}$ in the directed case. A path that starts and ends at the same node is called a \textit{cycle}.

\begin{definition}[Cycle]
    A cycle is a path of vertices such that only the first and last vertices are the same: $V_i - V_{i+1} - \cdots - V_{i + n} - V_i$ in the undirected case, or $V_i \to V_{i+1} \to \cdots \to V_{i + n} \to V_i$ in the directed case.
\end{definition}

Thus, when the cycle is traversed, it loops back to its beginning. The graph in Figure \ref{fig:directed_vs_undirected}a is an undirected cycle, while the graph in Figure \ref{fig:directed_vs_undirected}b is \textit{not} a directed cycle. Directed and undirected cycles (or, often, their \textit{absence}) play important roles throughout graph theory and probabilistic graphical modeling. We will see throughout this tutorial that \textit{acyclic} graphs are the focus of many inference and learning algorithms.

Many other relationships exist among groups of nodes (Figure \ref{fig:node_roles}). Let $U,V$ be two neighbors. When $U$ has a directed edge into $V$, we say that $U$ is the \textit{parent} and $V$ is the \textit{child}. If a third variable $W$ is also a parent of $V$, we say that $U$ and $W$ are \textit{spouses}. If $W$ is instead a child of $V$, we say that $U$ is the \textit{ancestor} of \textit{descendant} $W$. Ancestors and descendants generalize the parent-child relationship to paths of length greater than 1. When a node has no \textit{parents}, we say that it is a \textit{root}. When a node has no children, we say that it is a \textit{leaf}. In an undirected graph, a leaf is a node with degree 1. In general, we will use the notation $\mathbf{pa}(\cdot)$, $\mathbf{ch}(\cdot)$, $\mathbf{an}(\cdot)$, $\mathbf{de}(\cdot)$, $\mathbf{sp}(\cdot)$, $\mathbf{ne}(\cdot)$ to refer to the parent, child, ancestor, descendant, spouse, and neighbor sets of a given node, respectively.

\tikzset{
  prefix after node/.style={prefix after command=\pgfextra{#1}},
  /semifill/ang/.initial=45,
  /semifill/upper/.initial=none,
  /semifill/lower/.initial=none,
  semifill/.style={
    circle, draw,
    prefix after node={
      \pgfqkeys{/semifill}{#1}
      \path let \p1 = (\tikzlastnode.north), \p2 = (\tikzlastnode.center),
                \n1 = {\y1-\y2} in [radius=\n1]
            (\tikzlastnode.\pgfkeysvalueof{/semifill/ang}) 
            edge[
              draw=none,
              fill=\pgfkeysvalueof{/semifill/upper},
              to path={
                arc[start angle=\pgfkeysvalueof{/semifill/ang}, delta angle=180]
                -- cycle}] ()
            (\tikzlastnode.\pgfkeysvalueof{/semifill/ang}) 
            edge[
              draw=none,
              fill=\pgfkeysvalueof{/semifill/lower},
              to path={
                arc[start angle=\pgfkeysvalueof{/semifill/ang}, delta angle=-180,fill opacity=0.1]
                -- cycle}] ();}}}

\tikzset{
    side by side/.style 2 args={
        line width=2pt,
        #1,
        postaction={
            clip,postaction={draw,#2}
        }
    },
    circle node/.style={
        circle,
        draw,
        fill=white,
        minimum size=1.3cm
    }
}

\begin{figure}
    \centering

\begin{tikzpicture}[every edge quotes/.style = {font=\footnotesize, fill=white,sloped}]
  \node[circle,black,thick,draw,scale=0.7] (A) at (-2,0) {$A$};
  \node[circle,black,thick,draw,scale=0.7] (B) at (-1,1.5) {$B$};
  \node[circle,black,thick,draw,scale=0.7] (C) at (0,0) {$C$};
  \node[circle,black,thick,draw,scale=0.7] (D) at (1,1.5) {$D$};
  \node[circle,black,thick,draw,scale=0.7] (E) at (2,0) {$E$};
  \node[circle,black,thick,draw,scale=0.7] (F) at (0,-1.5) {$F$};
  %%%
  %\draw[-,thick,Maroon]  (X) edge[] node[yshift=5.5] {?} (Y);
  \draw[-,thick,black]  (A) edge[] (B);
  \draw[-,thick,black]  (A) edge[] (C);
  \draw[-,thick,black]  (A) edge[] (F);
  \draw[-,thick,black]  (B) edge[] (C);
  \draw[-,thick,black]  (C) edge[] (D);
  \draw[-,thick,black]  (D) edge[] (E);
  \draw[-,thick,black]  (C) edge[] (E);
  \draw[-,thick,black]  (E) edge[] (F);
  %%%
  \begin{pgfonlayer}{background}
    \fill[OrangeRed,opacity=0.3] \convexpath{A,B,C}{8pt};
    %\fill[OrangeRed,opacity=0.3] \convexpath{A,C}{8pt};
    %\fill[OrangeRed,opacity=0.3] \convexpath{B,C}{8pt};
    %%
    \fill[ProcessBlue,opacity=0.3] \convexpath{C,D,E}{8pt};
    %\fill[ProcessBlue,opacity=0.3] \convexpath{C,E}{8pt};
    %\fill[ProcessBlue,opacity=0.3] \convexpath{D,E}{8pt};
    %%
    \fill[Green,opacity=0.35] \convexpath{A,F}{8pt};
    \fill[Dandelion,opacity=0.3] \convexpath{E,F}{8pt};
    \end{pgfonlayer}
  %%%
  \node[] (label) at (0,-2.5) {\footnotesize\textsc{undirected graph} $\mathcal{G}$};
\end{tikzpicture}
\caption{Maximal cliques in $\mathcal{G}$ are $\{A,C,B\}$, $\{C,D,E\}$, $\{A,F\}$, and $\{E,F\}$.}
\label{fig:cliques}
\end{figure}

In undirected graphs, it is often useful to perform operations over \textit{fully connected subgraphs} called \textit{cliques}.

\begin{definition}[Clique] \label{def:clique}
    Let $\mathcal{G} = (\mathbf{V}, \mathbf{E})$ be an undirected graph. A \textit{clique} is a subset $\mathbf{C} \subseteq \mathbf{V}$ where every pair of nodes $V_i,V_j \in \mathbf{C}$ are neighbors.
\end{definition}

\begin{definition}[Maximal clique]
\label{def:maximal_clique}
    A \textit{maximal clique} is a clique to which no additional nodes in  $\mathbf{V}$ can be added without violating the definition of a clique.
\end{definition}

In Figure \ref{fig:cliques}, $\{A,B\}$, $\{A,C\}$, and $\{B,C\}$ form pairwise cliques, while $\{A,B,C\}$ forms a maximal clique. We can also define \textit{unary} cliques, consisting of a single node each. Later, cliques will be integral to our discussion of undirected graphs (Section \ref{sec:mrf}) and inference (Chapter \ref{sec:inference}).

We will introduce a final property of directed graphs: the \textit{topological ordering} or \textit{topological sort}.

\begin{definition}[Topological ordering] \label{def:topological_sort}
    Let $\mathcal{G} = (\mathbf{V}, \mathbf{E})$ be a directed graph. A topological ordering of the nodes in $\mathcal{G}$ is a linear ordering such that for every edge $V_i \to V_j \in \mathbf{E}$, $V_i$ precedes $V_j$ in the ordering.
\end{definition}

Topological orderings can convey significant structural information. In general, however, this information cannot uniquely specify the graph. It is possible for multiple directed graphs to satisfy the same topological ordering, and multiple orderings can describe the same graph. For example, $(U,V,W,X)$ and $(U,W,X,V)$ are both valid topological orderings for the graph in Figure \ref{fig:node_roles}g. We will return to this ambiguity in Section \ref{sec:structure_learning}. Nevertheless, we will see that topological orderings can be useful in both inference and learning.

\paragraph{Converting Directed Graphs to Undirected Graphs} In this tutorial, we will reference several methods and concepts that require the transformation of a directed graph to an undirected graph. These primarily rely on two different approaches, which serve different purposes. Most simply, we can remove directionality by deleting the arrowheads from each edge. This yields the \textit{undirected skeleton} of the directed graph. For example, Figure \ref{fig:directed_vs_undirected}a is the skeleton of Figure \ref{fig:directed_vs_undirected}b. However, some use cases require an additional manipulation: \textit{moralization} (Figure \ref{fig:moralization}).

When a node has two or more parents that are not adjacent to each other, we historically refer to this as an \textit{immorality} (an anachronistic and value-laden term). This brings us to our notion of a \textit{moral graph}.

\tikzset{
  prefix after node/.style={prefix after command=\pgfextra{#1}},
  /semifill/ang/.initial=45,
  /semifill/upper/.initial=none,
  /semifill/lower/.initial=none,
  semifill/.style={
    circle, draw,
    prefix after node={
      \pgfqkeys{/semifill}{#1}
      \path let \p1 = (\tikzlastnode.north), \p2 = (\tikzlastnode.center),
                \n1 = {\y1-\y2} in [radius=\n1]
            (\tikzlastnode.\pgfkeysvalueof{/semifill/ang}) 
            edge[
              draw=none,
              fill=\pgfkeysvalueof{/semifill/upper},
              to path={
                arc[start angle=\pgfkeysvalueof{/semifill/ang}, delta angle=180]
                -- cycle}] ()
            (\tikzlastnode.\pgfkeysvalueof{/semifill/ang}) 
            edge[
              draw=none,
              fill=\pgfkeysvalueof{/semifill/lower},
              to path={
                arc[start angle=\pgfkeysvalueof{/semifill/ang}, delta angle=-180,fill opacity=0.1]
                -- cycle}] ();}}}

\tikzset{
    side by side/.style 2 args={
        line width=2pt,
        #1,
        postaction={
            clip,postaction={draw,#2}
        }
    },
    circle node/.style={
        circle,
        draw,
        fill=white,
        minimum size=1.3cm
    }
}

\begin{figure}
    \centering

\begin{tikzpicture}
\node[circle,black,thick,draw,scale=0.7] (a) at (0,3) {$A$};
\node[circle,black,thick,draw,scale=0.7] (b) at (1,2) {$B$};
\node[circle,black,thick,draw,scale=0.7] (c) at (2,3) {$C$};
\node[circle,black,thick,draw,scale=0.7] (d) at  (1,1) {$D$};
\node[circle,black,thick,draw,scale=0.7] (g) at (0,0) {$G$};
\node[circle,black,thick,draw,scale=0.7] (e) at  (-1,1) {$E$};
\node[circle,black,thick,draw,scale=0.7] (f) at (-2,0) {$F$};
\draw[-{Stealth[width=5pt,length=5pt]},thick,black]  (a) edge[bend right=10] (d);
\draw[-{Stealth[width=5pt,length=5pt]},thick,black]  (b) edge[] (d);
\draw[-{Stealth[width=5pt,length=5pt]},thick,black]  (c) edge[bend left=10] (d);
\draw[-{Stealth[width=5pt,length=5pt]},thick,black]  (d) edge[] (g);
\draw[-{Stealth[width=5pt,length=5pt]},thick,black]  (e) edge[] (g);
\draw[-{Stealth[width=5pt,length=5pt]},thick,black]  (e) edge[] (f);
\node[] (label) at (0,-1) {\footnotesize \textbf{(a)} \textsc{digraph}};
\end{tikzpicture}
%%%%%%%%%%%%%%%%%%%%%%%%%%%%
\hspace{5mm}
%%%%%%%%%%%%%%%%%%%%%%%%%%%%
\begin{tikzpicture}
\node[circle,black,thick,draw,scale=0.7] (a) at (0,3) {$A$};
\node[circle,black,thick,draw,scale=0.7] (b) at (1,2) {$B$};
\node[circle,black,thick,draw,scale=0.7] (c) at (2,3) {$C$};
\node[circle,black,thick,draw,scale=0.7] (d) at  (1,1) {$D$};
\node[circle,black,thick,draw,scale=0.7] (g) at (0,0) {$G$};
\node[circle,black,thick,draw,scale=0.7] (e) at  (-1,1) {$E$};
\node[circle,black,thick,draw,scale=0.7] (f) at (-2,0) {$F$};
\draw[-,thick,black]  (a) edge[] (b);
\draw[-,thick,black]  (a) edge[] (c);
\draw[-,thick,black]  (a) edge[bend right=10] (d);
\draw[-,thick,black]  (b) edge[] (c);
\draw[-,thick,black]  (b) edge[] (d);
\draw[-,thick,black]  (c) edge[bend left=10] (d);
\draw[-,thick,black]  (d) edge[] (g);
\draw[-,thick,black]  (e) edge[] (g);
\draw[-,thick,black]  (e) edge[] (d);
\draw[-,thick,black]  (e) edge[] (f);

%\draw [black,thick] (28.04,-20.9) -- (35.36,-28.1);
%\draw [black,thick] (37.5,-21.8) -- (37.5,-27.2);
%\draw [black,thick] (47.6,-21.4) -- (39.76,-28.23);
%\draw [black,thick] (35.28,-32.21) -- (29.22,-37.69);
%\draw [black,thick] (18.63,-32.2) -- (24.77,-37.7);
%\draw [black,thick] (14.17,-32.2) -- (8.03,-37.7);
%\draw [black,thick] (27.049,-16.037) arc (150.27642:29.72358:12.034);
%\draw [black,thick] (28.9,-18.8) -- (34.5,-18.8);
%\draw [black,thick] (40.5,-18.8) -- (46.1,-18.8);
%\draw [black,thick] (19.4,-30.2) -- (34.5,-30.2);
%%%
  \begin{pgfonlayer}{background}
    %\fill[Salmon,opacity=0.3] \convexpath{a,b}{8pt};
    %\fill[Salmon,opacity=0.3] \convexpath{a,c}{8pt};
    %\fill[Salmon,opacity=0.3] \convexpath{a,d}{8pt};
    %\fill[Salmon,opacity=0.3] \convexpath{b,c}{8pt};
    %\fill[Salmon,opacity=0.3] \convexpath{b,d}{4pt};
    %\fill[Salmon,opacity=0.3] \convexpath{c,d}{8pt};
    \fill[Salmon,opacity=0.2] \convexpath{a,c,d}{8pt};
    %%
    %\fill[RoyalBlue,opacity=0.3] \convexpath{g,d}{8pt};
    %\fill[RoyalBlue,opacity=0.3] \convexpath{g,e}{8pt};
    %\fill[RoyalBlue,opacity=0.3] \convexpath{d,e}{8pt};
    \fill[RoyalBlue,opacity=0.35] \convexpath{g,e,d}{8pt};
    \fill[SpringGreen,opacity=0.3] \convexpath{e,f}{8pt};
    \end{pgfonlayer}
  %%%
  \node[] (label) at (0,-1) {\footnotesize \textbf{(b)} \textsc{moral graph}};
\end{tikzpicture}

    \caption{Directed graph \textbf{(a)} and its corresponding moral graph \textbf{(b)}, with shading denoting maximal cliques.}
    \label{fig:moralization}
\end{figure}
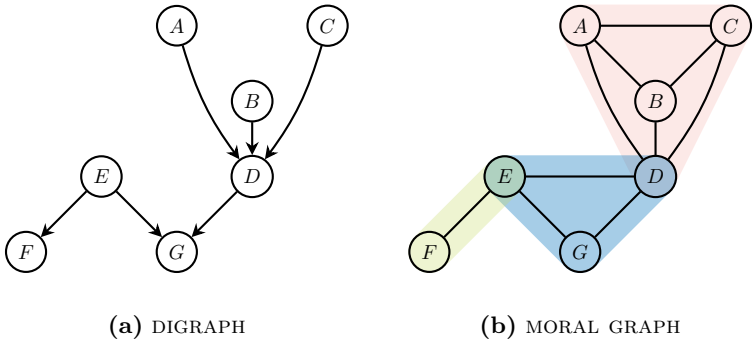

\begin{definition}[Moral graph]
\label{def:moral_graph}
    The moral graph corresponding to a directed acyclic graph $\mathcal{G}$ is an undirected graph with an edge for every adjacency in $\mathcal{G}$ and for every pair of non-adjacent spouses (i.e., nodes in $\mathcal{G}$ that share a child).
\end{definition}
 
Moralization ensures that each child node and all of its parents form a single clique in the resulting undirected graph \citep{bishop2006pattern}. To \textit{moralize} a directed graph $\mathcal{G}$, we proceed in two steps:
\begin{enumerate}
    \item For every pair of spouses in  $\mathcal{G}$ that are not already adjacent, add an undirected edge. This removes all immoralities.
    \item Remove the directionality for all edges in $\mathcal{G}$.
\end{enumerate}

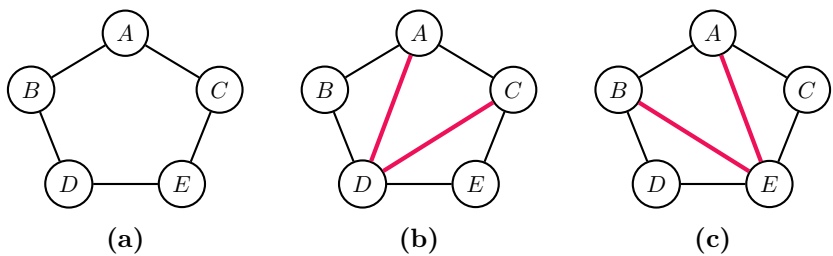
\begin{figure}[!t]
    \centering

\begin{tikzpicture}[scale=0.5]

    % Nodes.
    \node[draw,thick,circle,scale=0.8] (A) at (0,0) {$A$};
    \node[draw,thick,circle,scale=0.8] (B) at (-2.5,-1.5) {$B$};
    \node[draw,thick,circle,scale=0.8] (C) at (2.5,-1.5) {$C$};
    \node[draw,thick,circle,scale=0.8] (D) at (-1.5,-4) {$D$};
    \node[draw,thick,circle,scale=0.8] (E) at (1.5,-4) {$E$};

    % Edges.
    \draw[thick] (A) -- (B);
    \draw[thick] (A) -- (C);
    \draw[thick] (B) -- (D);
    \draw[thick] (C) -- (E);
    \draw[thick] (D) -- (E);

    % Labels.
    \node[] (label) at (0,-5.5) {\small\textbf{(a)}};
    
\end{tikzpicture}
%%%%%%%%%%%%%%%%%%%%%%%%%%%%%%
\hspace{5mm}
%%%%%%%%%%%%%%%%%%%%%%%%%%%%%%
\begin{tikzpicture}[scale=0.5]

    % Nodes.
    \node[draw,thick,circle,scale=0.8] (A) at (0,0) {$A$};
    \node[draw,thick,circle,scale=0.8] (B) at (-2.5,-1.5) {$B$};
    \node[draw,thick,circle,scale=0.8] (C) at (2.5,-1.5) {$C$};
    \node[draw,thick,circle,scale=0.8] (D) at (-1.5,-4) {$D$};
    \node[draw,thick,circle,scale=0.8] (E) at (1.5,-4) {$E$};

    % Edges.
    \draw[thick] (A) -- (B);
    \draw[thick] (A) -- (C);
    \draw[thick] (B) -- (D);
    \draw[thick] (C) -- (E);
    \draw[thick] (D) -- (E);
    \draw[ultra thick, OrangeRed] (D) -- (A);
    \draw[ultra thick, OrangeRed] (D) -- (C);

    % Labels.
    \node[] (label) at (0,-5.5) {\small\textbf{(b)}};
    
\end{tikzpicture}
%%%%%%%%%%%%%%%%%%%%%%%%%%%%%%
\hspace{5mm}
%%%%%%%%%%%%%%%%%%%%%%%%%%%%%%
\begin{tikzpicture}[scale=0.5]

    % Nodes.
    \node[draw,thick,circle,scale=0.8] (A) at (0,0) {$A$};
    \node[draw,thick,circle,scale=0.8] (B) at (-2.5,-1.5) {$B$};
    \node[draw,thick,circle,scale=0.8] (C) at (2.5,-1.5) {$C$};
    \node[draw,thick,circle,scale=0.8] (D) at (-1.5,-4) {$D$};
    \node[draw,thick,circle,scale=0.8] (E) at (1.5,-4) {$E$};

    % Edges.
    \draw[thick] (A) -- (B);
    \draw[thick] (A) -- (C);
    \draw[thick] (B) -- (D);
    \draw[thick] (C) -- (E);
    \draw[thick] (D) -- (E);
    \draw[ultra thick, OrangeRed] (E) -- (B);
    \draw[ultra thick, OrangeRed] (E) -- (A);

    % Labels.
    \node[] (label) at (0,-5.5) {\small\textbf{(c)}};
    
\end{tikzpicture}

    \caption{An undirected graph (\textbf{a}) and two possible chordalized graphs corresponding to it (\textbf{b}, \textbf{c}; chords in red).}
    \label{fig:chordal}
\end{figure}

In addition to moralization, we may wish to \textit{chordalize} the undirected skeleton of our directed graph $\mathcal{G}$ (Figure \ref{fig:chordal}). We can transform the undirected skeleton of $\mathcal{G}$ to a \textit{chordal graph} (or \textit{triangulated graph}) by adding edges such that the following definition is met.

\begin{definition}[Chordal graph, \citealt{vandenberghe2015chordal}]
\label{def:chordal_graph}
    An undirected graph is chordal if any cycle of length greater than three has a chord (i.e., an edge connecting any two nonconsecutive nodes). Thus, the longest minimal loop is a triangle. 
\end{definition}

Both moralization and chordalization are important steps in the junction tree algorithm, which we will explore in Chapter \ref{sec:inference}. Chordal graphs have played pivotal roles in combinatorial optimization, semidefinite optimization, nonlinear optimization, linear algebra, statistics, and signal processing. For further discussion of the theory and applications of chordal graphs, see \citet{vandenberghe2015chordal}. 

%Edge sequences can form special substructures as well. A \textit{walk} is... a \textit{path} is a sequence of edges between two nodes...

\subsection{Special Categories of Graphs}

There are certain categories of graphs that we will encounter frequently in graphical modeling. One such kind of graph is the \textit{directed acyclic graph} (DAG). This is simply a digraph with no cycles (e.g., Figure \ref{fig:directed_vs_undirected}b). DAGs can be used for both probabilistic and causal modeling, and have played a central role in modeling in the fundamental sciences. Another important kind of graph is the \textit{tree} (Figure \ref{fig:trees}). Many of the probabilistic inference algorithms that we will discuss apply to tree-structured graphs.

\begin{definition}[Undirected tree] \label{def:undirected_tree}
    An \textit{undirected tree} is a connected, undirected graph with no cycles. %A \textit{directed tree} or \textit{polytree} is a directed graph with no cycles whose undirected skeleton is a tree.
\end{definition}

The following properties apply to undirected trees. See Chapter 3 of \citet{deo1974graph} for proof of these theorems.

\begin{theorem}
    Any two nodes in a tree are connected by exactly one path. If any two nodes in a graph are connected by exactly one path, then that graph is a tree.
\end{theorem}

\begin{theorem}
    A tree with $n$ nodes has $n-1$ edges. If a connected graph with $n$ nodes has $n-1$ edges, then it is a tree.
\end{theorem}

\begin{theorem}
    A graph is a tree if and only if it is minimally connected (i.e., it is connected, and the removal of any edge disconnects the graph).
\end{theorem}

\definecolor{celadon}{rgb}{0.67, 0.88, 0.69}
\definecolor{asparagus}{rgb}{0.53, 0.66, 0.42}
\definecolor{darkseagreen}{rgb}{0.56, 0.74, 0.56}
\definecolor{mossgreen}{rgb}{0.68, 0.87, 0.68}
\definecolor{magicmint}{rgb}{0.67, 0.94, 0.82}

\begin{figure}
    \centering

\begin{tikzpicture}[every edge quotes/.style = {font=\footnotesize, fill=white,sloped}]
  \node[circle,LimeGreen,fill=LimeGreen!20,thick,draw,scale=0.7] (a) at (0,2) {\color{black}$A$};
  \node[circle,LimeGreen,fill=LimeGreen!20,thick,draw,scale=0.7] (b) at (-1,1) {\color{black}$B$};
  \node[circle,LimeGreen,fill=LimeGreen!20,thick,draw,scale=0.7] (c) at (1,1) {\color{black}$C$};
  \node[circle,LimeGreen,fill=LimeGreen!20,thick,draw,scale=0.7] (d) at (-1.5,0) {\color{black}$D$};
  \node[circle,LimeGreen,fill=LimeGreen!20,thick,draw,scale=0.7] (e) at (-0.5,0) {\color{black}$E$};
  \node[circle,LimeGreen,fill=LimeGreen!20,thick,draw,scale=0.7] (f) at (0.5,0) {\color{black}$F$};
  \node[circle,LimeGreen,fill=LimeGreen!20,thick,draw,scale=0.7] (g) at (1.5,0) {\color{black}$G$};
  %%%
  \draw[thick,black,-]  (a) edge[] (b);
  \draw[thick,black,-]  (a) edge[] (c);
  \draw[thick,black,-]  (b) edge[] (d);
  \draw[thick,black,-]  (b) edge[] (e);
  \draw[thick,black,-]  (c) edge[] (f);
  \draw[thick,black,-]  (c) edge[] (g);
  %%%
  \node[align=flush center,text width=3cm] (label) at (0,-1) {\footnotesize\textbf{(a)}}; 
\end{tikzpicture}
%%%%%%%%%%%%%%%%
\hspace{2mm}
%%%%%%%%%%%%%%%%
\begin{tikzpicture}[every edge quotes/.style = {font=\footnotesize, fill=white,sloped}]
  \node[circle,PineGreen,fill=PineGreen!20,thick,draw,scale=0.7] (a) at (0,2) {\color{black}$A$};
  \node[circle,PineGreen,fill=PineGreen!20,thick,draw,scale=0.7] (b) at (0,1) {\color{black}$B$};
  \node[circle,PineGreen,fill=PineGreen!20,thick,draw,scale=0.7] (c) at (-1,0) {\color{black}$C$};
  \node[circle,PineGreen,fill=PineGreen!20,thick,draw,scale=0.7] (d) at (0,0) {\color{black}$D$};
  \node[circle,PineGreen,fill=PineGreen!20,thick,draw,scale=0.7] (e) at (1,0) {\color{black}$E$};
  %%%
  \draw[thick,black,-{Stealth[width=5pt,length=5pt]}]  (a) edge[] (b);
  \draw[thick,black,-{Stealth[width=5pt,length=5pt]}]  (b) edge[] (c);
  \draw[thick,black,-{Stealth[width=5pt,length=5pt]}]  (b) edge[] (d);
  \draw[thick,black,-{Stealth[width=5pt,length=5pt]}]  (b) edge[] (e);
  %%%
  \node[align=flush center,text width=3cm] (label) at (0,-1) {\footnotesize\textbf{(b)}}; 
\end{tikzpicture}
%%%%%%%%%%%%%%%%
\hspace{2mm}
%%%%%%%%%%%%%%%%
\begin{tikzpicture}[every edge quotes/.style = {font=\footnotesize, fill=white,sloped}]
  \node[circle,celadon,fill=celadon!20,thick,draw,scale=0.7] (a) at (-1.5,2) {\color{black}$A$};
  \node[circle,celadon,fill=celadon!20,thick,draw,scale=0.7] (b) at (0,2) {\color{black}$B$};
  \node[circle,celadon,fill=celadon!20,thick,draw,scale=0.7] (c) at (1.5,2) {\color{black}$C$};
  \node[circle,celadon,fill=celadon!20,thick,draw,scale=0.7] (d) at (-0.5,1) {\color{black}$D$};
  \node[circle,celadon,fill=celadon!20,thick,draw,scale=0.7] (e) at (0.5,1) {\color{black}$E$};
  \node[circle,celadon,fill=celadon!20,thick,draw,scale=0.7] (f) at (-0.5,0) {\color{black}$F$};
  \node[circle,celadon,fill=celadon!20,thick,draw,scale=0.7] (g) at (0.5,0) {\color{black}$G$};
  %%%
  \draw[thick,black,-{Stealth[width=5pt,length=5pt]}]  (a) edge[] (d);
  \draw[thick,black,-{Stealth[width=5pt,length=5pt]}]  (b) edge[] (d);
  \draw[thick,black,-{Stealth[width=5pt,length=5pt]}]  (b) edge[] (e);
  \draw[thick,black,-{Stealth[width=5pt,length=5pt]}]  (c) edge[] (e);
  \draw[thick,black,-{Stealth[width=5pt,length=5pt]}]  (d) edge[] (f);
  \draw[thick,black,-{Stealth[width=5pt,length=5pt]}]  (e) edge[] (g);
  %%%
  \node[align=flush center,text width=3cm] (label) at (0,-1) {\footnotesize\textbf{(c)}}; 
\end{tikzpicture}

    \caption{\textbf{(a)} An undirected tree, \textbf{(b)} a directed tree, and \textbf{(c)} a polytree.}
    \label{fig:trees}
\end{figure}
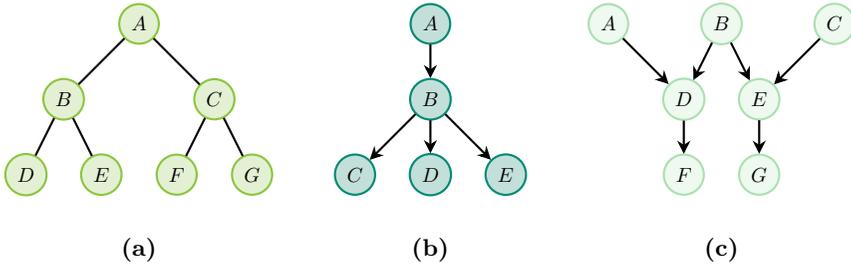

We can also define various tree-structured graphs with directed edges: \textit{directed trees} and \textit{polytrees}.

\begin{definition}[Directed tree] \label{def:directed_tree}
    A directed graph is a \textit{directed tree} if it is connected and if each node has at most one parent.
\end{definition}
%\citealt{koller2009probabilistic}

\begin{definition}[Polytree] \label{def:polytree}
    A \textit{polytree} is a DAG with no cycles in its undirected skeleton. A polytree is distinct from a tree, as nodes can have more than one parent.
\end{definition}

We will also make use of the concept of \textit{treewidth}, which is a number that characterizes how far a graph is from being a tree. The smaller the treewidth, the more ``tree-like'' the graph is, and true trees have a treewidth of 1.

\begin{definition}[Treewidth] \label{def:treewidth}
    A graph has \textit{treewidth} $k$ if it can be broken into overlapping groups of at most $k + 1$ vertices (called bags), arranged in a tree-like structure, such that: (a) every graph vertex is in at least one bag; (b) for every edge, both endpoints appear together in some bag; and (c) for each vertex, the bags containing it form a connected subtree of the tree-like structure over bags.
\end{definition}

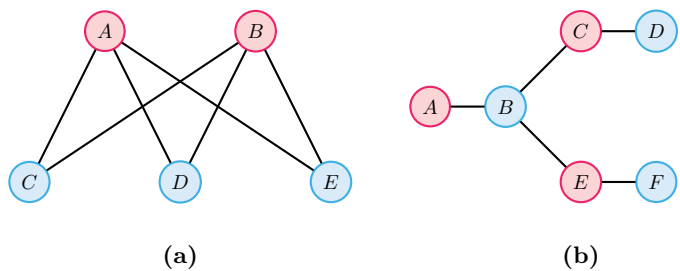
\begin{figure}
    \centering

\begin{tikzpicture}[every edge quotes/.style = {font=\footnotesize, fill=white,sloped}]
  \node[circle,WildStrawberry,fill=WildStrawberry!20,thick,draw,scale=0.7] (a) at (-1,2) {\color{black}$A$};
  \node[circle,WildStrawberry,fill=WildStrawberry!20,thick,draw,scale=0.7] (b) at (1,2) {\color{black}$B$};
  \node[circle,CornflowerBlue,fill=CornflowerBlue!20,thick,draw,scale=0.7] (c) at (-2,0) {\color{black}$C$};
  \node[circle,CornflowerBlue,fill=CornflowerBlue!20,thick,draw,scale=0.7] (d) at (0,0) {\color{black}$D$};
  \node[circle,CornflowerBlue,fill=CornflowerBlue!20,thick,draw,scale=0.7] (e) at (2,0) {\color{black}$E$};
  %%%
  \draw[thick,black,-]  (a) edge[] (c);
  \draw[thick,black,-]  (a) edge[] (d);
  \draw[thick,black,-]  (a) edge[] (e);
  \draw[thick,black,-]  (b) edge[] (c);
  \draw[thick,black,-]  (b) edge[] (d);
  \draw[thick,black,-]  (b) edge[] (e);
  %%%
  \node[align=flush center,text width=3cm] (label) at (0,-1) {\footnotesize\textbf{(a)}}; 
\end{tikzpicture}
%%%%%%%%%%%%%%%%
\hspace{5mm}
%%%%%%%%%%%%%%%%
\begin{tikzpicture}[every edge quotes/.style = {font=\footnotesize, fill=white,sloped}]
  \node[circle,WildStrawberry,fill=WildStrawberry!20,thick,draw,scale=0.7] (a) at (-2,1) {\color{black}$A$};
  \node[circle,CornflowerBlue,fill=CornflowerBlue!20,thick,draw,scale=0.7] (b) at (-1,1) {\color{black}$B$};
  \node[circle,WildStrawberry,fill=WildStrawberry!20,thick,draw,scale=0.7] (c) at (0,2) {\color{black}$C$};
  \node[circle,CornflowerBlue,fill=CornflowerBlue!20,thick,draw,scale=0.7] (d) at (1,2) {\color{black}$D$};
  \node[circle,WildStrawberry,fill=WildStrawberry!20,thick,draw,scale=0.7] (e) at (0,0) {\color{black}$E$};
  \node[circle,CornflowerBlue,fill=CornflowerBlue!20,thick,draw,scale=0.7] (f) at (1,0) {\color{black}$F$};
  %%%
  \draw[thick,black,-]  (a) edge[] (b);
  \draw[thick,black,-]  (b) edge[] (c);
  \draw[thick,black,-]  (b) edge[] (e);
  \draw[thick,black,-]  (c) edge[] (d);
  \draw[thick,black,-]  (e) edge[] (f);
  %%%
  \node[align=flush center,text width=3cm] (label) at (0,-1) {\footnotesize\textbf{(b)}}; 
\end{tikzpicture}

    \caption{\textbf{(a)} A complete bipartite graph and \textbf{(b)} an acyclic bipartite graph with partite sets in red and blue.}
    \label{fig:bipartite}
\end{figure}

Finally, we introduce the \textit{bipartite graph} (Figure \ref{fig:bipartite}). These will make an appearance in Chapter \ref{sec:representation} when we discuss factor graphs.

\begin{definition}[Bipartite graph] \label{def:bipartite_graph}
    A graph $\mathcal{G} = (\mathbf{V}, \mathbf{E})$ is bipartite if $\mathbf{V}$ can be partitioned into two disjoint sets $\mathbf{V}_1$, $\mathbf{V}_2$ (i.e., \textit{partite sets}), such that every edge in $\mathbf{E}$ has one endpoint in $\mathbf{V}_1$ and one endpoint in $\mathbf{V}_2$.
\end{definition}

Additionally, bipartite graphs have the following property. See \citet{harris2008combinatorics} for full proof of this theorem.

\begin{theorem} \label{prop:bipartite_no_odd_cycle}
    A graph containing at least two nodes is bipartite if and only if it contains no odd cycles.
\end{theorem}

Following from Theorem \ref{prop:bipartite_no_odd_cycle}, 
we can see that all trees are bipartite (as they contain no cycles).

\vspace{5mm}

\begin{reading}
%\textit{Texts on general graph theory}
    \begin{itemize}[leftmargin=*]
        \item \fullcite{deo1974graph}.
        \item \fullcite{harris2008combinatorics}.
    \end{itemize}
\end{reading}

%%%%%%%%%%%%%%%%%%%%%%%%
%% Representation
%%%%%%%%%%%%%%%%%%%%%%%%

\chapter{Representation}
\label{sec:representation}

Our journey in probabilistic graphical modeling begins with the topic of \textit{representation}. How do we choose a probability distribution to model some interesting aspect of the world? Coming up with a good model is not always easy: we have seen in the introduction that a naive model for spam classification would require us to specify a number of parameters that is exponential in the number of words in the English language. In this chapter, we will show that we can instead leverage the \textit{conditional independencies} present in our distribution to represent the joint distribution with fewer parameters. Thus, we can obtain compact, tractable models for many complex systems. Further, we can express the structure of our joint as a \textit{graphical model}: a mathematical object that visually encodes our conditional independence assumptions and enables the use of efficient graph-based algorithms for statistical inference and learning. Graphical models provide an intuitive visual language for reasoning over our system of interest, where complex computations used in learning and inference can be expressed as operations on the graph. %Graphical models can also enable reasoning in challenging statistical settings, such as missing data \citep{mohan2013graphical,mohan2021graphical}.

In this chapter, we will focus on effective and general techniques for parameterizing probability distributions with relatively few parameters. We will explore how the resulting models can be elegantly described as \textit{directed} and \textit{undirected} graphs (Figure \ref{fig:directed_vs_undirected}). We will explore connections between graphical structures and the assumptions made by the distributions that they describe. This will both make our modeling assumptions more explicit and help us design more efficient inference algorithms. We will discuss how the properties of different graphical representations can present advantages and shortcomings in particular settings, with a focus on Bayesian networks (Section \ref{sec:rep_bayesian_networks}), Markov random fields (Section \ref{sec:mrf}), factor graphs (Section \ref{sec:factor_graph}), and conditional random fields (Section \ref{sec:crf}). As a general accompaniment to this chapter, we recommend reading Chapters 3 and 4 in \citet{koller2009probabilistic} and Chapter 8 in \citet{bishop2006pattern}.

%%%%%%
%% Bayesian networks
%%%%%%

\section{Representing Distributions with Directed Graphs}

%\subsection{Probabilistic modeling with Bayesian networks}
\subsection{Bayesian Networks}
\label{sec:rep_bayesian_networks}

We begin with directed graphical models: a powerful class of graphs defined foremost by the fact that their edges have directionality. While directed graphs can contain cycles, we will restrict our attention to those that do not.
 %When discussing Bayesian networks, we generally care about forbidding directed cycles, but cycles in the corresponding undirected skeleton will sometimes make an appearance (e.g., in message-passing algorithms, Chapter \ref{sec:inference}).
Directed graphs that do not contain cycles are known as \textit{directed acyclic graphs} (DAGs; e.g., Figure \ref{fig:directed_vs_undirected}b,c). In probabilistic graphical modeling, these are used in models known as \textit{Bayesian networks}, \textit{Bayes nets}, or \textit{belief networks}.  In causal modeling, these are used in models often referred to as \textit{causal DAGs}. %Unlike undirected graphs, DAGs can also be used to model \textit{causal} relationships. When a causal interpretation is required, we often use the terms \textit{causal graphical models} or \textit{causal graphs}. 
%However, imposing a causal interpretation on DAGs generally requires special assumptions, which we explore in Section \ref{sec:structure_learning}. 

For now, we will primarily focus on the probabilistic interpretation of DAGs using the term \textit{Bayesian networks}. The power of Bayesian networks lies in their ability to enable reasoning in the presence of uncertainty. The following is just a short list of their useful properties.

\begin{enumerate}
    \item Bayesian networks are \textit{economical}: they allow us to represent probability distributions using compact parameterizations. 
    \item Unlike undirected graphs, Bayesian networks can admit both probabilistic and causal interpretations under sufficient conditions. Thus, Bayesian networks can potentially enable associational, interventional, and counterfactual insights \citep{pearl1995bayesian}.
    \item Though Bayesian networks are not \textit{innately} Bayesian, they can be combined with Bayesian statistical methods to elegantly incorporate prior knowledge and techniques for preventing overfitting \citep{heckerman1998tutorial}.
    %\item Bayesian networks readily enable reasoning in the presence of missing data \citep{mohan2013graphical,mohan2021graphical}.
    % VK: true for all graphical models in some sense? I would waybe skip this one
\end{enumerate}

%Bayesian networks are a family of probability distributions that admit a compact parameterization that can be naturally described by DAGs. 

\subsubsection{Bayesian Networks as Economical Models: Intuition}

How do Bayesian networks model probability distributions? What makes these representations \textit{concise} or \textit{economical}? We will begin by providing some intuition.

% i think copy-pasting some materials about Naive Bayes as a running example would be nice here
% see slides from generative modeling on lecture 2 for ideas; try pixel example, or just use spam example from introduction

The general idea behind the compact parameterization of Bayesian networks is surprisingly simple. Recall that by the chain rule (Definition \ref{def:chain_rule_variables}), we can write any probability $p$ as
\[
p(x_1, x_2, \dotsc, x_n) = p(x_1) p(x_2 \mid x_1) \cdots p(x_n \mid x_{n-1}, \dotsc, x_2, x_1).
\]
A compact Bayesian network is a distribution in which each factor on the right hand side depends only on a set of parent variables, denoted $\mathbf{pa}(x_i)$:
\[
p(x_i \mid x_{i-1}, \dotsc, x_1) = p(x_i \mid \mathbf{pa}(x_i)). 
\]
Often, the cardinality of $\mathbf{pa}(x_i)$ will be significantly less than $n$. For example, consider a model with five variables where node $x_5$ has parents $\mathbf{pa}(x_5) = \{x_4, x_3\}$. We can choose to write the factor $p(x_5 \mid x_4, x_3, x_2, x_1)$ as $p(x_5 \mid x_4, x_3)$ instead. 

When the variables are discrete (as is often the case in the problems we will discuss), we can think of the factors $p(x_i\mid \mathbf{pa}(x_i))$ as \textit{probability tables} in which rows correspond to assignments to $\mathbf{pa}(x_i)$, columns correspond to values of $x_i$, and entries contain the probabilities $p(x_i\mid \mathbf{pa}(x_i))$ (Figure \ref{fig:student_grades_dag}). If each variable takes $d$ values and has at most $k$ parents, then the entire table will contain at most $O(d^{k+1})$ entries. Since we have one table per variable, the entire probability distribution can be compactly described with only $O(nd^{k+1})$ parameters. Recall that we required $O(d^n)$ parameters under the naive approach.

\tikzset{
notice/.style  = { draw, rectangle callout, callout relative pointer={#1} }}
%https://texample.net/tikz/examples/energy-levels/

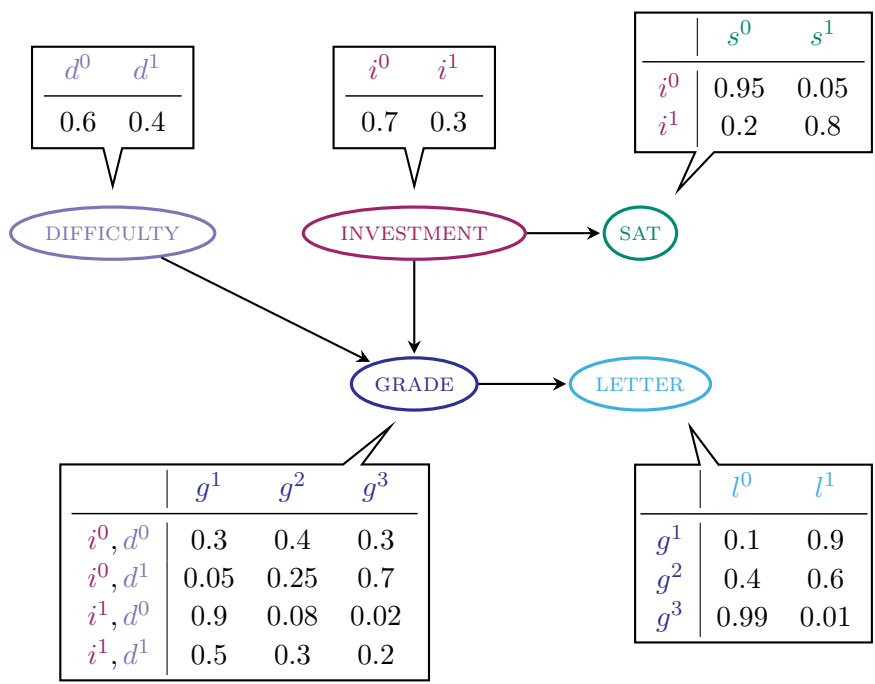
\begin{figure}[!t]
    \centering

\begin{tikzpicture}[
    node distance=1.5cm and 2.5cm,
    > = stealth, % arrow head style
    shorten > = 1pt, % don't touch arrow head to node
    auto,
    thick, % line style
    every node/.style={draw, rectangle, minimum height=2em, minimum width=2em, inner sep=2pt, align=center},
    every label/.append style={font=\tiny}
    ]

    % Nodes
    \node[ellipse,very thick,scale=0.9,Periwinkle] (Difficulty) at (-4,2) {\textsc{difficulty}};
    \node[ellipse,very thick,scale=0.9,RedViolet] (investment) at (0,2) {\textsc{investment}};
    \node[ellipse,very thick,scale=0.9,Blue] (Grade) at (0,0) {\textsc{grade}};
    \node[ellipse,very thick,scale=0.9,PineGreen] (SAT) at (3,2) {\textsc{sat}};
    \node[ellipse,very thick,scale=0.9,CornflowerBlue] (Letter) at (3,0) {\textsc{letter}};

    % Edges
    \path[-{Stealth[width=5pt,length=5pt]},thick] (Difficulty) edge (Grade);
    \path[-{Stealth[width=5pt,length=5pt]},thick] (investment) edge (Grade);
    \path[-{Stealth[width=5pt,length=5pt]},thick] (investment) edge (SAT);
    \path[-{Stealth[width=5pt,length=5pt]},thick] (Grade) edge (Letter);

    % Difficulty
    \node[notice={(0,-0.5)}, text width=2cm] (DTable) at (-4,3.8) {
        \begin{tabular}{cc}
            {\color{Periwinkle}$d^0$} & {\color{Periwinkle}$d^1$} \\ 
            \midrule
            0.6 & 0.4 \\
        \end{tabular}
    };

    % investment
    \node[notice={(0,-0.5)}, text width=2cm] (ITable) at (0,3.8) {
        \begin{tabular}{cc}
            {\color{RedViolet}$i^0$} & {\color{RedViolet}$i^1$} \\ 
            \midrule
            0.7 & 0.3 \\
        \end{tabular}
    };

    % Grade
    \node[notice={(0.5,0.5)}, text width=4.75cm] (GTable) at (-2.25,-2.5) {
        \begin{tabular}{c|ccc}
            & {\color{Blue}$g^1$} & {\color{Blue}$g^2$} & {\color{Blue}$g^3$} \\
            \midrule
            ${\color{RedViolet}i^0},{\color{Periwinkle}d^0}$ & 0.3 & 0.4 & 0.3 \\
            ${\color{RedViolet}i^0},{\color{Periwinkle}d^1}$ & 0.05 & 0.25 & 0.7 \\
            ${\color{RedViolet}i^1},{\color{Periwinkle}d^0}$ & 0.9 & 0.08 & 0.02 \\
            ${\color{RedViolet}i^1},{\color{Periwinkle}d^1}$ & 0.5 & 0.3 & 0.2 \\
        \end{tabular}
    };

    % SAT
    \node[notice={(-0.35,-0.5)}, text width=3cm] (STable) at (4.5,4) {
        \begin{tabular}{c|cc}
            & {\color{PineGreen}$s^0$} & {\color{PineGreen}$s^1$} \\
            \midrule
            ${\color{RedViolet}i^0}$ & 0.95 & 0.05 \\
            ${\color{RedViolet}i^1}$ & 0.2 & 0.8 \\
        \end{tabular}
    };

    % Letter
    \node[notice={(-0.25,0.5)}, text width=3cm] (LTable) at (4.5,-2.25) {
        \begin{tabular}{c|cc}
            & {\color{CornflowerBlue}$l^0$} & {\color{CornflowerBlue}$l^1$} \\ 
            \midrule
            {\color{Blue}$g^1$} & 0.1 & 0.9 \\
            {\color{Blue}$g^2$} & 0.4 & 0.6 \\
            {\color{Blue}$g^3$} & 0.99 & 0.01 \\
        \end{tabular}
    };

\end{tikzpicture}

    \caption{Bayesian network describing student academic performance. The joint distribution can be represented as a product of conditional probability distributions specified by the tables associated with each variable. Adapted from \citet{koller2009probabilistic}.}
    \label{fig:student_grades_dag}
\end{figure}

\paragraph{An Illustrative Example}

Consider a model of student performance~\citep{koller2009probabilistic}. In reality, student performance is determined by complex social dynamics. For ease of exposition, we will pretend that performance relies on a small number of straightforward determinants that we can observe. In our simplified model, grade $g$ depends on the exam's difficulty $d$ and the student's investment in studying $i$. In turn, this grade influences the quality $l$ of the reference letter from the teacher who taught the course. In addition, investment $i$ influences SAT score $s$. Variable $g$ takes 3 possible values. All other variables are binary. 

The graphical representation of this distribution is a DAG that visually specifies how the random variables depend on each other (shown in Figure \ref{fig:student_grades_dag}). Each variable is associated with a conditional probability table that is representative of the factor $p(x_i \mid \mathbf{pa}(x_i))$. We can read off these conditional distributions from the graphical structure to obtain a factorization for the joint distribution (following from Equation \ref{eq:factorization_bayesian_networks}). In this case, the joint probability distribution over the 5 variables naturally factorizes as
\[
p(l, g, i, d, s) = p(l \mid g)\, p(g \mid i, d)\, p(s \mid i) \, p(i)\, p(d).
\]
If we were provided the factorization first, we could easily reverse-engineer the DAG using the conditional independence assumptions that the factorization encodes. Thus, the graph and its corresponding factorization both clearly indicate the parent-child relationships in this distribution. %: the letter depends on the grade, which in turn depends on the student's investment and the difficulty of the exam, etc. 

Another way to interpret a directed graph is as a story for how the data were generated --- that is, a narrative to explain the \textit{data generating process}. In the above example, to determine the quality of the reference letter, we can first sample an investment level and an exam difficulty; then, a student's grade is sampled given these parameters; finally, the recommendation letter is generated based on that grade.

%\jm{Since this is taken directly from another source, I think it would be best to come up with our own example instead.}

\subsubsection{The Factorization of Bayesian Networks} 

We will now define Bayesian networks and their factorization more formally. Suppose we have a Bayesian network with an associated graph $\mathcal{G} = (\mathbf{V}, \mathbf{E})$. Each node $V_i \in \mathbf{V}$ corresponds to a random variable $X_i$. Each node is associated with one conditional probability distribution $p(x_i \mid \mathbf{pa}(x_i))$, specifying the probability of $x_i$ conditioned on the values taken by its parent set, $\mathbf{pa}(x_i)$. Thus, a Bayesian network defines a probability distribution $p$. Conversely, we say that a probability distribution \textit{factorizes} over a DAG $\mathcal{G}$ if it can be decomposed into a product of factors, as specified by $\mathcal{G}$. We can formalize these observations, beginning with the \textit{Markov condition}. 

\begin{definition}[Markov condition for Bayesian networks] \label{def:markov_condition}
    A node is conditionally independent of its non-descendants given its parents.
\end{definition}

%The Markov condition is closely related to the \textit{causal Markov condition}, a fundamental assumption in causal structure learning (Section \ref{sec:structure_learning}). Here, we do not require a causal interpretation. 
As a consequence of the Markov condition, we obtain the form for factorizing Bayesian networks.

\begin{definition}[Factorization of Bayesian networks] \label{def:factorization_bayesian_networks}
Let $p(\mathbf{x})$ be a joint probability distribution over random variables $\mathbf{X} = \{X_i\}_{i=1}^n$. Given the Markov condition, we can factorize the joint as
    \begin{align}
        p(\mathbf{x}) = \prod_{i=1}^n p(x_i \mid \mathbf{pa}(x_i)). \label{eq:factorization_bayesian_networks}
    \end{align}
\end{definition}

The conditional distribution describing a variable given its parent set is a \textit{factor} or \textit{module} that is invariant to all other factors. This fact explains the relationship between the factorization and graph given by Figure \ref{fig:intro_dag} in our introduction. Given this modularity, for all $i \neq j$, knowing $p(x_i \mid \mathbf{pa}(x_i))$ does not inform us about $p(x_j \mid \mathbf{pa}(x_j))$. Further, altering $p(x_i \mid \mathbf{pa}(x_i))$ does not alter $p(x_j \mid \mathbf{pa}(x_j))$ \citep{parascandolo2018learning}. %In the causal setting, these properties give rise to the Independence of Causal Mechanisms Assumption \citep{parascandolo2018learning}. 
These independence properties allow us to obtain compact factorizations for complex joint distributions, enabling efficient inference and learning. %\jm{Look into this.}

It is not hard to see that a probability represented by a Bayesian network will be valid: clearly, it will be nonnegative and one can show using an induction argument (and using the fact that the conditional probability distributions are valid probabilities) that the sum over all variable assignments will be 1. Conversely, we can also show by counter-example that when $\mathcal{G}$ contains cycles, its associated probability may not sum to 1.

\subsubsection{Independencies in Bayesian Networks}

To summarize, Bayesian networks represent probability distributions that can be formed via products of simpler, local conditional probability distributions (i.e., one \textit{factor} for each variable). By expressing a probability in this form, we introduce assumptions that certain variables are conditionally independent of each other.

This raises the question: exactly which independence assumptions are we making by using a Bayesian network model with a given structure? This question is important for two reasons. Firstly, we must clearly state the assumptions that we place on our model and be able to justify the extent to which we believe these assumptions are correct. Secondly, this information can help us design more efficient inference algorithms.

% also, maybe state high-level goal a bit more explicitly?

Let $I(p)$ denote the set of all independencies that hold for a joint distribution $p$. For example, if $p(x,y) = p(x) p(y)$, then we say that $x \ind y \in I(p)$. It turns out that a Bayesian network $\mathcal{G}$ very elegantly describes many independencies in $I(p)$. In actuality, Bayesian networks are imperfect representations that can fail to capture all independencies in some settings. We will return to this issue in Section \ref{sec:dags_imperfect}. For now, we will build our understanding of how these independencies can be recovered from the graph by introducing three primitive independence structures.

\begin{figure}[!t]
    \centering

\begin{tikzpicture}[scale=0.7,every edge quotes/.style = {font=\scriptsize, fill=white,sloped}]
  \node[circle,WildStrawberry,fill= Lavender!20,thick,draw,scale=0.7] (z) at (1,1.5) {\color{black}$Z$};
  \node[circle,black,thick,draw,scale=0.7] (x) at (0, 0) {$X$};
  \node[circle,black,thick,draw,scale=0.7] (y) at (2, 0) {$Y$};
  %%%
  \draw[thick,black,-{Stealth[width=5pt,length=5pt]}]  (z) edge[] (y);
  \draw[thick,black,-{Stealth[width=5pt,length=5pt]}]  (z) edge[] (x);
  %%%
  \node[align=flush center,text width=3cm] (label) at (1,-1.75) {\scriptsize\textbf{(a)} \textsc{fork} $p(z)p(x|z)p(y|z)$ {\color{RoyalBlue}$X \not\ind Y; X \ind Y \mid Z$}};
\end{tikzpicture}
%%%%%%%%%%%%%%%%
\hspace{-7mm}
%%%%%%%%%%%%%%%%
\begin{tikzpicture}[scale=0.7,every edge quotes/.style = {font=\scriptsize, fill=white,sloped}]
  \node[circle,WildStrawberry,fill= Lavender!20,thick,draw,scale=0.7] (z) at (1,1.5) {\color{black}$Z$};
  \node[circle,black,thick,draw,scale=0.7] (x) at (0, 0) {$X$};
  \node[circle,black,thick,draw,scale=0.7] (y) at (2, 0) {$Y$};
  %%%
  \draw[thick,black,{Stealth[width=5pt,length=5pt]}-]  (z) edge[] (y);
  \draw[thick,black,{Stealth[width=5pt,length=5pt]}-]  (z) edge[] (x);
  %%%
  \node[align=flush center,text width=3cm] (label) at (1,-1.75) {\scriptsize\textbf{(b)} \textsc{$v$-structure} $p(x)p(y)p(z|x,y)$ {\color{WildStrawberry}$X \ind Y ; X \not\ind Y \mid Z$}};
\end{tikzpicture} 
%%%%%%%%%%%%%%%%
\hspace{-7mm}
%%%%%%%%%%%%%%%%
\begin{tikzpicture}[scale=0.7,every edge quotes/.style = {font=\scriptsize, fill=white,sloped}] %style=framed, background rectangle/.append style= {thick,draw=black}, inner frame sep=0.1ex
  \node[circle,WildStrawberry,fill= Lavender!20,thick,draw,scale=0.7] (z) at (1,1.5) {\color{black}$Z$};
  \node[circle,black,thick,draw,scale=0.7] (x) at (0, 0) {$X$};
  \node[circle,black,thick,draw,scale=0.7] (y) at (2, 0) {$Y$};
  %%%
  \draw[thick,black,{Stealth[width=5pt,length=5pt]}-]  (x) edge[] (z);
  \draw[thick,black,{Stealth[width=5pt,length=5pt]}-]  (z) edge[] (y);
  %%%
  \node[align=flush center,text width=3cm] (label) at (1,-1.75) {\scriptsize\textbf{(c)} \textsc{cascade} $p(y)p(z|y)p(x|z)$ {\color{RoyalBlue}$X \not\ind Y; X \ind Y \mid Z$}};
\end{tikzpicture}
%%%%%%%%%%%%%%%%
\hspace{-7mm}
%%%%%%%%%%%%%%%%
\begin{tikzpicture}[scale=0.7,every edge quotes/.style = {font=\scriptsize, fill=white,sloped}]
  \node[circle,WildStrawberry,fill= Lavender!20,thick,draw,scale=0.7] (z) at (1,1.5) {\color{black}$Z$};
  \node[circle,black,thick,draw,scale=0.7] (x) at (0, 0) {$X$};
  \node[circle,black,thick,draw,scale=0.7] (y) at (2, 0) {$Y$};
  %%%
  \draw[thick,black,-{Stealth[width=5pt,length=5pt]}]  (x) edge[] (z);
  \draw[thick,black,-{Stealth[width=5pt,length=5pt]}]  (z) edge[] (y);
  %%%
  \node[align=flush center,text width=3cm] (label) at (1,-1.75) {\scriptsize\textbf{(d)} \textsc{cascade} $p(x)p(z|x)p(y|z)$ {\color{RoyalBlue}$X \not\ind Y; X \ind Y \mid Z$}};
\end{tikzpicture}

    \caption{Triple DAGs where $Z$ plays the role of common parent in a fork structure \textbf{(a)}, common child in a $v$-structure \textbf{(b)}, and mediator in a cascade or chain \textbf{(c, d)}. Note that \textbf{(a)}, \textbf{(c)}, and \textbf{(d)} all share the same statistical independencies, though their joint factorizations are unique.}
    \label{fig:triples}
\end{figure}
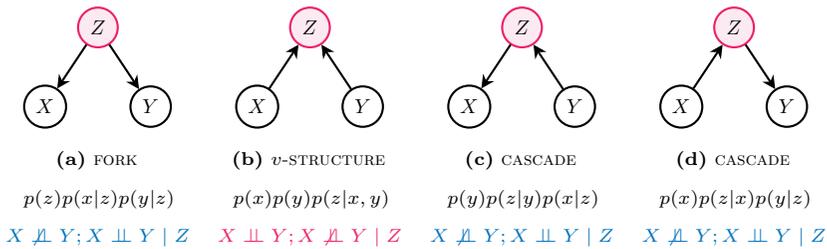

\subsubsection{The Triple DAG: Independence in Primitive Structures}

% introudce confoudner, causal connotation, briefly

% just say we stasrt with three nodes to get intuition
% define with we mean by latent vs observed
% Explain how we will use this later
% enumerating three node graphs with two edges
% As a special problem, we will use these four special graphs to characterize I(g) for all general graphs

We now introduce three primitive graphical structures that are fundamental building blocks for representation, learning, and inference on DAGs. Understanding these primitives will help us characterize how conditional independencies are encoded in a graph, which is an important step toward building models that support efficient inference procedures. For simplicity, consider a Bayesian network $\mathcal{G}$ with three nodes: $X$, $Y$, and $Z$. We will focus on node $Z$ for illustration. In Figure \ref{fig:triples} and Table \ref{tab:primitives_independencies}, we enumerate every role that $Z$ can play with respect to $\{X,Y\}$. Remark that $Z$ can take on one of three roles: \textit{common parent} (i.e., a special kind of \textit{confounder}), \textit{common child} (i.e., a special kind of \textit{collider}; see Example \ref{example:v_structure} for discussion), or \textit{mediator} (i.e., an intermediate variable on the directed path from $X$ to $Y$). These correspond to the three primitive structures in Bayesian networks: the \textit{fork}, the $v$-\textit{structure}, and the \textit{cascade} or \textit{chain} (of which there are two, one with root $X$ and one with root $Y$).
Each of the four resulting graphs corresponds to a different factorization of the joint probability distribution. However, only the $v$-structure differs in the conditional independencies that it encodes. This crucial observation propels much of graph structure learning (see Section \ref{sec:structure_learning}).

\begin{table}[!t]
    \centering
    \begin{adjustbox}{max width=\textwidth}
    \begin{tabular}{c c c c c p{5cm}}
    \toprule[1pt]
     \multicolumn{3}{c}{\small\textsc{Graphical Structure}} & \multicolumn{2}{c}{\small\textsc{Statistical Dependencies}} & \\
    \cmidrule(lr){1-3} \cmidrule(lr){4-5} 
     \textit{Primitive} & $\mathcal{G}$ & \textit{Role of} $Z$ & $\;\;\{X,Y\}\;\;$ & $\;\;\{X,Y,Z\}\;\;$ & \multicolumn{1}{c}{\textit{Intuition}} \\
    \midrule[1pt]
       \textsc{fork} & $X \leftarrow Z \rightarrow Y$ & \textsc{confounder} & {\color{RoyalBlue} $X \not\ind Y$} & {\color{RoyalBlue}$X \ind Y \mid Z$} & $X$ is not a parent of $Y$, and has no impact on the outcome of $Y$. The statistical association shared by $X$ and $Y$ is due to shared parent $Z$. Thus, knowing $X$ provides no additional information about $Y$ once we know $Z$. \\
       \midrule
       \textsc{cascade} & $X \rightarrow Z \rightarrow Y$ & \textsc{mediator} & {\color{RoyalBlue}$X \not\ind Y$} & {\color{RoyalBlue}$X \ind Y \mid Z$} & The outcome of $Y$ relies directly on $Z$ but only indirectly on $X$. As $Z$ mediates all the signal flowing from $X$ to $Y$, knowing $X$ provides no additional information about $Y$ once we know $Z$. \\
       \midrule
       $v$-\textsc{structure} & $X \rightarrow Z \leftarrow Y$ & \textsc{collider} & {\color{WildStrawberry}$X \ind Y$} & {\color{WildStrawberry}$X \not\ind Y \mid Z$} & $X$ and $Y$ are marginally independent, but knowing their common child $Z$ induces statistical association between them. Thus, knowing $X$ provides additional information about $Y$ only once we know $Z$.\\
    \bottomrule[1pt]
    \end{tabular}
    \end{adjustbox}
    \caption{Conditional independencies encoded by primitive DAG structures.}
    \label{tab:primitives_independencies}
\end{table}

\begin{figure}[!t]
    \centering

\begin{tikzpicture}[->, >={Stealth[length=5,width=5]}, node distance=2cm, every node/.style={draw, ellipse, minimum width=2cm, minimum height=0.75cm}]
    % Define nodes
    \node[thick,black] (Burglary) {\footnotesize\textsc{burglary}};
    \node[thick,black] (Earthquake) [right=4cm of Burglary] {\footnotesize\textsc{earthquake}};
    \node[thick,black] (Alarm) [below=1cm of $(Burglary)!0.5!(Earthquake)$] {\footnotesize\textsc{alarm}};
    
    % Define edges
    \path (Burglary) edge[thick,black] (Alarm);
    \path (Earthquake) edge[thick,black] (Alarm);
\end{tikzpicture}

    \caption{A $v$-structure in the classic \textsc{earthquake} network \citep{pearl1988probabilistic,koller2009probabilistic}, available in the \texttt{bnlearn} Bayesian network repository \citep{scutari2010learning}. This network encodes the belief that burglaries and earthquakes are probabilistically independent, yet both are causal for a security alarm sounding.}
    \label{fig:earthquake}
\end{figure}
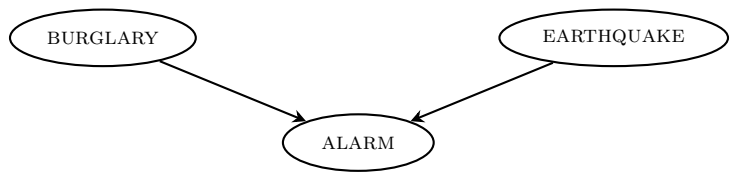

\begin{example}[Intuition for the $v$-structure]
\label{example:v_structure}
    The $v$-structure is perhaps the most unintuitive primitive. Given its centrality in graph structure learning, we will elaborate on it briefly. Consider the Bayesian network in Figure \ref{fig:earthquake}, where all variables are Bernoulli (taking values 0 = false or 1 = true). Assume a simplified world where there are only two reasons why a security alarm sounds (\textsc{alarm} = 1): a burglary took place or an earthquake occurred. This network encodes the belief that burglaries and earthquakes are probabilistically independent (i.e., \textsc{burglary} $\ind$ \textsc{earthquake}). Hence, there is no edge between \textsc{burglary} and \textsc{earthquake}. Additionally, the network encodes the belief that \textsc{burglary} and \textsc{earthquake} are both  causal for a security alarm sounding, and are therefore probabilistically dependent on \textsc{alarm} (i.e., \textsc{burglary} $\nind$ \textsc{alarm} and \textsc{earthquake} $\nind$ \textsc{alarm}). And yet, when conditioned on whether the alarm sounded, \textsc{burglary} and \textsc{earthquake} are now probabilistically \textit{dependent} (i.e., \textsc{burglary} $\nind$ \textsc{earthquake} $\mid$ \textsc{alarm}). This means that knowledge of the values that \textsc{alarm} and one of its parents take can provide knowledge of the value that its other parent takes. For example, if \textsc{alarm} = 1 and \textsc{burglary} = 0, then we know that \textsc{earthquake} must equal 1.
\end{example}

\paragraph{Markov Equivalence}

The simplified setting of the triple DAG also spotlights the idea of \textit{equivalence} among DAGs. As discussed, the fork and cascade share the same statistical dependencies, though their joint factorizations and graphical structures differ. In fact, these two primitives are \textit{Markov equivalent}. Formally, we can define an equivalence class over structures as follows.

\begin{definition}[Markov equivalence class (MEC),  \citealt{andersson1997characterization}]
\label{def:markov_equivalence_class}
    Two DAGs are Markov equivalent if and only if they share the same undirected skeleton and the same $v$-structures.
\end{definition}

We can also easily see from this definition that the $v$-structure primitive is \textit{not} Markov equivalent to the fork and cascade primitives: these share the same undirected skeleton, but not the same $v$-structures. %The MEC arises frequently in graphical modeling, as we will see for constraint-based and score-based structure learning (Section \ref{sec:structure_learning}). 

% Removing exmaple below. I don't think this is a good example because in real life, you wouldn't run your sprinkler if it was raining (hence this is not a v structure), among other reasons.
%Suppose that $Z$ is a Boolean variable that indicates whether our lawn is wet one morning. $X$ and $Y$ are two explanations for it being wet: either it rained (indicated by $X$) or the sprinkler turned on (indicated by $Y$). If we know that the grass is wet ($Z$ is true) and the sprinkler didn't go on ($Y$ is false), then the probability that $X$ is true must be 1, because that is the only other possible explanation. Hence, $X$ and $Y$ are not independent given $Z$.

\subsubsection{Independence in General Graphs} 

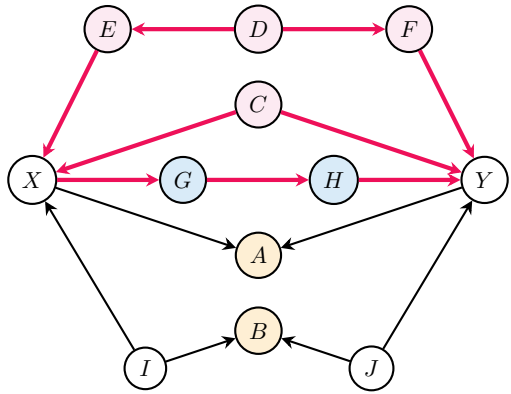
\begin{figure}[!t]
    \centering

\begin{tikzpicture}[every edge quotes/.style = {font=\footnotesize, fill=white,sloped},>={Stealth[width=5pt,length=5pt]}]
  \node[circle,black,thick,draw,scale=0.8] (x) at (-3, 1) {$X$};
  \node[circle,black,thick,draw,scale=0.8] (y) at (3, 1) {$Y$};
  \node[circle,fill=Dandelion!20,thick,draw,scale=0.8] (a) at (0, 0) {$A$};
  \node[circle,fill=Dandelion!20,thick,draw,scale=0.8] (b) at (0, -1) {$B$};
  \node[circle,fill=Lavender!20,thick,draw,scale=0.8] (c) at (0, 2) {$C$};
  \node[circle,fill=Lavender!20,thick,draw,scale=0.8] (d) at (0, 3) {$D$};
  \node[circle,fill=Lavender!20,thick,draw,scale=0.8] (e) at (-2, 3) {$E$};
  \node[circle,fill=Lavender!20,thick,draw,scale=0.8] (f) at (2, 3) {$F$};
  \node[circle,fill=CornflowerBlue!20,thick,draw,scale=0.8] (g) at (-1, 1) {$G$};
  \node[circle,fill=CornflowerBlue!20,thick,draw,scale=0.8] (h) at (1, 1) {$H$};
  \node[circle,black,thick,draw,scale=0.8] (i) at (-1.5, -1.5) {$I$};
  \node[circle,black,thick,draw,scale=0.8] (j) at (1.5, -1.5) {$J$};
  %%%
  \draw[ultra thick,OrangeRed,->]  (x) edge[] (g);
  \draw[ultra thick,OrangeRed,->]  (g) edge[] (h);
  \draw[ultra thick,OrangeRed,->]  (h) edge[] (y);
  \draw[thick,black,->]  (x) edge[] (a);
  \draw[thick,black,->]  (y) edge[] (a);
  \draw[thick,black,->]  (i) edge[] (x);
  \draw[thick,black,->]  (i) edge[] (b);
  \draw[thick,black,->]  (j) edge[] (y);
  \draw[thick,black,->]  (j) edge[] (b);
  %\draw[thick,black,<->]  (x) edge[bend right=30] (b);
  %\draw[thick,black,<->]  (y) edge[bend left=30] (b);
  \draw[ultra thick,OrangeRed,->]  (c) edge[] (x);
  \draw[ultra thick,OrangeRed,->]  (c) edge[] (y);
  \draw[ultra thick,OrangeRed,->]  (e) edge[] (x);
  \draw[ultra thick,OrangeRed,->]  (f) edge[] (y);
  \draw[ultra thick,OrangeRed,->]  (d) edge[] (e);
  \draw[ultra thick,OrangeRed,->]  (d) edge[] (f);
\end{tikzpicture}

    \caption{Colliders $\{A,B\}$, confounders $\{C,D,E,F\}$, and mediators $\{G, H\}$ with respect to nodes $X$ and $Y$. %Bidirected edges denote the existence of latent confounders. 
    Red paths are \textit{active} for $\{X,Y\}$ with respect to the empty set, while black paths are \textit{inactive} with respect to the empty set.} %Path $X \leftarrow C \rightarrow Y$ and $X \leftarrow E \leftarrow D \rightarrow F \rightarrow Y$ are backdoor paths for $X$ and $Y$.}
    \label{fig:confounder_collider_mediator}
\end{figure}

We can extend our three primitives to general networks by applying them recursively over larger graphs. To explore this in greater formality, we will introduce the concepts of $d$-separation and $d$-connection \citep{pearl1988probabilistic}.
%Here, $d$ refers to \textit{dependence}. \jm{I have seen the d attributed to directed and dependence in authoritative sources, so not sure what to say here.}
To lay the groundwork, we will first introduce notions of \textit{active paths} and \textit{inactive paths} (Figure \ref{fig:confounder_collider_mediator}). These concepts require us to generalize the notions of colliders, confounders, and mediators from DAGs with three nodes to the setting of general graphs of arbitrary complexity.

Note that colliders, confounders, and mediators are all defined with respect to a specified pair of variables. In the causal setting, this might be with respect to a cause-effect pair (also known as an exposure-outcome or treatment-outcome pair). In the simplest case, colliders are \textit{common descendants} or \textit{common effects} of the pair of interest (e.g., node $A$ with respect to pair $X$ and $Y$ in Figure \ref{fig:confounder_collider_mediator}). This generalizes the \textit{common child} relationship shown in Figure \ref{fig:triples}b to DAGs of arbitrary size. However, a collider can also take the form of node $B$ in Figure \ref{fig:confounder_collider_mediator}. In this case, the pair of interest shares confounders $I$ and $J$ with $B$, rather than being ancestral to $B$. Given its characteristic $M$-shape, this important kind of collider structure is often referred to as an $M$-\textit{structure}  \citep{ding2015adjust,pearl2015comment}.

The formal definition of an active path relies on this general notion of a collider. We have seen basic examples of active paths already: the cascade and fork when no intermediate variable is conditioned on (Figure \ref{fig:triples}c,d; Table \ref{tab:primitives_independencies}). Now we provide a general notion, which we define this with respect to the undirected skeleton of a directed path and a node set $\mathbf{S}$.

% VK: maybe introduce general notion first? or just a chain first?
% Maybe just remove this for now and say paths that go through confounders
%\begin{definition} [Backdoor path, \citealt{pearl_causal_2009}]
%    Let $\{X,Y\}$ denote a pair of variables. A backdoor path with respect to $\{X,Y\}$ is any path between them with an edge pointing into $X$ ($\cdots \to X$) and an edge pointing into $Y$ ($\cdots \to Y$)
    %when it is 1) not blocked by a non-collider that \textit{is} conditioned on or 2) not blocked by a collider that is \textit{not} conditioned on \citep{neal_introduction_2020}.
%    \label{def:backdoor_path}
%\end{definition}

%See Figure \ref{fig:confounder_collider_mediator} for examples of confounders on backdoor paths. Backdoor paths are a special kind of \textit{active path} that plays a fundamental role in causal inference on observational data, as it transmits \textit{non-causal association} between the two variables of interest. But what is an active path, more generally? 

\begin{definition}
    [Active path, \citealt{spirtes2001causation}]  \label{def:active}
    %Let $\mathbf{S}$ denote a node set and $X,Y \not\in \mathbf{S}$ denote a variable pair of interest.
    An undirected path %between $X$ and $Y$ 
    is \textit{active} relative to a node set $\mathbf{S}$ if every node on this path is active relative to $\mathbf{S}$. A node $V \in \mathbf{V}$ is active on a path relative to $\mathbf{S}$ if
    \begin{enumerate}[noitemsep,topsep=0pt]
        \item $V \notin \mathbf{S}$ is not a collider on the corresponding directed path,
        \item $V\in \mathbf{S}$ is a collider, or 
        \item $V\notin \mathbf{S}$ is a collider and at least one of its descendants is in $\mathbf{S}$.
    \end{enumerate}
\end{definition}

We then define an \textit{inactive} or \textit{blocked} path as one that is not active (e.g., due to existence of a collider $\notin \mathbf{S}$ on that path). As the definitions of active and inactive are with respect to $\mathbf{S}$, we assume $\mathbf{S} = \emptyset$ unless otherwise stated. For example, when  $\mathbf{S} = \emptyset$, the red paths between $X$ and $Y$ in Figure \ref{fig:confounder_collider_mediator} are active. The black paths between $X$ and $Y$ are inactive when  $\mathbf{S} = \emptyset$, as the flow of association is blocked by colliders $A$ and $B$. When $\mathbf{S} = \{A,B\}$, the black paths $X \rightarrow A \leftarrow Y$ and $X \leftrightarrow B \leftrightarrow Y$ are rendered active. To block the red path $X \leftarrow E \leftarrow D \rightarrow F \rightarrow Y$, we must condition on at least one variable in $\{E,D,F\}$. For the red path $X \to G \to H \to Y$, conditioning on at least one of the mediators $G$ or $H$ would render this path inactive.

With this context in mind, we finally come to define the fundamental concept of $d$-separation.

\begin{definition} [$d$-separation] \label{def:d_sep}
    Let $\mathcal{G} = (\mathbf{V}, \mathbf{E})$ be a directed graph. Let $\mathbf{X}$, $\mathbf{Y}$, $\mathbf{Z}$ be disjoint subsets of $\mathbf{V}$. If there is no active path between any $X \in \mathbf{X}$ and $Y \in \mathbf{Y}$ given $\mathbf{Z}$, then $\mathbf{X}$ and $\mathbf{Y}$ are $d$-separated given $\mathbf{Z}$. \label{def:d-sep}
\end{definition}

If two variables are $d$-separated relative to a set of variables $\mathbf{Z}$ in a directed graph, then they are independent conditional on $\mathbf{Z}$ in all probability distributions that factorize over the graph. Roughly, variables $X$ and $Y$ are independent conditional on $\mathbf{Z}$ if knowledge of $X$ provides no extra information about $Y$ once you have knowledge of $\mathbf{Z}$. 

We can then define $d$-connection as the negation of $d$-separation. Variables $X$ and $Y$ are $d$-connected by set $\mathbf{Z}$ if and only if there exists an undirected path between $X$ and $Y$ such that for every collider on the path, the collider or a descendent of it is in $\mathbf{Z}$, and no non-collider on the path is in $\mathbf{Z}$.

%

%\begin{definition}[$D$-separation]
%    Let $Q$, $W$, and $O$ be three sets of nodes in a Bayesian Network $\mathcal{G}$. We say that $Q$ and $W$ are $d$-separated given $O$ (i.e., the variables $O$ are observed) if $Q$ and $W$ are not connected by an \textit{active path}. An undirected path in $\mathcal{G}$ is called \textit{active} given observed variables $O$ if for every consecutive triple of variables $X,Y,Z$ on the path, one of the following holds:
%\begin{itemize}
%    \item $X \leftarrow Y \leftarrow Z$, and $Y$ is unobserved $Y \not\in O$
%    \item $X \rightarrow Y \rightarrow Z$, and $Y$ is unobserved $Y \not\in O$
%    \item $X \leftarrow Y \rightarrow Z$, and $Y$ is unobserved $Y \not\in O$
%    \item $X \rightarrow Y \leftarrow Z$, and $Y$ or any of its descendants are observed.
%\end{itemize}
%\end{definition}

%{% include marginfigure.html id="dp2" url="assets/img/dsep2.png" description="In this example, $X_1$ and $X_6$ are $d$-separated given $X_2, X_3$." %}{% include marginfigure.html id="dp1" url="assets/img/dsep1.png" description="However, $X_2, X_3$ are not $d$-separated given $X_1, X_6$. There is an active pass which passed through the $v$-structure created when $X_6$ is observed." %}

%For example, in the graph below, $X_1$ and $X_6$ are $d$-separated given $X_2, X_3$. However, $X_2, X_3$ are not $d$-separated given $X_1, X_6$, because we can find an active path $(X_2, X_6, X_5, X_3)$

The notion of $d$-separation is useful, because it lets us describe a large fraction of the dependencies that hold in our model. Let $I(\mathcal{G}) = \{(X \ind Y \mid Z) : \text{$X,Y$ are $d$-separated given $Z$}\}$ be a set of variables that are $d$-separated in $\mathcal{G}$. 

%{% include sidenote.html id="note_imap" note="We will not formally prove this, but the intuition is that if $X,Y$ and $Y,Z$ are mutually dependent, so are $X,Z$. Thus we can look at adjacent nodes and propagate dependencies according to the local dependency structures outlined above." %}:

\begin{definition}[$I$-map]
    If $p$ factorizes over $\mathcal{G}$, then $I(\mathcal{G}) \subseteq I(p)$. In this case, we say that $\mathcal{G}$ is an $I$-map (independence map) for $p$.
\end{definition}

%\jm{Define I-map better.}

In other words, all the independencies encoded in $\mathcal{G}$ are sound: variables that are $d$-separated in $\mathcal{G}$ are truly independent in $p$. However, the converse is not true: a distribution may factorize over $\mathcal{G}$, yet have independencies that are not captured in $\mathcal{G}$.

In a way this is almost a trivial statement. If $p(x,y) = p(x)p(y)$, then this distribution still factorizes over the graph $y \rightarrow x$, since we can always write it as $p(x,y) = p(x\mid y)p(y)$ with a conditional probability distribution $p(x\mid y)$ in which the probability of $x$ does not actually vary with $y$. However, we can construct a graph that matches the structure of $p$ by simply removing that unnecessary edge.

%\paragraph{Markov Blankets}

%From this definition, we arrive at the following useful property.
%\begin{theorem}
%    Given its Markov blanket, node $X$ is independent of all other nodes in $\mathcal{G}$.
%\end{theorem}

\subsubsection{Markov Blankets in Bayesian Networks}

Per the Markov condition, we know that a variable $X$ is conditionally independent of its non-descendants given its parents (Definition \ref{def:markov_condition}). Another important conditional independence fact in Bayesian networks follows from the definition of a \textit{Markov blanket} (\citealt{pearl1988probabilistic}). 

\begin{definition}[Markov blankets in Bayesian networks] \label{def:markov_blanket_dag}
    Let $\mathcal{G} = (\mathbf{V}, \mathbf{E})$ and $X \in \mathbf{V}$ be a graph and node of interest, respectively. When $\mathcal{G}$ is a Bayesian network, the Markov blanket for $X$ is the union of the parents, children, and spouses of $X$.
\end{definition}

Following from this definition, a node $X$ is conditionally independent of all other nodes in $\mathcal{G}$, given its Markov blanket. As node $X$ is $d$-separated from the remainder of the graph given its Markov blanket, we can use the Markov blanket as a parsimonious set of features to explain $X$ \citep{aliferis2010local}. As we will see in Chapter \ref{sec:approximate_inference}, Markov blankets are useful objects for approximate inference. Note that Markov blankets are defined differently in undirected graphs (see Definition \ref{def:markov_blanket_mrf} and  Figure \ref{fig:markov_blanket}). %In Section \ref{sec:structure_learning}, we will see how learning Markov blankets can facilitate feature selection in machine learning.

\subsubsection{Powerful Yet Imperfect Representations}
\label{sec:dags_imperfect}

In summary, we have shown that a Bayesian network $\mathcal{G}$ representing joint probability distribution $p$ specifies
\begin{enumerate}
    \item a factorization of $p$ that is a product of conditional distributions, each describing a variable given its parent set; and
    \item a set of conditional independencies that must be satisfied by any $p$ factorizing according to $\mathcal{G}$.
\end{enumerate}

This raises our last and perhaps most important question: can DAGs express \textit{all} the independencies of any distribution $p$? More formally, given a distribution $p$, can we construct a graph $\mathcal{G}$ such that $I(\mathcal{G}) = I(p)$?

\begin{figure}[!t]
    \centering

\begin{tikzpicture}[every edge quotes/.style = {font=\footnotesize, fill=white,sloped}]
  \node[circle,black,thick,draw,scale=0.8] (a) at (0, 0) {$A$};
  \node[circle,black,thick,draw,scale=0.8] (b) at (2, 0) {$B$};
  \node[circle,black,thick,draw,scale=0.8] (c) at (4, 0) {$C$};
  \node[circle,black,thick,draw,scale=0.8] (d) at (6, 0) {$D$};
  %%%
  \draw[thick,black,-{Stealth[width=5pt,length=5pt]}]  (a) edge[] (b);
  \draw[thick,black,-{Stealth[width=5pt,length=5pt]}]  (a) edge[bend right=30] (c);
  \draw[thick,black,-{Stealth[width=5pt,length=5pt]}]  (a) edge[bend right=30] (d);
  \draw[thick,black,-{Stealth[width=5pt,length=5pt]}]  (b) edge[] (c);
  \draw[thick,black,-{Stealth[width=5pt,length=5pt]}]  (b) edge[bend left=30] (d);
  \draw[thick,black,-{Stealth[width=5pt,length=5pt]}]  (c) edge[] (d);
  %%%
  %\node[] (label) at (1,-0.75) {\footnotesize};
\end{tikzpicture}

    \caption{A fully connected Bayesian network over four variables. There are no independencies in this model, and it is an $I$-map for any distribution.}
    \label{fig:fully_connected}
\end{figure}
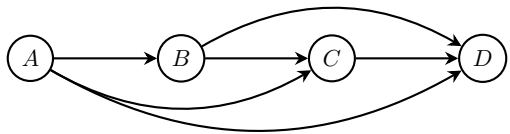

First, note that it is easy to construct a DAG $\mathcal{G}$ such that $I(\mathcal{G}) \subseteq I(p)$. A fully connected DAG $\mathcal{G}$ is an $I$-map for any distribution since $I(\mathcal{G}) = \emptyset$ (Figure \ref{fig:fully_connected}). A more interesting question is whether we can find a \textit{minimal} $I$-map $\mathcal{G}$ for $p$. That is, we might wish to find an $I$-map $\mathcal{G}$ such that the removal of even a single edge from $\mathcal{G}$ will result in it no longer being an $I$-map. To do this, we could start with a fully connected $\mathcal{G}$ and iteratively remove edges until $\mathcal{G}$ is no longer an $I$-map. 

However, what we are truly interested in determining is whether \textit{any} probability distribution $p$ always admits a \textit{perfect} map $\mathcal{G}$ for which $I(p)=I(\mathcal{G})$. Unfortunately, the answer is no. For example, consider the following distribution $p$ over three variables $X,Y,Z$. We can sample Bernoulli random variables $X,Y \sim \text{Ber}(0.5)$ and set
\[
Z = (X \text{ xor } Y) \text{ xor } \epsilon,
\]
where $\epsilon$ is a noise term representing a biased coin flip (also Bernoulli) and ``xor'' denotes the \textit{exclusive or} logical operator (which is true if and only if its two arguments differ). This data generating process is referred to as the \textit{noisy-xor example}. One can check using some algebra $\{X \ind Y, Z \ind Y, X \ind Z\} \in I(p)$ but $Z \ind \{Y,X\} \not\in I(p)$. Thus, $ X \rightarrow Z \leftarrow Y $ is an $I$-map for $p$, but none of the 3-node graph structures that we discussed perfectly describes $I(p)$, and hence this distribution doesn't have a perfect map. The noisy-xor example is a classic case of a \textit{faithfulness violation}, where faithfulness dictates that conditional independence in $p$ implies $d$-separation in $\mathcal{G}$ \citep{marx2021weaker}. We will revisit faithfulness briefly in Section \ref{sec:learning_directed_models}, as it is a common assumption in graph structure learning.

A related question is whether perfect maps are unique when they exist. Again, this is not the case. As a counterexample, $X \rightarrow Y$ and $X \leftarrow Y$ encode the same independencies, yet form different graphs. This invokes again the concept of Markov equivalence (Definition \ref{def:markov_equivalence_class}). Indeed, this phenomenon presents a fundamental challenge for learning graphical structures using tests of conditional independence alone, and we will see in Section \ref{sec:constraint_based} that such methods cannot provide graphical information beyond the MEC \citep{spirtes2001causation}. %More generally, we say that two Bayesian networks $\mathcal{G}_1, \mathcal{G}_2$ are $I$\textit{-equivalent} if they encode the same dependencies $I(\mathcal{G}_1) = I(\mathcal{G}_2)$. \jm{Differentiate this from Markov equivalence.}

%When are two Bayesian networks $I$-equivalent? To answer this, let's return to a simple example with three variables. Each of the graphs in Figure \ref{fig:triples} have the same undirected skeleton. The cascade-type structures (a,b) are clearly symmetric and the directionality of arrows does not matter. In fact, (a,b,c) encode exactly the same dependencies. We can change the directions of the arrows as long as we don't turn them into a $v$-structure (d). When we do have a $v$-structure, however, we cannot change any arrows: structure (d) is the only one that describes the dependency $X \not\ind Y \mid Z$. These examples provide intuition for the following general result on $I$-equivalence.
%\begin{theorem}
%    If $\mathcal{G},\mathcal{G}'$ have the same skeleton and the same $v$-structures, then $I(\mathcal{G}) = I(\mathcal{G}').$
%\end{theorem}
%\jm{This is basically markov equivalence, need to address this.}

%Again, it is easy to understand intuitively why this is true. Two graphs are $I$-equivalent if the $d$-separation between variables is the same. We can flip the directionality of any edge, unless it forms a $v$-structure, and the $d$-connectivity of the graph will be unchanged. We refer the reader to  \citet{koller2009probabilistic} for a full proof in Theorem 3.7 (page 77).

\vspace{2mm}

\begin{reading}
    \begin{itemize}[leftmargin=*]
        \item \fullcite{heckerman1998tutorial}.
        \item \fullcite{pearl1995causal}.
        \item \fullcite{pearl1995bayesian}.
        \item \fullcite{cooper1990computational}.
    \end{itemize}
\end{reading}

\vspace{2mm}

\section{Representing Distributions with Undirected Graphs}

While Bayesian networks can compactly represent many interesting probability distributions, we have seen that some distributions cannot be perfectly represented by models in this class. To prevent false independencies among the variables of our model, it may seem that such cases require a less compact representation. Under a naive approach, this could potentially result in a graph with superfluous edges --- and, consequently, unnecessary parameters in the model. This could make it more difficult to learn the parameters of our model and to make predictions.

However, Bayesian networks are not the only technique for compactly representing and visualizing probability distributions. In this section, we review three representational forms for undirected graphical models: \textit{Markov random fields}, \textit{conditional random fields}, and \textit{factor graphs}.

\subsection{Markov Random Fields}
\label{sec:mrf}

We begin our discussion of undirected graphical models with  \textit{Markov random fields} (MRFs), also known as \textit{Markov networks}. This class of models can compactly represent independence assumptions that directed models cannot represent. MRFs have featured prominently in image analysis and computer vision \citep{geman1986markov, li2009markov, kato2012markov}, spatiotemporal statistics \citep{clifford1990markov}, models of term dependency in text analysis \citep{metzler2005markov}, and more. %models of physical texture \citep{hassner1981use}, and more. 
In this section, we will explore the advantages and drawbacks of MRFs for various problem settings.

Formally, an MRF is a probability distribution $p$ over variables $\mathbf{X} = \{X_i\}_{i=1}^n$ defined by an undirected graph $\mathcal{G}$, in which nodes correspond to variables $X_i$. The probability distribution $p$ takes the form
\begin{align}
    p(\mathbf{x}) = \frac{1}{Z} \prod_{c \in \mathbf{C}} \phi_c(\mathbf{x}_c), \label{eq:mrf}
\end{align}
where $\mathbf{C}$ denotes the set of cliques in $\mathcal{G}$ (Definition \ref{def:clique}; Figure \ref{fig:cliques}), \textit{partition function} $Z$ is a normalizing constant, and each \textit{factor} $\phi_c$ is a nonnegative function $\phi_c : \text{Val}(\mathbf{X}_c) \mapsto \mathbb{R}$ over the variables in a clique. As $p(\mathbf{x})$ is a product of nonnegative functions, $p(\mathbf{x})$ is also guaranteed to be nonnegative. The clique constitutes the \textit{scope} (i.e., the domain) of factor $\phi_c$ (e.g., $Scope[\phi_c] = \mathbf{X}_c$).

Factors $\phi_c$ are also referred to as \textit{potential functions} or \textit{clique potentials}, and we often prefer to define these over \textit{maximal} cliques (Definition \ref{def:maximal_clique}). However, it is equally valid for $p(\mathbf{x})$ to contain factors whose scope is \textit{any} clique in $\mathcal{G}$ (e.g., pairwise potentials). Counter to what we have seen for the factorization of Bayesian networks (Definition \ref{def:factorization_bayesian_networks}), MRF potentials are not necessarily restricted to marginal or conditional distributions. This lack of restriction can result in a product of potentials that does not sum to 1. This explains the importance of partition function
\[
Z \coloneqq \sum_{x_1, \dotsc, x_n} \prod_{c \in \mathbf{C}} \phi_c(\mathbf{x}_c), 
\]
which ensures that $p(\mathbf{x})$  is a properly normalized probability distribution summing to 1. Note that we can generalize the formula for $Z$ to accommodate continuous random variables by replacing summation with integration, where appropriate.

In summary, a distribution $p$ can be expressed as a normalized product of factors, each with a scope over any clique in the corresponding undirected graph $\mathcal{G}$. Thus, the scope of a potential can be a single node, an edge, a triangle, etc. Furthermore, we do not need to specify a factor for each clique. The cliques over which we define our factors is a modeling choice, as we will illustrate in the following example.

\subsubsection{An Illustrative Example}

As a motivating example, suppose that we are modeling voting preferences among individuals. Let each person receive a label $\in \{A,B,C,D\}$. Let's say that $\{A,B\}$, $\{B,C\}$, $\{C,D)$, and $\{D,A\}$ are friends. From domain knowledge, we assume that friends tend to have similar voting preferences. These influences can be naturally represented by an undirected graph, like that in Figure \ref{fig:voting}.

One way to define a probability over the joint voting decision of $\{A,B,C,D\}$ is to assign scores to each assignment to these variables, and then define a probability as a normalized score. A score can be any function, but here we will define it to be 
\[
\tilde p(a,b,c,d)  = \phi(a,b)\phi(b,c)\phi(c,d)\phi(d,a),
\]
where $\phi(x,y)$ is a factor that assigns more weight to consistent votes among friends $X,Y$. For example,
\begin{align*}
\phi(x,y) =
\begin{cases}
10 & \text{if } X = Y = 1 \\
5  & \text{if } X = Y = 0 \\
1  & \text{otherwise}.
\end{cases}
\end{align*}
The final probability is then defined as
\[
p(a,b,c,d)  = \frac{1}{Z} \tilde p(a,b,c,d) ,
\]
where $ Z = \sum_{a,b,c,d} \tilde p(a,b,c,d)  $ is a normalizing constant that ensures that the distribution sums to 1.

\tikzset{
  prefix after node/.style={prefix after command=\pgfextra{#1}},
  /semifill/ang/.initial=45,
  /semifill/upper/.initial=none,
  /semifill/lower/.initial=none,
  semifill/.style={
    circle, draw,
    prefix after node={
      \pgfqkeys{/semifill}{#1}
      \path let \p1 = (\tikzlastnode.north), \p2 = (\tikzlastnode.center),
                \n1 = {\y1-\y2} in [radius=\n1]
            (\tikzlastnode.\pgfkeysvalueof{/semifill/ang}) 
            edge[
              draw=none,
              fill=\pgfkeysvalueof{/semifill/upper},
              to path={
                arc[start angle=\pgfkeysvalueof{/semifill/ang}, delta angle=180]
                -- cycle}] ()
            (\tikzlastnode.\pgfkeysvalueof{/semifill/ang}) 
            edge[
              draw=none,
              fill=\pgfkeysvalueof{/semifill/lower},
              to path={
                arc[start angle=\pgfkeysvalueof{/semifill/ang}, delta angle=-180,fill opacity=0.1]
                -- cycle}] ();}}}

\tikzset{
    side by side/.style 2 args={
        line width=2pt,
        #1,
        postaction={
            clip,postaction={draw,#2}
        }
    },
    circle node/.style={
        circle,
        draw,
        fill=white,
        minimum size=1.3cm
    }
}

\begin{figure}
    \centering

\begin{tikzpicture}%[every edge quotes/.style = {auto, font=\small, sloped,inner sep=10pt}]
%[every edge quotes/.style = {font=\footnotesize, fill=white,sloped}]
  \node[circle,black,thick,draw,scale=0.7] (A) at (0,2) {$A$};
  \node[circle,black,thick,draw,scale=0.7] (B) at (1,1) {$B$};
  \node[circle,black,thick,draw,scale=0.7] (C) at (0,0) {$C$};
  \node[circle,black,thick,draw,scale=0.7] (D) at (-1,1) {$D$};
  \node[] (label) at (1.75,1.75) {\color{OrangeRed}$\phi(a,b)$};
  \node[] (label) at (-1.75,1.75) {\color{SkyBlue}$\phi(a,d)$};
  \node[] (label) at (-1.75,0.25) {\color{LimeGreen}$\phi(d,c)$};
  \node[] (label) at (1.75,0.25) {\color{Dandelion}$\phi(b,c)$};
  %%%
  \draw[-,thick,black]  (A) edge[] (B);
  \draw[-,thick,black]  (A) edge[] (D);
  \draw[-,thick,black]  (B) edge[] (C);
  \draw[-,thick,black]  (D) edge[] (C);
  %%%
  \begin{pgfonlayer}{background}
    \fill[OrangeRed,opacity=0.3] \convexpath{A,B}{8pt};
    \fill[SkyBlue,opacity=0.3] \convexpath{A,D}{8pt};
    \fill[Dandelion,opacity=0.3] \convexpath{B,C}{8pt};
    \fill[LimeGreen,opacity=0.3] \convexpath{D,C}{8pt};
    \end{pgfonlayer}
  %%%
  %\node[] (label) at (0,-1) {\footnotesize\textsc{undirected graph} $\mathcal{G}$};
\end{tikzpicture}

\caption{MRF representing a joint probability distribution of voting preferences over four individuals. Highlighting indicates the potential functions ($\phi$) associated with each clique.}
\label{fig:voting}
\end{figure}
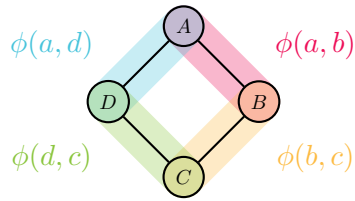

When normalized, we can view $\phi(a,b)$ as an interaction that pushes $B$'s vote closer to that of $A$. The term $\phi(b,c)$ pushes $B$'s vote closer to $C$, and the most likely vote will require reconciling these conflicting influences.

In this example, a sensible modeling choice was to define a factor over each edge (i.e., a clique of two nodes). Technically, we could have chosen to additionally specify \textit{unary factors} (i.e., cliques over single nodes). However, this would not have been a useful choice for this setting, as our assumptions concern only relations between friends, not individual voting tendencies.

Unlike in the directed case, the undirected model we describe here is not saying anything about how one variable is generated from another set of variables (as a conditional probability distribution would do). We have simply indicated a level of coupling between dependent variables in the graph. In a sense, this requires less prior knowledge, as we no longer have to specify a full generative story of how the vote of $B$ is constructed from the vote of $A$ (which we would need to do if we had a factor of the form $p(b\mid a)$). Instead, we simply identify dependent variables and define the strength of their interactions. This in turn defines an \textit{energy landscape} over the space of possible assignments, and we convert this energy to a probability via the normalizing constant.

\subsubsection{Independencies in Markov Random Fields}

Recall that in the case of Bayesian networks, we defined a set of independencies $I(\mathcal{G})$ that were described by a directed graph $\mathcal{G}$. We then showed how these describe true independencies that must hold in a distribution $p$ that factorizes over the directed graph (i.e., $I(\mathcal{G}) \subseteq I(p)$).

% VK: below, it would be nice to state things in terms of the notation I(G) and I(p) more

\tikzset{
  prefix after node/.style={prefix after command=\pgfextra{#1}},
  /semifill/ang/.initial=45,
  /semifill/upper/.initial=none,
  /semifill/lower/.initial=none,
  semifill/.style={
    circle, draw,
    prefix after node={
      \pgfqkeys{/semifill}{#1}
      \path let \p1 = (\tikzlastnode.north), \p2 = (\tikzlastnode.center),
                \n1 = {\y1-\y2} in [radius=\n1]
            (\tikzlastnode.\pgfkeysvalueof{/semifill/ang}) 
            edge[
              draw=none,
              fill=\pgfkeysvalueof{/semifill/upper},
              to path={
                arc[start angle=\pgfkeysvalueof{/semifill/ang}, delta angle=180]
                -- cycle}] ()
            (\tikzlastnode.\pgfkeysvalueof{/semifill/ang}) 
            edge[
              draw=none,
              fill=\pgfkeysvalueof{/semifill/lower},
              to path={
                arc[start angle=\pgfkeysvalueof{/semifill/ang}, delta angle=-180,fill opacity=0.1]
                -- cycle}] ();}}}

\tikzset{
    side by side/.style 2 args={
        line width=2pt,
        #1,
        postaction={
            clip,postaction={draw,#2}
        }
    },
    circle node/.style={
        circle,
        draw,
        fill=white,
        minimum size=1.3cm
    }
}

\begin{figure}
    \centering

\begin{tikzpicture}[every edge quotes/.style = {font=\footnotesize, fill=white,sloped}]
  \node[circle,black,thick,draw,scale=0.8] (A) at (-2,1) {$A$};
  \node[circle,black,thick,draw,scale=0.8] (B) at (-1,2) {$B$};
  \node[circle,OrangeRed,fill=Lavender!30,thick,draw,scale=0.8] (C) at (0,2) {\color{black}$C$};
  \node[circle,black,thick,draw,scale=0.8] (D) at (1,2) {$D$};
  \node[circle,black,thick,draw,scale=0.8] (E) at (2,1) {$E$};
  \node[circle,black,thick,draw,scale=0.8] (F) at (-1,0) {$F$};
  \node[circle,OrangeRed,fill=Lavender!30,thick,draw,scale=0.8] (G) at (0,0) {\color{black}$G$};
  \node[circle,black,thick,draw,scale=0.8] (H) at (1,0) {$H$};
  %%%
  \draw[-,thick,black]  (A) edge[] (B);
  \draw[-,thick,black]  (A) edge[] (F);
  \draw[-,thick,black]  (B) edge[] (C);
  \draw[-,thick,black]  (C) edge[] (D);
  \draw[-,thick,black]  (D) edge[] (E);
  \draw[-,thick,black]  (C) edge[] (G);
  \draw[-,thick,black]  (F) edge[] (G);
  \draw[-,thick,black]  (G) edge[] (H);
  \draw[-,thick,black]  (H) edge[] (E);
  %%%
  %\begin{pgfonlayer}{background}
   % \fill[Red,opacity=0.3] \convexpath{C,G}{8pt};
    %\end{pgfonlayer}
  %%%
  %\node[] (label) at (0,-2.5) {\footnotesize\textsc{undirected graph} $\mathcal{G}$};
\end{tikzpicture}
\caption{As $\{C,G\}$ forms a cutset of nodes for this graph, $\{A,B,F\}$ and $\{D,E,H\}$ are conditionally independent given $\{C,G\}$.}
\label{fig:cut_set}
\end{figure}
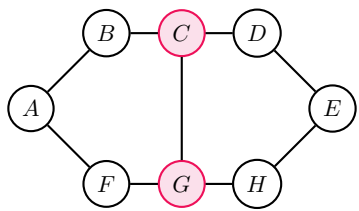

What independencies can be described by an MRF? The answer here is very simple and intuitive. Take MRF $\mathcal{G} = (\mathbf{V}, \mathbf{E})$. Variables $X,Y \in \mathbf{V}$ are dependent if they are connected by a path of unobserved variables. Now assume that every neighbor of $X$, denoted as set $\mathbf{ne}(X)$, is observed. Then, $X$ is conditionally independent of $\mathbf{V} \setminus \mathbf{ne}(X)$ given $\mathbf{ne}(X)$. This is because the neighbors of $X$ \textit{directly} influence $X$, but all remaining variables \textit{indirectly} influence $X$ via its neighbors. Thus, we can \textit{block} their influence by conditioning on $\mathbf{ne}(X)$.

We can also consider independence in MRFs in terms of  \textit{cutsets}, where a cutset is a set of nodes (or a set of edges) whose removal from $\mathcal{G}$ disconnects $\mathcal{G}$ into at least two connected components. Given the nodes in a cutset, the nodes in one connected component will be conditionally independent of those in another component. In Figure \ref{fig:cut_set}, $\{C,G\}$ is a cutset of nodes that partitions the graph into two disjoint subsets. 

Finally, we can revisit \textit{Markov blankets} (Definition \ref{def:markov_blanket_dag}). As highlighted in Figure \ref{fig:markov_blanket}, the Markov blanket for a node in a directed graph can differ from its Markov blanket in the corresponding undirected skeleton.

\begin{definition}[Markov blankets in undirected graphs] \label{def:markov_blanket_mrf}
    Let $\mathcal{G} = (\mathbf{V}, \mathbf{E})$ and $X \in \mathbf{V}$ be a graph and node of interest, respectively. When $\mathcal{G}$ is an undirected graph, the Markov blanket for a node $X$ is the set of all neighbors for $X$, i.e., $\mathbf{ne}(X)$.
\end{definition}

Thus, $X$ is independent from the remaining nodes in $\mathcal{G}$ when its Markov blanket is observed. The utility of this fact will become more clear when we explore approximate inference in Chapter \ref{sec:inference}. 

 \tikzset{
  prefix after node/.style={prefix after command=\pgfextra{#1}},
  /semifill/ang/.initial=45,
  /semifill/upper/.initial=none,
  /semifill/lower/.initial=none,
  semifill/.style={
    circle, draw,
    prefix after node={
      \pgfqkeys{/semifill}{#1}
      \path let \p1 = (\tikzlastnode.north), \p2 = (\tikzlastnode.center),
                \n1 = {\y1-\y2} in [radius=\n1]
            (\tikzlastnode.\pgfkeysvalueof{/semifill/ang}) 
            edge[
              draw=none,
              fill=\pgfkeysvalueof{/semifill/upper},
              to path={
                arc[start angle=\pgfkeysvalueof{/semifill/ang}, delta angle=180]
                -- cycle}] ()
            (\tikzlastnode.\pgfkeysvalueof{/semifill/ang}) 
            edge[
              draw=none,
              fill=\pgfkeysvalueof{/semifill/lower},
              to path={
                arc[start angle=\pgfkeysvalueof{/semifill/ang}, delta angle=-180,fill opacity=0.1]
                -- cycle}] ();}}}

\tikzset{
    side by side/.style 2 args={
        line width=2pt,
        #1,
        postaction={
            clip,postaction={draw,#2}
        }
    },
    circle node/.style={
        circle,
        draw,
        fill=white,
        minimum size=1.3cm
    }
}

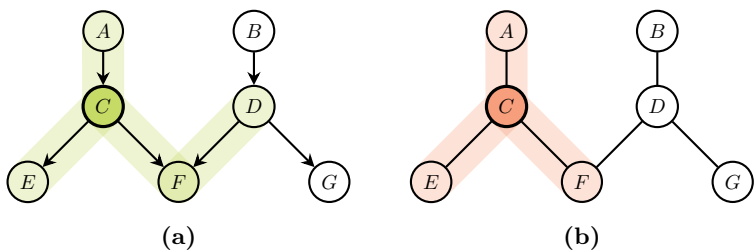
\begin{figure}
    \centering

\begin{tikzpicture}[every edge quotes/.style = {font=\footnotesize, fill=white,sloped}]
  \node[circle,black,thick,draw,scale=0.7] (A) at (-1,2) {$A$};
  \node[circle,black,thick,draw,scale=0.7] (B) at (1,2) {$B$};
  \node[circle,black,fill=SpringGreen,very thick,draw,scale=0.7] (C) at (-1,1) {\color{black}$C$};
  \node[circle,black,thick,draw,scale=0.7] (D) at (1,1) {$D$};
  \node[circle,black,thick,draw,scale=0.7] (E) at (-2,0) {$E$};
  \node[circle,black,thick,draw,scale=0.7] (F) at (0,0) {$F$};
  \node[circle,black,thick,draw,scale=0.7] (G) at (2,0) {$G$};
  %%%
  \draw[-{Stealth[width=5pt,length=5pt]},thick,black]  (A) edge[] (C);
  \draw[-{Stealth[width=5pt,length=5pt]},thick,black]  (C) edge[] (E);
  \draw[-{Stealth[width=5pt,length=5pt]},thick,black]  (C) edge[] (F);
  \draw[-{Stealth[width=5pt,length=5pt]},thick,black]  (B) edge[] (D);
  \draw[-{Stealth[width=5pt,length=5pt]},thick,black]  (D) edge[] (F);
  \draw[-{Stealth[width=5pt,length=5pt]},thick,black]  (D) edge[] (G);
  %%%
  \begin{pgfonlayer}{background}
    \fill[SpringGreen,opacity=0.3] \convexpath{A,C}{8pt};
    \fill[SpringGreen,opacity=0.3] \convexpath{C,E}{8pt};
    \fill[SpringGreen,opacity=0.3] \convexpath{C,F}{8pt};
    \fill[SpringGreen,opacity=0.3] \convexpath{D,F}{8pt};
    \end{pgfonlayer}
  %%%
  \node[] (label) at (0,-0.75) {\footnotesize \textbf{(a)}};
\end{tikzpicture}
%%%%%%%%%%%%%%%%%%%%%%%
\hspace{5mm}
%%%%%%%%%%%%%%%%%%%%%%%
\begin{tikzpicture}[every edge quotes/.style = {font=\footnotesize, fill=white,sloped}]
  \node[circle,black,thick,draw,scale=0.7] (A) at (-1,2) {$A$};
  \node[circle,black,thick,draw,scale=0.7] (B) at (1,2) {$B$};
  \node[circle,black,very thick,fill=Melon,draw,scale=0.7] (C) at (-1,1) {\color{black}$C$};
  \node[circle,black,thick,draw,scale=0.7] (D) at (1,1) {$D$};
  \node[circle,black,thick,draw,scale=0.7] (E) at (-2,0) {$E$};
  \node[circle,black,thick,draw,scale=0.7] (F) at (0,0) {$F$};
  \node[circle,black,thick,draw,scale=0.7] (G) at (2,0) {$G$};
  %%%
  \draw[-,thick,black]  (A) edge[] (C);
  \draw[-,thick,black]  (C) edge[] (E);
  \draw[-,thick,black]  (C) edge[] (F);
  \draw[-,thick,black]  (B) edge[] (D);
  \draw[-,thick,black]  (D) edge[] (F);
  \draw[-,thick,black]  (D) edge[] (G);
  %%%
  \begin{pgfonlayer}{background}
    \fill[Melon,opacity=0.3] \convexpath{A,C}{8pt};
    \fill[Melon,opacity=0.3] \convexpath{C,E}{8pt};
    \fill[Melon,opacity=0.3] \convexpath{C,F}{8pt};
    \end{pgfonlayer}
  %%%
  \node[] (label) at (0,-0.75) {\footnotesize \textbf{(b)}};
\end{tikzpicture}

\caption{Highlighted nodes belong to the Markov blanket for variable $C$ in a directed graph \textbf{(a)} and an undirected graph \textbf{(b)}.}
\label{fig:markov_blanket}
\end{figure}

\subsubsection{Comparison to Bayesian Networks}

As with Bayesian networks, MRFs are powerful but imperfect modes of representation. In the directed case, we found that $I(\mathcal{G}) \subseteq I(p)$ but yet there were distributions $p$ whose independencies could not be exactly described by $\mathcal{G}$. In the undirected case, the same holds.

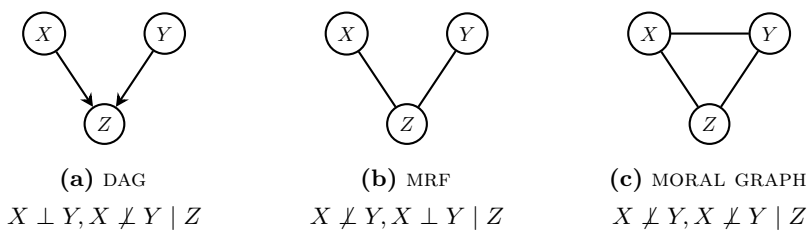
\begin{figure}[!t]
    \centering

\begin{tikzpicture}[scale=0.8,every edge quotes/.style = {font=\footnotesize, fill=white,sloped}]
  \node[circle,black,thick,draw,scale=0.7] (z) at (1,0) {$Z$};
  \node[circle,black,thick,draw,scale=0.7] (x) at (0, 1.5) {$X$};
  \node[circle,black,thick,draw,scale=0.7] (y) at (2, 1.5) {$Y$};
  %%%
  \draw[thick,black,{Stealth[width=5pt,length=5pt]}-]  (z) edge[] (y);
  \draw[thick,black,{Stealth[width=5pt,length=5pt]}-]  (z) edge[] (x);
  %%%
  \node[align=flush center,text width=3cm] (label) at (1,-1.25) {\footnotesize \textbf{(a)} \textsc{dag} $X \perp Y, X \not\perp Y \mid Z$}; %{\color{Mulberry}$X \perp Y ; X \not\perp Y \mid Z$}};
\end{tikzpicture}
%%%%%%%%%%%%%%%%%%%%%%%
\hspace{5mm}
%%%%%%%%%%%%%%%%%%%%%%%
\begin{tikzpicture}[scale=0.8,every edge quotes/.style = {font=\footnotesize, fill=white,sloped}]
  \node[circle,black,thick,draw,scale=0.7] (z) at (1,0) {$Z$};
  \node[circle,black,thick,draw,scale=0.7] (x) at (0, 1.5) {$X$};
  \node[circle,black,thick,draw,scale=0.7] (y) at (2, 1.5) {$Y$};
  %%%
  \draw[thick,black,-]  (z) edge[] (y);
  \draw[thick,black,-]  (z) edge[] (x);
  %%%
  \node[align=flush center,text width=3cm] (label) at (1,-1.25) {\footnotesize \textbf{(b)} \textsc{mrf} $X \not\perp Y, X \perp Y \mid Z$}; %{\color{Mulberry}$X \perp Y ; X \not\perp Y \mid Z$}};
\end{tikzpicture}
%%%%%%%%%%%%%%%%%%%%%%%
\hspace{5mm}
%%%%%%%%%%%%%%%%%%%%%%%
\begin{tikzpicture}[scale=0.8,every edge quotes/.style = {font=\footnotesize, fill=white,sloped}]
  \node[circle,black,thick,draw,scale=0.7] (z) at (1,0) {$Z$};
  \node[circle,black,thick,draw,scale=0.7] (x) at (0, 1.5) {$X$};
  \node[circle,black,thick,draw,scale=0.7] (y) at (2, 1.5) {$Y$};
  %%%
  \draw[thick,black,-]  (z) edge[] (y);
  \draw[thick,black,-]  (z) edge[] (x);
  \draw[thick,black,-]  (y) edge[] (x);
  %%%
  \node[align=flush center,text width=3cm] (label) at (1,-1.25) {\footnotesize \textbf{(c)} \textsc{moral graph} $X \not\perp Y, X \not\perp Y \mid Z$}; %{\color{Mulberry}$X \perp Y ; X \not\perp Y \mid Z$}};
\end{tikzpicture}

    \caption{The $v$-structure (Figure \ref{fig:triples}) is an example of a probability distribution that has a perfect directed graphical representation but no undirected representation.}
    \label{fig:undirected_v}
\end{figure}

So, what independencies \textit{cannot} be described by an MRF? Consider a probability distribution described by a DAG $\mathcal{G}$ that takes the form of a $v$-structure (Figure \ref{fig:undirected_v}). Neither the undirected skeleton of $\mathcal{G}$ nor its corresponding moral graph can describe the same independence assumptions encoded by the original DAG.

\begin{figure}[!t]
    \centering

\begin{tikzpicture}[scale=0.8]
  \node[circle,black,thick,draw,scale=0.7] (A) at (0,2) {$A$};
  \node[circle,black,thick,draw,scale=0.7] (B) at (1,1) {$B$};
  \node[circle,black,thick,draw,scale=0.7] (C) at (0,0) {$C$};
  \node[circle,black,thick,draw,scale=0.7] (D) at (-1,1) {$D$};
  %%%
  \draw[-{Stealth[width=5pt,length=5pt]},thick,black]  (A) edge[] (B);
  \draw[-{Stealth[width=5pt,length=5pt]},thick,black]  (A) edge[] (D);
  \draw[-{Stealth[width=5pt,length=5pt]},thick,black]  (B) edge[] (C);
  \draw[-{Stealth[width=5pt,length=5pt]},thick,black]  (D) edge[] (C);
  \node[] (label) at (0,-1) {\footnotesize \textbf{(a)}};
\end{tikzpicture}
%%%%%%%%%%%%%%%%%%%%%%%
\hspace{5mm}
%%%%%%%%%%%%%%%%%%%%%%%
\begin{tikzpicture}[scale=0.8]
  \node[circle,black,thick,draw,scale=0.7] (A) at (0,2) {$A$};
  \node[circle,black,thick,draw,scale=0.7] (B) at (1,1) {$B$};
  \node[circle,black,thick,draw,scale=0.7] (C) at (0,0) {$C$};
  \node[circle,black,thick,draw,scale=0.7] (D) at (-1,1) {$D$};
  %%%
  \draw[-{Stealth[width=5pt,length=5pt]},thick,black]  (B) edge[] (A);
  \draw[-{Stealth[width=5pt,length=5pt]},thick,black]  (B) edge[] (C);
  \draw[-{Stealth[width=5pt,length=5pt]},thick,black]  (D) edge[] (A);
  \draw[-{Stealth[width=5pt,length=5pt]},thick,black]  (D) edge[] (C);
  \node[] (label) at (0,-1) {\footnotesize \textbf{(b)}};
\end{tikzpicture}
%%%%%%%%%%%%%%%%%%%%%%%
\hspace{5mm}
%%%%%%%%%%%%%%%%%%%%%%%
\begin{tikzpicture}[scale=0.8]
  \node[circle,black,thick,draw,scale=0.7] (A) at (0,2) {$A$};
  \node[circle,black,thick,draw,scale=0.7] (B) at (1,1) {$B$};
  \node[circle,black,thick,draw,scale=0.7] (C) at (0,0) {$C$};
  \node[circle,black,thick,draw,scale=0.7] (D) at (-1,1) {$D$};
  %%%
  \draw[-{Stealth[width=5pt,length=5pt]},thick,black]  (A) edge[] (B);
  \draw[-{Stealth[width=5pt,length=5pt]},thick,black]  (C) edge[] (B);
  \draw[-{Stealth[width=5pt,length=5pt]},thick,black]  (D) edge[] (A);
  \draw[-{Stealth[width=5pt,length=5pt]},thick,black]  (D) edge[] (C);
  \node[] (label) at (0,-1) {\footnotesize \textbf{(c)}};
\end{tikzpicture}

    \caption{Examples of directed models for our four-variable voting example. None can accurately express our prior knowledge about the dependency structure among the variables, such that all variables are marginally dependent but $A \ind C \mid \{B,D\}$ and $B \ind D \mid \{A,C\}$.}
    \label{fig:voting_dags}
\end{figure}
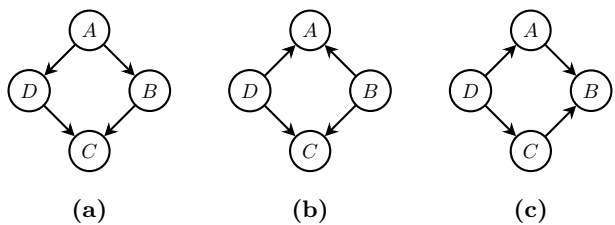

In our voting example, we had a distribution over $\{A,B,C,D\}$ that satisfied $A \ind C \mid \{B,D\}$ and $B \ind D \mid \{A,C\}$. Intuitively, this is reasonable: we assumed that only friends can directly influence each others' vote, and so a person's voting preference is independent of their non-friends' preferences given their friends' preferences. The MRF turns out to be a perfect map for this distribution (Figure \ref{fig:voting}). However, we can demonstrate by counterexample that these independencies cannot be perfectly represented by a Bayesian network (Figure \ref{fig:voting_dags}). Given the acyclicity constraint in Bayesian networks, there must be at least one sink node (in this case, colliders with no descendants). Thus, $A \ind C \mid \{B,D\}$ and $B \ind D \mid \{A,C\}$ cannot both be satisfied as at least one conditioning set will open up an active path (Definition \ref{def:active}).

More generally, MRFs have several advantages over directed models in certain settings. These include:

\begin{enumerate}
    \item MRFs can provide compact representations for a wider range of problems in which there is no natural directionality associated with variable dependencies.
    \item MRFs can succinctly express certain dependencies that DAGs cannot easily describe (although the converse is also true). 
\end{enumerate}

However, MRFs also possess several important drawbacks relative to Bayesian networks. These include:

\begin{enumerate}
    \item Computing the normalizing constant $Z$ requires summing over a potentially exponential number of assignments. In the general case, this will be NP-hard. Thus, many MRFs will be intractable and will require approximation techniques.
    \item Undirected models can be difficult to interpret.
    \item It is much easier to generate data from a Bayesian network, which is important in some applications.
    \item MRFs do not admit a causal interpretation.
\end{enumerate}

It is not hard to see that Bayesian networks are a special case of MRFs with (1) a very specific type of clique factor, which corresponds to a conditional probability distribution and implies a directed acyclic structure in the graph; and (2) a normalizing constant of 1. Recall our prior discussion on \textit{moralization} (Figure \ref{fig:moralization}): by ``marrying'' non-adjacent nodes that share children (i.e., removing \textit{immoralities}), a Bayesian network can always be converted into an undirected network with normalizing constant 1 (e.g., Figure \ref{fig:undirected_v}). The converse is also possible. However, this may yield a very large (e.g., fully connected) directed graph, making inference computationally intractable
% VK: not sure i understood about the converse

Thus, in some senses, MRFs have more power than Bayesian networks. Nevertheless, they are more difficult to deal with computationally and they limit certain forms of interpretation. A general rule of thumb is to use Bayesian networks whenever possible, opting for MRFs only if there is no natural way to model the problem with a directed graph (e.g., in our voting example).

\vspace{2mm}
\begin{reading}
    \begin{itemize}[leftmargin=*]
        \item \fullcite{kato2012markov}.
    \end{itemize}
\end{reading}

\subsection{Factor Graphs}
\label{sec:factor_graph}

\begin{figure}
    \centering

\begin{tikzpicture}[every edge quotes/.style = {font=\footnotesize, fill=white,sloped}]
  \node[circle,black,thick,draw,scale=0.7] (a) at (-0.75,0.75) {$A$};
  \node[circle,black,thick,draw,scale=0.7] (b) at (0.75,0.75) {$B$};
  \node[circle,black,thick,draw,scale=0.7] (c) at (0,-0.75) {$C$};
  \node[black,fill=black,thick,draw,scale=0.7] (factor) at (0, 0) {};
  \node[scale=0.7] (factor_label) at (0.75,0) {$\phi(a,b,c)$};
  %%%
  \draw[thick,black,-]  (a) edge[] (factor);
  \draw[thick,black,-]  (b) edge[] (factor);
  \draw[thick,black,-]  (c) edge[] (factor);
  %%%
  \node[] (label) at (0,-1.5) {\footnotesize\textbf{(a)} \textsc{single factor}};
\end{tikzpicture}
%%%%%%%%%%%%%%%%%%
\hspace{3mm}
%%%%%%%%%%%%%%%%%%
\begin{tikzpicture}[every edge quotes/.style = {font=\footnotesize, fill=white,sloped}]
  \node[circle,black,thick,draw,scale=0.7] (a) at (0,0) {$A$};
  \node[black,fill=black,thick,draw,scale=0.7] (f1) at (-1, 0) {};
  \node[black,below=of f1,scale=0.7,yshift=1cm] {$\phi(a)$};
  \node[circle,black,thick,draw,scale=0.7] (b) at (2,0) {$B$};
  \node[black,fill=black,thick,draw,scale=0.7] (f2) at (1, 0) {};
  \node[black,below=of f2,scale=0.7,yshift=1cm] {$\phi(b|a)$};
  \node[circle,black,thick,draw,scale=0.7] (c) at (4,0) {$C$};
  \node[black,fill=black,thick,draw,scale=0.7] (f3) at (3, 0) {};
  \node[black,below=of f3,scale=0.7,yshift=1cm] {$\phi(c|b)$};
  %%%
  \draw[thick,black,-]  (f1) edge[] (a);
  \draw[thick,black,-]  (a) edge[] (b);
  \draw[thick,black,-]  (b) edge[] (c);
  %%%
  \node[] (label) at (1.5,-1.5) {\footnotesize\textbf{(b)} \textsc{markov chain}};
\end{tikzpicture}
%%%%%%%%%%%%%%%%%%
\\
\vspace{5mm}
%%%%%%%%%%%%%%%%%%
\begin{tikzpicture}[every edge quotes/.style = {font=\footnotesize, fill=white,sloped}]
  \node[circle,black,thick,draw,scale=0.7] (a) at (0,0) {$A$};
  \node[black,fill=black,thick,draw,scale=0.7] (f1) at (-1, 0) {};
  \node[black,below=of f1,scale=0.7,yshift=0.8cm] {$\phi(a)$};
  \node[circle,black,thick,draw,scale=0.7] (b) at (2,0) {$B$};
  \node[black,fill=black,thick,draw,scale=0.7] (f2) at (1, 0) {};
  \node[black,below=of f2,scale=0.7,yshift=0.8cm] {$\phi(b|a)$};
  \node[circle,black,thick,draw,scale=0.7] (c) at (4,0) {$C$};
  \node[black,fill=black,thick,draw,scale=0.7] (f3) at (3, 0) {};
  \node[black,below=of f3,scale=0.7,yshift=0.8cm] {$\phi(c|a,b)$};
  \node[circle,black,thick,draw,scale=0.7] (d) at (6,0) {$D$};
  \node[black,fill=black,thick,draw,scale=0.7] (f4) at (5, 0) {};
  \node[black,below=of f4,scale=0.7,yshift=0.8cm] {$\phi(d|a,b,c)$};
  %%%
  \draw[thick,black,-]  (f1) edge[] (a);
  \draw[thick,black,-]  (a) edge[] (b);
  \draw[thick,black,-]  (b) edge[] (c);
  \draw[thick,black,-]  (c) edge[] (d);
  \draw[thick,black,-]  (a) edge[bend right=30] (f3);
  \draw[thick,black,-]  (a) edge[bend left=30] (f4);
  \draw[thick,black,-]  (b) edge[bend left=30] (f4);
  %%%
  \node[] (label) at (2.5,-1.5) {\footnotesize\textbf{(c)} \textsc{factorization by chain rule}};
\end{tikzpicture}

    \caption{Example factor graphs with factors $\phi$. \textbf{(a)} A probability distribution represented by a single factor (i.e., the trivial case). \textbf{(b)}  A Markov chain. \textbf{(c)} The factor graph resulting from applying the chain rule of probability to a distribution (Definition \ref{def:chain_rule_variables}). Adapted from \citet{kschischang2001factor}.}
    \label{fig:factor_graphs}
\end{figure}
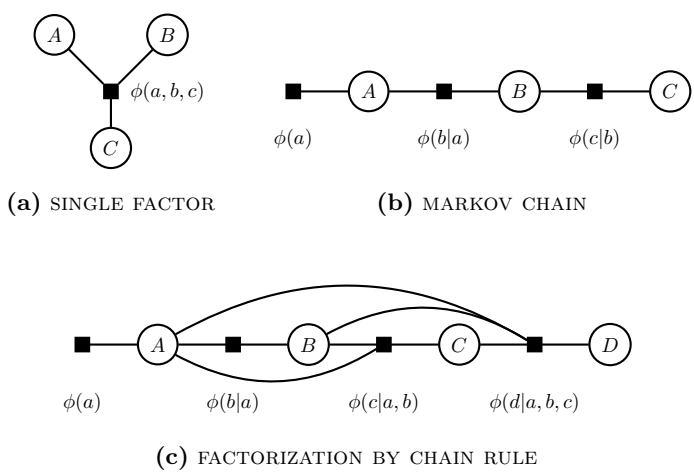

It is often useful to view MRFs in a way where factors and variables are explicit and separate in the representation. A \textit{factor graph} is one such way to do this. A factor graph is a \textit{bipartite graph} (Definition \ref{def:bipartite_graph}). Nodes representing variables in the distribution constitute one partition, while nodes representing the factors defined on these variables constitute the second partition. In other words, all edges go between factors and the variables that those factors depend on (Figure \ref{fig:factor_graphs}).

\begin{figure}
    \centering

\begin{tikzpicture}[every edge quotes/.style = {font=\footnotesize, fill=white,sloped}]
  \node[circle,black,thick,draw,scale=0.7] (a) at (-1,1) {$A$};
  \node[circle,black,thick,draw,scale=0.7] (b) at (1,1) {$B$};
  \node[circle,black,thick,draw,scale=0.7] (c) at (0,0) {$C$};
  \node[circle,black,thick,draw,scale=0.7] (d) at (0,-1) {$D$};
  %%%
  \draw[thick,black,-{Stealth[width=5pt,length=5pt]}]  (a) edge[] (c);
  \draw[thick,black,-{Stealth[width=5pt,length=5pt]}]  (b) edge[] (c);
  \draw[thick,black,-{Stealth[width=5pt,length=5pt]}]  (c) edge[] (d);
  %%%
  \node[] (label) at (0,-2) {\footnotesize\textbf{(a)} \textsc{dag} $\mathcal{G}$};
\end{tikzpicture}
%%%%%%%%%%%%%%%%%%%%
\hspace{5mm}
%%%%%%%%%%%%%%%%%%%%
\begin{tikzpicture}[every edge quotes/.style = {font=\footnotesize, fill=white,sloped}]
  \node[circle,black,thick,draw,scale=0.7] (a) at (-1,1) {$A$};
  \node[black,fill=black,thick,draw,scale=0.7] (fa) at (-1,0.5) {};
  \node[scale=0.8] (fa_label) at (-1,0.2) {$\phi(a)$};
  \node[circle,black,thick,draw,scale=0.7] (b) at (1,1) {$B$};
  \node[black,fill=black,thick,draw,scale=0.7] (fb) at (1,0.5) {};
  \node[scale=0.8] (fb_label) at (1,0.2) {$\phi(b)$};
  \node[black,fill=black,thick,draw,scale=0.7] (f1) at (0, 0.5) {};
  \node[scale=0.8] (f1_label) at (0,1) {$\phi(c|a,b)$};
  \node[circle,black,thick,draw,scale=0.7] (c) at (0,0) {$C$};
  \node[black,fill=black,thick,draw,scale=0.7] (f2) at (0,-0.5) {};
  \node[scale=0.8] (f2_label) at (-0.6,-0.5) {$\phi(d|c)$};
  \node[circle,black,thick,draw,scale=0.7] (d) at (0,-1) {$D$};
  %%%
  \draw[thick,black,-]  (a) edge[] (f1);
  \draw[thick,black,-]  (a) edge[] (fa);
  \draw[thick,black,-]  (b) edge[] (f1);
  \draw[thick,black,-]  (b) edge[] (fb);
  \draw[thick,black,-]  (f1) edge[] (c);
  \draw[thick,black,-]  (c) edge[] (d);
  %%%
  \node[] (label) at (0,-2) {\footnotesize\textbf{(a)} \textsc{factor graph} $\mathcal{F}$};
\end{tikzpicture}

    \caption{Factor graph for a simple DAG representing a probability distribution that factorizes as $p(a,b,c,d) = p(a)p(b)p(c|a,b)p(d|c)$.}
    \label{fig:factor_dag}
\end{figure}
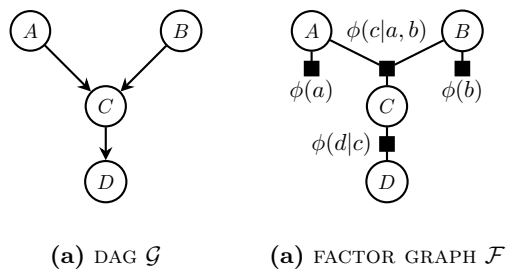

This graphical representation allows us to more readily see the factor dependencies between variables. We can obtain factor graphs for undirected graphs or for directed graphs (Figure \ref{fig:factor_dag}). As MRFs can admit multiple valid factorizations, multiple factor graphs can correspond to the same distribution. We will see in Chapter \ref{sec:inference} that factor graphs are a useful tool in inference, particularly in the family of message-passing algorithms.

\vspace{2mm}

\begin{reading}
    \begin{itemize}[leftmargin=*]
        \item \fullcite{kschischang2001factor}.
    \end{itemize}
\end{reading}

\vspace{2mm}

\subsection{Conditional Random Fields}
\label{sec:crf}

An important special case of MRFs arises when MRFs are applied to model a conditional probability distribution $p(\mathbf{y} \mid \mathbf{x})$, where $\mathbf{x} \in \mathcal{X}$ and $\mathbf{y} \in \mathcal{Y}$ are vector-valued variables. \textit{Conditional random fields} (CRFs) were introduced for segmenting and labeling sequence data, and are essentially MRFs with clique potentials that are conditioned on a set of features \citep{lafferty2001conditional}. CRFs were designed to address several drawbacks of hidden Markov models, stochastic grammars, and other generative models commonly used in domains that model sequence data (e.g., computational linguistics and computational biology).

Conditional distributions $p(\mathbf{y} \mid \mathbf{x})$ are common in supervised learning settings in which we are given $\mathbf{x}$ and want to predict $\mathbf{y}$. This setting is also known as \textit{structured prediction}, and CRFs can be viewed as a structured output extension of logistic regression \citep{murphy2012machine}. CRFs are more statistically efficient than MRFs when all we care to model is the distribution of labels given the data (just as discriminative classifiers can be efficient relative to generative classifiers).

The conditional distribution specified by a CRF can be expressed as a normalized product of potentials
\[
p(\mathbf{y} \mid \mathbf{x}) = \frac{1}{Z(\mathbf{x})} \prod_{c \in \mathbf{C}} \phi_c(\mathbf{x}_c,\mathbf{y}_c) 
\]
with partition function
\[
Z(\mathbf{x}) = \sum_{\mathbf{y} \in \mathcal{Y}} \prod_{c \in \mathbf{C}} \phi_c(\mathbf{x}_c,\mathbf{y}_c). 
\]
%\jm{This form is different than most other texts I am seeing; not necessarily wrong, but curious where you got this one.}

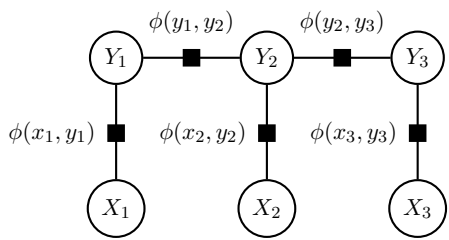
\begin{figure}
    \centering

\begin{tikzpicture}[every edge quotes/.style = {font=\footnotesize, fill=white,sloped}]

  % X
  \node[circle,black,thick,draw,scale=0.8] (x1) at (0,0) {$X_1$};
  \node[circle,black,thick,draw,scale=0.8] (x2) at (2,0) {$X_2$};
  \node[circle,black,thick,draw,scale=0.8] (x3) at (4,0) {$X_3$};

  % Y
  \node[circle,black,thick,draw,scale=0.8] (y1) at (0,2) {$Y_1$};
  \node[circle,black,thick,draw,scale=0.8] (y2) at (2,2) {$Y_2$};
  \node[circle,black,thick,draw,scale=0.8] (y3) at (4,2) {$Y_3$};
  
  % Factors
  \node[black,fill=black,thick,draw,scale=0.8] (f_x1y1) at (0,1) {};
  \node[black,fill=black,thick,draw,scale=0.8] (f_x2y2) at (2,1) {};
  \node[black,fill=black,thick,draw,scale=0.8] (f_x3y3) at (4,1) {};
  \node[black,fill=black,thick,draw,scale=0.8] (f_y1y2) at (1,2) {};
  \node[black,fill=black,thick,draw,scale=0.8] (f_y2y3) at (3,2) {};
  \node[left=0.05cm of f_x1y1,scale=0.8] (fa_label) {$\phi(x_1,y_1)$};
  \node[left=0.05cm of f_x2y2,scale=0.8] (fa_label) {$\phi(x_2,y_2)$};
  \node[left=0.05cm of f_x3y3,scale=0.8] (fa_label) {$\phi(x_3,y_3)$};
  \node[above=0.1cm of f_y1y2,scale=0.8] (fa_label) {$\phi(y_1,y_2)$};
  \node[above=0.1cm of f_y2y3,scale=0.8] (fa_label) {$\phi(y_2,y_3)$};
  
  % Edges
  \draw[thick,black,-]  (y1) edge[] (y2);
  \draw[thick,black,-]  (y2) edge[] (y3);
  \draw[thick,black,-]  (y1) edge[] (x1);
  \draw[thick,black,-]  (y2) edge[] (x2);
  \draw[thick,black,-]  (y3) edge[] (x3);
\end{tikzpicture}

    \caption{A factor graph representation of a linear-chain CRF.}
    \label{fig:crf_factor_graph}
\end{figure}

Note that in this case, the normalizing constant now depends on $\mathbf{x}$ (therefore, we say that it is a function). This is not surprising: $p(\mathbf{y} \mid \mathbf{x})$ is a probability over $\mathbf{y}$ that is parameterized by $\mathbf{x}$. That is, it encodes a different probability function for each $\mathbf{x}$. In that sense, a CRF results in an instantiation of a new MRF for each input $\mathbf{x}$.

CRFs are often represented as factor graphs (e.g., Figure \ref{fig:crf_factor_graph}). Every conditional distribution $p(\mathbf{y} \mid \mathbf{x})$ is a CRF for some factor graph, though that graph may be trivial \citep{sutton_introduction_2012}. Formally, we can say that a CRF meets the following condition.
\begin{definition}[Conditional random field (CRF), \citealt{sutton_introduction_2012}]
    Let $\mathcal{G}$ be a factor graph over random vector $\mathbf{Y}$. If the distribution $p(\mathbf{y} \mid \mathbf{x})$ for any fixed $\mathbf{x}$ factorizes according to $\mathcal{G}$, then $p(\mathbf{y} \mid \mathbf{x})$ is a CRF.
\end{definition}

\subsubsection{An Illustrative Example}

As a motivating example, consider the problem of recognizing a word from a sequence of black-and-white character images $\mathbf{x}_i \in [0, 1]^{d\times d}$ given to us in the form of pixel matrices. The output of our predictor is a sequence of alphabet letters $\mathbf{y}_i \in$ \{$\mathcal{A}$, $\mathcal{B}$, $\dotsc$, $\mathcal{Z}$\}.

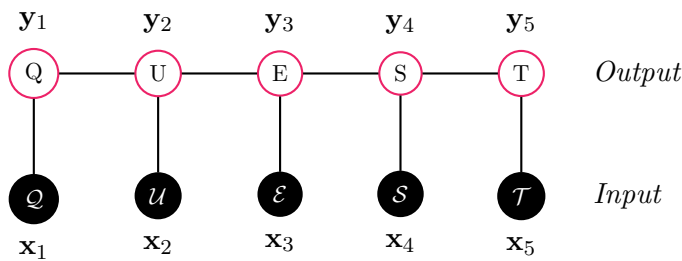
\begin{figure}
    \centering

\begin{tikzpicture}[>=stealth, node distance=1cm]

    % Nodes for y
    \node[draw, thick,circle, WildStrawberry, align=center,scale=0.8] (y1) {\color{black}Q};
    \node[draw, thick,circle, WildStrawberry, align=center, right=of y1,scale=0.8] (y2) {\color{black}U};
    \node[draw, thick,circle, WildStrawberry, align=center, right=of y2,scale=0.8] (y3) {\color{black}E};
    \node[draw, thick,circle, WildStrawberry, align=center, right=of y3,scale=0.8] (y4) {\color{black}S};
    \node[draw, thick,circle, WildStrawberry, align=center, right=of y4,scale=0.8] (y5) {\color{black}T};

    % Nodes for x
    \node[draw, thick,circle, fill=black,below=1cm of y1, align=center,scale=0.8] (x1) {\color{white}$\mathcal{Q}$};
    \node[draw, thick,circle, fill=black,below=1cm of y2, align=center,scale=0.8] (x2) {\color{white}$\mathcal{U}$};
    \node[draw, thick,circle, fill=black,below=1cm of y3, align=center,scale=0.8] (x3) {\color{white}$\mathcal{E}$};
    \node[draw, thick,circle, fill=black,below=1cm of y4, align=center,scale=0.8] (x4) {\color{white}$\mathcal{S}$};
    \node[draw, thick,circle, fill=black,below=1cm of y5, align=center,scale=0.8] (x5) {\color{white}$\mathcal{T}$};

    % Edges between y's
    \draw[-,thick] (y1) -- (y2);
    \draw[-,thick] (y2) -- (y3);
    \draw[-,thick] (y3) -- (y4);
    \draw[-,thick] (y4) -- (y5);

    % Edges between x's and y's
    \draw[-,thick] (x1) -- (y1);
    \draw[-,thick] (x2) -- (y2);
    \draw[-,thick] (x3) -- (y3);
    \draw[-,thick] (x4) -- (y4);
    \draw[-,thick] (x5) -- (y5);

    % Labels
    \node[align=center, right=0.5cm of y5] (output) {\textit{Output}};
    \node[align=center, right=0.5cm of x5] (input) {\textit{Input}};
    \node[above=0.1cm of y1] (y1_label) {$\mathbf{y}_1$};
    \node[above=0.1cm of y2] (y2_label) {$\mathbf{y}_2$};
    \node[above=0.1cm of y3] (y3_label) {$\mathbf{y}_3$};
    \node[above=0.1cm of y4] (y4_label) {$\mathbf{y}_4$};
    \node[above=0.1cm of y5] (y5_label) {$\mathbf{y}_5$};
    \node[below=0.1cm of x1] (x1_label) {$\mathbf{x}_1$};
    \node[below=0.1cm of x2] (x2_label) {$\mathbf{x}_2$};
    \node[below=0.1cm of x3] (x3_label) {$\mathbf{x}_3$};
    \node[below=0.1cm of x4] (x4_label) {$\mathbf{x}_4$};
    \node[below=0.1cm of x5] (x5_label) {$\mathbf{x}_5$};

\end{tikzpicture}

    \caption{A chain-structured CRF for optical character recognition.}
    \label{fig:optical_character_recognition}
\end{figure}

We could in principle train a classifier to separately predict each $\mathbf{y}_i$ from $\mathbf{x}_i$. However, since the letters together form a word, the predictions across different $i$ ought to inform each other. Consider the example CRF shown in Figure~\ref{fig:optical_character_recognition}, which illustrates a prediction of the word \textit{``QUEST''}. In this example, say it was ambiguous whether the second letter was a $\mathcal{U}$ or a $\mathcal{V}$. Since we can tell with high confidence that its neighbors are $\mathcal{Q}$ and $\mathcal{E}$, we can infer that $\mathcal{U}$ is the most likely true label. CRFs enable us to perform this prediction jointly.

More formally, suppose $p(\mathbf{y} \mid \mathbf{x})$ is a chain CRF with two types of factors. First, we have image factors $\phi(\mathbf{x}_i, \mathbf{y}_i)$ for $i = 1, \dotsc, n$. These assign higher values to $\mathbf{y}_i$ that are consistent with an input $\mathbf{x}_i$. We can also think of the $\phi(\mathbf{x}_i,\mathbf{y}_i)$ as probabilities $p(\mathbf{y}_i\mid \mathbf{x}_i)$ given by, say, standard (unstructured) softmax regression. Second, we have pairwise factors $\phi(\mathbf{y}_i, \mathbf{y}_{i+1})$ for $i = 1, \dotsc, n-1$. These pairwise factors can be seen as empirical frequencies of letter co-occurrences obtained from a large corpus of English text (e.g., Wikipedia).

Given a model of this form, we can jointly infer the structured label $\mathbf{y}$ using MAP inference, which we will discuss further in Chapter \ref{sec:approximate_inference}:
\[
\hat{\mathbf{y}} = 
\argmax_\mathbf{y} \phi(\mathbf{y}_1, \mathbf{y}_1) \prod_{i=2}^n \phi(\mathbf{y}_{i-1}, \mathbf{y}_i) \phi(\mathbf{x}_i, \mathbf{y}_i).
\]

\subsubsection{CRF Features}

In most practical applications of CRFs, we further assume that the factors $\phi_c(\mathbf{x}_c,\mathbf{y}_c)$ are of the form
\[
\phi_c(\mathbf{x}_c,\mathbf{y}_c) = \exp(\mathbf{w}_c^T \mathbf{f}_c(\mathbf{x}_c, \mathbf{y}_c)),
\]
where $\mathbf{w}$ denotes a weight vector and  $\mathbf{f}_c(\mathbf{x}_c, \mathbf{y}_c)$ is a set of \textit{features} describing the compatibility between $\mathbf{x}_c$ and $\mathbf{y}_c$.

In our optical character recognition example, we can introduce features $\mathbf{f}(\mathbf{x}_i, \mathbf{y}_i)$ that encode the compatibility of the letter $\mathbf{y}_i$ with the pixels $\mathbf{x}_i$. For example, $\mathbf{f}(\mathbf{x}_i, \mathbf{y}_i)$ may be the probability of letter $\mathbf{y}_i$  produced by logistic regression (or a deep neural network) evaluated on pixels $\mathbf{x}_i$. In addition, we can introduce features $\mathbf{f}(\mathbf{y}_i, \mathbf{y}_{i+1})$ between adjacent letters. These may be indicators of the form
\[
\mathbf{f}(\mathbf{y}_i, \mathbf{y}_{i+1}) = \mathbb{I}(\mathbf{y}_i = \ell_1, \mathbf{y}_{i+1} = \ell_2),
\]
where $\ell_1, \ell_2$ are two letters of the alphabet. The CRF would then learn weights $\mathbf{w}$ that assign more weight to more common letter sequences $(\ell_1, \ell_2)$, while at the same time making sure that the predicted $\mathbf{y}_i$ are consistent with the input $\mathbf{x}_i$. This process would allow us to determine $\mathbf{y}_i$ in cases where $\mathbf{x}_i$ is ambiguous, like in our above example.

The most important thing to know about CRF features is that they can be arbitrarily complex. In fact, we can define an optical character recognition model with factors $\phi(\mathbf{x},\mathbf{y}_i) = \exp(\mathbf{w}_i^T \mathbf{f}(\mathbf{x}, \mathbf{y}_i))$ that depend on the entire input $\mathbf{x}$. This does not affect computational performance at all: at inference time, $\mathbf{x}$ is always observed and our decoding problem involves maximizing
\[
\phi(\mathbf{x}, \mathbf{y}_1) \prod_{i=2}^n \phi(\mathbf{y}_{i-1}, \mathbf{y}_i) \phi(\mathbf{x}, \mathbf{y}_i) = \phi'(\mathbf{y}_1) \prod_{i=2}^n \phi(\mathbf{y}_{i-1}, \mathbf{y}_i) \phi'(\mathbf{y}_i),
\]
where $\phi'(\mathbf{y}_i) = \phi(\mathbf{x},\mathbf{y}_i)$. Using global features only changes the values of the factors but not their scope, which possesses the same type of chain structure. We will see in the next chapter that this structure is all that is needed to ensure a tractable solution to this optimization problem.

This observation may be interpreted in a slightly more general form. If we were to model $p(\mathbf{x},\mathbf{y})$ using an MRF (viewed as a single model over $\mathbf{x},\mathbf{y}$ with normalizing constant $Z = \sum_{\mathbf{x},\mathbf{y}} \Tilde{p}(\mathbf{x},\mathbf{y})$), then we would need to fit two distributions to the data: $p(\mathbf{y} \mid \mathbf{x})$ and $p(\mathbf{x})$. However, if all we are interested in is predicting $\mathbf{y}$ given $\mathbf{x}$, then modeling $p(\mathbf{x})$ is unnecessary. In fact, it may be statistically disadvantageous to do so. For example, we might not have enough data to fit both $p(\mathbf{y} \mid \mathbf{x})$ and $p(\mathbf{x})$; since the models have shared parameters, fitting one may not result in the best parameters for the other. Further, it might not be a good idea computationally: we would need to make simplifying assumptions so that $p(\mathbf{x})$ can be handled tractably. CRFs forgo this assumption, and thus often perform better on prediction tasks.

\begin{reading}
    \begin{itemize}[leftmargin=*]
        \item \fullcite{sutton_introduction_2012}.
    \end{itemize}
\end{reading}

%%%%%%%%%%%%%%%%%%%%%%%%
%% Inference
%%%%%%%%%%%%%%%%%%%%%%%%

\chapter{Exact Inference}
\label{sec:inference}

\begin{figure}
    \centering

\begin{tikzpicture}[every edge quotes/.style = {font=\footnotesize, fill=white,sloped}]
  \node[circle,black,scale=0.8,thick,draw] (x) at (0,1.5) {$X$};
  \node[circle,black,scale=0.8,thick,draw] (y) at (0, 0) {$Y$};
  %%%
  \draw[thick,black,-{Stealth[width=5pt,length=5pt]}]  (x) edge[] (y);
  %%%
  \node[black,below=of y,yshift=0.6cm] {\footnotesize\textbf{(a)} $p(x,y) = p(x)p(y \mid x)$};
\end{tikzpicture}
%%%%%%%%%%%%
\hspace{1mm}
%%%%%%%%%%%%
\begin{tikzpicture}[every edge quotes/.style = {font=\footnotesize, fill=white,sloped}]
  \node[circle,black,scale=0.8,thick,draw] (x) at (0,1.5) {$X$};
  \node[circle,black,scale=0.8,thick,draw,fill=SeaGreen!20] (y) at (0, 0) {$Y$};
  %%%
  \draw[thick,black,-{Stealth[width=5pt,length=5pt]}]  (x) edge[] (y);
  %%%
  \node[black,below=of y,yshift=0.6cm] {\footnotesize\textbf{(b)} Observe  $Y = y$};
\end{tikzpicture}
%%%%%%%%%%%%
\hspace{1mm}
%%%%%%%%%%%%
\begin{tikzpicture}[every edge quotes/.style = {font=\footnotesize, fill=white,sloped}]
  \node[circle,black,scale=0.8,thick,draw,fill=SeaGreen!20] (x) at (0,1.5) {$X$};
  \node[circle,black,scale=0.8,thick,draw,fill=SeaGreen!20] (y) at (0, 0) {$Y$};
  %%%
  \draw[thick,black,-{Stealth[width=5pt,length=5pt]}]  (y) edge[] (x);
  %%%
  \node[black,below=of y,yshift=0.6cm] {\footnotesize\textbf{(c)} $p(x,y) = p(y)p(x \mid y)$};
\end{tikzpicture}

    \caption{A simple case of inference over a graphical representation of Bayes' rule (figure and caption adapted from  \citealt{bishop2006pattern}, Chapter 8.4). %Figure 8.37
    Let $p(x,y)$ be a joint distribution over variables $X$ and $Y$. We can factorize this joint as $p(x)p(y \mid x)$ by applying the chain rule (Definition \ref{def:chain_rule_variables}), as graphically represented by \textbf{(a)}. Next, we observe the value taken by $Y$ (denoted by shading) and treat $p(x)$ as a prior over the latent variable $X$ \textbf{(b)}. We now wish to infer the posterior distribution $p(x \mid y)$. By the sum and product rules, $p(y) = \sum_{x'} p(y  \mid  x')p(x')$. By Bayes' rule (Definition \ref{def:bayes_rule}), $p(x \mid y) = p(y \mid x)p(x)/p(y)$. Thus, we can rearrange the joint factorization as $p(y)p(x \mid y)$, as represented by \textbf{(c)}.}
    \label{fig:bayes_theorem_inference}
\end{figure}
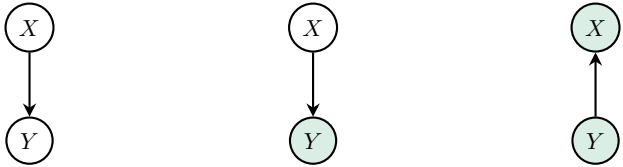

We now turn our attention to the problem of \textit{inference} in graphical models (Figure \ref{fig:bayes_theorem_inference}).  Generally speaking, inference is the process of drawing conclusions from available evidence. Inference can also be framed as a ``reasoned change in view'' \citep{harman1986change} or ``reasoning with beliefs'' \citep{boghossian2014inference}. Through this lens, inference can be a dynamic process where beliefs are iteratively challenged and updated as new evidence is introduced.

Given a probabilistic model (such as a Bayesian network or MRF), we perform inference to answer useful questions about the world. Our model can help us answer questions like:
\begin{itemize}
    \item \textit{What is the posterior distribution over a variable of interest?}
    \item \textit{What is the most probable explanation for a given hypothesis?}
    \item \textit{How certain are we that this explanation holds?}
\end{itemize}

In this chapter, we present means of answering such questions. We will mainly consider the problem of computing posterior probabilities of the form $p(\mathbf{y} \mid \mathbf{x})$ for \textit{query variables} $\mathbf{y}$ given \textit{observed evidence} $\mathbf{x} = (x_1, \ldots, x_n)$ \citep{zhang1996exploiting}. Formally, we will focus on two types of queries:

\begin{enumerate}
    \item \textit{Marginal inference.} What is the probability of a given variable in our model after we sum everything else out, e.g.,
    \[
    p(y=1) = \sum_{x_1} \sum_{x_2} \cdots \sum_{x_n} p(y=1, x_1, x_2, \dotsc, x_n).
    \]
    \item \textit{Maximum a posteriori (MAP) inference.} What is the most likely assignment to the variables in the model, e.g.,
    \[
    \hat{\mathbf{x}} =
    \argmax_{x_1, \dotsc, x_n} p(y=1, x_1, \dotsc, x_n).
    \]
\end{enumerate}

%Returning to our spam example, we might be interested in determining the probability that a given email is spam. We can answer such questions through \textit{statistical inference}: the process of drawing conclusions about a population $\mathcal{P}$ from observed data that describe a random sample from $\mathcal{P}$. Drawing conclusions from finite data samples can be challenging, as we will see throughout this tutorial. \textit{Bayesian inference} is a form of statistical inference in which we update the probability assigned to a hypothesis based on new evidence, with updates formulated according to Bayes' rule (Definition \ref{def:bayes_rule}; see \citealt{gelman1995bayesian}). In probabilistic graphical modeling, we employ statistical inference -- sometimes from a Bayesian perspective -- to answer marginal and MAP queries and reason under uncertainty (Figure \ref{fig:bayes_theorem_inference}). 

In this chapter, we place our initial focus on \textit{exact} probabilistic inference in graphical models. This is a challenging task that is NP-hard in the general case \citep{cooper1990computational}. Nevertheless, tractable solutions can be obtained for certain kinds of problems. %Though probabilistic inference on Bayesian networks is NP-hard in the general case \citep{cooper1990computational}...
As illustrated throughout this chapter, the tractability of an inference problem depends heavily on the structure of the graph that describes the probability of interest.

To begin, we will introduce three fundamental and inter-related algorithms for exact inference: variable elimination (Section \ref{sec:variable_elimination}), belief propagation for tree-structured graphs (Section \ref{sec:belief_propagation}), and the junction tree algorithm (Section \ref{sec:junction_tree}).
We will then discuss the computational complexity of MAP versus marginal inference and present a couple of algorithms for efficient MAP inference (Section \ref{sec:exact_map}).
In Chapter \ref{sec:approximate_inference}, we will continue our discussion in the context of \textit{approximate inference} for the setting where exact inference is intractable.
% \williex{update:}~review the problem of MAP inference (Section \ref{sec:exact_map}). We will then

%%%%%%
%% Variable elimination
%%%%%%

\section{Variable Elimination}
\label{sec:variable_elimination}

We first introduce \textit{variable elimination}, a fundamental exact inference method for answering queries to MRFs and Bayesian networks \citep{zhang1994simple}. Variable elimination is a general approach that can be used for both marginal and MAP inference. %Also known as \textit{bucket elimination} \citep{dechter1998bucket}.
It efficiently computes quantities of interest by serially \textit{eliminating} variables that are irrelevant to the query.
Given a model over variables $\mathbf{X}$, with joint distribution $p(\mathbf{x})$, variable elimination can be used to answer queries such as conditional probabilities --- for example, computing a distribution of the form
\[
p(\mathbf{x}_q \mid \mathbf{x}_e),
\]
where $\mathbf{X}_q$ is a set of query variables, $\mathbf{X}_e$ is a (potentially empty) set of evidence variables, and the remaining variables in $\mathbf{X}$ are marginalized out. Note that for MAP inference, this query will take a different form. At a high-level, variable elimination exploits two convenient observations~\citep{koller2009probabilistic}:
\begin{enumerate}
    \item By the factorization of Bayesian networks (Definition~\ref{def:factorization_bayesian_networks}), factors often depend on a relatively small subset of variables in the joint. We can reduce the number of parameters in our model by leveraging this fact (see Section~\ref{sec:rep_bayesian_networks}).
    \item With dynamic programming, we can compute subexpressions in the joint and memoize to prevent exponential increases in computation.
\end{enumerate}
This procedure repeatedly performs two factor operations: \textit{product} and \textit{marginalization}.  Conveniently, we can interchange product and sum operations to reduce the number of computations. For example, consider three factors: $\phi_x$ with scope $\mathbf{X}$, and $\phi_y$ with scope $\mathbf{Y}$, and $\phi_{yz}$ with scope $\mathbf{Y} \cup \mathbf{Z}$. Because $\mathbf{Y}$ is in scope for $\phi_y$ and $\phi_{yz}$ only, we can interchange sum and product as follows:
\[
\sum_{\mathbf{y}} \phi_x \phi_y \phi_{yz} = \phi_x \sum_{\mathbf{y}} \phi_y \phi_{yz}.
\]
This simple rearrangement, whereby we \textit{push} the summation deeper into the product, is the crux of variable elimination.

For ease of exposition, we will assume for the remainder of this chapter that all random variables are discrete variables taking $k$ possible values each. While the principles behind variable elimination also extend to many continuous distributions (e.g., Gaussians), we will not discuss these extensions here.

We present the full variable elimination procedure in Algorithm \ref{alg:variable_elimination}. To lay the groundwork, we first describe the special case of marginal inference in chain-structured graphs. Building on this, we then present the general case, known as the \textit{sum-product variable elimination algorithm}, and describe how it can be used for both marginal and MAP inference queries, as well as queries that condition on observed evidence.

% first introduce the building blocks: (1) the product and marginalization operations and (2) elimination orderings. We then introduce marginal inference and the notion of conditioning on observed evidence.

%Assume that we are given a graphical model as a product of factors $\phi_c(\mathbf{x}_c)$:
%\[
%p(x_1, \dotsc, x_n) = \prod_{c \in \mathbf{C}} \phi_c(\mathbf{x}_c). 
%\]
%Recall that we can view a factor as a multi-dimensional table assigning a value to each assignment of a set of variables $\mathbf{x}_c$. In a Bayesian network, the factors correspond to conditional probability distributions. In an MRF, the factors encode an unnormalized distribution. To compute marginals for the MRF, we perform the following steps:
%\begin{enumerate}
%    \item Calculate the partition function (also using variable elimination).
%    \item Compute marginals using the unnormalized distribution.
%    \item Divide the result by the partition constant to construct a valid marginal probability.
%\end{enumerate}

% VK: would be nice to start with a simple example first (<1 page for a really simple but representative graph) and then expand the more general version

\subsection{Marginal Inference in a Chain-Structured Graph}

% \vk{I would start with this. If I had to choose one recurrent piece of feedback it would be to start with simple examples first and then do the general abstract exposition}

We will first use a simple chain-structured graph as an illustrative example for several useful concepts. 

%\subsubsection{An Illustrative Example}

\begin{figure}[!t]
    \centering
%\begin{tikzpicture}[scale=0.12]
%\tikzstyle{every node}+=[inner sep=0pt]
%\draw [black,thick] (23.2,-29.3) circle (3);
%\draw (23.2,-29.3) node {$X_1$};
%\draw [black,thick] (35.2,-29.2) circle (3);
%\draw (35.2,-29.2) node {$X_2$};
%\draw [black,thick] (47.4,-29.2) circle (3);
%\draw (47.4,-29.2) node {$\,\cdots$};
%\draw [black,thick] (60,-29.2) circle (3);
%\draw (60,-29.2) node {$X_n$};
%%%%%%%%
%\draw [black,thick,-{Stealth[width=5pt,length=5pt]}] (26.2,-29.28) -- (32.2,-29.22);
%\draw [black,thick,-{Stealth[width=5pt,length=5pt]}] (38.2,-29.2) -- (44.4,-29.2);
%\draw [black,thick,-{Stealth[width=5pt,length=5pt]}] (50.4,-29.2) -- (57,-29.2);
%\end{tikzpicture}

\begin{tikzpicture}[->, >=stealth, node distance=1.5cm, thick]
    % Define nodes
    \node[circle, draw, thick, scale=0.7] (X1) at (0,0) {$X_1$};
    \node[circle, draw, thick, scale=0.7] (X2) at (1.5,0) {$X_2$};
    \node[circle, draw, thick, scale=0.7] (dots) at (3,0) {$\cdots$};
    \node[circle, draw, thick, scale=0.7] (Xn) at (4.5,0) {$X_n$};
    
    % Draw arrows
    \draw[->,thick] (X1) -- (X2);
    \draw[->,thick] (X2) -- (dots);
    \draw[->,thick] (dots) -- (Xn);
\end{tikzpicture}

    \caption{A chain-structured Bayesian network with $n$ variables.}
    \label{fig:basic_chain}
\end{figure}
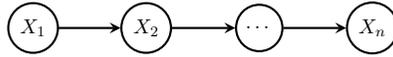

Suppose for simplicity that we are given a Bayesian network that resembles Figure \ref{fig:basic_chain}. This network represents a joint probability distribution of the form
\[
p(x_1, \dotsc, x_n) = p(x_1) \prod_{i=2}^n p(x_i \mid x_{i-1}). 
\]
Assume that we are interested in computing the marginal probability $p(x_n)$. The naive way of calculating this is to sum the probability over all $k^{n-1}$ assignments to $x_1, \dotsc, x_{n-1}$:
\[
p(x_n) = \sum_{x_1} \cdots \sum_{x_{n-1}} p(x_1, \dotsc, x_n). 
\]
However, we can do much better by leveraging the factorization of our joint probability distribution. This factorization allows us to rewrite the sum in a way that pushes certain variables deeper into the product.
\begin{align*}
p(x_n)
& = \sum_{x_1} \cdots \sum_{x_{n-1}} p(x_1) \prod_{i=2}^n p(x_i \mid x_{i-1}) \\
& = \sum_{x_{n-1}} p(x_n \mid x_{n-1}) \sum_{x_{n-2}} p(x_{n-1} \mid x_{n-2}) \cdots \sum_{x_1} p(x_2 \mid x_1) p(x_1).
\end{align*}
We sum the inner terms first, starting from $x_1$ and ending with $x_{n-1}$. Concretely, we start by computing an intermediary factor
\[
\tau(x_2) = \sum_{x_1} p(x_2 \mid x_1) p(x_1)
\]
by summing out $x_1$. This takes $O(k^2)$ time because we must sum over $x_1$ for each assignment to $x_2$. The resulting factor $\tau(x_2)$ can be thought of as a table of $k$ values (though not necessarily probabilities), with one entry for each assignment to $x_2$ (just as factor $p(x_1)$ can be represented as a table). We can then rewrite the marginal probability using $\tau$ as
\[
p(x_n) = \sum_{x_{n-1}} p(x_n \mid x_{n-1}) \sum_{x_{n-2}} p(x_{n-1} \mid x_{n-2}) \cdots \sum_{x_2} p(x_3 \mid x_2) \tau(x_2).
\]
Note that this has the same form as the initial expression, except that we are summing over one fewer variable. Thus, we are sequentially \textit{eliminating} one variable at a time, which gives the algorithm its name.

%{% include sidenote.html id=``note-dp'' note=``This technique is a special case of \textit{dynamic programming}, a general algorithm design approach in which we break apart a larger problem into a sequence of smaller ones.'' %}. 

We can therefore compute another factor
\[
\tau(x_3) = \sum_{x_2} p(x_3 \mid x_2) \tau(x_2),
\]
and repeat the process until we are only left with $x_n$, which yields a table of probabilities representing $p(x_n)$. Since each step takes $O(k^2)$ time, and we perform $O(n)$ steps, inference now takes $O(n k^2)$ time, which is much better than our naive $O(k^n)$ solution.

%\paragraph{Potentials} 

%\jm{Expand this section.}
%\jm{Cite Abeel course notes.}

\subsection{The Sum-Product Variable Elimination Algorithm}

While the previous example illustrated marginal inference in a chain-structured graph, we now present variable elimination as a general procedure applicable to arbitrary graphical models with factorized representations --- including both Bayesian networks and MRFs. Later, we show how, in addition to marginal inference, this procedure can be used for MAP inference and computing conditional distributions.
% We then write out the full procedure formally in Algorithm~\ref{alg:variable_elimination}.

%\jm{Improve this explanation.}
% \vk{maybe start here? i thought the intro was a bit abstract; but need to make sure notation is defined}
We first focus here on the \textit{sum-product formulation} of variable elimination. %Conveniently, we can interchange sum and product operations. %For example, consider three factors $\phi_x$ with scope $\mathbf{X}$, and $\phi_y$ with scope $\mathbf{Y}$, and $\phi_{yz}$ with scope $\{\mathbf{Y}, \mathbf{Z}\}$. Because $\mathbf{Y}$ is in scope for $\phi_y$ and $\phi_{yz}$ only, we can interchange sum and product as follows:
%\[
%\sum_{\mathbf{y}} \phi_x \phi_y \phi_{yz} = \phi_x \sum_{\mathbf{y}} \phi_y \phi_{yz}.
%\]
This procedure repeatedly performs two factor operations: \textit{product} and \textit{marginalization}. Given two factors $\phi_1$ and $\phi_2$, the factor product operation simply defines the product $\phi_3 := \phi_1 \times \phi_2$ as
\[
\phi_3(\mathbf{x}) = \phi_1(\mathbf{x}^{(1)}) \times \phi_2(\mathbf{x}^{(2)}). 
\]

The scope of $\phi_3$ is defined as the union of the variables in the scopes of $\phi_1$ and $\phi_2$. We use $\mathbf{x}^{(i)}$ to denote an assignment to the variables in the scope of $\phi_i$ defined by the restriction of $\mathbf{x}$ to that scope. For example, we define $\phi_3(a,b,c) := \phi_1(a,b) \times \phi_2(b,c)$.
% $\phi_3(\mathbf{a},\mathbf{b},\mathbf{c}) := \phi_1(\mathbf{a},\mathbf{b}) \times \phi_2(\mathbf{b},\mathbf{c})$

Next, the marginalization operation ``locally'' eliminates a set of variables from a factor. If we have a factor $\phi(\mathbf{x},\mathbf{y})$ over two sets of variables $\mathbf{X}$ and $\mathbf{Y}$, marginalizing $\mathbf{Y}$ produces a new factor
\[
\tau(\mathbf{x}) = \sum_\mathbf{y} \phi(\mathbf{x},\mathbf{y}), 
\]
where the sum is over all joint assignments to the set of variables $\mathbf{Y}$. We use $\tau$ to refer to the marginalized factor. It is important to understand that this factor does not necessarily correspond to a probability distribution, even if $\phi$ was a conditional probability distribution.

% Given these two operations, the full sum-product variable elimination algorithm proceeds as follows.
% Suppose that we have a graphical model $\mathcal{G} = (\mathbf{V}, \mathbf{E})$, with an associated set of $d$ factors $\Phi \coloneqq \{\phi_i\}_{i=1}^d$, and we want to perform marginal inference on query variables $\mathbf{X}_q \in \mathbf{V}$, i.e., we want to compute $p(\mathbf{x_q})$.
% We therefore aim to marginalize out the variables $\mathbf{X}_m := \mathbf{V} \setminus \mathbf{X}_q$. 
% Suppose also that we choose an ordering of the variables $\mathbf{X}_m$, referred to as an \textit{elimination ordering} (described in Section~\ref{sec:elimination_orderings}), one example being the topological ordering of a Bayesian network.
% Given this ordering, we iterate over each of the variables in $\mathbf{X}_m$, and perform a product factor operation and then marginalization factor operation.
% Each step produces a new factor (with an updated scope), which we can use to replace a subset of factors in $\Phi$; this sequentially reduces the size of $\Phi$, as it \textit{eliminates} (marginalizes) each variable in $\mathbf{X}_m$.
% Intuitively, this corresponds to choosing a sum and pushing it as far as possible inside the product of the factors, as we did in the chain example.
% We show the pseudocode for this sum-product variable elimination algorithm in Algorithm \ref{alg:variable_elimination}.

Given the definitions of the product and marginalization operations over factors, the sum-product variable elimination algorithm proceeds as follows.
Suppose we are given a graphical model $\mathcal{G} = (\mathbf{V}, \mathbf{E})$ with an associated set of $d$ factors $\Phi \coloneqq \{\phi_i\}_{i=1}^d$, and we wish to compute the marginal distribution over a subset of query variables $\mathbf{X}_q \subseteq \mathbf{V}$, i.e., we aim to compute $p(\mathbf{x}_q)$.
To do this, we must eliminate the remaining variables $\mathbf{X}_m \coloneqq \mathbf{V} \setminus \mathbf{X}_q$ by summing them out.
We begin by selecting an \textit{elimination ordering} over the variables in $\mathbf{X}_m$ (see Section~\ref{sec:elimination_orderings}); for instance, in a Bayesian network, a topological ordering is one possible choice.
Then, for each variable in the ordering, we identify the subset of factors in $\Phi$ that mention the variable, and on these we perform the product factor operation and then the marginalization factor operation (which sums out the given variable).
The result is a new factor that replaces this subset of factors in $\Phi$, and the process continues.
This sequentially reduces the number of factors and eliminates variables one by one.
Conceptually, the algorithm pushes summations as far inside the product of factors as possible, as illustrated in the chain-structured example.
At the end of the procedure, a single factor remains, defined over the query variables $\mathbf{X}_q$, which represents the unnormalized marginal distribution $p(\mathbf{x}_q)$.
The full pseudocode for the sum-product variable elimination algorithm is provided in Algorithm~\ref{alg:variable_elimination}.

\begin{algorithm}[!t]
\caption{\textit{Sum-Product Variable Elimination}, adapted from \citet{zhang1996exploiting} and \citet{koller2009probabilistic}} \label{alg:variable_elimination}
\begin{algorithmic}[1]

\INPUT{~} \\ 
\textbullet~$\Phi \coloneqq \{\phi_i\}_{i=1}^d$: Set of factors implied by graph $\mathcal{G} = (\mathbf{V}, \mathbf{E})$ \\
\textbullet~$\mathbf{X}_q \in \mathbf{V}$: Set of query variables \\
\textbullet~$\mathcal{O}$: Elimination ordering for $\mathbf{V} \setminus \mathbf{X}_q$ \\
\OUTPUT{$\phi^* = p(\mathbf{x}_q)$}
\vspace{2mm}
\FOR{$X \in \mathcal{O}$, following the elimination ordering}
    % \STATE{$\mathcal{O} \gets \mathcal{O} \setminus \{X\}$}
    \STATE{Remove variable $X$ from $\mathcal{O}$}
    \STATE{Remove from $\Phi$ all $\{\phi_j\}_{j=1}^k$ where $X \in Scope[\phi_j]$}
    \STATE{Add new factor $\sum_{x} \prod_{j=1}^k \phi_j$ to $\Phi$}
\ENDFOR
\STATE{$\phi^* \gets \prod_{i=1}^d \phi_i$}
\RETURN{$\phi^*$}
\end{algorithmic}
\end{algorithm}

\subsubsection{Two Illustrative Examples}

To make Algorithm~\ref{alg:variable_elimination} concrete, we revisit the chain-structured example from Figure~\ref{fig:basic_chain} to illustrate how these steps play out.
% To make Algorithm~\ref{alg:variable_elimination} concrete, we can return to our running example of the chain in Figure \ref{fig:basic_chain} to illustrate how these steps play out.
In this case, the chosen ordering was simply the topological ordering $x_1, x_2, \dotsc, x_{n-1}$. Starting with $x_1$, we collected all the factors involving $x_1$, which were $p(x_1)$ and $p(x_2 \mid x_1)$. We then used them to construct a new factor
\[
\tau(x_2) = \sum_{x_1} p(x_2 \mid x_1) p(x_1).
\]
This can be seen as the results of Steps 1 and 2 of variable elimination in Algorithm \ref{alg:variable_elimination}: first we form a large factor $p(x_2, x_1) = p(x_2 \mid x_1) p(x_1)$, then we eliminate $x_1$ from that factor to produce $\tau$. Next, we repeat the same procedure for $x_2$, except that the factors are now $p(x_3 \mid x_2), \tau(x_2)$.

As a second illustrative example of the variable elimination algorithm, we walk through a slightly more complex application of Algorithm~\ref{alg:variable_elimination}.
Recall the graphical model of student performance that we introduced earlier (Figure \ref{fig:student_grades_dag}).
The probability specified by the model is of the form
\[
p(l, g, i, d, s) = p(l \mid g) p(s \mid i) p(i) p(g \mid i, d) p(d). 
\]

Let's suppose that we are computing $p(l)$ and are eliminating variables in their topological ordering in the graph. First, we eliminate $d$, which corresponds to creating a new factor
\[
\tau_1(g,i) = \sum_d p(g \mid i, d) p(d).
\]
Next, we eliminate $i$ to produce a factor
\[
\tau_2(g,s) = \sum_i \tau_1(g,i) p(i) p(s \mid i).
\]
Then, we eliminate $s$ yielding $\tau_3(g) = \sum_s \tau_2(g,s)$, and so on. Note that these operations are equivalent to summing out the factored probability distribution as follows:

\[
p(l) = \sum_g p(l \mid g) \sum_s \sum_i p(s\mid i) p(i) \sum_d p(g \mid i, d) p(d).
\]

This example requires computing at most $k^3$ operations per step: each factor is at most over 2 variables, and one variable is summed out at each step (the dimensionality $k$ in this example is either 2 or 3).

\subsubsection{Computational Complexity}

Though variable elimination is often faster than the naive approach of inference by enumeration, it is still NP-hard.  In short, the computational complexity of variable elimination is exponential in the size of the largest factor. However, it is very important to understand that the running time of variable elimination depends heavily on the structure of the graph. More concretely, the running time depends on the following structural properties of the Bayesian network or MRF $\mathcal{G}$. 

\begin{itemize}
    \item \textit{Total variables in $\mathcal{G}$}. Let $n = |\mathbf{V}|$ denote the number of variables in the system.
    \item \textit{Treewidth}. Treewidth $w$, which measures how close the graph is to a tree (Definition~\ref{def:treewidth}). In this setting, the treewidth can be shown to be equivalent to the maximum size of any factor $\tau$ formed during the variable elimination algorithm.
    \item \textit{Domain size of variables in $\mathbf{V}$}. Let $k_i$ denote the size of the domain (i.e., the number of possible values) for each $V_i \in \mathbf{V}$. For ease of exposition, we can assume that all variables have $k$ states.
\end{itemize}

Thus, we can express the time complexity of variable elimination as $O(nk^{w+1})$. As $w = 1$ in trees (including chains), the computational complexity of variable elimination on tree-structured graphs is linear in $n$: $O(nk^2)$. For graphs with large treewidth, variable elimination can become intractable for large $n$.

% \vk{student example is very far now, not sure if we can use it here}

% Returning to our previous example of a graph depicting student performance (Figure \ref{fig:student_grades_dag}), suppose we had eliminated $g$ first. Then, we would have had to transform the factors $p ( g \mid i , d )$ , $\phi ( l \mid g )$ into a big factor $\tau ( d , i , l )$ over 3 variables, which would require $O( k^4 )$ time to compute: $k$ times for each of three conditional variables, and $k$ times for each value of $g$. If we had a factor corresponding to adjacency $S \to G$, then we would have had to eliminate $p ( g \mid s )$ as well, producing a single giant factor $\tau( d , i , l , s )$ in $O( k^5 )$ time. Then, eliminating any variable from this factor would require almost as much work as if we had started with the original distribution, since all the variables have become coupled. Clearly some orderings are more efficient than others.
% %In fact, the running time of variable elimination is $O ( n k^{w + 1} )$ , where $w$ is the maximum size of any factor $\tau$ formed during the elimination process and $n$ is the number of variables.

\subsection{Introducing Evidence}

Above we have described the variable elimination algorithm for the case of marginal inference. A closely related and equally important problem is the computation of conditional probabilities of the form
\[
p(\mathbf{x}_q \mid \mathbf{x}_e = \mathbf{e}) = \frac{p(\mathbf{x}_q, \mathbf{x}_e = \mathbf{e})}{p(\mathbf{x}_e = \mathbf{e})},
\]
where $p(\mathbf{x}_q,\mathbf{x}_e,\mathbf{x}_m)$ is a probability distribution over \textit{query} variables $\mathbf{X}_q$, \textit{observed evidence} variables $\mathbf{X}_e$, and \textit{unobserved} variables $\mathbf{X}_m$.

We can compute this probability by performing variable elimination once on $p(\mathbf{x}_q, \mathbf{x}_e = \mathbf{e})$ and then once more on $p(\mathbf{x}_e = \mathbf{e})$. To compute $p(\mathbf{x}_q, \mathbf{x}_e = \mathbf{e})$, we simply take every factor $\phi(\mathbf{x}'_m, \mathbf{x}'_q, \mathbf{x}'_e)$ which has scope over variables $\mathbf{X}'_e \subseteq \mathbf{X}_e$, and we set their values to $\mathbf{e}$. Then we perform standard variable elimination to obtain a factor over $\mathbf{X}_q$ only.

\subsection{Elimination Orderings}
\label{sec:elimination_orderings}

The variable elimination algorithm requires an \textit{ordering} over the variables, according to which variables will be eliminated. In our chain example, we simply took the ordering implied by the DAG. However, it is important to note that the time complexity of variable elimination is sensitive to the order in which variables are eliminated. Further, it is NP-hard to find the optimal ordering.

In practice, we can resort to the following heuristics to obtain an ordering for variable elimination. These methods often result in reasonably good performance in many interesting settings~\citep{koller2009probabilistic}.

\begin{itemize}
    \item \textit{Min-neighbors}, where we choose the next variable with the fewest dependent variables.
    \item \textit{Min-weight}, where we choose variables to minimize the product of the cardinalities of dependent variables.
    \item \textit{Min-fill}, where we choose vertices to minimize the size of the factor that will be added to the graph.
\end{itemize}

\subsection{Generalizing Factor Operations}
\label{sec:generalizing_factor_operations}

We have so far described the sum-product formulation of variable elimination for marginal (and conditional) inference, involving the \textit{product} and \textit{marginalization} factor operators.
We now generalize these operations, enabling alternative inference procedures with the variable elimination algorithm, including MAP inference.

The premise of variable elimination rests on the properties of \textit{commutative semirings}: a set $\mathbf{S}$ and two binary operations that are associative, commutative, and satisfy identity laws. Commutative semirings satisfy the distributive law, whereby 
\begin{align}
    (x \times y) + (x \times z) = x \times (y + z) \label{eq:distribution}
\end{align}
for all $x,y,z \in \mathbf{S}$. Note that the distributive law reduced the number of computations in Equation \ref{eq:distribution} from three (on the lefthand side) to two (on the righthand side). See \citet{golan2013semirings} for a formal discussion of commutative semirings.

\begin{table}[!t]
\centering
\begin{tabular}{llll}
\toprule
 & \textsc{Domain} & \textbf{+} & \textbf{$\times$} \\
\midrule
\textit{Sum-product} & $[0, \infty)$ & $(+, 0)$ & $(\times, 1)$ \\
\textit{Max-product} & $[0, \infty)$ & $(\max, 0)$ & $(\times, 1)$ \\
\textit{Min-sum}  & $(-\infty, \infty]$ & $(\min, \infty)$ & $(+, 0)$ \\
\textit{Boolean satisfiability} & $\{T, F\}$ & $(\vee, F)$ & $(\wedge, T)$  \\
\bottomrule
\end{tabular}
\caption{Examples of common commutative semirings. Table from \citet{murphy2012machine}.}
\label{tab:semirings}
\end{table}

The associative, commutative, and distributive properties of these binary operations allow us to perform more efficient inference. When the binary operations of a commutative semiring are taken to be addition and multiplication, we obtain the sum-product formulation of variable elimination:
\[
p(\mathbf{x}_q \mid \mathbf{x}_e) \propto \sum_{\mathbf{x}} \prod_{c \in \mathbf{C}} \phi_c (\mathbf{x}_c),
\]
where $\mathbf{X}_q$ is a set of query variables, $\mathbf{C}$ is a set of cliques $c$, and $\mathbf{X}_e$ is a set of evidence variables whose values are fixed rather than marginalized over \citep{murphy2012machine}.
Notably, replacing summation with the \textit{max} operation in variable elimination yields a version of the algorithm for MAP inference.
% We can also replace addition with the max operation to carry out MAP inference with the variable elimination algorithm.
See Table \ref{tab:semirings} for additional commutative semirings.

\subsection{Limiting Extraneous Computation}

We have seen how the variable elimination algorithm can answer marginal queries of the form $p(\mathbf{x}_q \mid \mathbf{x}_e = \mathbf{e})$ for both directed and undirected networks. However, this algorithm has an important shortcoming: if we want to ask the model for another query (e.g., $p(\mathbf{x}'_q \mid \mathbf{x}'_e = \mathbf{e}')$), we need to restart the algorithm from scratch. This is very wasteful and computationally burdensome.

Fortunately, it turns out that this problem is also easily avoidable. When computing marginals, variable elimination produces many intermediate factors $\tau$ as a side-product of the main computation. These factors turn out to be the same as the ones that we need to answer other marginal queries. By caching them after a first run of variable elimination, we can easily answer new marginal queries at essentially no additional cost.

\section{Belief Propagation in Trees}
\label{sec:belief_propagation}

We now introduce \textit{belief propagation}: a family of exact inference algorithms based on recursive \textit{message passing}. Belief propagation was first introduced by Judea Pearl \citep{pearl1982,pearl1986fusion}, but the more general concept of message passing takes a wide range of important algorithms as special cases: the forward/backward algorithm, the Viterbi algorithm, turbo decoding, the Kalman filter, and more \citep{kschischang2001factor, yedidia2003understanding, mceliece1998turbo}. Belief propagation performs exact inference over tree-structured graphs, both undirected and directed (Definitions \ref{def:undirected_tree}, \ref{def:directed_tree}). In Section \ref{sec:junction_tree}, we will introduce the junction tree algorithm: an extension of belief propagation that works for general networks by transforming the input graph into a tree whose nodes represent cliques. 

Belief propagation first executes two runs of the variable elimination algorithm to initialize a data structure that holds a set of pre-computed factors. Once the structure is initialized, it can answer marginal queries in $O(1)$ time. You can think of variable elimination and belief propagation as two flavors of the same technique: top-down dynamic programming versus bottom-up table filling, respectively. Just like in computing the $n^{th}$ Fibonacci number $F_n$, top-down dynamic programming (i.e., variable elimination) computes \textit{just} that number, but bottom-up (i.e., belief propagation) will create a filled table of all $F_i$ for $i \leq n$. Moreover, the two-pass nature of belief propagation is a result of the underlying dynamic programming on bi-directional trees, while the Fibonacci numbers' relation is a uni-directional tree. %\jm{Changed reference to junction trees to BP instead, since we discuss this first.}

In parallel with the sum-product and max-product variants of variable elimination, we now introduce two corresponding forms of belief propagation for tree-structured graphs:
\begin{enumerate}
    \item The \textit{sum-product algorithm} for efficient marginal inference. That is, computing marginals of the form $p(x_i)$. 
    \item The \textit{max-product algorithm} for MAP inference. That is, computing queries of the form $\max_{x_1, \dotsc, x_n} p(x_1, \dotsc, x_n)$.
\end{enumerate}

\subsubsection{Variable Elimination as Message Passing}

% \vk{this is pretty nice, but i think it should be consolidated with the previous discussion; some version of this could be the explanation of the pseudocode}
% \vk{also, a picture would better explain post-order}
% \vk{by the way, i would put graph cuts last in this section, the current material is much more important}

First, consider what happens if we run the variable elimination algorithm on a tree in order to compute a marginal $p(x_i)$. We can find an optimal ordering for this problem by rooting the tree at $x_i$ and iterating through the nodes in \textit{post-order} --- where a post-order traversal of a rooted tree is one that starts from the leaves and goes up the tree, such that each node is always visited after all of its children. The root is visited last.

\begin{figure}
    \centering

\begin{tikzpicture}[every edge quotes/.style = {font=\footnotesize, fill=white,sloped}]
  \node[circle,CornflowerBlue,fill=CornflowerBlue!10,thick,draw,scale=1] (w) at (-4, 2) {\color{black}$W$};
  \node[circle,Blue,fill=Blue!10,thick,draw,scale=1] (x) at (4, 2) {\color{black}$X$};
  \node[circle,RubineRed,fill=RubineRed!10,thick,draw,scale=1] (y) at (0, 2) {\color{black}$Y$};
  \node[circle,YellowGreen,fill=YellowGreen!10,thick,draw,scale=1] (z) at (0, -1) {\color{black}$Z$};
  %%%
  \draw[thick,black,-]  (x) edge[] (y);
  \draw[thick,black,-]  (w) edge[] (y);
  \draw[thick,black,-]  (y) edge[] (z);
  %%%
  \node[] (zy_up) at (-0.25,0.25) {\color{RubineRed}$\bigg\uparrow$};
  \node[] (zy_down) at (0.25,0.25) {\color{YellowGreen}$\bigg\downarrow$};
  \node[] (wy_left) at (-2,1.75) {\color{CornflowerBlue}$\longleftarrow$};
  \node[] (wy_right) at (-2,2.25) {\color{RubineRed}$\longrightarrow$};
  \node[] (xy_left) at (2,1.75) {\color{RubineRed}$\longleftarrow$};
  \node[] (xy_right) at (2,2.25) {\color{Blue}$\longrightarrow$};
  %%%
  \node[rotate=90] (m_zy_up) at (-0.75,0.25) {\color{RubineRed}$m_{z \to y}(y)$};
  \node[rotate=270] (m_zy_down) at (0.75,0.25) {\color{YellowGreen}$m_{y \to z}(z)$};
  \node[] (m_wy_left) at (-2,1.25) {\color{CornflowerBlue}$m_{y \to w}(w)$};
  \node[] (m_wy_right) at (-2,2.75) {\color{RubineRed}$m_{w \to y}(y)$};
  \node[] (m_xy_left) at (2,1.25) {\color{RubineRed}$m_{x \to y}(y)$};
  \node[] (m_xy_right) at (2,2.75) {\color{Blue}$m_{y \to x}(x)$};
  %%%
  %\node[align=flush center,text width=3cm] (label) at (1,-1.25) {\footnotesize\textbf{(d)} \textsc{cascade} $p(x)p(z|x)p(y|z)$}; 
\end{tikzpicture}

    \caption{Message passing for belief propagation in an MRF. Each node pair shares messages in both direction. A message $m_{x \to y}(y)$ goes from variable $X$ to variable $Y$, passing information about the state of $Y$. Messages are colored in correspondence with the recipient node.}
    \label{fig:message_passing}
\end{figure}
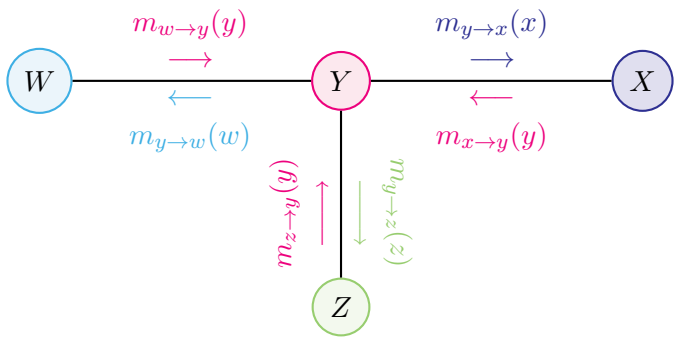

This ordering is optimal because the largest clique formed during variable elimination will be of size two. At each step, we will eliminate a node $x_j$. This will involve computing the factor
\[
\tau_k(x_k) = \sum_{x_j} \phi(x_k, x_j) \tau_j(x_j),
\]
where $x_k$ is the parent of $x_j$ in the tree. At a later step, $x_k$ will be eliminated, and $\tau_k(x_k)$ will be passed up the tree to the parent $x_l$ of $x_k$ in order to be multiplied by the factor $\phi(x_l, x_k)$, before being marginalized out. The factor $\tau_j(x_j)$ can be thought of as a \textit{message} that $x_j$ sends to $x_k$, which summarizes all of the information from the subtree rooted at $x_j$. We can visualize this transfer of information using arrows on a tree (e.g., Figure \ref{fig:message_passing}).

At the end of variable elimination, $x_i$ receives messages from all of its immediate children, marginalizes them out, and we obtain the final marginal. Now suppose that after computing $p(x_i)$, we want to compute $p(x_k)$ as well. We would again run variable elimination with $x_k$ as the root, waiting until $x_k$ receives all messages from its children. The key insight: the messages $x_k$ received from $x_j$ now will be the same as those received when $x_i$ was the root. Further intuition for why this is true lies in the fact that there is only a single path connecting any two nodes in the tree. Thus, if we store the intermediary messages of the variable elimination algorithm, we can quickly compute other marginals as well.

\subsubsection{A Message Passing Algorithm}

A key question here is, \textit{how exactly do we produce all the messages that we need to compute all marginal distributions?}
% Notice, for example, that the messages to $x_k$ from the side of $x_i$ will need to be recomputed.

The answer is very simple: a node $x_i$ sends a message to a neighbor $x_j$ whenever it has received messages from all nodes besides $x_j$. It's a fun exercise to the reader to show that in a tree, there will always be a node with a message to send, unless all the messages have been sent out. This will happen after precisely $2 \vert \mathbf{E} \vert$ steps, since each edge can receive messages only twice: once from $x_i \to x_j$, and once more in the opposite direction. Finally, this algorithm will be correct because our messages are defined as the intermediate factors in the variable elimination algorithm.

\subsection{Sum-Product Message Passing}

We are now ready to formally define the belief propagation algorithm. We begin with the sum-product formulation.

The sum-product message passing algorithm is defined as follows: while there is a node $x_i$ ready to transmit to $x_j$, send the message
\[
m_{i\to j}(x_j) = \sum_{x_i} \phi(x_i) \phi(x_i,x_j) \prod_{x_\ell \in \mathbf{ne}(x_i) \setminus x_j} m_{\ell \to i}(x_i).
\]
The notation $\mathbf{ne}(x_i) \setminus x_j$ refers to the set of nodes that are neighbors of $x_i$, excluding $x_j$. Again, observe that this message is precisely the factor $\tau$ that $x_i$ would transmit to $x_j$ during a round of variable elimination with the goal of computing $p(x_j)$.

Because of this observation, after we have computed all messages, we can answer any marginal query over $x_i$ in constant time using the equation
\[
p(x_i) \propto \phi(x_i) \prod_{x_\ell \in \mathbf{ne}(x_i)} m_{\ell \to i}(x_i). 
\]

\subsubsection{Message Passing for Factor Trees}
In addition to tree graphs, message passing can also be applied to factor graphs (Section~\ref{sec:factor_graph}), specifically in the case where the factor graph forms a tree (referred to as a \textit{factor tree}), using a slight modification of the above procedure. Recall that a factor graph is a bipartite graph with edges going between factors and the variables in their scope. On factor graphs, we have two types of messages: variable-to-factor messages $\nu$ and factor-to-variable messages $\mu$. Both messages require taking a product, but only the factor-to-variable messages $\mu$ require a sum:

\begin{align*}
    \nu_{x_i \to f_s}(x_i) &= \prod_{f_t\in \mathbf{ne}(x_i) \setminus f_s}\mu_{f_t \to x_i}(x_i) \\
    \mu_{f_s \to x_i}(x_i) &= \sum_{\mathbf{ne}(f_s) \setminus x_i} f_s(\mathbf{ne}(f_s)) \prod_{x_j\in \mathbf{ne}(f_s)\setminus x_i}\nu_{x_j\to f_s}(x_j)
\end{align*}

The algorithm proceeds in the same way as with undirected graphs: as long as there is a factor (or variable) ready to transmit to a variable (or factor), send the appropriate factor-to-variable (or variable-to-factor) message as defined above.

%\jm{Get figure of factor graph for BP.}

%\input{algorithms/algorithm_belief_propagation}

\subsection{Max-Product Message Passing}

We now introduce a second variant of the belief propagation algorithm: \textit{max-product message passing}, which can be used to perform MAP inference queries of the form 
\begin{align*}
p^* = \max_{x_1, \dotsc, x_n} p(x_1, \dotsc, x_n)
\hspace{4mm}\text{or}\hspace{4mm}
\hat{\mathbf{x}} = \argmax_{x_1, \dotsc, x_n} p(x_1, \dotsc, x_n),
\end{align*}
where the first expression returns the maximum value of the PDF, and the second expression instead returns the arguments that achieve this maximal value (or, equivalently, returns the MAP point estimate).

The framework that we have introduced for marginal inference allows us easily perform MAP inference as well. The key observation is that the sum and max operators both distribute over products. In general, the max operator only distributes over products of non-negative factors. By definition, however, MRF factors are non-negative. Thus, by replacing sums in marginal inference with maxes, and following the prior belief propagation algorithm, we are able to solve the MAP inference problem.

As an example, in a chain-structured MRF, we can compute the partition function (a marginal inference query) as follows:
\begin{align*}
Z
&= \sum_{x_1} \cdots \sum_{x_n} \phi(x_1) \prod_{i=2}^n \phi(x_i, x_{i-1}) \\
&= \sum_{x_n} \sum_{x_{n-1}} \phi(x_n, x_{n-1}) \sum_{x_{n-2}} \phi(x_{n-1}, x_{n-2}) \cdots \sum_{x_1} \phi(x_2 , x_1) \phi(x_1).
\end{align*}

Instead, to compute the maximum value $\Tilde{p}^*$ of $\Tilde{p}(x_1, \dotsc, x_n)$ (a MAP inference query), we simply replace sums with maxes:
\begin{align*}
\Tilde{p}^*
&= \max_{x_1} \cdots \max_{x_n} \phi(x_1) \prod_{i=2}^n \phi(x_i, x_{i-1}) \\
&= \max_{x_n} \max_{x_{n-1}} \phi(x_n, x_{n-1}) \max_{x_{n-2}} \phi(x_{n-1}, x_{n-2}) \cdots \max_{x_1} \phi(x_2 , x_1) \phi(x_1).
\end{align*}
\noindent Since both problems decompose in the same way, we can reuse all of the machinery developed for marginal inference and apply it directly to MAP inference. Note that this also applies to factor trees.

There is a small caveat: often, we want not only the maximum value of a distribution (i.e., $\max_x p(x)$), but also its most probable assignment (i.e., $\argmax_x p(x)$). This problem can be easily solved by keeping \textit{back-pointers} during the optimization procedure. A back-pointer is a reference to the argument (or input value) that led to the optimal value at a given step. In the above example, we would keep a back-pointer to the best assignment to $x_1$ given each assignment to $x_2$, a pointer to the best assignment to $x_2$ given each assignment to $x_3,$ and so on.

\subsubsection{Computational Complexity} 

An incredible property of belief propagation on trees is that exact marginals can be computed in a number of operations that scales linearly with respect to the number of nodes in the graph. As we will see in Section \ref{sec:junction_tree}, this remarkable efficiency does not hold when the input graph is no longer tree-structured.

\section{The Junction Tree Algorithm}
\label{sec:junction_tree}

In our discussion on belief propagation, we assumed that the graph describing our distribution of interest is a tree. What if that is not the case? Then, inference will not be tractable. Nevertheless, we are not without options: we can transform the original graph into a tree-like form, and then run message passing on this graph.

At a high-level, the junction tree algorithm partitions the graph into clusters of variables. Internally, the variables within a cluster could be highly coupled. However, interactions \textit{among} clusters will have a tree structure (i.e., a cluster will only be directly influenced by its neighbors in the tree). This leads to tractable global solutions if the local (cluster-level) problems can be solved exactly.

\subsubsection{An Illustrative Example}

% \vk{this was a bit hard for me to follow}
% \vk{overall the JT tree part could also use another review}

Before we define the full junction tree algorithm, we begin with an example. Suppose that we aim to perform marginal inference on an MRF of the form
\[
p(x_1, \dotsc, x_n) = \frac{1}{Z} \prod_{c \in \mathbf{C}} \phi_c(\mathbf{x}_c),
\]
where each factor is defined on a clique $\mathbf{x}_c$. Crucially, we assume that the cliques satisfy an assumption known as the \textit{running~intersection~property}. This means there exists an ordering of the cliques $\mathbf{x}_c^{(1)}, \dotsc, \mathbf{x}_c^{(n)}$ such that for any variable $x_i$, if $x_i \in \mathbf{x}_c^{(j)}$ and $x_i \in \mathbf{x}_c^{(k)}$, then $x_i$ must also appear in every clique $\mathbf{x}_c^{(\ell)}$ that lies on the path between $\mathbf{x}_c^{(j)}$ and $\mathbf{x}_c^{(k)}$.

%{% include maincolumn_img.html src='assets/img/junctionpath.png' caption='A chain MRF whose cliques are organized into a chain structure. Round nodes represent cliques and the variables in their scope; rectangular nodes indicate sepsets, which are variables forming the intersection of the scopes of two neighboring cliques.' %}

For instance, suppose that we are interested in computing the marginal probability $p(x_1)$ in a chain-structured MRF with five nodes, where each clique corresponds with a pair of nodes connected by an edge.
Note that this graph satisfies the running intersection property, as each variable is contained in (at most) two adjacent cliques.
% Each pairwise clique $(x_i, x_{i+1)}$ shares exactly one variable with its neighbor --- the variable $x_{i+1}$ with the next clique and $x_i$ with the previous one. This ensures that any variable appearing in multiple cliques appears in a contiguous sequence of cliques, forming a connected subtree — thus satisfying the running intersection property.
We can again use a form of variable elimination to push certain variables deeper into the product of cluster potentials:
\[
\phi(x_1) \sum_{x_2} \phi(x_1,x_2) \sum_{x_3} \phi(x_2,x_3) \sum_{x_4} \phi(x_3,x_4) \sum_{x_5} \phi(x_4,x_5).
\]
We first sum over $x_5$, which creates a factor
\[
\tau(x_3, x_4) = \phi(x_3,x_4) \sum_{x_5} \phi(x_4,x_5).
\]
Next, $x_4$ gets eliminated, and so on. At each step, each cluster marginalizes out the variables that are not in the scope of its neighbor. This marginalization can also be interpreted as computing a message over the variables that the cluster shares with the neighbor.

% Suppose that we are interested in computing the marginal probability $p(x_1)$ in the above example. Given our assumptions, we can again use a form of variable elimination to push certain variables deeper into the product of cluster potentials:
% \[
% \phi(x_1) \sum_{x_2} \phi(x_1,x_2) \sum_{x_3} \phi(x_1,x_2,x_3) \sum_{x_5} \phi(x_2,x_3,x_5) \sum_{x_6} \phi(x_2,x_5,x_6).
% \]
% We first sum over $x_6$, which creates a factor
% \[
% \tau(x_2, x_3, x_5) = \phi(x_2,x_3,x_5) \sum_{x_6} \phi(x_2,x_5,x_6).
% \]
% Next, $x_5$ gets eliminated, and so on. At each step, each cluster marginalizes out the variables that are not in the scope of its neighbor. This marginalization can also be interpreted as computing a message over the variables that the cluster shares with the neighbor.

The running intersection property is what enables us to push sums in all the way to the last factor. We can eliminate $x_5$ because we know that only the last cluster will carry this variable: since it is not present in the neighboring cluster, it cannot be anywhere else in the graph without violating the running intersection property.

\begin{figure}[!t]
    \centering
    \includegraphics[width=0.7\linewidth]{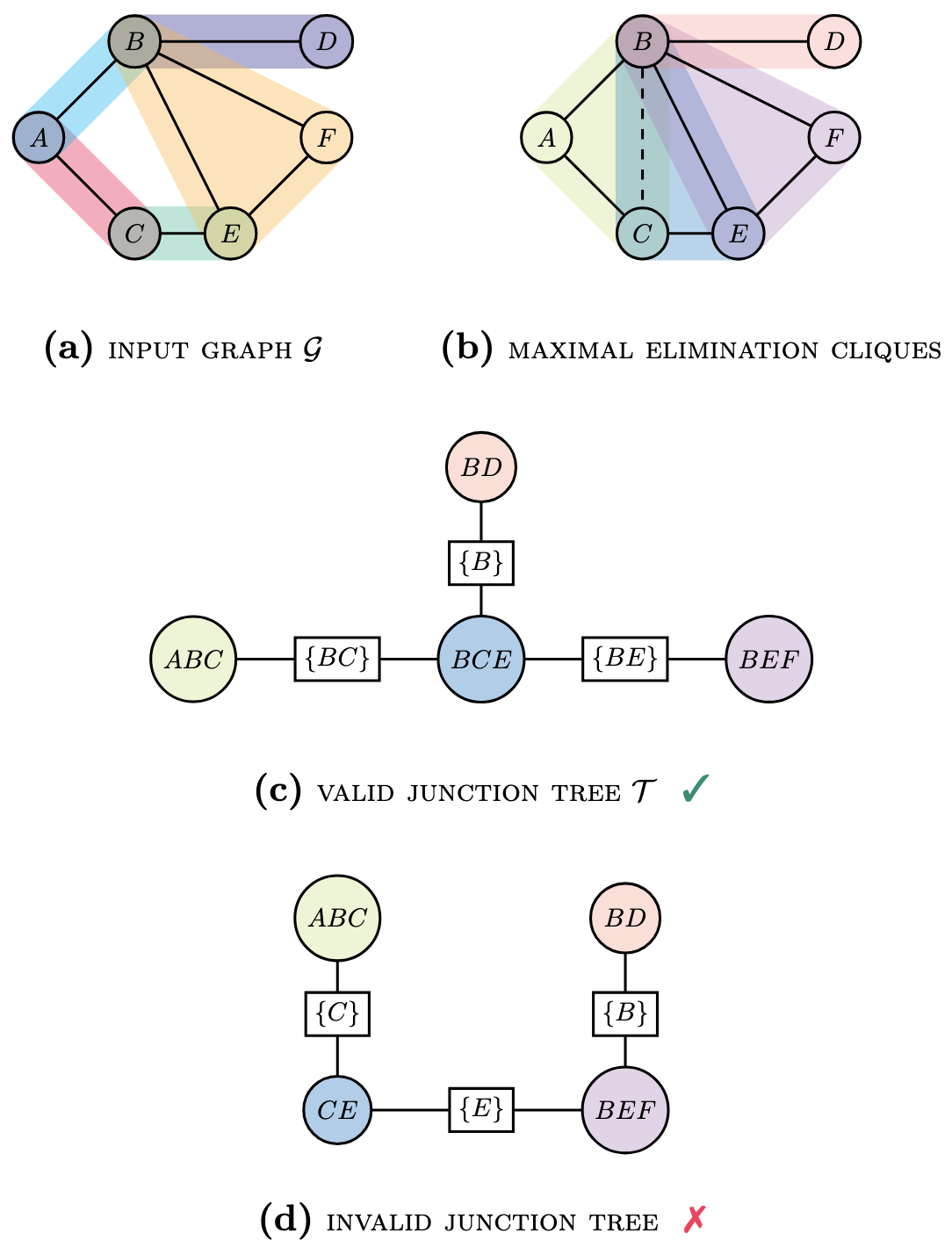}
    \caption{MRF with graph $\mathcal{G}$ and corresponding junction tree $\mathcal{T}$. In MRF \textbf{(a)},  potentials are denoted by colored highlights. In chordalized graph \textbf{(b)}, highlights now correspond to the maximal elimination cliques obtained by variable elimination. Graph \textbf{(c)} demonstrates a \textit{valid} junction tree for $\mathcal{G}$. Circular nodes in tree $\mathcal{T}$ denote maximal elimination cliques. Rectangular nodes in $\mathcal{T}$ denote \textit{sepsets} (i.e., ``separation sets''), which are sets of variables shared by neighboring clusters. Graph \textbf{(d)} represents an \textit{invalid} junction tree for $\mathcal{G}$, as it violates the running intersection property with respect to variable $B$. Adapted in part from notes by Mark Paskin. %\href{https://ai.stanford.edu/~paskin/gm-short-course/lec3.pdf}{https://ai.stanford.edu/~paskin/gm-short-course/lec3.pdf}
    }
    \label{fig:junction_tree_potentials_highlighted}
\end{figure}

\subsubsection{Junction Trees}

The core idea of the junction tree algorithm is to turn a graph into a tree of clusters that are amenable to the variable elimination algorithm (Figure \ref{fig:junction_tree_potentials_highlighted}). 
%(Figure \ref{fig:junction_tree_tree}). 
Then, we can simply perform message passing on this tree.

Suppose we have an undirected graphical model $\mathcal{G}$. If the model is directed, we consider the corresponding (chordalized) moral graph (Figure \ref{fig:asia}). A junction tree $\mathcal{T}=(\mathbf{C}, \mathbf{E}_\mathcal{T})$ over $\mathcal{G} = (\mathbf{X}, \mathbf{E}_\mathcal{G})$ is a tree whose nodes $c \in \mathbf{C}$ represent node clusters or subsets $\mathbf{x}_c \subseteq \mathbf{X}$. More specifically, these clusters are \textit{maximal  cliques} (Definition \ref{def:maximal_clique}).\footnote{We consider the moral graph when the initial graph is directed because moralization ensures that all variables which participate in a given factor belong to a single clique.} When we visualize junction trees, we often represent these clusters along with their \textit{sepsets} (i.e., separation sets), which are the variables forming the intersection of the scopes of two neighboring cliques. A valid junction tree must satisfy the following properties:
\begin{enumerate}
    \item \textit{Family preservation}. For each factor $\phi$, there is a cluster $c$ such that $\text{Scope}[\phi] \subseteq \mathbf{x}_c$.
    \item \textit{Running intersection}. For every pair of clusters $c^{(i)}, c^{(j)}$, every cluster on the path between $c^{(i)}, c^{(j)}$ contains $\mathbf{x}_c^{(i)} \cap \mathbf{x}_c^{(j)}$. This is sometimes referred to as the \textit{junction tree property}.
\end{enumerate}
Thus, we reach the following definition.

\begin{definition}[Junction tree]
\label{def:junction_tree}
    A graph $\mathcal{T} = (\mathbf{C}, \mathbf{E}_{\mathcal{T}})$ is a junction tree for graph $\mathcal{G} = (\mathbf{X}, \mathbf{E}_{\mathcal{G}})$ if and only if
    \begin{enumerate}
        \item $\mathcal{T}$ is tree-structured.
        \item $\mathbf{C}$ is the set of maximal cliques of $\mathcal{G}$.
        \item The running intersection property holds.
    \end{enumerate}
\end{definition}

Note that we can always find a trivial junction tree with one node containing all the variables in the original graph. However, such trees are not useful for inference because they will not result in efficient marginalization algorithms.

\subsubsection{Finding Good Junction Trees} 

Optimal junction trees are those that make the clusters as small and modular as possible. Unfortunately, it is NP-hard to find the optimal junction tree in the general case. A special case where we \textit{can} efficiently identify the optimal junction tree is when $\mathcal{G}$ is itself a tree. Then, we can define a cluster for each edge in the tree. It is not hard to check that the result satisfies the above definition. 

Although finding optimal junction trees is NP-hard, there are practical approaches for finding \textit{good} junction trees for the general case.
\begin{itemize}
    \item \textit{By hand}. Typically, our models will have a very regular structure for which there will be an obvious solution. For example, if our model is a grid, then the clusters will be associated with pairs of adjacent rows (or columns) in the grid.
    \item \textit{Using variable elimination}. The variable elimination algorithm can be used to identify the maximal cliques that will become the nodes in our junction tree. 
\end{itemize}

\begin{figure}[!t]
    \centering
    \includegraphics[width=0.9\linewidth]{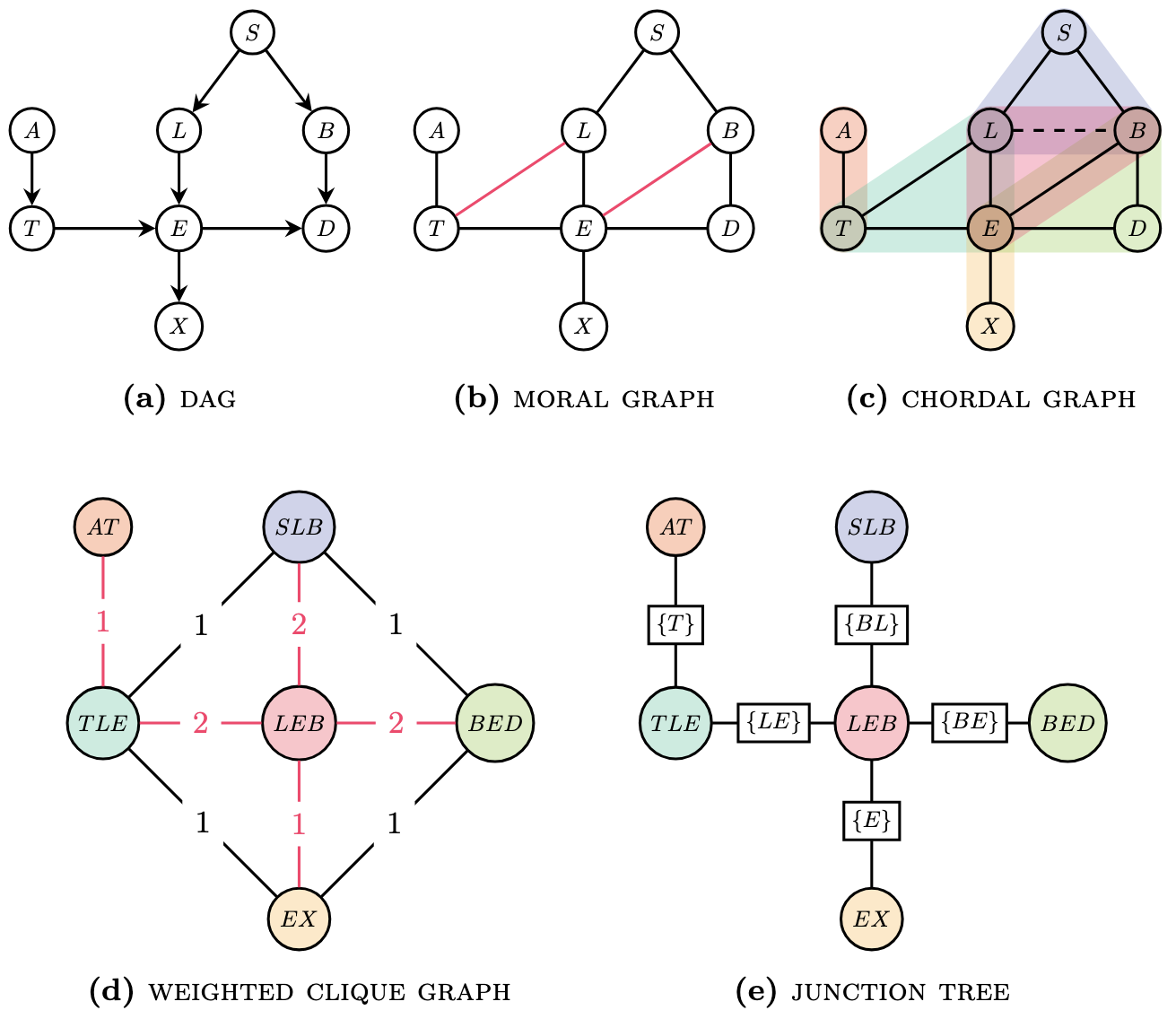}
    \caption{Transforming the \textsc{Asia} DAG \citep{lauritzen1988local} into a junction tree, per Algorithm \ref{alg:get_junction_tree}. We begin with the original DAG (\textbf{a}) and obtain the undirected moral graph (\textbf{b}). We then chordalize the moral graph to identify maximal  cliques (\textbf{c}; chord denoted by a dashed edge; cliques in highlights). We transform the chordal graph into a weighted clique graph, where each node represents a maximal elimination clique and edge weights are assigned according to the cardinality of the sepset (\textbf{d}). From this, we can obtain the maximum weight spanning tree (pink edges). Finally, we have our junction tree (\textbf{e}).}
    \label{fig:asia}
\end{figure}

We will elaborate on the latter approach, starting with the following propositions. Recall the notion of a \textit{chordal graph}, where the longest minimal cycle is a triangle (Definition \ref{def:chordal_graph}).

\begin{proposition}
\label{prop:chordal_junction}
    Any chordal graph has a corresponding junction tree. Furthermore, any graph with a junction tree must be chordal. %(\href{https://ocw.mit.edu/courses/6-438-algorithms-for-inference-fall-2014/a49d895264467b7e118bf0602616cc37_MIT6_438F14_Lec14.pdf}{cite}).
\end{proposition}

\begin{proposition}
\label{prop:chordal_n_cliques}
    Any chordal graph with $n$ nodes has at most $n$ maximal cliques. Further, the chordal graph with $n$ nodes and $n$ maximal cliques is the graph with no edges, and a connected chordal graph has at most $n-1$ maximal cliques.
\end{proposition}

For proof of Propositions \ref{prop:chordal_junction} and \ref{prop:chordal_n_cliques}, see Chapter 3 of \citet{vandenberghe2015chordal}. Following from these observations, we can see that it is possible to  identify the maximal cliques (and thus the junction tree) of a chordal graph in polynomial time. In fact, every maximal clique is also an \textit{elimination clique} for a given elimination ordering in variable elimination. Thus, we can run variable elimination on a chordal graph to efficiently discover the maximal cliques.

But what if our original graph is not chordal? Fortunately, running variable elimination on a non-chordal graph will \textit{chordalize} it along the way. After variable elimination, the reconstituted graph is guaranteed to be chordal --- and, therefore, guaranteed to have a corresponding junction tree. From there, finding a junction tree equates to finding a \textit{maximum weight spanning tree} over the complete cluster graph of the maximal elimination cliques (Figure \ref{fig:asia}). Maximum weight spanning trees can be efficiently identified using various algorithms (e.g., Kruskal’s algorithm \citep{kruskal1956shortest} or Prim’s algorithm \citep{prim1957shortest}; see \citealt{kleinberg2006algorithm}).

Note that multiple valid elimination orderings exist, and the maximum weight spanning tree can be non-unique. Thus, multiple valid junction trees can result, depending on the selected elimination order and the selected spanning tree.

In sum, we can obtain a valid junction tree using the procedure outlined in Algorithm \ref{alg:get_junction_tree}. In Figure \ref{fig:junction_tree_potentials_highlighted}, we illustrate both a valid and invalid junction tree for a given MRF.

\subsubsection{The Junction Tree Algorithm}

We now define the junction tree algorithm and explain why it works. At a high-level, this algorithm implements a form of message passing on the junction tree, which will be equivalent to variable elimination (in the same way that belief propagation is equivalent to variable elimination).

More precisely, let us define the potential $\psi_c(\mathbf{x}_c)$ of each cluster $c$ as the product of all the factors $\phi$ in $\mathcal{G}$ that have been assigned to $c$. By the family preservation property, this is well-defined, and we can assume that our distribution is of the form
\[
p(x_1, \dotsc, x_n) = \frac{1}{Z} \prod_{c \in \mathbf{C}} \psi_c(\mathbf{x}_c). 
\]

Then, at each step of the algorithm, we choose a pair of adjacent clusters $c^{(i)}, c^{(j)}$ in $\mathcal{T}$ and compute a message whose scope is the sepset $\mathbf{S}_{ij}$ between the two clusters:
\[
m_{i\to j}(\mathbf{S}_{ij}) = \sum_{\mathbf{x}_c \backslash \mathbf{S}_{ij}} \psi_c(\mathbf{x}_c) \prod_{x_\ell \in \mathbf{ne}(x_i) \backslash x_j} m_{\ell \to i}(\mathbf{S}_{\ell i}).
\]

\begin{algorithm}[!t]
\caption{\textit{Obtaining a junction tree}} \label{alg:get_junction_tree} 
\small
    \begin{algorithmic}[1]
    \STATE If the original graph is directed, obtain the moral graph (Definition \ref{def:moral_graph}). \\
    \STATE Select a node ordering. Chordalize the graph (Definition \ref{def:chordal_graph}) by running variable elimination on this ordering (Algorithm \ref{alg:variable_elimination}). \\
    %\STATE Obtain a set of maximal elimination cliques from the chordal graph. \\
    \STATE Construct a complete cluster graph over the maximal elimination cliques resulting from variable elimination. \\
    \STATE Weight each edge by the cardinality of the sepset corresponding to it. This yields a weighted clique graph. \\
    \STATE Construct a maximum weight spanning tree from the weighted clique graph. The resulting graph is a junction tree. 
    \end{algorithmic}
\end{algorithm}

\begin{algorithm}[!t]
\caption{\textit{Junction tree algorithm}} \label{alg:junction_tree_algorithm} 
\small
    \begin{algorithmic}[1]
    \STATE Input a valid junction tree $\mathcal{T}$ corresponding to original graph $\mathcal{G}$ (e.g., as obtained by Algorithm \ref{alg:get_junction_tree}).
    \STATE Define potentials on $\mathcal{T}$ (i.e.,  singleton potentials and edge potentials).
    \STATE Run the sum-product algorithm on $\mathcal{T}$ (Algorithm \ref{alg:variable_elimination}).
    \STATE Compute marginals for each node in $\mathcal{G}$ using the maximal clique marginals obtained at Line 3.
    \end{algorithmic}
\end{algorithm}

We choose $c^{(i)}, c^{(j)}$ only if $c^{(i)}$ has received messages from all of its neighbors except $c^{(j)}$. Just as in belief propagation, this procedure will terminate in exactly $2 \lvert \mathbf{E}_\mathcal{T} \rvert$ steps. After it terminates, we will define the belief of each cluster based on all the messages that it receives
\[
\beta_c(\mathbf{x}_c) = \psi_c(\mathbf{x}_c) \prod_{x_\ell \in \mathbf{ne}(x_i)} m_{\ell \to i}(\mathbf{S}_{\ell i}).
\]

These updates are often referred to as \textit{Shafer-Shenoy} \citep{shafer1990probability}. After all the messages have been passed, beliefs will be proportional to the marginal probabilities over their scopes, i.e., $\beta_c(\mathbf{x}_c) \propto p(\mathbf{x}_c)$. We can answer queries of the form $\Tilde{p}(x)$ for $x \in \mathbf{x}_c$ by marginalizing out the variable in its belief
\[
\Tilde{p}(x) = \sum_{\mathbf{x}_c \backslash x} \beta_c(\mathbf{x}_c).
\]
Readers familiar with combinatorial optimization will recognize this as a special case of dynamic programming on a tree decomposition of a graph with bounded treewidth. To get the actual (normalized) probability, we divide by the partition function $Z$ which is computed by summing all the beliefs in a cluster, $Z = \sum_{\mathbf{x}_c} \beta_c(\mathbf{x}_c)$.

Note that this algorithm makes it obvious why we want small clusters: the running time will be exponential in the size of the largest cluster (if only because we can need to marginalize out variables from the cluster, which often must be done using brute force). This is why a junction tree of a single node containing all the variables is not useful: it amounts to performing full brute-force marginalization.

Once we have obtained a junction tree via Algorithm \ref{alg:get_junction_tree}, we can perform the full junction tree algorithm, as in Algorithm \ref{alg:junction_tree_algorithm}. %\jm{Pseudocode: \href{https://www.stats.ox.ac.uk/~evans/gms/_book/jt.html#fig:junc}{link 1}, \href{https://ocw.mit.edu/courses/6-438-algorithms-for-inference-fall-2014/a49d895264467b7e118bf0602616cc37_MIT6_438F14_Lec14.pdf}{link 2}.}

\subsubsection{Variable Elimination Over a Junction Tree}

Why does this method work? First, let us convince ourselves that running variable elimination with a certain ordering is equivalent to performing message passing on the junction tree. Then, we will see that the junction tree algorithm is just a way of precomputing these messages and using them to answer queries.

Suppose we are performing variable elimination to compute $\Tilde{p}(x')$ for some variable $x'$, where $\Tilde{p} = \prod_{c \in \mathbf{C}} \psi_c$. Let $c^{(i)}$ be a cluster containing $x'$. We will perform variable elimination with the ordering given by the structure of the tree rooted at $c^{(i)}$. For the MRF in Figure \ref{fig:junction_tree_potentials_highlighted}, say that we choose to eliminate variable $B$, and we set $\{A,B,C\}$ as the root cluster.

First, we pick a set of variables $x_{-k}$ in a leaf $c^{(j)}$ of $\mathcal{T}$ that does not appear in the sepset $S_{kj}$ between $c^{(j)}$ and its parent $c^{(k)}$ (if there is no such variable, we can multiply $\psi(\mathbf{x}_c^{(j)})$ and $\psi(\mathbf{x}_c^{(k)})$ into a new factor with a scope not larger than that of the initial factors). In our example, we can pick the variable $f$ in the factor $\{B,E,F\}$.

Then we marginalize out $x_{-k}$ to obtain a factor $m_{j \to k}(\mathbf{S}_{ij})$. We multiply $m_{j \to k}(\mathbf{S}_{ij})$ with $\psi(\mathbf{x}_c^{(k)})$ to obtain a new factor $\tau(\mathbf{x}_c^{(k)})$. Doing so, we have effectively eliminated the factor $\psi(\mathbf{x}_c^{(j)})$ and the unique variables it contained. In the running example, we can sum out $f$ and the resulting factor over $\{B, E\}$ may be folded into $\{B,C,E\}$.

Note that the messages computed in this case are exactly the same as those of junction tree. In particular, when $c^{(k)}$ is ready to send its message, it will have been multiplied by $m_{\ell \to k}(\mathbf{S}_{ij})$ from all neighbors except its parent, which is exactly how junction tree sends its message.

Repeating this procedure eventually produces a single factor $\beta(\mathbf{x}_c^{(i)})$, which is our final belief. Since variable elimination implements the messages of the junction tree algorithm, $\beta(\mathbf{x}_c^{(i)})$ will correspond to the junction tree belief. Assuming we have convinced ourselves in the previous section that variable elimination works, we know that this belief will be valid.

Formally, we can prove correctness of the junction tree algorithm through an induction argument on the number of factors $\psi$; we will leave this as an exercise to the reader. The key property that makes this argument possible is the running intersection property: it assures us that it's safe to eliminate a variable from a leaf cluster that is not found in that cluster's sepset. By the running intersection property, this cannot occur anywhere except that one cluster.

The important thing to note is that if we now set $c^{(k)}$ to be the root of the tree (e.g., if we set $\{B,C,E\}$ to be the root), the message it will receive from $c^{(j)}$ (or from $\{B,E,F\}$ in our example) will not change. Hence, the caching approach we used for the belief propagation algorithm extends immediately to junction trees. The algorithm we formally defined above implements this caching.

\subsubsection{Computational Complexity of the Junction Tree Algorithm}

Let $\mathbf{X}$ be the nodes in our original graph $\mathcal{G}$. Both the number of cliques in the corresponding junction tree and the number of messages that we must compute in Algorithm \ref{alg:junction_tree_algorithm} are $O(|\mathbf{X}|)$. The time complexity and space complexity of the junction tree algorithm are dominated by the treewidth (i.e., the size of the largest clique in the junction tree over all possible elimination orderings). In the general case, these will be exponential in the treewidth. Recall that the largest maximal clique size of a chordal graph will depend on the elimination ordering, such that some orderings are more favorable than others.

%%%%%
% MAP
%%%%%

\section{Exact MAP Inference}
\label{sec:exact_map}

In the final section of this chapter, we shift focus from general-purpose inference algorithms to the specific problem of MAP inference in graphical models. While earlier sections introduced algorithms that are applicable to both marginal and MAP inference, there exist special cases in which MAP inference can be performed more efficiently. This section discusses the computational complexity of MAP inference and presents an example of an efficient exact algorithm based on graph cuts.

As an example, consider MAP inference in an MRF with distribution $p$ (Equation \ref{eq:mrf}), which corresponds to the following optimization problem:
\[
\max_\mathbf{x} \log p(\mathbf{x}) = \max_\mathbf{x} \sum_c \theta_c(\mathbf{x}_c) - \log Z,
\]
where $\theta_c(\mathbf{x}_c) = \log\phi_c(\mathbf{x}_c)$. 
We will now consider several efficient methods for solving this optimization problem. See Chapter 13 in \citet{koller2009probabilistic} for a thorough discussion of MAP inference.

\subsection{The Challenges of MAP Inference}

In a way, MAP inference is easier than marginal inference. One reason for this is that the intractable partition constant $\log Z$ does not depend on $\mathbf{x}$ and can be ignored. In the example above, for instance, we only need to compute
\[
\argmax_\mathbf{x} \sum_c \theta_c(\mathbf{x}_c). 
\]

Marginal inference can also be seen as computing and summing all assignments to the model, one of which is the MAP assignment. If we replace summation with maximization, we can also find the assignment with the highest probability. However, there exist more efficient methods than this sort of enumeration-based approach.

Note, however, that MAP inference is still not an easy problem in the general case. The above optimization objective includes many intractable problems as special cases~\citep{shimony1994finding}, such as 3-SAT. We can reduce 3-SAT to MAP inference by constructing for each clause $ c = (x \lor y \lor \neg z)$ a factor $\theta_c (x, y, z)$ that equals one if $x, y, z$ satisfy clause $c$, and equals zero otherwise. Then, the 3-SAT instance is satisfiable if and only if the value of the MAP assignment equals the number of clauses. We can also use a similar construction to prove that marginal inference is NP-hard. The high-level idea is to add an additional variable $X$ that equals $1$ when all the clauses are satisfied, and zero otherwise. The marginal probability will be greater than zero if and only if the 3-SAT instance is satisfiable.
%\jm{I am not sure this paragraph is clear for the general reader.} 
% \vk{also not sure if clear, consider adding reference/citation or clarifying or removing}

Nonetheless, we will see that the MAP problem is easier than general inference, in the sense that there are some models in which MAP inference can be solved in polynomial time while general inference is NP-hard.

\subsubsection{Illustrative Examples}

Many interesting examples of MAP inference are instances of \textit{structured prediction}, which involves doing inference in a CRF $p(\mathbf{y} \mid \mathbf{x})$:
\[
\arg\max_\mathbf{y} \log p(\mathbf{y}|\mathbf{x}) = \arg\max_\mathbf{y} \sum_c \theta_c(\mathbf{y}_c, \mathbf{x}_c). 
\]

We discussed structured prediction in detail when we covered CRFs in Section \ref{sec:crf}. Recall that our main example was handwriting recognition, in which we are given images of characters in the form of pixel matrices $\mathbf{x}_i \in [0, 1]^{d\times d}$ (Figure \ref{fig:optical_character_recognition}). MAP inference in this setting amounts to jointly recognizing the most likely word $(y_i)_{i=1}^n$ encoded by the images.

\begin{figure}[!t]
    \centering
    \includegraphics[width=0.3\linewidth]{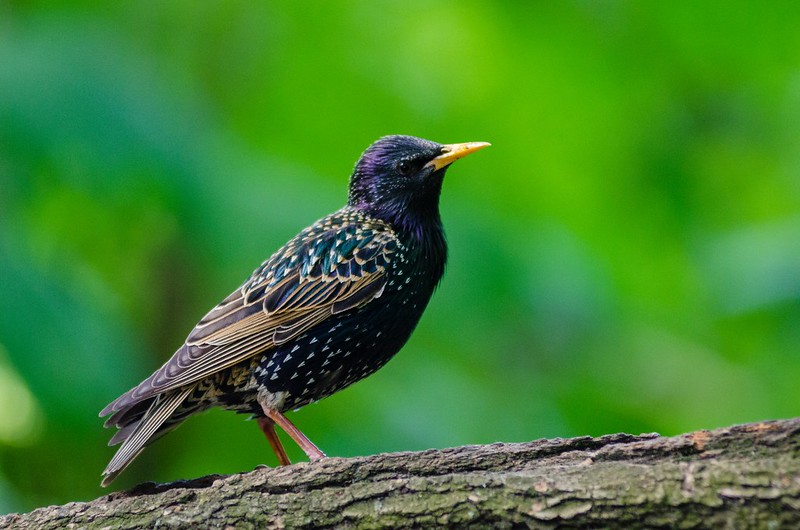}
    \includegraphics[width=0.3\linewidth]{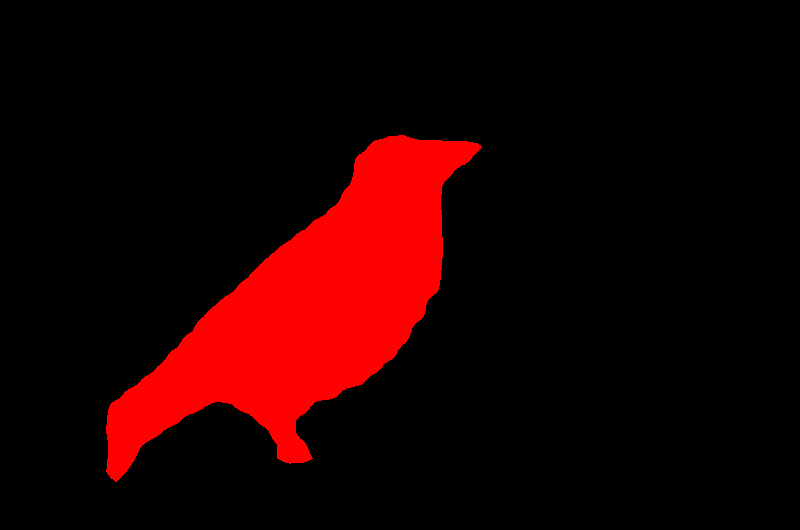}
    \includegraphics[width=0.3\linewidth]{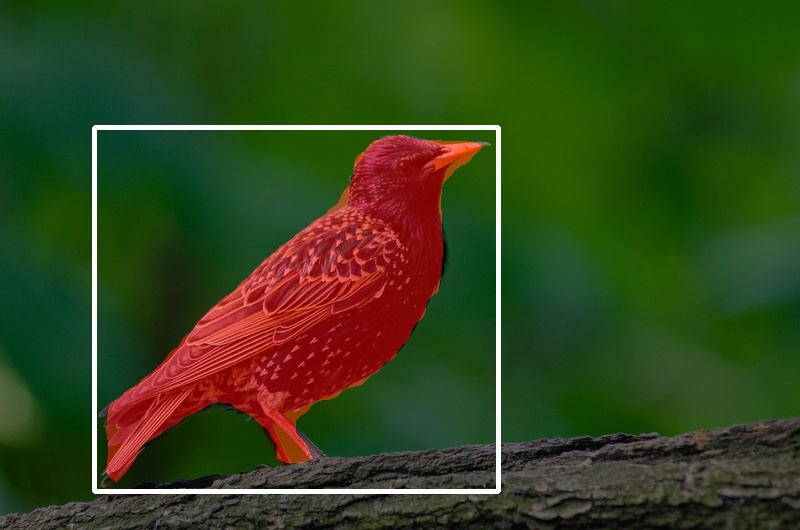}
    \caption{An illustration of the image segmentation problem, where a segmentation mask is used to partition the image into \textit{bird} and \textit{background}. Original photograph taken by Mathias Appel (public domain) with segmentation performed in Python.} %Custom Python code.
    \label{fig:image_segmentation}
\end{figure}

Another example of MAP inference is image segmentation. Image segmentation is used frequently in computer vision applications (e.g., medical image analysis). Here, we are interested in locating an entity in an image and labeling all its pixels (e.g., the bird in Figure \ref{fig:image_segmentation}). Our input $\mathbf{x}_i \in [0, 1]^{d\times d}$ is a matrix of image pixels, and our task is to predict the label $\mathbf{y}_i \in \{0, 1\}^{d\times d}$, indicating whether each pixel encodes the object we want to recover. Intuitively, neighboring pixels should have similar values in $\mathbf{y}_i$ (e.g., pixels associated with the bird should form one continuous shape, rather than white noise).

%{% include marginfigure.html id=``segmentation'' url=``assets/img/imagesegmentation.png'' description=``An illustration of the image segmentation problem.'' %}

%This prior knowledge can be naturally modeled in the language of graphical models via a Potts model. As in our first example, we can introduce potentials $\phi(y_i,x)$ that encode the likelihood that any given pixel is from our subject. We then augment them with pairwise potentials $\phi(y_i, y_j)$ for neighboring $y_i, y_j$, which will encourage adjacent $y$'s to have the same value with higher probability.
%\jm{Define Potts model.}

\subsection{Exact MAP Inference with Graph Cuts}
\label{sec:exact_map_graph_cuts}

We start our discussion with an efficient exact MAP inference algorithm called \textit{graph cuts} for certain  models over binary-valued variables $(X_i \in \{0, 1\})$. Unlike some methods (e.g., the junction tree algorithm), this algorithm is tractable even when the model has large treewidth.

A graph cut in an undirected graph $\mathcal{G} = (\mathbf{V}, \mathbf{E})$ is a partition of $\mathbf{V}$ into two disjoint sets $\mathbf{V}_s$ and $\mathbf{V}_t$. When each edge $(v_1, v_2) \in \mathbf{E}$ is associated with a nonnegative cost, $cost(v_1, v_2)$, the cost of a graph cut is the sum of the costs of the edges that cross between the two partitions:
\[
cost(\mathbf{V}_s, \mathbf{V}_t) = \sum_{v_1 \in \mathbf{V}_s,\ v_2 \in \mathbf{V}_t} cost(v_1, v_2). 
\]

\begin{figure}[!t]
    \centering

\begin{tikzpicture}[scale=0.8,every edge quotes/.style = {font=\footnotesize,fill=white,text= RubineRed}]

% Nodes
\node[draw,circle,thick,black,scale=0.7] (A) at (-2.5,0) {$A$};
\node[draw,circle,thick,black,scale=0.7] (B) at (-1.5,1.5) {$B$};
\node[draw,circle,thick,black,scale=0.7] (C) at (-1.5,-1.5) {$C$};
\node[draw,circle,thick,black,scale=0.7] (D) at (0,0) {$D$};
\node[draw,circle,thick,black,scale=0.7] (E) at (1.5,1.5) {$E$};
\node[draw,circle,thick,black,scale=0.7] (F) at (1.5,-1.5) {$F$};
\node[draw,circle,thick,black,scale=0.7] (G) at (2.5,0) {$G$};

% Edges
\draw[thick] (A) edge["$7$"] (B);
\draw[thick] (A) edge["$4$"] (C);
\draw[thick] (B) edge["$2$"] (D);
\draw[thick] (B) edge["$1$"] (E);
\draw[thick] (C) edge["$1$"] (D);
\draw[thick] (C) edge["$5$"] (F);
\draw[thick] (D) edge["$5$"] (E);
\draw[thick] (D) edge["$2$"] (F);
\draw[thick] (E) edge["$4$"] (G);
\draw[thick] (F) edge["$1$"] (G);

% Label
\node[] (label) at (0,-2.5) {\small \textbf{(a)}};

\end{tikzpicture}
%%%%%%%%%%%%%%%%%%%%%%%%%%%%%%
\hspace{10mm}
\begin{tikzpicture}[scale=0.8,every edge quotes/.style = {font=\footnotesize,fill=white,text= RubineRed}]

% Nodes
\node[draw,circle,thick,black,scale=0.7,fill=Melon!30] (A) at (-2.5,0) {$A$};
\node[draw,circle,thick,black,scale=0.7,fill=Melon!30] (B) at (-1.5,1.5) {$B$};
\node[draw,circle,thick,black,scale=0.7,fill=Melon!30] (C) at (-1.5,-1.5) {$C$};
\node[draw,circle,thick,black,scale=0.7,fill=SeaGreen!30] (D) at (0,0) {$D$};
\node[draw,circle,thick,black,scale=0.7,fill=SeaGreen!30] (E) at (1.5,1.5) {$E$};
\node[draw,circle,thick,black,scale=0.7,fill=Melon!30] (F) at (1.5,-1.5) {$F$};
\node[draw,circle,thick,black,scale=0.7,fill=SeaGreen!30] (G) at (2.5,0) {$G$};

% Edges
\draw[thick] (A) edge["$7$"] (B);
\draw[thick] (A) edge["$4$"] (C);
\draw[thick,dotted,gray] (B) edge["$2$"] (D);
\draw[thick,dotted,gray] (B) edge["$1$"] (E);
\draw[thick,dotted,gray] (C) edge["$1$"] (D);
\draw[thick] (C) edge["$5$"] (F);
\draw[thick] (D) edge["$5$"] (E);
\draw[thick,dotted,gray] (D) edge["$2$"] (F);
\draw[thick] (E) edge["$4$"] (G);
\draw[thick,dotted,gray] (F) edge["$1$"] (G);

% Label
\node[] (label) at (0,-2.5) {\small \textbf{(b)}};

\end{tikzpicture} \\
%%%%%%%%%%%%%%%%%%%%%%%%%%%%%%
\vspace{5mm}
\begin{tikzpicture}[scale=0.8,every edge quotes/.style = {font=\footnotesize,fill=white,text= RubineRed}]

% Nodes
\node[draw,circle,thick,black,scale=0.7,fill=Melon!30] (A) at (-2.5,0) {$A$};
\node[draw,circle,thick,black,scale=0.7,fill=Melon!30] (B) at (-1.5,1.5) {$B$};
\node[draw,circle,thick,black,scale=0.7,fill=Melon!30] (C) at (-1.5,-1.5) {$C$};
\node[draw,circle,thick,black,scale=0.7,fill=SeaGreen!30] (D) at (0,0) {$D$};
\node[draw,circle,thick,black,scale=0.7,fill=SeaGreen!30] (E) at (1.5,1.5) {$E$};
\node[draw,circle,thick,black,scale=0.7,fill=Melon!30] (F) at (1.5,-1.5) {$F$};
\node[draw,circle,thick,black,scale=0.7,fill=SeaGreen!30] (G) at (2.5,0) {$G$};

% Edges
\draw[thick] (A) -- (B);
\draw[thick] (A) -- (C);
\draw[thick] (C) -- (F);
\draw[thick] (D) -- (E);
\draw[thick] (E) -- (G);

% Highlights
\begin{pgfonlayer}{background}
    \fill[Melon,opacity=0.3] \convexpath{A,C}{8pt};
    \fill[Melon,opacity=0.3] \convexpath{A,B}{8pt};
    \fill[Melon,opacity=0.3] \convexpath{C,F}{8pt};
    \fill[SeaGreen,opacity=0.3] \convexpath{D,E}{8pt};
    \fill[SeaGreen,opacity=0.3] \convexpath{E,G}{8pt};
    \end{pgfonlayer}

% Label
\node[] (label) at (0,-2.5) {\small \textbf{(c)}};

\end{tikzpicture}

    \caption{Min-cut in an MRF $\mathcal{G} = (\mathbf{V}, \mathbf{E})$. Using the edge weights in $\mathcal{G}$ (\textbf{a}), we can identify the cut with minimum cost (dotted edges in \textbf{b}). In this MRF, the resulting node partition is $\mathbf{V}_s = \{A,B,C,F\}$ and $\mathbf{V}_t = \{D,E,G\}$ (\textbf{c}).}
    \label{fig:min_cut_mrf}
\end{figure}
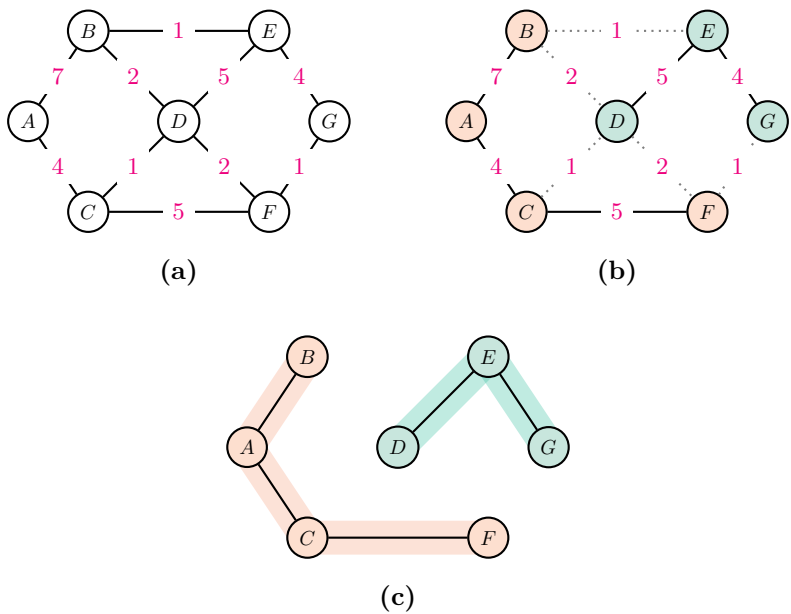

The \textit{min-cut} problem is to find the partition $\mathbf{V}_s, \mathbf{V}_t$ that minimizes the cost of the graph cut (Figure \ref{fig:min_cut_mrf}). The fastest algorithms for computing min-cuts in a graph take $O(\lvert \mathbf{E} \rvert \; \lvert \mathbf{V} \rvert \; \log \lvert \mathbf{V} \rvert)$ or $O(\lvert \mathbf{V} \rvert^3)$ time, and we refer readers to algorithms textbooks for details on their implementation \citep{cormen2022introduction}.

Now, we show a reduction of MAP inference on a particular class of MRFs to the min-cut problem. Suppose we are given an MRF over binary variables with pairwise factors in which edge energies (i.e., negative log-edge factors) have the form
\[
E_{uv}(x_u, x_v) =
\begin{cases}
0 & \text{if } x_u = x_v \\
\lambda_{uv} & \text{if } x_u \neq x_v
\end{cases}
\]
where $\lambda_{uv} \geq 0$ is a cost that penalizes edge mismatches. Assume also that each node $u$ has a unary potential described by an energy function $E_u(x_u)$. Thus, the full distribution is
\[
p(\mathbf{x}) = \frac{1}{Z} \exp\left[ -\sum_u E_u(x_u) - \sum_{u,v \in \mathbf{E}} E_{uv}(x_u, x_v) \right].
\]
Thus, MAP inference is equivalent to minimizing the energy:
\[
\argmax_\mathbf{x} p(\mathbf{x}) = \argmin_\mathbf{x} \left[ \sum_u E_u(x_u) + \sum_{u,v \in \mathbf{E}} E_{uv}(x_u, x_v) \right].
\]
For each node $u$, we can normalize its energies such that $E_u \geq 0$, and either $E_u(0) = 0$ or $E_u(1) = 0$. Specifically, we replace $E_u$ with $E_u' = E_u - \min E_u$, which is equivalent to multiplying the unnormalized probability distribution by a nonnegative constant $e^{\min E_u}$. This does not change the probability distribution. For example, we would replace
\[
\begin{array}{cc}
 x_u & E_u(x_u) \\
 \hline
 0 & 4 \\
 1 & -5
\end{array}
\quad\to\quad
\begin{array}{cc}
 x_u & E_u'(x_u) \\
 \hline
 0 & 9 \\
 1 & 0
\end{array}.
\]

The motivation for this model comes from image segmentation. We are looking for an assignment that minimizes the energy, which (among other things) tries to reduce discordance between adjacent variables. 

We can formulate energy minimization in this type of model as a min-cut problem in an augmented graph $\mathcal{G}'$:
\begin{itemize}
    \item We construct $\mathcal{G}'$ by adding special source and sink nodes $s,t$ to our PGM graph.
    \item The node $s$ is connected to nodes $u$ with $E_u(0) = 0$ by an edge with weight $E_u(1)$.
    \item The node $\mathcal{T}$ is connected to nodes $v$ with $E_v(1) = 0$ by an edge with weight $E_v(0)$.
    \item Finally, all the edges of the original graph get $E_{uv}$ as their weight.
\end{itemize}

%{% include marginfigure.html id=``mincut'' url=``assets/img/mincut.png'' description=``Formulating the segmentation task in a 2x2 MRF as a graph cut problem. Dashed edges are part of the min-cut. (Source: Machine Learning: A Probabilistic Perspective).'' %}

By construction, the cost of a minimal cut in this graph equals the minimum energy in the model. In particular, all nodes on the $s$ side of the cut receive an assignment of 0, and all nodes on the $\mathcal{T}$ side receive an assignment of 1. The edges between the nodes that disagree are precisely the ones in the minimal cut.

Similar techniques can be applied in slightly more general types of models with a certain type of edge potentials that are called \textit{submodular}. We refer the reader to Section 13.6 in \citet{koller2009probabilistic} for more details. In Chapter \ref{sec:approximate_inference}, we will revisit MAP inference in the context of approximation methods.

%\vspace{5mm}
\clearpage

\begin{reading}
    \begin{itemize}[leftmargin=*]
        %\item \fullcite{zhang1994simple}.
        %\item \fullcite{zhang1996exploiting}.
        \item \fullcite{lepar1998comparison}.
        %\item \fullcite{kschischang2001factor}.
        \item \fullcite{yedidia2003understanding}.
        \item Chapter 8.4 in \fullcite{bishop2006pattern}.
        %\item Section 2.5 in: \fullcite{wainwright_graphical_2008}.
        \item Chapter 9 in \fullcite{koller2009probabilistic}.
        \item Chapter 20 in \fullcite{murphy2012machine}.
    \end{itemize}
\end{reading}

\chapter{Approximate Inference}
\label{sec:approximate_inference}

%The general problem of exact probabilistic inference in Bayesian networks is NP-hard \citep{cooper1990computational}. Thus, exact inference is computationally intractable for many problems, especially when data is large. In this chapter, we explore the rich body of \textit{approximate inference} methods that trade accuracy for computational efficiency. Still, approximating probability distributions remains a challenging task: even approximate probabilistic inference in Bayesian networks is NP-hard in the general case \citep{dagum1993approximating}. Nevertheless, many clever solutions have been proposed, and special cases can be polynomial time with respect to input size.
%In this chapter, we review approximate inference approaches that generally fall under one of two categories: \textit{variational methods} and \textit{sampling methods}. 

% In practice, probabilistic models are often quite complex. 
In many settings, simple algorithms like variable elimination may be unreasonably slow. The general problem of exact probabilistic inference in Bayesian networks is NP-hard \citep{cooper1990computational}, and many interesting classes of models may not admit exact polynomial-time solutions at all. For this reason, much research effort in machine learning is spent on developing algorithms that yield \textit{approximate} solutions to the inference problem. Approximate inference algorithms generally forgo the accuracy of exact inference in exchange for computational efficiency. Still, approximating probability distributions remains a challenging task: even approximate probabilistic inference in Bayesian networks is NP-hard in the general case \citep{dagum1993approximating}, though special cases can be polynomial-time with respect to input size. 

There exist two main families of approximate algorithms: \textit{sampling} methods, which produce answers by repeatedly generating random numbers from a distribution of interest, and \textit{variational} methods, which formulate inference as an optimization problem. Sampling methods can be used to perform both marginal and MAP inference queries. In addition, they can compute various interesting quantities, such as expectations $\mathbb{E}[f(X)]$ of random variables distributed according to a given probabilistic model. Sampling methods have historically been the main way of performing approximate inference, although variational methods have emerged as viable (and often computationally superior) alternatives. Variational inference methods (sometimes called \textit{variational Bayes}) take their name from the \textit{calculus of variations}, which deals with optimizing functions that take other functions as arguments. 

We begin this chapter by exploring general approximate inference methods.
First, we introduce \textit{sampling-based approaches}, covering popular Monte Carlo and Markov methods (Section \ref{sec:sampling_methods}).
We then provide intuition for \textit{inference as optimization}, reviewing the foundations of variational inference (Section \ref{sec:variational_methods}).
Finally, we conclude by revisiting special cases of \textit{MAP inference} through the lens of approximate inference strategies (Section \ref{sec:approximate_map}).

%%%%%%
%% Sampling-based inference
%%%%%%

\section{Sampling-Based Approximate Inference}
\label{sec:sampling_methods}

\subsection{Sampling from a Probability Distribution}

As a warm-up, let's think for a moment about how we might sample from a multinomial distribution with $k$ possible outcomes and associated probabilities $\theta_1, \dotsc, \theta_k$.

Sampling, in general, is not an easy problem. Our computers can only generate samples from very simple distributions such as the uniform distribution over $[0,1]$. Even those samples are not truly random. They are actually taken from a deterministic sequence whose statistical properties (e.g., running averages) are indistinguishable from a truly random one. We call such sequences \textit{pseudorandom}. All sampling techniques involve calling some kind of simple subroutine multiple times in a clever way.

In our case, we may reduce sampling from a multinomial variable to sampling a single uniform variable by subdividing a unit interval into $k$ regions with region $i$ having size $\theta_i$ (Figure \ref{fig:sampling_uniform}). We then sample uniformly from $[0,1]$ and return the value of the region in which our sample falls.

\definecolor{bananamania}{rgb}{0.98, 0.91, 0.71}
\definecolor{babypink}{rgb}{0.96, 0.76, 0.76}
\definecolor{powderblue}{rgb}{0.69, 0.88, 0.9}
\definecolor{mistyrose}{rgb}{1.0, 0.89, 0.88}
\definecolor{magicmint}{rgb}{0.67, 0.94, 0.82}
\definecolor{mintcream}{rgb}{0.96, 1.0, 0.98}
\definecolor{periwinkle}{rgb}{0.8, 0.8, 1.0}

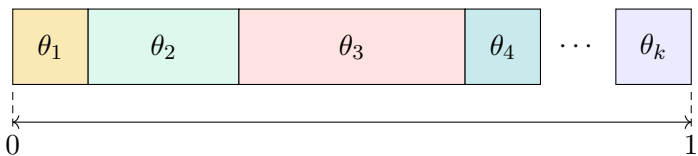
\begin{figure}[!t]
    \centering
\begin{tikzpicture}
    % Define the sections and colors
    \draw[white,fill=white] (7,0) rectangle (8,1) node[midway] {\color{black}$\cdots$};
    \draw[fill=bananamania] (0,0) rectangle (1,1) node[midway] {$\theta_1$};
    \draw[fill=magicmint!40] (1,0) rectangle (3,1) node[midway] {$\theta_2$};
    \draw[fill=mistyrose] (3,0) rectangle (6,1) node[midway] {$\theta_3$};
    \draw[fill=powderblue!70] (6,0) rectangle (7,1) node[midway] {$\theta_4$};
    \draw[fill=periwinkle!40] (8,0) rectangle (9,1) node[midway] {$\theta_k$};

    % Add labels for 0 and 1
    \node at (0, -0.8) {$0$};
    \node at (9, -0.8) {$1$};
    \draw[<->] (0,-0.5) --++ (9,0);
    \foreach\i in {0,9}
        \draw[densely dashed,very thin] (\i,-0.1) --++ (0,-0.5);
\end{tikzpicture}
    \caption{Reducing the problem of sampling from a multinomial distribution to sampling a uniform distribution in $[0,1]$ by subdividing a unit interval into $k$ regions, where each region $i$ is size $\theta_i$.}
    \label{fig:sampling_uniform}
\end{figure}

%{% include maincolumn_img.html src=``assets/img/multinomial-sampling.png'' caption=``Reducing sampling from a multinomial distribution to sampling a uniform distribution in [0,1].'' %}

\subsubsection{Forward Sampling}

Our technique for sampling from multinomials naturally extends to Bayesian networks with multinomial-distributed variables, via a method called \textit{ancestral} (or \textit{forward}) sampling. In particular, to sample from a probabilistic model $p(\mathbf{x}) = p(x_1, \dotsc, x_n)$ specified by a Bayesian network, we can sample variables in topological order. We start by sampling the variables with no parents; then we sample from the next generation by conditioning these variables' conditional probability distributions on values sampled at the first step. We proceed like this until all $n$ variables have been sampled. Importantly, in a Bayesian network over $n$ variables, forward sampling allows us to sample from the joint distribution $\mathbf{x} \sim p(\mathbf{x})$ in linear, $O(n)$, time by taking exactly 1 multinomial sample from each conditional probability distribution.

As an example, take our earlier model of student grades (Figure \ref{fig:student_grades_dag}). Here, we would first sample an exam difficulty $d'$ and an investment level $i'$. Then, once we have samples $d'$ and $i'$, we generate a student grade $g'$ from $p(g \mid d', i')$. At each step, we simply perform standard multinomial sampling.

%A former CS228 student has created an [interactive web simulation](http://pgmlearning.herokuapp.com/samplingApp) for visualizing Bayesian network forward sampling methods. Feel free to play around with it and, if you do, please submit any feedback or bugs through the Feedback button on the web app.

% \vk{I didn't follow the part below}

Forward sampling can also be performed efficiently on undirected graphical models if the model can be represented as a junction tree (Definition~\ref{def:junction_tree}) with a small number of variables per node.
We begin by performing message passing, until each clique holds a potential proportional to the marginal distribution over the variables in that clique, conditioned on any observed evidence.
We then choose one of the cliques in the junction tree to act as the root. Sampling proceeds in a top-down manner through the tree.
Starting at the root clique, we sample its variables sequentially, conditioning each sample on the previously sampled variables within the clique.
After sampling all variables in the root clique, we pass the sampled values of shared variables to a neighboring child clique. In that child clique, we condition on these shared variables to sample the remaining (unseen) variables in the clique.
This process continues: at each clique, we condition on previously sampled variables, sample the remaining variables in the clique, and pass the samples of shared variables to the next neighboring cliques in the tree.

% We first calibrate the clique tree, which gives us the marginal distribution over each node. We then choose a node to be the root. Then, we  marginalize over variables in the root node to get the marginal for a single variable. Let $E$ denote \textit{evidence}. Once the marginal for a single variable $x_1 \sim p(X_1 \mid E=e)$ has been sampled from the root node, the newly sampled value $X_1 = x_1$ can be incorporated as evidence. We then finish sampling other variables from the same node, each time incorporating the newly sampled nodes as evidence (i.e., $x_2 \sim p(X_2=x_2 \mid X_1=x_1,E=e)$ and $x_3 \sim p(X_3=x_3 \mid X_1=x_1,X_2=x_2,E=e)$ and so on). When moving down the tree to sample variables from other nodes, each node must send an updated message containing the values of the sampled variables.

\subsection{Monte Carlo Estimation}

Sampling from a distribution lets us perform many useful tasks, including marginal inference, MAP inference, and computing integrals of the form
\[
\mathbb{E}_{\mathbf{x} \sim p}[f(\mathbf{x})] = \sum_\mathbf{x} f(\mathbf{x}) p(\mathbf{x}).
\]
If $f(\mathbf{x})$ does not have special structure that matches the Bayesian network structure of $p$, this integral will typically be impossible to perform analytically. Instead, we will approximate it using a large number of samples from $p$. Algorithms that construct solutions based on a large number of samples from a given distribution are referred to as \textit{Monte Carlo}\footnote{The name \textit{Monte Carlo} refers to a famous casino in the city of Monaco. The term was originally coined as a codeword by physicists working on the atomic bomb as part of the secret Manhattan project.} (MC) methods.

Monte Carlo integration is an important instantiation of the general Monte Carlo principle. This technique approximates a target expectation with
\[
\mathbb{E}_{\mathbf{x} \sim p}[f(\mathbf{x})] \approx I_T = \frac{1}{T} \sum_{t=1}^T f(\mathbf{x}^t), 
\]
where $\mathbf{x}^1, \dotsc, \mathbf{x}^T$ are samples drawn according to $p$. It can be shown that
\begin{align*}
\mathbb{E}_{\mathbf{x}^1, \dotsc, \mathbf{x}^T \, \stackrel{iid}{\sim} p} [I_T] &= \mathbb{E}_{\mathbf{x} \sim p}[f(\mathbf{x})] \\
\text{Var}_{\mathbf{x}^1, \dotsc, \mathbf{x}^T \, \stackrel{iid}{\sim} p} [I_T] &= \frac{1}{T} \text{Var}_{\mathbf{x} \sim p} [f(\mathbf{x})].
\end{align*}
The first equation says that the MC estimate $I_T$ is an unbiased estimator for $\mathbb{E}_{\mathbf{x} \sim p}[f(\mathbf{x})]$. The two equations together show that $I_T \to \mathbb{E}_{\mathbf{x} \sim p}[f(\mathbf{x})]$ as $T \to \infty$. In particular, the variance of $I_T$ can be made arbitrarily small with enough samples.

\subsubsection{Rejection Sampling}

A special case of Monte Carlo integration is rejection sampling. We may use it to compute the area of a region $\gamma$ by sampling in a larger region with a known area and recording the fraction of samples that falls within $\gamma$.

For example, suppose we have a Bayesian network over the set of variables $\mathbf{X} = \mathbf{Z} \cup \mathbf{E}$. We may use rejection sampling to compute marginal probabilities $p(\mathbf{E}=\mathbf{e})$. We can rewrite the probability as
\[
p(\mathbf{E}=\mathbf{e}) = \sum_{\mathbf{z}} p(\mathbf{Z}=\mathbf{z}, \mathbf{E}=\mathbf{e}) = \sum_\mathbf{x} p(\mathbf{x}) \mathbb{I}(\mathbf{E}=\mathbf{e}) = \mathbb{E}_{\mathbf{x} \sim p}[\mathbb{I}(\mathbf{E}=\mathbf{e})] 
\]
and then take the Monte Carlo approximation. In other words, we draw many samples from $p$ and report the fraction of samples that are consistent with the value of the marginal.

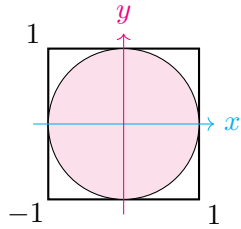
\begin{figure}[!t]
    \centering
\begin{tikzpicture}
    % Draw the square
    \draw[thick] (-1,-1) rectangle (1,1);
    
    % Draw the circle
    \filldraw[fill=Lavender!30, draw=black] (0,0) circle(1);
    
    % Draw the x and y axes
    \draw[->,ProcessBlue] (-1.2,0) -- (1.2,0) node[right] {$x$};
    \draw[->,RubineRed] (0,-1.2) -- (0,1.2) node[above] {$y$};
    
    % Label the ticks
    \node at (-1.3, -1.2) {$-1$};
    \node at (1.2, -1.2) {$1$};
    \node at (-1.2, 1.2) {$1$};
\end{tikzpicture}
    \caption{Graphical illustration of rejection sampling. We may compute the area of circle by drawing uniform samples from the square. The fraction of points that fall in the circle represents its area. This method breaks down if the size of the circle is small relative to the size of the square.}
    \label{fig:rejection_sampling}
\end{figure}

\subsubsection{Importance Sampling}

Unfortunately, rejection sampling can be very wasteful. If $p(\mathbf{E}=\mathbf{e})$ equals, say, 1\%, then we will discard 99\% of all samples.

A better way of computing such integrals uses \textit{importance sampling}. The main idea is to sample from a distribution $q$ (hopefully with $q(\mathbf{x})$ roughly proportional to $f(\mathbf{x}) \cdot p(\mathbf{x})$), and then \textit{reweigh} the samples in a principled way, so that their sum still approximates the desired integral.

More formally, suppose we are interested in computing $\mathbb{E}_{\mathbf{x} \sim p}[f(\mathbf{x})]$. We may rewrite this integral as
\begin{align*}
\mathbb{E}_{\mathbf{x} \sim p}[f(\mathbf{x})]
&= \sum_\mathbf{x} f(\mathbf{x}) p(\mathbf{x}) \\
&= \sum_\mathbf{x} f(\mathbf{x}) \frac{p(\mathbf{x})}{q(\mathbf{x})} q(\mathbf{x}) \\
&= \mathbb{E}_{\mathbf{x} \sim q}[ f(\mathbf{x})w(\mathbf{x}) ] \\
&\approx \frac{1}{T} \sum_{t=1}^T f(\mathbf{x}^t) w(\mathbf{x}^t)
\end{align*}
where $w(\mathbf{x}) = \frac{p(\mathbf{x})}{q(\mathbf{x})}$ and the samples $\mathbf{x}^t$ are drawn independently from $q$. In other words, we may instead take samples from $q$ and reweigh them with $w(\mathbf{x})$. The expected value of this Monte Carlo approximation will be the original integral. The variance of this new estimator is
\[
\text{Var}_{\mathbf{x} \sim q}[ f(\mathbf{x})w(\mathbf{x}) ] = \mathbb{E}_{\mathbf{x} \sim q} [f^2(\mathbf{x}) w^2(\mathbf{x})] - \mathbb{E}_{\mathbf{x} \sim q} [f(\mathbf{x}) w(\mathbf{x})]^2 \geq 0 .
\]
Note that we can set the variance equal to zero by choosing
\[
q(\mathbf{x}) = \frac{\lvert f(\mathbf{x}) \rvert p(\mathbf{x})}{\int \lvert f(\mathbf{x}) \rvert p(\mathbf{x}) d\mathbf{x}}.
\]
If we can sample from this $q$ (and evaluate the corresponding weight), then we only need a single Monte Carlo sample to compute the true value of our integral. Of course, sampling from such a $q$ is NP-hard in general (its denominator $\mathbb{E}_{\mathbf{x} \sim p}[\lvert f(\mathbf{x}) \vert]$ is basically the quantity we're trying to estimate in the first place), but this at least gives us an indication for what to strive for.

In the context of our previous example for computing $p(\mathbf{E}=\mathbf{e})$, we may take $q$ to be the uniform distribution and apply importance sampling as follows:
\begin{align*}
p(\mathbf{E}=\mathbf{e})
&= \mathbb{E}_{\mathbf{z}\sim p}[p(\mathbf{e} \mid \mathbf{z})] \\
&= \mathbb{E}_{\mathbf{z}\sim q}\left[ p(\mathbf{e} \mid \mathbf{z})\frac{p(\mathbf{z})}{q(\mathbf{z})} \right] \\
&= \mathbb{E}_{\mathbf{z}\sim q}\left[\frac{p(\mathbf{e},\mathbf{z})}{q(\mathbf{z})} \right] \\
&= \mathbb{E}_{\mathbf{z}\sim q} [w_\mathbf{e}(\mathbf{z})] \\
&\approx \frac{1}{T} \sum_{t=1}^T w_\mathbf{e}(\mathbf{z}^t)
\end{align*}
where $w_\mathbf{e}(\mathbf{z}) = p(\mathbf{e}, \mathbf{z})/q(\mathbf{z})$, and each $\mathbf{z}^t$ is sampled independently from $q$. Unlike rejection sampling, this will use all of the samples in our calculation. If $p(\mathbf{z} \mid \mathbf{e})$ is not too far from uniform, this will converge to the true probability after only a very small number of samples.

\subsubsection{Normalized Importance Sampling}

Unfortunately, unnormalized importance sampling is not suitable for estimating conditional probabilities of the form
\[
p(\mathbf{Z}_i=\mathbf{z}_i \mid \mathbf{E}=\mathbf{e}) = \frac{p(\mathbf{Z}_i=\mathbf{z}_i, \mathbf{E}=\mathbf{e})}{p(\mathbf{E}=\mathbf{e})},
\]
where $\mathbf{Z}_i \subseteq \mathbf{Z}$. Note that, using unnormalized importance sampling, we could estimate the numerator as
% \begin{align*}
% \mathbb{P}(X_i=x_i, E=e)
% &= \sum_z \delta(z) p(e, z) \\
% &= \sum_z \delta(z) w_e(z) q(z) \\
% &= \mathbb{E}_{z \sim q}[ \delta(z) w_e(z) ] \\
% &\approx \frac{1}{T} \sum_{t=1}^T \delta(z^t) w_e(z^t).
% \end{align*}
\begin{align*}
p(\mathbf{Z}_i=\mathbf{z}_i, \mathbf{E}=\mathbf{e})
&= \sum_{\mathbf{z}} \delta(\mathbf{z}) p(\mathbf{e}, \mathbf{z}) \\
&= \sum_{\mathbf{z}} \delta(\mathbf{z}) w_\mathbf{e}(\mathbf{z}) q(\mathbf{z}) \\
&= \mathbb{E}_{\mathbf{z} \sim q}[ \delta(\mathbf{z}) w_\mathbf{e}(\mathbf{z}) ] \\
&\approx \frac{1}{T} \sum_{t=1}^T \delta(\mathbf{z}^t) w_\mathbf{e}(\mathbf{z}^t).
\end{align*}
where
\[
\delta(\mathbf{z}) = \begin{cases}1 & \text{if $\mathbf{z}$ is consistent with $\mathbf{Z}_i = \mathbf{z}_i$} \\ 0 & \text{otherwise}. \end{cases}
\]
While the denominator is the same as the result we derived earlier:
\[
p(\mathbf{E}=\mathbf{e}) \approx \frac{1}{T} \sum_{t=1}^T w_\mathbf{e}(\mathbf{z}^t),
\]
where the $\mathbf{z}^t$ are drawn independently from q.
If we estimate the numerator $p(\mathbf{Z}_i=\mathbf{z}_i, \mathbf{E}=\mathbf{e})$ and the denominator $p(\mathbf{E}=\mathbf{e})$ with different and independent samples of $\mathbf{z}^t \sim q$, then the errors in the two approximations may compound. For example, if the numerator is an under-estimate and the denominator is an over-estimate, the final probability could be a severe under-estimate.

However, if we instead use the same set of $T$ samples $\mathbf{z}^1, \dotsc, \mathbf{z}^T \sim q$ for both the numerator and denominator, we avoid this issue of compounding errors. Thus, the final form of the normalized importance sampling estimate can be written
\[
\hat{p}_T(\mathbf{Z}_i=\mathbf{z}_i \mid \mathbf{E}=\mathbf{e})
= \frac{\frac{1}{T} \sum_{t=1}^T \delta(\mathbf{z}^t) w_\mathbf{e}(\mathbf{z}^t)}
       {\frac{1}{T} \sum_{t=1}^T w_\mathbf{e}(\mathbf{z}^t)},
\]
where we use the notation $\hat{p}_T$ to denote the sampling-based approximation to the density of interest $p$, given $T$ samples.

Unfortunately, there is one drawback to the normalized importance sampling estimator, which is that it is \textit{biased}. If $T = 1$, then we have
\[
\mathbb{E}_{\mathbf{z} \sim q} [\hat{p}_1(\mathbf{Z}_i=\mathbf{z}_i \mid \mathbf{E}=\mathbf{e})]
    = \mathbb{E}_{\mathbf{z} \sim q} [\delta(\mathbf{z})]
    \neq p(\mathbf{Z}_i=\mathbf{z}_i \mid \mathbf{E}=\mathbf{e})
\]
Fortunately, because the numerator and denominator are both unbiased, the normalized importance sampling estimator remains \textit{asymptotically unbiased}, meaning that
\[
\lim_{T \to \infty} \mathbb{E}_{\mathbf{z} \sim q} [\hat{p}_T(\mathbf{Z}_i=\mathbf{z}_i \mid \mathbf{E}=\mathbf{e})] = p(\mathbf{Z}_i=\mathbf{z}_i \mid \mathbf{E}=\mathbf{e}). 
\]

% \jm{Are we using Big $\mathbb{P}$ and little $p$ consistently throughout this chapter?}
% \vk{i'm not sure, there is also the broader issue of using hats that I don't fully understand (let's double check if we defined hats); i'm not currently sure why the boldface is needed}
% \vk{this section is a bit informal (see e.g., Owens for something more formal but still not too long), but that could be okay}

\subsection{Markov Chain Monte Carlo}

We now turn our attention from computing expectations to performing marginal and MAP inference using sampling. We will solve these problems using a technique called \textit{Markov chain Monte Carlo} (MCMC).

MCMC is another algorithm that was developed during the Manhattan project and eventually republished in the scientific literature some decades later. It is so impactful that it was recently named as one of the ten most important algorithms of the 20$^{th}$ century \citep{cipra2000best}.

\subsubsection{Markov Chains}

\begin{figure}[!t]
    \centering
    \begin{tikzpicture}[every edge quotes/.style = {font=\footnotesize,fill=white,sloped}]
    \node[circle,black,thick,draw,scale=1.0,fill=Lavender!30] (1) at (0,0) {$1$};
    \node[circle,black,thick,draw,scale=1.0,fill=Lavender!30] (2) at (3,0) {$2$};
    \node[circle,black,thick,draw,scale=1.0,fill=Lavender!30] (3) at (6,0) {$3$};
    %%%
    \draw[-{Stealth[width=5pt,length=5pt]},thick]  (1) edge["\color{RubineRed}$1.0$"] (2);
    \draw[-{Stealth[width=5pt,length=5pt]},thick]  (2) edge["\color{RubineRed}$0.5$"] (3);
    \draw[-{Stealth[width=5pt,length=5pt]},thick,black]  (3) edge[bend right=40,"\color{RubineRed}$1.0$"] (1);
    \draw[-{Stealth[width=5pt,length=5pt]},thick,black]  (2) edge[bend left=50, "\color{RubineRed}$0.5$"] (1);
\end{tikzpicture}
    \caption{A Markov chain over three states. The weighted directed edges indicate probabilities of transitioning to a different state.}
    \label{fig:markov_chain}
\end{figure}
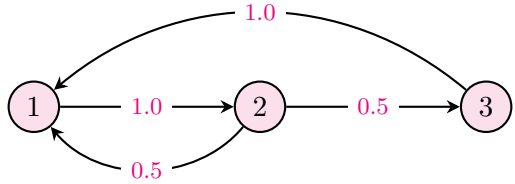

%ProcessBlue
%RubineRed
%Lavender

A key concept in MCMC is that of a \textit{Markov chain}. A (discrete-time) Markov chain is a sequence of random variables $(S_0, S_1, S_2, \ldots)$ with each random variable $S_i \in \{1,2,\ldots,d\}$ taking one of $d$ possible values, intuitively representing the state of a system. The initial state is distributed according to a probability $P(S_0)$. All subsequent states are generated from a conditional probability distribution that depends only on the previous random state. That is, $S_i$ is distributed according to $P(S_i \mid S_{i-1})$.

The probability $P(S_i \mid S_{i-1})$ is the same at every step $i$. This means that the transition probabilities at any time in the entire process depend only on the given state and not on the history of how we got there. This is called the \textit{Markov assumption}.

%{% include marginfigure.html id=``mc'' url=``assets/img/markovchain.png'' description=``A Markov chain over three states. The weighted directed edges indicate probabilities of transitioning to a different state.'' %}

It is very convenient to represent the transition probability as a $d \times d$ matrix $\mathbf{T}$ where
\[
\mathbf{T}_{ij} = P(S_\text{new} = i \mid S_\text{prev} = j). 
\]
If the initial state $S_0$ is drawn from a vector of probabilities $p_0$, we may represent the probability $p_t$ of ending up in each state after $t$ steps as
\[
p_t = \mathbf{T}^t p_0, 
\]
where $\mathbf{T}^t$ denotes matrix exponentiation (i.e., we apply the matrix operator $t$ times). The limit
\[
\pi = \lim_{t \to \infty} p_t
\]
(when it exists) is called a \textit{stationary distribution} of the Markov chain. We will construct below a Markov chain with a stationary distribution $\pi$ that exists and is the same for all $p_0$. We will refer to such $\pi$ as \textit{the} stationary distribution of the chain.

\subsubsection{Sufficient Conditions for Stationary Distributions}

A sufficient condition for the existence of a stationary distribution for a Markov chain with transition probability $\mathbf{T}$ is \textit{detailed balance}, defined as:
\[
\pi(x') \mathbf{T}(x \mid x') = \pi(x) \mathbf{T}(x' \mid x) \quad \forall x.
\]
It is easy to show that such a $\pi$ must form a stationary distribution by summing both sides of the equation over $x$ and simplifying.
% \vk{consider taking proof from my DGM slides}
However, the reverse may not hold and indeed it is possible to have MCMC without satisfying detailed balance \citep{suwa2010markov}.

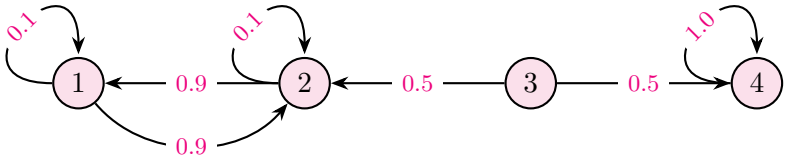
\begin{figure}[!t]
    \centering
    \begin{tikzpicture}[every edge quotes/.style = {font=\footnotesize,fill=white,sloped},>=Stealth]
    \node[circle,black,thick,draw,scale=1.0,fill=Lavender!30] (1) at (0,0) {$1$};
    \node[circle,black,thick,draw,scale=1.0,fill=Lavender!30] (2) at (3,0) {$2$};
    \node[circle,black,thick,draw,scale=1.0,fill=Lavender!30] (3) at (6,0) {$3$};
    \node[circle,black,thick,draw,scale=1.0,fill=Lavender!30] (4) at (9,0) {$4$};
    %%%
    \draw[->,thick]  (3) edge["\color{RubineRed}$0.5$"] (2);
    \draw[->,thick]  (3) edge["\color{RubineRed}$0.5$"] (4);
    \draw[->,thick,black]  (1) edge[bend right=50,"\color{RubineRed}$0.9$"] (2);
    \draw[->,thick,black]  (2) edge["\color{RubineRed}$0.9$"] (1);
    %% self loops
    \draw[->,thick,black] 
    (1) edge[out=180,in=90,loop,"\color{RubineRed}$0.1$"] (1);
    \draw[->,thick,black] (2) edge[out=180,in=90,loop,"\color{RubineRed}$0.1$"] (2);
    \draw[->,thick,black] (4) edge[out=180,in=90,loop,"\color{RubineRed}$1.0$"] (4);
\end{tikzpicture}
    \caption{A reducible Markov chain over four states. %\jm{Was this borrowed from somewhere?}
    }
    \label{fig:markov_chain_reducible}
\end{figure}

The high-level idea of MCMC will be to construct a Markov chain whose states will be joint assignments to the variables in the model and whose stationary distribution will equal a probability distribution of interest. Suppose we have defined a Markov chain with a stationary distribution $\pi$.
In addition to the existence of a stationary distribution, we need to ensure that the chain will converge to this stationary distribution from any initial state (as required in MCMC).
In particular, we desire the Markov chain to be ergodic, i.e., to converge to a unique stationary distribution $\pi$ over time.
This turns out to be true under two sufficient conditions:
% In order to construct such a chain, we first need to understand when stationary distributions exist.
% \vk{how does this relate to the above detailed balance condition?}
\begin{enumerate}
    \item \textit{Irreducibility}: It is possible to get from any state $x$ to any other state $x'$ with probability > 0 in a finite number of steps.
    \item \textit{Aperiodicity}: It is possible to return to any state at any time. That is, there exists an $n$ such that for all $i$ and all $n' \geq n$, 
    \[
    P(s_{n'}=i \mid s_0 = i) > 0.
    \]
\end{enumerate}
The first condition is meant to prevent \textit{absorbing states} (i.e., states from which we can never leave). In the example in Figure~\ref{fig:markov_chain_reducible}, if we start in states $1$ or $2$, we will never reach state 4. Conversely, if we start in state 4, then we will never reach states $1$ or $2$. If we start the chain in the middle (in state 3), then clearly it cannot have a single limiting distribution. The second condition is necessary to rule out transition operators such as
\[
\mathbf{T} =
\begin{bmatrix}
0 & 1 \\
1 & 0
\end{bmatrix}.
\]
Note that this chain alternates forever between states 1 and 2 without ever settling in a stationary distribution.

%{% include maincolumn_img.html src='assets/img/reducible-chain.png' caption='A reducible Markov chain over four states.' %}

%ProcessBlue
%RubineRed
%Lavender

\begin{theorem}
    An irreducible and aperiodic finite-state Markov chain has a stationary distribution.
\end{theorem}

In the case of continuous state spaces, ensuring that a Markov chain is ergodic typically requires stronger conditions than just irreducibility and aperiodicity. For generality, we will simply assume that our Markov chains are ergodic.

\subsubsection{High-Level Procedure}

MCMC algorithms construct a Markov chain over the assignments to a probability function $p$. The chain will have a stationary distribution equal to $p$ itself. By running the chain for some number of time steps, we will thus sample from $p$.

At a high level, MCMC algorithms will have the following structure. They take as argument a transition operator $\mathcal{T}$ specifying a Markov chain whose stationary distribution is $p$, and an initial assignment $\mathbf{x}_0$ to the variables of $p$. We then perform the following steps.
\begin{enumerate}
    \item Run the Markov chain from $\mathbf{x}_0$ for $B$ \textit{burn-in} steps.
    \item Run the Markov chain for $N$ \textit{sampling} steps and collect all the states that it visits.
\end{enumerate}

Assuming $B$ is sufficiently large, the latter collection of states will form samples from $p$. We may then use these samples for Monte Carlo integration (or importance sampling). We may also use them to produce Monte Carlo estimates of marginal probabilities. Finally, we may take the sample with the highest probability and use it as an estimate of the mode (i.e., to perform MAP inference).

\subsubsection{The Metropolis-Hastings Algorithm}

The Metropolis-Hastings (MH) algorithm is our first way to construct Markov chains within MCMC. The MH method constructs a transition operator $\mathcal{T}(\mathbf{x}' \mid \mathbf{x})$ from two components:
\begin{enumerate}
    \item A transition kernel $\mathcal{Q}(\mathbf{x}'\mid \mathbf{x})$, specified by the user.
    \item An acceptance probability for moves proposed by $\mathcal{Q}$, specified by the algorithm as
    \[
    \mathcal{A}(\mathbf{x}' \mid \mathbf{x}) = \min \left(1, \frac{p(\mathbf{x}')\mathcal{Q}(\mathbf{x} \mid \mathbf{x}')}{p(\mathbf{x})\mathcal{Q}(\mathbf{x}' \mid \mathbf{x})} \right).
    \]
\end{enumerate}
At each step of the Markov chain, we choose a new point $\mathbf{x}'$ according to $\mathcal{Q}$. Then, we either accept this proposed change (with probability $\alpha$), or we remain at our current state with probability $1-\alpha$. Notice that the acceptance probability encourages us to move towards more likely points in the distribution $p$ (imagine for example that $\mathcal{Q}$ is uniform). When $\mathcal{Q}$ suggests that we move into a low-probability region, we follow that move only a certain fraction of the time. In practice, the distribution $\mathcal{Q}$ is taken to be something simple, like a Gaussian centered at $\mathbf{x}$ if we are dealing with continuous variables.

Given any $\mathcal{Q}$, the MH algorithm will ensure that $p$ will be a stationary distribution of the resulting Markov chain. More precisely, $p$ will satisfy the detailed balance condition with respect to the MH Markov chain. To see that, first observe that if $\mathcal{A}(\mathbf{x}' \mid \mathbf{x}) < 1$, then
\[
\frac{p(\mathbf{x})\mathcal{Q}(\mathbf{x}' \mid \mathbf{x})}{p(\mathbf{x}')\mathcal{Q}(\mathbf{x} \mid \mathbf{x}')} > 1
\]
and thus $\mathcal{A}(\mathbf{x} \mid \mathbf{x}') = 1$. When $\mathcal{A}(\mathbf{x}' \mid \mathbf{x}) < 1$, this lets us write:
\begin{align*}
\mathcal{A}(\mathbf{x}' \mid \mathbf{x}) &= \frac{p(\mathbf{x}')\mathcal{Q}(\mathbf{x} \mid \mathbf{x}')}{p(\mathbf{x})\mathcal{Q}(\mathbf{x}' \mid \mathbf{x})} \\
p(\mathbf{x}')\mathcal{Q}(\mathbf{x} \mid \mathbf{x}')\mathcal{A}(\mathbf{x} \mid \mathbf{x}') &= p(\mathbf{x})\mathcal{Q}(\mathbf{x}' \mid \mathbf{x}) \mathcal{A}(\mathbf{x}' \mid \mathbf{x}) \\
p(\mathbf{x}')\mathcal{T}(\mathbf{x} \mid \mathbf{x}') &= p(\mathbf{x})\mathcal{T}(\mathbf{x}' \mid \mathbf{x}),
\end{align*}
% \begin{alignat*}{2}
% & \hfil \mathcal{A}(\mathbf{x}' \mid \mathbf{x}) &&= \frac{P(\mathbf{x}')\mathcal{Q}(\mathbf{x} \mid \mathbf{x}')}{P(\mathbf{x})\mathcal{Q}(\mathbf{x}' \mid \mathbf{x})} \\
% &\implies P(\mathbf{x}')\mathcal{Q}(\mathbf{x} \mid \mathbf{x}')\mathcal{A}(\mathbf{x} \mid \mathbf{x}') &&= P(\mathbf{x})\mathcal{Q}(\mathbf{x}' \mid \mathbf{x}) \mathcal{A}(\mathbf{x}' \mid \mathbf{x}) \\
% &\implies P(\mathbf{x}')\mathcal{T}(\mathbf{x} \mid \mathbf{x}')&&= P(\mathbf{x})\mathcal{T}(\mathbf{x}' \mid \mathbf{x}),
% \end{alignat*}
which is simply the detailed balance condition. We used $\mathcal{T}(\mathbf{x} \mid \mathbf{x}')$ to denote the full transition operator of MH (obtained by applying both $\mathcal{Q}$ and $\mathcal{A}$). Thus, if the MH Markov chain is ergodic, its stationary distribution will be $p$.

\subsubsection{Gibbs Sampling}

A widely-used special case of MH methods is Gibbs sampling. Given an ordered set of variables $(x_1,\ldots,x_n)$ and a starting configuration $\mathbf{x}^0 = (x_1^0,\ldots,x_n^0)$, Gibbs sampling iteratively samples each variable in turn from its conditional distribution, given the current values of all other variables. At iteration $t$, the algorithm updates each variable $x_i$ in sequence by drawing
\begin{align*}
x_i^{(t)} \sim p\left( x_i \mid x_1^{(t)}, \ldots, x_{i-1}^{(t)}, x_{i+1}^{(t-1)}, \ldots, x_n^{(t-1)} \right)
\end{align*}
i.e., conditioning on the most recent values of all other variables. We outline the full Gibbs sampling procedure in Algorithm \ref{alg:gibbs_sampling}.

We use $\mathbf{x}_{-i}$ to denote all variables in $\mathbf{x}$ except $x_i$. It is often very easy to performing each sampling step, since we only need to condition $x_i$ on its Markov blanket (Definitions \ref{def:markov_blanket_dag}, \ref{def:markov_blanket_mrf}), which is typically small. Note that when we update $x_i$, we \textit{immediately} use its new value for sampling other variables $x_j$. 

\begin{algorithm}[!t]
\caption{\textit{Gibbs Sampling}} \label{alg:gibbs_sampling} 
\vspace{3mm}
Repeat until convergence for $t = 1, 2,\dots$:
\begin{enumerate}
    \item Set $\mathbf{x} \leftarrow \mathbf{x}^{t-1}$.
    \item For each variable $x_i$ in the order we fixed:
	\begin{enumerate}
	    \item Sample $x'_i \sim p(x_i \mid \mathbf{x}_{-i})$
        \item Update $\mathbf{x} \leftarrow (x_1, \dotsc, x'_i, \dotsc, x_n).$
	\end{enumerate}
 \item Set $\mathbf{x}^t \leftarrow \mathbf{x}$
\end{enumerate}
\end{algorithm}

Gibbs sampling can be seen as a special case of MH with proposal
\[
\mathcal{Q}(x_i', \mathbf{x}_{-i} \mid x_i, \mathbf{x}_{-i}) = p(x_i' \mid \mathbf{x}_{-i}). 
\]
It is easy check that the acceptance probability simplifies to one. Assuming the right transition operator, both Gibbs sampling and MH will eventually produce samples from their stationary distribution, which by construction is $p$. There exist simple ways of ensuring that this will be the case.
\begin{enumerate}
    \item To ensure \textit{irreducibility}, the transition operator $\mathcal{Q}$ with MH should be able to move to every state. In the case of Gibbs sampling, we would like to make sure that every $x_i'$ can be sampled from $p(x_i \mid \mathbf{x}_{-i})$.
    \item To ensure \textit{aperiodicity}, it is enough to let the chain transition stay in its state with some probability.
\end{enumerate}
In practice, it is not difficult to ensure that these requirements are met.

\subsubsection{Running Time of MCMC}

A key parameter to this algorithm is the number of burn-in steps $B$. Intuitively, this corresponds to the number of steps needed to converge to our limit (stationary) distribution. This is called the \textit{mixing time} of the Markov chain.

%{% include sidenote.html id=``note-mixing'' note=``There is a technical definition of this quantity, which we will not cover here.'' %}.

Unfortunately, this time may vary dramatically  and may sometimes take (essentially) forever. For example, if the transition matrix is
\[
\mathbf{T} =
\begin{bmatrix}
1-\epsilon & \epsilon \\
\epsilon & 1-\epsilon
\end{bmatrix},
\]
then for small $\epsilon$ it will take a very long time to reach the stationary distribution, which is close to $(0.5, 0.5)$. At each step, we will stay in the same state with overwhelming probability. Very rarely, we will transition to another state, and then stay there for a very long time. The average of these states will converge to $(0.5, 0.5)$, but the convergence will be very slow.

This problem will also occur with complicated distributions that have two distinct and narrow modes. With high probability, the algorithm will sample from a given mode for a very long time. These examples are indications that sampling is a hard problem in general, and MCMC does not give us a free lunch. Nonetheless, for many real-world distributions, sampling will produce very useful solutions.

Another – and perhaps more important – problem is that we may not know when to end the burn-in period, even if it is theoretically not very long. There exist many heuristics to determine whether a Markov chain has \textit{mixed}. However, these heuristics typically involve plotting certain quantities and estimating them by eye. Even the quantitative measures are not significantly more reliable than this approach.

In summary, even though MCMC is able to sample from the right distribution (which in turn can be used to solve any inference problem), doing so may sometimes require a very long time. Furthermore, there is no easy way to judge the amount of computation that we will need to spend to find a good solution.

There are also a number of more-advanced MCMC methods that can improve on mixing performance by incorporating gradient information about the target distribution.
These gradient-based MCMC algorithms include Langevin Monte Carlo (LMC), which augments random-walk proposals with steps in the direction of the gradient of the log-density, and Hamiltonian Monte Carlo (HMC), which simulates trajectories under a physical system that uses position and momentum variables to explore the space more efficiently.
LMC and HMC can make more informed steps in high-dimensional parameter spaces and often achieve faster convergence with improved mixing.
For more information on gradient-based MCMC, see~\citet{brooks2011handbook, betancourt2017conceptual}.

% \williex{TODO: add a sentence about gradient-based MCMC (Langevin Monte Carlo, Hamiltonian Monte Carlo), which are quite popular in practice, and have good mixing performance, give a few citations for futher reading. Also allude to modern autodiff/deep learning for LMC/HMC.}

\begin{reading}
    \textit{Approximate Inference with Sampling}
    \begin{itemize}[leftmargin=*]
        \item Chapter 11 in  \fullcite{bishop2006pattern}.
        \item Chapter 12 in \fullcite{koller2009probabilistic}.
        \item Chapters 23 \& 24 in  \fullcite{murphy2012machine}.
        \item \fullcite{lindsten2013backward}.
        \item \fullcite{angelino_patterns_2016}.
        \item \fullcite{naesseth_elements_2019}.
        \item \fullcite{brooks2011handbook}.
    \end{itemize}
\end{reading}

%%%%%%
%% Variational inference
%%%%%%

\section{Variational Methods for Approximate Inference}
\label{sec:variational_methods}

We have seen that inference in probabilistic models is often intractable, and we have learned about some algorithms that provide approximate solutions to the inference problem %(e.g., marginal inference) 
by using subroutines involving sampling random variables. Unfortunately, these sampling-based methods have several important shortcomings.
\begin{enumerate}
    \item Although MCMC methods are guaranteed to find a globally optimal solution given enough time, it is difficult to tell how close they are to a good solution given the finite amount of time that they have in practice.
    \item In order to quickly reach a good solution, MCMC methods require choosing an appropriate sampling technique (e.g., a good proposal in MH). Choosing this technique can be an art in itself.
\end{enumerate}
In this section, we will discuss an alternative approach to approximate inference called the \textit{variational inference} family of algorithms.

\subsection{Inference as Optimization}

The main idea of variational inference methods is to cast inference as an optimization problem. Suppose we are given an intractable probability distribution $p$. Variational techniques will try to solve an optimization problem over a class of tractable distributions $\mathcal{Q}$ in order to find a $q \in \mathcal{Q}$ that is most similar to $p$. We will then query $q$ (rather than $p$) in order to get an approximate solution.

The main differences between sampling and variational techniques include the following:
\begin{enumerate}
    \item Unlike sampling-based methods, variational approaches will almost never find the globally optimal solution.
    \item However, we will always know if they have converged. In some cases, we will even have bounds on their accuracy.
    \item In practice, variational inference methods often scale better and are more amenable to techniques like stochastic gradient optimization, parallelization over multiple processors, and GPU acceleration.
\end{enumerate}

Although sampling methods were historically invented first (in the 1940s), variational inference techniques have grown increasingly popular in recent years and are now commonly used in many applications.

\subsection{The Kullback-Leibler Divergence}

\begin{figure}[!t]
    \centering
    \includegraphics[width=0.4\textwidth]{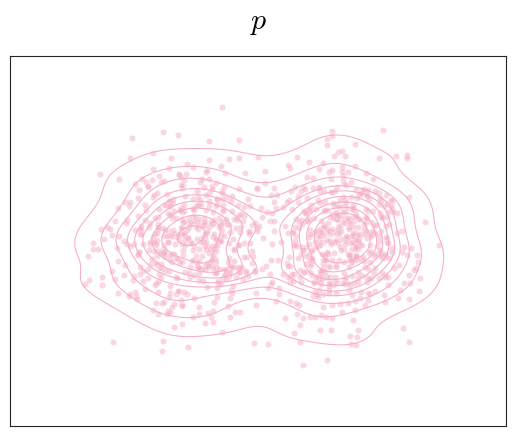} 
    \includegraphics[width=0.4\textwidth]{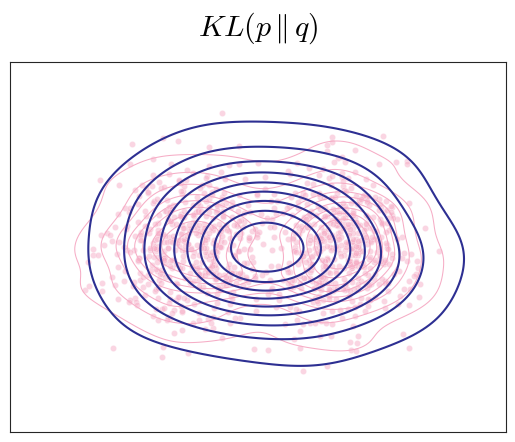} \\
    \includegraphics[width=0.4\textwidth]{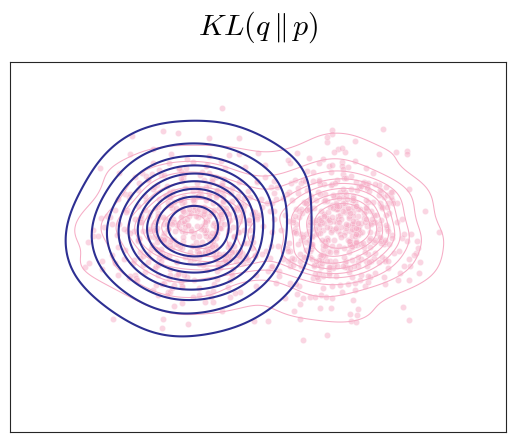}
    \includegraphics[width=0.4\textwidth]{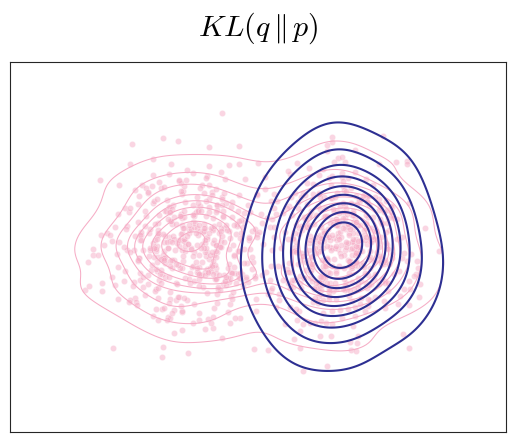}
    \caption{Fitting a unimodal approximating distribution $q$ (blue) to a multimodal $p$ (pink). Using $\text{KL}(p \, \| \, q)$ leads to a $q$ that tries to cover both modes. Using $\text{KL}(q \, \| \,p)$ forces $q$ to choose one of the two modes of $p$. } 
    \label{fig:kl_both_directions}
\end{figure}

To formulate inference as an optimization problem, we need to choose an approximating family $\mathcal{Q}$ and an optimization objective $J(q)$. This objective needs to capture the similarity between $q$ and $p$. The field of information theory provides us with a tool for this called the \textit{Kullback-Leibler divergence}.

\begin{definition}[Kullback-Leibler (KL) divergence] \label{def:kl_divergence}
    The KL divergence between two distributions $q$ and $p$ with discrete support is defined as
\begin{align}
    \text{KL}(q \,\|\, p) = \sum_\mathbf{x} q(\mathbf{x}) \log \frac{q(\mathbf{x})}{p(\mathbf{x})}.
\end{align}
\end{definition}
In information theory, this function is used to measure differences in information contained within two distributions. The KL divergence has the following properties that make it especially useful in our setting:
\begin{proposition} 
    $\text{KL}(q \,\|\, p) \geq 0 $ for all $q,p$.
\end{proposition}
\begin{proposition} 
    $\text{KL}(q \,\|\, p) = 0 $ if and only if $ q = p $.
\end{proposition}
These can be proven as an exercise. Note that $\text{KL}(q \,\|\, p) \neq \text{KL}(p \,\|\, q)$, i.e., the KL divergence is not symmetric. This is why we say that it's a \textit{divergence}, but not a distance. We will come back to this distinction shortly.

\subsection{The Variational Lower Bound}

How do we perform variational inference with a KL divergence? First, let's fix a form for $p$. We'll assume that $p$ is a general (discrete, for simplicity) undirected model of the form
\[
p(x_1,\ldots,x_n; \theta) = \frac{\Tilde{p}(x_1,\ldots,x_n ; \theta)}{Z(\theta)} =\frac{1}{Z(\theta)} \prod_{k} \phi_k(\mathbf{x}_k; \theta),
\]
where the $\phi_k$ are the factors and $Z(\theta)$ is the normalization constant. This formulation captures virtually all the distributions in which we might want to perform approximate inference, such as conditional distributions of directed models $p(\mathbf{x} \mid \mathbf{e}) = p(\mathbf{x}, \mathbf{e})/p(\mathbf{e})$ with evidence $\mathbf{e}$.

Given this formulation, optimizing $\text{KL}(q \,\|\, p)$ directly is not possible because of the potentially intractable normalization constant $Z(\theta)$. In fact, even evaluating $\text{KL}(q \,\|\, p)$ is not possible, because we need to evaluate $p$.

Instead, we will work with the following objective, which has the same form as the KL divergence, but only involves the unnormalized probability $\Tilde{p}(\mathbf{x}) = \prod_{k} \phi_k(\mathbf{x}_k; \theta) $:
\[
J(q) = \sum_\mathbf{x} q(\mathbf{x}) \log \frac{q(\mathbf{x})}{\Tilde{p}(\mathbf{x})}. 
\]
This function is not only tractable, it also has the following important property.
\begin{proposition}
    \begin{align*}
    J(q)
    &= \sum_\mathbf{x} q(\mathbf{x}) \log \frac{q(\mathbf{x})} {\Tilde{p}(\mathbf{x})} \\
    &= \sum_\mathbf{x} q(\mathbf{x}) \log \frac{q(\mathbf{x})}{p(\mathbf{x})} - \log Z(\theta) \\
    &= \text{KL}(q \,\|\, p) - \log Z(\theta).
    \end{align*}
\end{proposition}
\noindent Since $\text{KL}(q \,\|\, p) \geq 0 $, we get by rearranging terms that
\[
\log Z(\theta) = \text{KL}(q \,\|\, p) - J(q) \geq -J(q).
\]
Thus, $-J(q)$ is a \textit{lower bound} on the log partition function $\log Z(\theta)$. In many cases, $Z(\theta)$ has an interesting interpretation. For example, we may be trying to compute the conditional (posterior) probability $p(\mathbf{x} \mid \mathcal{D}) = p(\mathbf{x},\mathcal{D}) / p(\mathcal{D})$ of variables $\mathbf{x}$ given observed data $\mathcal{D}$ that plays the role of evidence. We assume that $p(\mathbf{x},\mathcal{D})$ is directed. In this case, minimizing $J(q)$ amounts to maximizing a lower bound on the log-likelihood $\log p(\mathcal{D})$ of the observed data.

Because of this property, $-J(q)$ is called the \textit{variational lower bound} or the \textit{evidence lower bound}. It is often written in the following form.
\begin{definition}[Evidence Lower Bound (ELBO)] \label{def:elbo}
    \begin{align}
        \log Z(\theta) \geq \mathbb{E}_{q(\mathbf{x})} [ \log \Tilde{p}(\mathbf{x}) - \log q(\mathbf{x}) ].
    \end{align}
\end{definition}
Crucially, the difference between $\log Z(\theta)$ and $-J(q)$ is precisely $\text{KL}(q \,\|\, p)$. Thus, by maximizing the evidence lower bound, we are minimizing $\text{KL}(q \,\|\, p)$ by ``squeezing'' it between $-J(q)$ and $\log Z(\theta)$.

\subsection{On the Choice of KL Divergence}

To recap, we have just defined an optimization objective for variational inference (the variational lower bound) and we have shown that maximizing this lower bound leads to minimizing the divergence $\text{KL}(q \,\|\, p)$.

Recall that $\text{KL}(q \,\|\, p) \neq \text{KL}(p \,\|\, q)$. Both divergences equal zero when $q = p$, but assign different penalties when $q \neq p$. This raises the question: why did we choose one over the other and how do they differ?

Perhaps the most important difference is computational: optimizing $\text{KL}(q \,\|\, p)$ involves an expectation with respect to $q$, while $\text{KL}(p \,\|\, q)$ requires computing expectations with respect to $p$, which is typically intractable even to evaluate.

However, choosing this particular divergence affects the returned solution when the approximating family $\mathcal{Q}$ does not contain the true $p$. Observe that $\text{KL}(q \,\|\, p)$ --- which is called the \textit{I-projection} or \textit{information projection} --- is infinite if $p(\mathbf{x}) = 0$ and $q(\mathbf{x}) > 0$.
This means that if $p(\mathbf{x}) = 0$ we must have $q(\mathbf{x}) = 0$. We say that $\text{KL}(q \,\|\, p)$ is zero-forcing for $q$ and it will typically under-estimate the support of $p$.

On the other hand, $\text{KL}(p \,\|\, q)$ --- known as the \textit{M-projection} or the \textit{moment projection} --- is infinite if $q(\mathbf{x}) = 0$ and $p(\mathbf{x}) > 0$. Thus, if $p(\mathbf{x}) > 0$ we must have $q(\mathbf{x}) > 0$. We say that $\text{KL}(p \,\|\, q)$ is zero-avoiding for $q$ and it will typically over-estimate the support of $p$. Figure \ref{fig:kl_both_directions} illustrates this phenomenon graphically.

Due to the above properties, we often call $\text{KL}(p \,\|\, q)$ the \textit{inclusive} KL divergence, and $\text{KL}(q \,\|\, p)$ the \textit{exclusive} KL divergence.

\subsection{Mean Field Inference}

The next step in our development of variational inference concerns the choice of approximating family $\mathcal{Q}$. The machine learning literature contains dozens of proposed ways to parametrize this class of distributions. These include exponential families, neural networks, Gaussian processes, latent variable models, and many other types of models.

However, one of the most widely used classes of distributions is simply the set of fully-factored $q(\mathbf{x}) = q_1(x_1) q_2(x_2) \cdots q_n(x_n)$. Here, each $q_i(x_i)$ is a categorical distribution over a one-dimensional discrete variable, which can be described as a one-dimensional table.

This choice of $\mathcal{Q}$ turns out to be easy to optimize over and works surprisingly well. It is perhaps the most popular choice when optimizing the variational bound. Variational inference with this choice of $\mathcal{Q}$ is called \textit{mean field} variational inference. It consists in solving the following optimization problem:
\[
\hat{q} = \argmin_{q_1, \ldots, q_n} J(q). 
\]
The standard way of performing this optimization problem is via coordinate descent over the $q_j$: we iterate over $j=1,2,\ldots,n$ and for each $j$ we optimize $\text{KL}(q \,\|\, p)$ over $q_j$ while keeping the other ``coordinates'' $q_{-j} = \prod_{i \neq j} q_i$ fixed.

Interestingly, the optimization problem for one coordinate has a simple closed form solution:
\[
\log q_j(x_j) \gets \mathbb{E}_{q_{-j}} \left[ \log \Tilde{p}(\mathbf{x}) \right] + \textrm{const.}
\]
Notice that both sides of the above equation contain univariate functions of $x_j$: we are thus replacing $q(x_j)$ with another function of the same form. The constant term is a normalization constant for the new distribution.

Notice also that on the right-hand side, we are taking an expectation of a sum of factors
\[
\log \Tilde{p}(\mathbf{x}) = \sum_k \log \phi(\mathbf{x}_k).
\]
Of these, only factors belonging to the Markov blanket of $x_j$ are a function of $x_j$, by the definition of the Markov blanket (Definitions \ref{def:markov_blanket_dag}, \ref{def:markov_blanket_mrf}). The rest are constant with respect to $x_j$ and can be pushed into the constant term.

This leaves us with an expectation over a much smaller number of factors. If the Markov blanket of $x_j$ is small (as is often the case), we are able to analytically compute $q(x_j)$. For example, if the variables are discrete with $K$ possible values, and there are $F$ factors and $N$ variables in the Markov blanket of $x_j$, then computing the expectation takes $O(K F K^N)$ time: for each value of $x_j$ we sum over all $K^N$ assignments of the $N$ variables, and in each case, we sum over the $F$ factors.

The result of this is a procedure that iteratively fits a fully-factored $q(\mathbf{x}) = q_1(x_1) q_2(x_2) \cdots q_n(x_n)$ that approximates $p$ in terms of $\text{KL}(q \,\|\, p)$. After each step of coordinate descent, we increase the variational lower bound, tightening it around $\log Z(\theta)$.

In the end, the factors $q_j(x_j)$ will not quite equal the true marginal distributions $p(x_j)$, but they will often be good enough for many practical purposes, such as determining $\max_{x_j} p(x_j)$.

\subsection{Loopy Belief Propagation}
\label{sec:loopy_bp}

As we have seen, the junction tree algorithm (Section \ref{sec:junction_tree}) has a running time that is potentially exponential in the size of the largest cluster (since we need to marginalize all the cluster's variables). For many graphs, it will be difficult to find a good junction tree and applying the algorithm will not be possible. However, some use cases do not require the exact solution that the junction tree algorithm provides. Instead, we may be satisfied with a quick approximate solution. \textit{Loopy belief propagation} (LBP) is a special case of variational inference that obtains approximate solutions for complex (non-tree structured) graphs \citep{frey1997revolution}. Below, we briefly describe LBP.

\subsubsection{Definition for Pairwise Models}

Suppose that we are given an MRF with pairwise potentials. The main idea of LBP is to disregard loops in the graph and perform message passing anyway. In other words, given an ordering on the edges, at each time $\mathcal{T}$ we iterate over a pair of adjacent variables $x_i, x_j$ in that order and simply perform the update
\[
m^{t+1}_{i\to j}(x_j) = \sum_{x_i} \phi(x_i) \phi(x_i,x_j) \prod_{\ell \in N(i) \setminus j} m^{t}_{\ell \to i}(x_i).
\]
We keep performing these updates for a fixed number of steps or until convergence (the messages don't change). Messages are typically initialized uniformly.

%{% include sidenote.html id=``note-factorgraphs'' note=``Arbitrary potentials can be handled using an algorithm called LBP on \textit{factor graphs}. We will include this material at some point in the future.'' %}. 

\subsubsection{Properties}

This heuristic approach often works surprisingly well in practice. In general, however, it may not converge and its analysis is still an area of active research. We know for example that it provably converges on trees and on graphs with at most one cycle. If the method does converge, its beliefs may not necessarily equal the true marginals, although  they will often be close in practice.

%{% include marginfigure.html id=``lbp'' url=``assets/img/lbp-performance.png'' description=``Marginals obtained via LBP compared to true marginals obtained from the junction tree algorithm on an intensive care monitoring task. Results are close to the diagonal, hence very similar.'' %} 

%\jm{Address this note in original text: We will return to this algorithm later and try to explain it as a special case of \textit{variational inference} algorithms.}

\begin{reading}
    \textit{Variational Inference}
    \begin{itemize}[leftmargin=*]
        \item \fullcite{jordan1999introduction}.
        \item \fullcite{murphy1999loopy}.
        %\item Chapters 10 in  \fullcite{bishop2006pattern}.
        \item \fullcite{wainwright_graphical_2008}.
        %\item Chapters 21 \& 22 in  \fullcite{murphy2012machine}.
        \item \fullcite{hoffman2013stochastic}.
        \item \fullcite{blei2017variational}.
    \end{itemize}
\end{reading}

%%%%%%
%% MAP inference
%%%%%%

\section{Approximate MAP Inference}
\label{sec:approximate_map}

\subsection{Linear Programming for Approximate MAP Inference}
We conclude this chapter by describing efficient approximate inference procedures for special-cases of MAP inference---paralleling the final section of the previous chapter, which focused on efficient \textit{exact} inference algorithms for MAP queries.

Although graph cut-based methods recover the exact MAP assignment (Section~\ref{sec:exact_map_graph_cuts}), they are only applicable in certain restricted classes of MRFs. The algorithms we see next solve the MAP problem approximately, but apply to much larger classes of graphical models.

Our first approximate inference strategy consists of reducing MAP inference to integer linear programming. Linear programming (LP) --- also known as linear optimization --- refers to a class of problems of the form
\[
\begin{array}{ll}
\text{minimize (over $\mathbf{x}$)} & \bf{c} \cdot \mathbf{x} \\
\text{subject to} & A \mathbf{x} \preceq \bf{b}
\end{array}
\]
where $\mathbf{x} \in \mathbb{R}^n$ is the optimization variable, and $\mathbf{c}, \mathbf{b} \in \mathbb{R}^n$ and $A\in \mathbb{R}^{n\times n}$ are problem parameters.

Problems of this form are found in almost every field of science and engineering. They have been extensively studied since the 1930s, which has led to extensive theory. A major breakthrough of applied mathematics in the 1980s was the development of polynomial-time algorithms for linear programming \citep{karmarkar1984new,adler1989implementation} %https://en.wikipedia.org/wiki/Karmarkar%27s_algorithm
and practical tools that can solve very large LP instances in a reasonable time (e.g., 100,000 variables or more; \citealt{manual1987ibm}). %https://en.wikipedia.org/wiki/CPLEX

Integer linear programming (ILP) is an extension of linear programming in which we also require that $\mathbf{x} \in \{0, 1\}^n$. Unfortunately, this makes optimization considerably more difficult, and ILP is in general NP-hard. Nonetheless, there are many heuristics for solving ILP problems in practice, and commercial solvers can handle instances with thousands of variables or more.

One of the main techniques for solving ILP problems is \textit{rounding}. In rounding, we relax the requirement that $\mathbf{x} \in \{0, 1\}^n$ into $0 \leq \mathbf{x} \leq 1$ and solve the resulting LP. Finally, we round the LP solution to its nearest integer value. This approach works surprisingly well in practice and has theoretical guarantees for some classes of ILPs.

\subsubsection{Formulating MAP Inference as ILP}

For simplicity, let's look at MAP in pairwise MRFs. Consider the MRF $\mathcal{G} = (\mathbf{V}, \mathbf{E})$. We can reduce the MAP objective to integer linear programming by introducing two types of indicator variables:
\begin{enumerate}
    \item A variable $\mu_i(x_i)$ for each $i \in \mathbf{V}$ and state $x_i$.
    \item A variable $\mu_{ij}(x_i,x_j)$ for each edge $(i,j) \in \mathbf{E}$ and pair of states $x_i,x_j$.
\end{enumerate}

We can rewrite the MAP objective in terms of these variables as
\[
\max_\mu \sum_{i \in \mathbf{V}} \sum_{x_i} \theta_i (x_i) \mu_i(x_i) + \sum_{i,j \in \mathbf{E}} \sum_{x_i, x_j} \theta_{ij} (x_i, x_j) \mu_{ij}(x_i, x_j).
\]
We would like to optimize over these $\mu$'s. For that, we also need to introduce constraints. First, we need to force each cluster to choose a local assignment:
\begin{align*}
\mu_i(x_i) & \in \{0,1\} \; \forall \, i, x_i \\
\sum_{x_i} \mu_i(x_i) & = 1\; \forall \, i \\
\mu_{ij}(x_i,x_j) & \in \{0,1\} \; \forall \, i,j \in \mathbf{E}, x_i, x_j \\
\sum_{x_i, x_j} \mu_{ij}(x_i, x_j) & = 1 \; \forall \, i,j \in \mathbf{E}.
\end{align*}
These assignments must also be consistent:
\begin{align*}
\sum_{x_i} \mu_{ij}(x_i, x_j) & = \mu_j(x_j) \; \forall \, i,j \in \mathbf{E}, x_j \\
\sum_{x_j} \mu_{ij}(x_i, x_j) & = \mu_i(x_i) \; \forall \, i,j \in \mathbf{E}, x_i.
\end{align*}
Together, these constraints along with the MAP objective yield an integer linear program whose solution equals the MAP assignment. This ILP is still NP-hard, but we have a simple way to transform this into an (easy to solve) LP via relaxation. This is the essence of the linear programming approach to MAP inference.

In general, this method will only give approximate solutions. An important special case are tree-structured graphs, in which the relaxation is guaranteed to always return integer solutions, which are in turn optimal~\citep{koller2009probabilistic}.
%{% include sidenote.html id=``note-tree'' note=``See e.g., the textbook of Koller and Friedman for a proof and a more detailed discussion.'' %}.

\subsection{Dual Decomposition}

Let us now look at another way to transform the MAP objective into a more amenable optimization problem. Suppose that we are dealing with an MRF of the form
\[
\max_\mathbf{x} \sum_{i \in \mathbf{V}} \theta_i (x_i) + \sum_{f \in \mathcal{F}} \theta_f (\mathbf{x}_f), 
\]
where $f \in \mathcal{F}$ denote arbitrary factors (e.g., the edge potentials in a pairwise MRF). Let us use $p^*$ to denote the optimal value of this objective and let $\mathbf{x}^*$ denote the optimal assignment.
The above objective is difficult to optimize because the potentials are coupled. Consider for a moment an alternative objective where we optimize the potentials separately:
\[
\sum_{i \in \mathbf{V}} \max_{x_i} \theta_i (x_i) + \sum_{f \in \mathcal{F}} \max_{\mathbf{x}^f} \theta_f (\mathbf{x}^f).
\]
This would be easy to optimize, but would only give us an upper bound on the value of the true MAP assignment. To make our relaxation tight, we would need to introduce constraints that encourage consistency between the potentials:
\[
x^f_i - x_i = 0 \; \forall f, \forall i \in f. 
\]
The dual decomposition approach consists in softening these constraints in order to achieve a middle ground between the two optimization objective defined above.

We will achieve this by first forming the \textit{Lagrangian} for the constrained problem, which is
\[
L(\delta, \mathbf{x}^f, \mathbf{x}) = \sum_{i \in \mathbf{V}} \theta_i (x_i) + \sum_{f \in \mathcal{F}} \theta_f (\mathbf{x}^f) + \sum_{f \in \mathcal{F}} \sum_{i \in f} \sum_{x'_i} \delta_{fi}(x_i')\left( \mathbb{I}_{x'_i = x_i} - \mathbb{I}_{x'_i = x^f_i} \right).
\]
Here, the $\delta$ variables are called \textit{Lagrange multipliers}. Each of them is associated with a constraint, and they allow us to move the hard consistency constraints into the optimization objective. The Lagrangian is a standard technique from constrained optimization that converts a constrained problem into an unconstrained one via these multipliers.
Intuitively, each multiplier $\delta_{fi}(x_i')$ adjusts the objective to penalize disagreement between the local variable copies $x^f_i$ and the global variable $x_i$.
By optimizing the Lagrangian over both the original variables and the multipliers, dual decomposition seeks a solution that approximately satisfies the original constraints while breaking the problem into smaller, tractable components.

%{% include sidenote.html id=``note-cvxopt'' note=``There is a very deep and powerful theory of constrained optimization centered around Lagrangians. We refer the reader to a course on [convex optimization](http://stanford.edu/class/ee364a/) for a thorough discussion.'' %}. 

Observe that $\mathbf{x}, \mathbf{x}^f = \mathbf{x}^*$ is a valid assignment to the Lagrangian. Its value equals $p^*$ for any $\delta$, since the Lagrange multipliers are simply multiplied by zero. This shows that the Lagrangian is an upper bound on $p^*$:
\[
L(\delta) := \max_{\mathbf{x}^f, \mathbf{x}} L(\delta, \mathbf{x}^f, \mathbf{x}) \geq p^* \; \forall \delta.
\]
In order to get the tightest such bound, we may optimize $L(\delta)$ over $\delta$. It turns out that by the theory of Lagarange duality, at the optimal $\delta^*$, this bound will be exactly tight, i.e.,
\[
L(\delta^*) = p^*. 
\]
It is actually not hard to prove this in our particular setting. To see that, note that we can reparametrize the Lagrangian as:
\begin{align*}
L(\delta)
&\coloneqq \sum_{i \in \mathbf{V}} \max_{x_i} \left( \theta_i (x_i) + \sum_{f:i \in f} \delta_{fi}(x_i) \right) + \sum_{f \in \mathcal{F}} \max_{\mathbf{x}^f} \left( \theta_f (\mathbf{x}^f) + \sum_{i \in f} \delta_{fi}(x_i) \right) \\
&= \sum_{i \in \mathbf{V}} \max_{x_i} \bar \theta_{i}^\delta (x_i) + \sum_{f \in \mathcal{F}} \max_{\mathbf{x}^f} \bar \theta_{f}^\delta (\mathbf{x}^f).
\end{align*}

Suppose we can find dual variables $\bar \delta$ such that the local maximizers of $\bar \theta_{i}^{\bar \delta} (x_i)$ and $\bar \theta_{f}^{\bar \delta} (\mathbf{x}^f)$ agree. In other words, we can find a $\bar{\mathbf{x}}$ such that $\bar x_i \in \arg\max_{x_i} \bar \theta_{i}^{\bar \delta} (x_i)$ and $\bar{\mathbf{x}}^f \in \arg\max_{\mathbf{x}^f} \bar \theta_{f}^{\bar \delta} (\mathbf{x}^f)$. Then we have that
\[
L(\bar \delta)
= \sum_{i \in \mathbf{V}} \bar \theta_{i}^{\bar\delta} (\bar x_i) + \sum_{f \in \mathcal{F}} \bar \theta_{f}^{\bar\delta} (\bar{\mathbf{x}}^f)
= \sum_{i \in \mathbf{V}} \theta_i (\bar x_i) + \sum_{f \in \mathcal{F}} \theta_f (\bar{\mathbf{x}}^f).
\]
The first equality follows by definition of $L(\delta)$, while the second follows because terms involving Lagrange multipliers cancel out when $\mathbf{x}$ and $\mathbf{x}^f$ agree.

On the other hand, we have by the definition of $p^*$ that
\[
\sum_{i \in \mathbf{V}} \theta_i (\bar x_i) + \sum_{f \in \mathcal{F}} \theta_f (\bar{\mathbf{x}}^f) \leq p^* \leq L(\bar\delta)
\]
which implies that $L(\bar\delta) = p^*$.

This argument has shown two things:
\begin{enumerate}
    \item The bound given by the Lagrangian can be made tight for the right choice of $\delta$.
    \item To compute $p^*$, it suffices to find a $\delta$ at which the local sub-problems agree with each other. This happens surprisingly often in practice.
\end{enumerate}

\subsubsection{Minimizing the Objective}

There exist several ways of computing $L(\delta^*)$, of which we will give a brief overview.

Since the objective $L(\delta)$ is continuous and convex, we may minimize it using subgradient descent. Let 
\[
\bar x_i \in \argmax_{x_i} \bar \theta_{i}^{\delta} (x_i)
\]
and let 
\[
\bar{\mathbf{x}}^f \in \argmax_{\mathbf{x}^f} \bar \theta_{f}^\delta (\mathbf{x}^f).
\]
It can be shown that the gradient $g_{fi}(x_i)$ of $L(\delta)$ w.r.t. $\delta_{fi}(x_i)$ equals $1$ if $\bar x_i^f \neq \bar x_i$ and zero otherwise; similarly, $g_{fi}(x_i^f)$ equals $-1$ if $\bar x_i^f \neq \bar x_i$ and zero otherwise. This expression has the effect of decreasing $\bar \theta_{i}^{\delta} (\bar x_i)$ and increasing $\bar \theta_{f}^\delta (\bar{\mathbf{x}}^f)$, thus bringing them closer to each other.

%{% include sidenote.html id=``note-affine'' note=``The objective is a pointwise max of a set of affine functions.'' %}, 

To compute these gradients we need to perform the operations $\bar x_i \in \argmax_{x_i} \bar \theta_{i}^{\delta} (x_i)$ and $\bar{\mathbf{x}}^f \in \argmax_{\mathbf{x}^f} \bar \theta_{f}^\delta (\mathbf{x}^f)$. This is possible if the scope of the factors is small, if the graph has small treewidth, if the factors are constant on most of their domain, and in many other useful special cases.

An alternative way of minimizing $L(\delta)$ is via block coordinate descent. A typical way of forming blocks is to consider all the variables $\delta_{fi}(x_i)$ associated with a fixed factor $f$. This results in updates that are very similar to loopy max-product belief propagation. In practice, this method may be faster than subgradient descent, is guaranteed to decrease the objective at every step, and does not require tuning a step-size parameter. Its drawback is that it does not find the global minimum (since the objective is not \textit{strongly} convex).

\subsubsection{Recovering the MAP Assignment}

As we have seen above, if a solution $\mathbf{x}, \mathbf{x}^f$ agrees on the factors for some $\delta$, then we can guarantee that this solution is optimal.

If the optimal $\mathbf{x}, \mathbf{x}^f$ do not agree, finding the MAP assignment from this solution is still NP-hard. However, this is usually not a big problem in practice. From the point of view of theoretical guarantees, if each $\bar \theta_i^{\delta^*}$ has a unique maximum, then the problem will be decodable. If this guarantee is not met by all variables, we can clamp the ones that can be uniquely decoded to their optimal values and use exact inference to find the remaining variables' values.

\subsection{Other Methods}

\subsubsection{Local Search}

A more heuristic-type solution consists in starting with an arbitrary assignment and perform ``moves'' on the joint assignment that locally increase the probability. This technique has no guarantees. However, we can often use prior knowledge to come up with highly effective moves. Therefore, in practice, local search may perform extremely well.

\subsubsection{Branch and Bound}

Alternatively, one may perform exhaustive search over the space of assignments, while pruning branches that can be provably shown not to contain a MAP assignment. The LP relaxation or its dual can be used to obtain upper bounds useful for pruning trees.

\subsubsection{Simulated Annealing}

A third approach is to use sampling methods (e.g., Metropolis-Hastings) to sample from $p_t(\mathbf{x}) \propto \exp(\frac{1}{t} \sum_{c \in \mathbf{C}} \theta_c (\mathbf{x}_c ))$. The parameter $T$ is called the temperature. When $t \to \infty $, $p_t$ is close to the uniform distribution, which is easy to sample from. As $t \to 0$, $p_t$ places more weight on $\arg\max_\mathbf{x} \sum_{c \in \mathbf{C}} \theta_c (\mathbf{x}_c )$, the quantity we want to recover. However, since this distribution is highly peaked, it is also very difficult to sample from.

The idea of simulated annealing is to run a sampling algorithm starting with a high $T$, and gradually decrease it, as the algorithm is being run. If the ``cooling rate'' is sufficiently slow, we are guaranteed to eventually find the mode of our distribution. In practice, however, choosing the rate requires a lot of tuning. This makes simulated annealing somewhat difficult to use in practice.

\begin{reading}
    \textit{Approximate MAP Inference}
    \begin{itemize}[leftmargin=*]
        \item Chapter 8 in \fullcite{sra2011optimization}.
        \item Chapter 13 in \fullcite{koller2009probabilistic}.
        \item \fullcite{wainwright_graphical_2008}.
    \end{itemize}
\end{reading}

%%%%%%%%%%%%%%%%%%%%%%%%
%% Learning
%%%%%%%%%%%%%%%%%%%%%%%%

\chapter{Learning}
\label{sec:learning}

We now turn our attention to the third and last part of this tutorial: \textit{learning}. Given a dataset, we would like to fit a model that will make useful predictions on various tasks that we care about.

A graphical model has two components: (1) the graph structure and (2) the parameters of the factors induced by this graph. These components lead to two different learning tasks:
\begin{enumerate}
    \item \textit{Structure learning}, where we want to estimate the graph (i.e., determine from data how the variables depend on each other).
    \item \textit{Parameter learning}, where the graph structure is known and we want to estimate the factors.
\end{enumerate}
The chapter begins in Section \ref{sec:learning_theory_basics} with a basic review of learning theory. In Section \ref{sec:learning_directed_models} we address parameter learning in directed models, while Section \ref{sec:learning_undirected_models} explores the undirected case. It turns out that the former will admit easy closed form solutions, while the latter will involve potentially intractable numerical optimization techniques. In Section \ref{sec:learning_latent_models}, we consider latent variable models, which contain unobserved hidden variables that succinctly model the event of interest. Section \ref{sec:learning_bayesian_models} concludes the chapter with an overview of Bayesian learning, a general statistical framework that offers advantages over more-conventional approaches. 

\section{Learning Theory Basics}
\label{sec:learning_theory_basics}

\subsection{The Learning Task}
\label{sec:learning_theory_basics_the_learning_task}

Before, we start our discussion of learning, let's first reflect on what it means to fit a model and what is a desirable objective for this task.

Let's assume that the domain is governed by some underlying distribution $p^*$. We are given a dataset $\mathcal{D}$ of $N$ samples from $p^*$. The standard assumption is that the data instances are independent and identically distributed (iid). We are also given a family of models $\mathcal{M}$, and our task is to learn some ``good'' model in $\mathcal{M}$ defined by a distribution $p$. For example, a family $\mathcal{M}$ might consist of all Bayesian networks with a given graph structure and all possible choices of the corresponding conditional probability distribution tables.

The goal of learning is to return a model that accurately captures the distribution $p^*$ from which our dataset $\mathcal{D}$ was sampled. Note that this is not achievable in general because limited data may only provide a rough approximation of the true underlying distribution. 
Still, we want to somehow select the ``best'' approximation to the underlying distribution $p^*$.

What is ``best'' in this case? It depends on what we want to do. Tasks of interest could include:
\begin{itemize}
    \item \textit{Density estimation.} We are interested in the full distribution (so later we can compute whatever conditional probabilities we want).
    \item \textit{Specific prediction tasks.} We are using the distribution to make a prediction (e.g., is this email spam or not?).
    \item \textit{Structure or knowledge discovery.} We are interested in the model itself (e.g., how do some genes interact with each other?).
\end{itemize}

\subsubsection{Maximum Likelihood}

Let's assume that we want to learn the full distribution so that later we can answer any probabilistic inference query. In this setting we can view the learning problem as \textit{density estimation}. We want to construct a $p$ as ``close'' as possible to $p^*$. How do we evaluate ``closeness''? We will again use the KL divergence (Definition \ref{def:kl_divergence}), which we explored in our discussion of variational inference (Section \ref{sec:variational_methods}). In particular, we aim to find a $p$ that minimizes
\[
\text{KL}(p^* \,\|\, p) = \sum_\mathbf{x} p^*(\mathbf{x}) \log \frac{p^*(\mathbf{x})}{p(\mathbf{x})} = -H(p^*) - \mathbb{E}_{\mathbf{x} \sim p^*} [ \log p(\mathbf{x}) ].
\]
The first term does not depend on $p$; hence, minimizing the KL divergence is equivalent to maximizing the expected log-likelihood
\[
\mathbb{E}_{\mathbf{x} \sim p^*} [ \log p(\mathbf{x}) ]. 
\]
This objective asks that $p$ assign high probability to instances sampled from $p^*$, so as to reflect the true distribution. Although we can now compare models, since we are not computing $H(p^*)$, we don’t know how close we are to the optimum.

However, there is still a problem: in general we do not know $p^*$. We will thus approximate the expected log-likelihood $\mathbb{E}_{\mathbf{x} \sim p^*} [ \log p(\mathbf{x}) ]$ with the empirical log-likelihood (a Monte Carlo estimate):
\[
\mathbb{E}_{\mathbf{x} \sim p^*} [ \log p(\mathbf{x}) ] \approx \frac{1}{|\mathcal{D}|} \sum_{\mathbf{x} \in \mathcal{D}} \log p(\mathbf{x}),
\]
where $\mathcal{D}$ is a dataset drawn iid from $p^*$. Maximum likelihood learning is then defined as producing the estimate
\[
\hat{p} = \argmax_{p \in \mathcal{M}} \frac{1}{|\mathcal{D}|} \sum_{\mathbf{x} \in \mathcal{D}} \log p(\mathbf{x}).
\]

\subsubsection{A Step-by-Step Proof with the Empirical Distribution}

How do we know that the estimate defined above will perform maximum likelihood learning? In particular, we'd like this estimate to maximize the expected likelihood (or log-likelihood). Here we will give a step-by-step proof using the notion of an \textit{empirical distribution}.

% \vk{it seems a bit unusual to introduce this twice? do we need this subsection?}

The Monte Carlo samples provide empirical samples from the true data distribution. Suppose we define a distribution using the $N$ samples from $\mathcal{D}$, called the empirical distribution $\tilde{p}$. This empirical distribution is defined such that if we find some new data point $\mathbf{x}$, its probability is proportional to the number of times it is found in the training data set $\mathcal{D}$, which may be zero times:
\[
\tilde{p}(\mathbf{x}) = \frac{1}{N}\sum\limits_{i=1}^N \delta(\mathbf{x}, \bar{\mathbf{x}}_i),
\]
where $\bar{\mathbf{x}}_i \in \mathcal{D}$ denotes a sample in our dataset, and $\delta$ is a function that evaluates to 1 if its two arguments are the same and 0 otherwise (similar to the Kronecker delta).

% \jm{What is $\delta(x, \bar{x}_i)$? This is not defined.} \vk{its some kind of indicator or kernel function that we should define}

Now suppose we have a parameterized family of distributions $p(\mathbf{x} ; \theta)$, specified by a parameter $\theta \in \Theta$.
The KL Divergence between the density defined by the empirical distribution $\tilde{p}$ and a distribution $p(\mathbf{x} ; \theta)$ from our model family is
\begin{align*}
    \text{KL}(\tilde{p}(\mathbf{x})  \; || \;  p(\mathbf{x} ; \theta)) &= \sum_\mathbf{x} \tilde{p}(\mathbf{x}) \log \frac{\tilde{p}(\mathbf{x})}{p(\mathbf{x} ; \theta)} \\
&= \sum_\mathbf{x} \tilde{p}(\mathbf{x}) \log \tilde{p}(\mathbf{x}) - \sum_\mathbf{x} \tilde{p}(\mathbf{x}) \log  p(\mathbf{x} ; \theta).
\end{align*}
We want to \textit{minimize this divergence} over $\theta \in \Theta$.  We ignore the left term (equivalent to $-H(\tilde{p})$, as shown earlier) because it does not depend on our model parameters $\theta$. Our minimization problem then becomes
\[
\argmin_{\theta \in \Theta} \text{KL}(\tilde{p}(\mathbf{x}) \; || \; p(\mathbf{x} ; \theta)) = \argmin_{\theta \in \Theta} - \sum_\mathbf{x} \tilde{p}(\mathbf{x}) \log p(\mathbf{x} ; \theta).
\]
We can then convert the problem to a maximization problem by swapping the sign:
\[
\argmax_{\theta \in \Theta} - \text{KL}(\tilde{p}(\mathbf{x}) \; || \; p(\mathbf{x} ; \theta)) = \argmax_{\theta \in \Theta} \sum_\mathbf{x} \tilde{p}(\mathbf{x}) \log  p(\mathbf{x} ; \theta).
\]
Plugging in our definition of the empirical distribution $\tilde{p}$, we see that this then becomes
\[
\argmax_{\theta \in \Theta} \sum_\mathbf{x} \frac{1}{N} \sum\limits_{i=1}^N \delta(\mathbf{x}, \bar{\mathbf{x}}_i) \log p(\mathbf{x} ; \theta).
\]
Swapping the order of two finite sums is always legal, and the sum over $\mathbf{x}$ only provides non-zero components when $\mathbf{x}=\bar{\mathbf{x}}_i$. The optimization problem thus reduces to maximizing the expected log-likelihood:
\[
\argmax_{\theta \in \Theta} \frac{1}{N} \sum\limits_{i=1}^N \log  p(\bar{\mathbf{x}}_i \hspace{1pt};\hspace{1pt} \theta).
\]

\paragraph{An Illustrative Example}

As a simple example, consider estimating the outcome of a biased coin. Our dataset is a set of tosses and our task is to estimate the probability of heads ($h$) or tails ($t$) on the next flip. We assume that the process is controlled by a probability distribution $p^*(x)$ where $x \in \{h,t\}$. Our class of models $\mathcal{M}$ is going to be the set of all probability distributions over $\{h,t\}$.

How should we choose $p$ from $\mathcal{M}$ if 60 out of 100 tosses are heads? Let's assume that $p(x=h)=\theta$ and $p(x=t)=1-\theta$. If our observed data are $\mathcal{D} = \{h,h,t,h,t\}$, our likelihood becomes
\[
\prod_i p(x_i ) = \theta \cdot \theta \cdot (1 - \theta) \cdot \theta \cdot (1 - \theta)
\]
and maximizing this yields $\theta = 60\%$.  More generally, our log-likelihood function is simply
\[
\ell(\theta) = \text{\# heads} \cdot \log(\theta) + \text{\# tails} \cdot \log(1 - \theta),
\]
for which the optimal solution is
\[
\hat{\theta} = \frac{\text{\# heads}}{\text{\# heads} + \text{\# tails}}. 
\]

\subsection{Likelihood, Loss, and Risk}

We may now generalize this by introducing the concept of a \textit{loss function}. A loss function $L(\mathbf{x},p)$ measures the loss that a model distribution $p$ makes on a particular instance $\mathbf{x}$. Assuming instances are sampled from some distribution $p^*$, our goal is to
find the model that minimizes the \textit{expected loss} or \textit{risk},
\[
\mathbb{E}_{\mathbf{x} \sim p^*} [ L(\mathbf{x}, p) ] \approx \frac{1}{|\mathcal{D}|} \sum_{\mathbf{x} \in \mathcal{D}} L(\mathbf{x},p).
\]
Notice that the loss function which corresponds to maximum likelihood estimation is the log loss,  $L(\mathbf{x},p) = -\log p(\mathbf{x})$.

Another example of a loss is the conditional log-likelihood. Suppose we want to predict a set of variables $\mathbf{y}$ given $\mathbf{x}$ (e.g., for segmentation or stereo vision).
We concentrate on predicting $p(\mathbf{y} \mid \mathbf{x})$, and use a conditional loss function $L(\mathbf{x},\mathbf{y},p) = -\log p(\mathbf{y} \mid \mathbf{x})$.
Since the loss function only depends on $p(\mathbf{y} \mid \mathbf{x})$, it suffices to estimate the conditional distribution, not the joint.
This is the objective function we use to train CRFs (Section \ref{sec:crf}).

Suppose next that our ultimate goal is structured prediction: given $\mathbf{x}$ we predict $\mathbf{y}$ via $\argmax_\mathbf{y} p(\mathbf{y} \mid \mathbf{x})$. What loss function should we use to measure error in this setting? One reasonable choice would be the classification error:
\[
\mathbb{E}_{(\mathbf{x},\mathbf{y})\sim p^*} [\mathbb{I}\{ \exists \, \mathbf{y}' \neq \mathbf{y} : p(\mathbf{y}' \mid \mathbf{x}) \geq p(\mathbf{y} \mid \mathbf{x}) \}].
\]
This is the probability that we predict the wrong assignment over all $(\mathbf{x}, \mathbf{y})$ pairs sampled from $p^*$. A somewhat better choice might be the Hamming loss, which counts the number of variables in which the MAP assignment differs from the ground truth label. There also exists a fascinating line of work on generalizations of the hinge loss to CRFs, which leads to a class of models called \textit{structured support vector machines} \citep{tsochantaridis2004support,tsochantaridis2005large,finley2008training}.

The moral of the story here is that it often makes sense to choose a loss that is appropriate to the task at hand (e.g., prediction rather than full density estimation).

\subsection{Empirical Risk and Overfitting}

\begin{figure}[!t]
    \centering
    \includegraphics[width=0.9\linewidth]{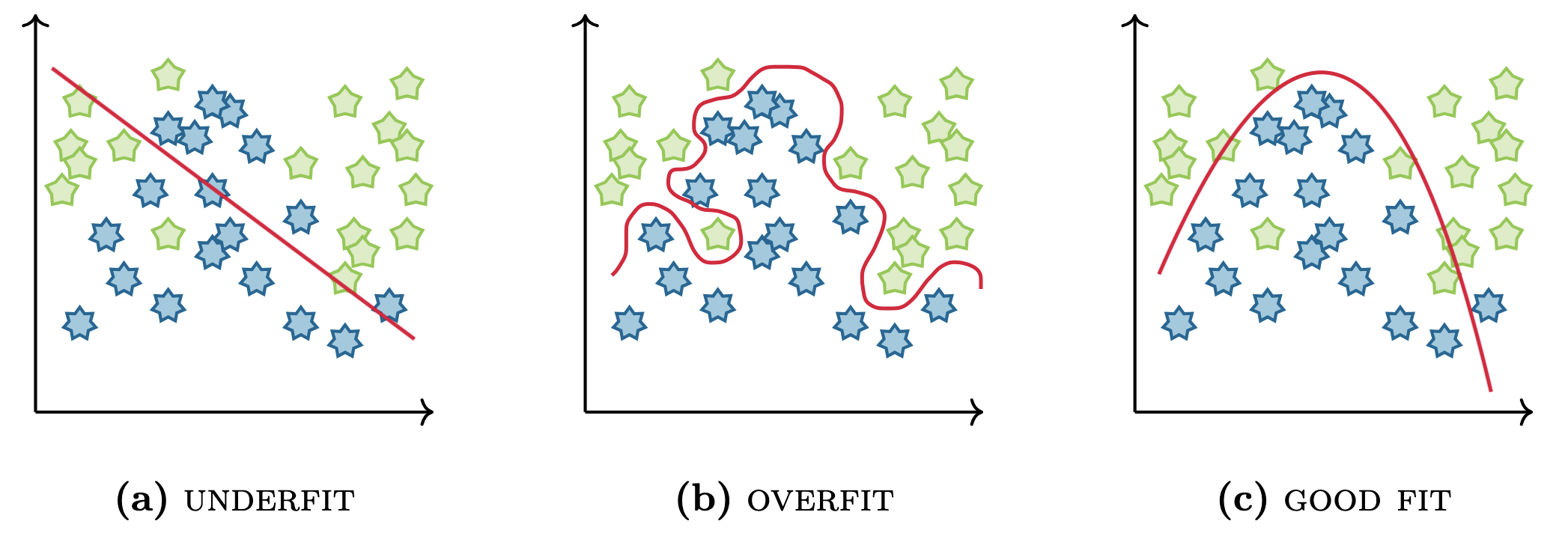}
    \caption{Underfit (\textbf{a}), overfit (\textbf{b}), and well fit (\textbf{c}) decision boundaries in a toy classification problem. The underfit linear model displays large bias, corresponding to high error rates at both training and test time. The overfit  model is prone to high variance as it learns spurious patterns in the training data, resulting in low training error but high test error. A good fit will display greater robustness, offering relatively low training and test error.}
    \label{fig:overfit_underfit}
\end{figure}

Empirical risk minimization can easily \textit{overfit} the observed data, overshooting our goal of learning the true generating distribution and instead learning spurious patterns in our random sample. Conversely, we do not want to \textit{underfit} with an overly simplistic model that is not sufficiently informative. Thus, one of the main challenges of machine learning is to choose a model that is sufficiently rich to be useful, yet not so complex as to overfit the training set (Figure \ref{fig:overfit_underfit}). In tackling this challenge, we must grapple with the fact that the data that we have access to is only a finite sample: there is generally a vast amount of samples that we have never seen, and we cannot generally verify the extent to which our sample is representative of the population. Nevertheless, we aim for some degree of \textit{generalizability}: ideally, a machine learning model demonstrates robustness by generalizing well to such ``never-seen'' samples. In this subsection, we briefly cover problems related to overfitting and generalization.

\paragraph{Generalization Error}

At training, we minimize the empirical loss
\[
\frac{1}{|\mathcal{D}|} \sum_{\mathbf{x} \in \mathcal{D}} \log p(\mathbf{x}). 
\]
However, we are actually interested in minimizing
\[
\mathbb{E}_{\mathbf{x} \sim p^*} [ \log p(\mathbf{x}) ]. 
\]
We cannot guarantee with certainty the quality of our learned model. This is because the data $\mathcal{D}$ is sampled stochastically from $p^*$, and we might get an unlucky sample. The goal of learning theory is to prove that the model is approximately correct: for most $\mathcal{D}$, the learning procedure returns a model whose error is low. There exists a large literature that quantifies the probability of observing a given error between the empirical and the expected loss under a particular type of model and a particular dataset size. %\jm{Find citations.}

\renewcommand\fbox{\fcolorbox{white}{white}}

\begin{figure}[!t]
    \centering

\fbox{
\begin{tabular}{c}
\begin{tikzpicture}[scale=0.8]
    % Draw the square
    \draw[thick, CornflowerBlue, fill=CornflowerBlue!20] (0,0) rectangle (4,4);
    
    % Draw the circle inside the square
    \draw[thick, CarnationPink, fill=CarnationPink!10] (2,2) circle (1);

    % Draw points
    \foreach \x/\y in {2/2, 1.8/2.3, 2.5/1.8, 2.1/2.5, 2.2/1.9, 2.3/2.3, 1.8/1.7, 1.9/1.8, 1.7/1.8, 1.6/2.1} {
        \node[star, star points=8, thick, draw=OrangeRed, fill=OrangeRed!60, inner sep=2pt] at (\x,\y) {};
    }
\end{tikzpicture} \\
\footnotesize \textsc{low bias $|$ low variance} 
\end{tabular}
}
%%%
\fbox{

\begin{tabular}{c}
\begin{tikzpicture}[scale=0.8]
    % Draw the square
    \draw[thick, CornflowerBlue, fill=CornflowerBlue!20] (0,0) rectangle (4,4);
    
    % Draw the circle inside the square
    \draw[thick, CarnationPink, fill=CarnationPink!10] (2,2) circle (1);

    % Draw points
    \foreach \x/\y in {3/2, 2.8/1.3, 2.5/1.8, 2.1/0.8, 1.2/1.5, 1.3/1.2, 2.8/1.7, 2.6/2.8, 1.4/2.8, 1.3/2.1} {
        \node[star, star points=8, thick, draw=OrangeRed, fill=OrangeRed!60, inner sep=2pt] at (\x,\y) {};
    }
\end{tikzpicture} \\
\footnotesize \textsc{low bias $|$ high variance} 
\end{tabular}

} \\
\vspace{5mm}
%%%
\fbox{
\begin{tabular}{c}
\begin{tikzpicture}[scale=0.8]
    % Draw the square
    \draw[thick, CornflowerBlue, fill=CornflowerBlue!20] (0,0) rectangle (4,4);
    
    % Draw the circle inside the square
    \draw[thick, CarnationPink, fill=CarnationPink!10] (2,2) circle (1);

    % Draw points
    \foreach \x/\y in {2/0.2, 1.8/0.3, 2.5/0.7, 2.1/0.5, 2.2/0.4, 2.3/0.3, 1.8/0.7, 1.6/0.5, 1.4/0.8, 1.5/0.2} {
        \node[star, star points=8, thick, draw=OrangeRed, fill=OrangeRed!60, inner sep=2pt] at (\x,\y) {};
    }
\end{tikzpicture} \\
\footnotesize \textsc{high bias $|$ low variance} 
\end{tabular}
}
%%%
\fbox{
\begin{tabular}{c}
\begin{tikzpicture}[scale=0.8]
    % Draw the square
    \draw[thick, CornflowerBlue, fill=CornflowerBlue!20] (0,0) rectangle (4,4);
    
    % Draw the circle inside the square
    \draw[thick, CarnationPink, fill=CarnationPink!10] (2,2) circle (1);

    % Draw points
    \foreach \x/\y in {0.5/3, 0.8/2.3, 1.5/3.2, 2.1/3.4, 0.2/2.9, 0.3/2.2, 0.8/2.7, 2.6/3.7, 0.4/3.8, 0.3/2.6} {
        \node[star, star points=8, thick, draw=OrangeRed, fill=OrangeRed!60, inner sep=2pt] at (\x,\y) {};
    }
\end{tikzpicture} \\
\footnotesize \textsc{high bias $|$ high variance} 
\end{tabular}
}

    \caption{Samples with low to high bias and variance when the true distribution is centered on the pink circle.}
    \label{fig:bias_variance}
\end{figure}
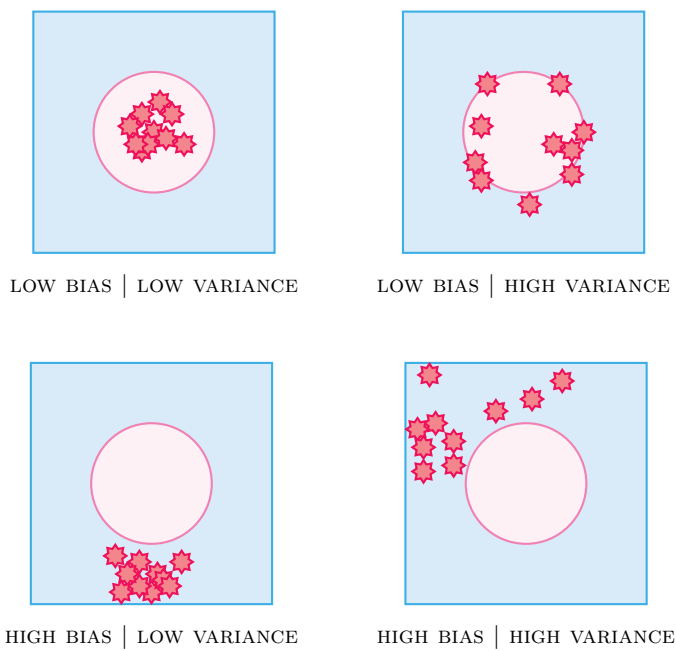

\paragraph{The Bias-Variance Tradeoff}

In machine learning, we want to learn salient patterns in the observed training data while also generalizing well to unseen examples. These two objectives are often seen to compete, resulting in the so-called \textit{bias-variance tradeoff}. Note that model error can be decomposed into \textit{bias}, \textit{variance}, and \textit{noise}, defined as follows:
\begin{itemize}
    \item \textit{Bias}: We typically restrict the hypothesis space of distributions that we search over. If the hypothesis space is very limited, it might not be able to represent the target distribution $p^*$, even with unlimited data. The assumptions encoded in our hypothesis space ``bias'' the model toward certain solutions (e.g., linear models assume linearity). Bias is a measure of how well our model learns $p^*$, and is an inherent error that cannot be resolved by increasing data size. Averaging performance across different training sets, an \textit{unbiased} model will accurately learn the target distribution.
    \item \textit{Variance}: If we select a highly expressive hypothesis class, we might represent the data better. Variance is a measure of how much the model fluctuates (or \textit{varies}) when trained on different samples $\mathcal{D}$. When our model has high variance, small perturbations on $\mathcal{D}$ can result in very different estimates. Models that generalize well to unseen data will closely approximate $p^*$ with low variance across different training sets.
    \item \textit{Noise}: This component of model error captures how much random measurement noise is present in the data sample.
\end{itemize}
See Figure \ref{fig:bias_variance} for an illustration of bias and variance. There is a general notion that larger hypothesis classes come with higher model variance but lower bias, as a larger hypothesis space is more likely to contain hypotheses close to or equal to the true distribution $p^*$. Thus, there is an inherent bias-variance tradeoff when selecting the hypothesis class. 

High bias can be avoided by increasing the capacity of the model. We can avoid high variance using several approaches. We can impose hard constraints, such as selecting a less expressive hypothesis class: e.g., Bayesian networks with at most $d$ parents, or pairwise (rather than arbitrary-order) MRFs. In general, more complex models are more prone to overfitting, while simpler models are more prone to underfitting. We can  introduce a soft preference for ``simpler'' models by adding a regularization term $R(p)$ to the loss $L(\mathbf{x},p)$, which penalizes overly complex $p$.

%%%%%%%%%%%%%%%%%%%%%%%%%%%%%%
%% Learning in directed models
%%%%%%%%%%%%%%%%%%%%%%%%%%%%%%

\section{Learning in Directed Models}
\label{sec:learning_directed_models}

\subsection{Maximum Likelihood Learning in Bayesian Networks}

Let us now apply this discussion to a particular problem of interest: parameter learning in Bayesian networks. Suppose that we are given a Bayesian network with $n$ variables,
\[
p(\mathbf{x}) = \prod^n_{i=1} p(x_i \mid \textbf{pa}(x_i) ; \theta_i).
\]
This is the same formulation of a Bayesian network as given in Definition~\ref{def:factorization_bayesian_networks}, but we have explicitly shown the dependence on model parameters $\theta_i$ in each factor. Recall that in the case of a discrete variables, each factor represents a vector or table of probabilities.

Suppose also that we have a set of iid samples $\mathcal{D}=\{\mathbf{x}^{(1)},\mathbf{x}^{(2)},\ldots,\mathbf{x}^{(m)}\}$. We wish to know the maximum likelihood estimate of the parameters --- that is, to learn the conditional probability distributions. We express the likelihood as
\[
L(\theta; \mathcal{D}) = \prod_{j=1}^m \prod_{i=1}^n p( x_i^{(j)} \mid \textbf{pa}(x_i^{(j)}) ; \theta_i).
\]
Taking logs and combining like terms, in the case of discrete variables, this becomes
\[
\log L(\theta; \mathcal{D}) = \sum_{i=1}^n \sum_{\textbf{pa}(x_i)} \sum_{x_i} \# (x_i, \textbf{pa}(x_i)) \cdot \log p(x_i \mid \textbf{pa}(x_i); \theta_i)
\]
where $\# (x_i, \textbf{pa}(x_i))$ denotes the number of times we see the joint assignment of a given $x_i$ and $\textbf{pa}(x_i)$ in the data $\mathcal{D}$.
% \jm{What exactly is this?}. \vk{it's the number of times we see that joint assignment in the data}
Thus, maximization of the (log) likelihood function decomposes into separate maximizations for the local conditional distributions! This is essentially the same as the head/tails example that we saw earlier (except with more categories). It is a simple calculus exercise to formally show that the maximum likelihood estimate for each factor is
\[
 \hat{\theta}_i(x_i \mid \textbf{pa}(x_i)) = \frac{\#\; (x_i, \textbf{pa}(x_i))}{\# \; (\textbf{pa}(x_i))},
\]
where $\hat{\theta}_i(x_i \mid \textbf{pa}(x_i))$ denotes the estimate of a table of conditional probabilities for values $x_i$ given parents $\textbf{pa}(x_i)$.
We can therefore draw the convenient conclusion that the maximum likelihood estimate has a closed-form solution for Bayesian networks with discrete variables. Even when the variables are not discrete, the task is nearly as simple: the log-factors are linearly separable, and hence the log-likelihood reduces to estimating each of them separately. The simplicity of learning is one of the most convenient features of Bayesian networks.

%%%%%%%%%%%%%%%%%%%%%%%%%%%%%%
%% Structure learning
%%%%%%%%%%%%%%%%%%%%%%%%%%%%%%

\subsection{Structure Learning for Bayesian Networks}
\label{sec:structure_learning}

We now review the task of structure learning from observational data. As the search space of DAGs grows super-exponentially with respect to the cardinality of the node set, exhaustive search for the correct DAG is intractable in the general case (Table \ref{tab:total_dags}). %\footnote{See Entry A003024 of \textit{The On-Line Encyclopedia of Integer Sequences} (\href{https://oeis.org/A003024}{https://oeis.org/A003024}; \citealt{sloane2003line}).}
Thus, structure learning typically entails some form of heuristic search. Even still, DAG structure learning is NP-hard under various problem formulations \citep{chickering1996learning,chickering2004large}. Constraining the search space can potentially enable polynomial-time solutions (e.g., imposing sparsity constraints that limit the number of parents per node; \citealt{claassen2013learning}), though such constraints are sometimes unrealistic for real-world data. Consequently, several popular and influential structure learning algorithms still display worst-case exponential time complexity.

\begin{table}[!t]
    \centering
    \begin{tabular}{c c c c}
        \toprule
        \textsc{total nodes} & \textsc{total possible dags} & \textsc{power of 10}\\
        \midrule
        1 & 1 \\
        2 & 3 \\
        3 & 25 \\
        4 & 543 & $> 10^2$ \\
        5 & 29281 & $> 10^4$ \\
        6 & 3781503 & $> 10^6$ \\
        7 & 1138779265 & $> 10^9$ \\
        8 & 783702329343 & $> 10^{11}$ \\
        8 & 1213442454842881 & $> 10^{15}$  \\
        10 & 4175098976430598143 & $> 10^{18}$ \\
        \bottomrule 
    \end{tabular}
    \caption{The total number of possible DAGs with $n$ labeled nodes grows super-exponentially with respect to $n$. When $n = 10$, this number already exceeds 4 quintillion \citep{sloane1995encyclopedia,sloane2003line}.}
    \label{tab:total_dags}
\end{table}

There are two dominant paradigms for structure learning in DAGs: \textit{score-based} methods \citep{huang2018generalized} and \textit{constraint-based} methods \citep{spirtes2001causation}. Structure learning can also entail \textit{hybrid} methods that draw from both traditions \citep{ogarrio2016hybrid}. Still other methods do not fall neatly into any of these categories (e.g., \citealt{janzing2012information,janzing2015justifying}). In this subsection, we provide a high-level overview of common structure learning approaches.

This discussion will also circle us back to concepts in causal graphical modeling. Structure learning is often deployed in causal inference settings, where certain assumptions can allow us to interpret the graphical structure as a representation of the causal mechanisms that explain the data generating process. When structure learning is causal, we refer to this task as \textit{causal discovery} \citep{glymour2019review}. In addition to score-based, constraint-based, and hybrid causal discovery methods, there is also a rich class of algorithms based on \textit{functional causal models} that make parametric assumptions on the data generating process in exchange for other mathematical conveniences \citep{hoyer2008nonlinear,zhang2009identifiability}. Throughout this section, we will reference both causal discovery methods and non-causal structure learning approaches.

\paragraph{Global, Local, and Time-Series Structures} This section mainly focuses on \textit{global} structure learning, where all relations in the graph are of interest. In practice, we might only be interested in substructures of the global graph, such that learning the full graph would require extraneous computation. \textit{Local} structure learning can provide efficient solutions for such problems. This includes the bivariate case, where we are interested in learning whether $X$ causes $Y$ or $Y$ causes $X$ \citep{mooij2016distinguishing}. Another common case of local structure learning is \textit{Markov blanket learning}, which can be used for causal feature selection in machine learning \citep{aliferis2010local,yu2020causality}. %Using the Markov blanket (Definitions \ref{def:markov_blanket_dag},  \ref{def:markov_blanket_mrf}) of the target variable can provide parsimony, predictivity, and interpretability relative to non-causal feature selection \citep{aliferis2010local}. 
Additionally, when data have a temporal dimension, we can apply \textit{time-series} structure learning approaches. This chapter will not address the time-series setting, though we refer the reader to external resources (e.g., \citealt{assaad2022survey}).

\subsection{Score-Based Methods}
\label{sec:score_based}

We can formulate score-based structure learning as a search problem that consists of two parts: 
\begin{enumerate}
    \item \textit{The goodness-of-fit metric.} First, we define a criterion to evaluate how well the Bayesian network fits the observed data.
    \item \textit{The search algorithm.} Next, we define a procedure that searches over the space of DAGs (or a constrained subspace) for a structure achieving the maximal score.
\end{enumerate}
Putting these two parts together, we aim to search among a set of candidate graphs $\mathbf{G}$ for the graph $\mathcal{G}$ that best fits our data $\mathcal{D}$ according to our score metric: 
\[
\argmax_{\mathcal{G} \in \mathbf{G}} \; \text{Score}(\mathcal{G}, \mathcal{D}).
\]
Score-based structure learning is traditionally treated as a combinatorial
optimization problem, though recent innovations have also given rise to continuous optimization approaches \citep{zheng2018dags}.

\subsubsection{Score Metrics}

\newcommand*\mystrut[1]{\vrule width0pt height0pt depth#1\relax}

Score metrics for a structure $\mathcal{G}$ and data $\mathcal{D}$ generally take the form
\[
\text{Score}(\mathcal{G}, \mathcal{D}) = \underbrace{\mystrut{1.5ex} \ell(\mathcal{G}; \mathcal{D})}_{\text{log-likelihood}} - \underbrace{\mystrut{1.5ex} \phi(|\mathcal{D}|) \,  \|\mathcal{G}\|}_{\text{regularization}}.
\]
Here $\ell(\mathcal{G}; \mathcal{D})$ refers to the log-likelihood of the data under the graph structure $\mathcal{G}$. The parameters in the Bayesian network $\mathcal{G}$ are estimated based on MLE and the log-likelihood score is calculated based on the estimated parameters. If the score function only consisted of the log-likelihood term, then the optimal graph would be a complete graph, which is probably overfitting the data. We can mitigate the potential for overfitting by introducing the regularization term $\phi(\lvert \mathcal{D} \rvert) \, \lVert \mathcal{G} \rVert$, where $\lvert \mathcal{D} \rvert$ is the number of data samples and $\|\mathcal{G}\|$ is the number of parameters in the graph $\mathcal{G}$. This kind of regularization penalizes models with many parameters, favoring simpler models.

When $\phi(t) = 1$, this score function is known as the \textit{Akaike Information Criterion} (AIC; \citealt{akaike1998information,cavanaugh2019akaike}). When $\phi(t) = \log(t)/2$, this score function is known as the \textit{Bayesian Information Criterion} (BIC; \citealt{schwarz1978estimating,neath2012bayesian}). With the BIC, the influence of model complexity decreases as $\lvert \mathcal{D} \rvert$ grows, allowing the log-likelihood term to eventually dominate the score. The AIC and BIC differ in their properties and assumptions, and should be selected based on the needs of the experiment at hand \citep{vrieze2012model}. 

Many other score functions have since been proposed. Another family of Bayesian score functions is known as the \textit{Bayesian Dirichlet (BD) score}. The BD score first defines the probability of data $\mathcal{D}$ conditional on the graph structure $\mathcal{G}$ as
\[
p(\mathcal{D} ; \mathcal{G})=\int p(\mathcal{D} \mid \Theta_{\mathcal{G}} ;  \mathcal{G})\, p(\Theta_{\mathcal{G}} ; \mathcal{G}) \, d \Theta_{\mathcal{G}},
\]
where $p(\mathcal{D} \mid \Theta_{\mathcal{G}} ; \mathcal{G})$ is the probability of the data given the network structure $\mathcal{G}$ and parameters $\Theta_{\mathcal{G}}$, and $p(\Theta_{\mathcal{G}} ; \mathcal{G})$ is the prior probability of the parameters. When the prior probability is specified as a Dirichlet distribution,
\[
p(\mathcal{D} \mid \Theta_{\mathcal{G}})
= \prod_i \prod_{\pi_i} \left[ \frac{\Gamma(\sum_j N'_{i,\pi_i,j})}{\Gamma(\sum_j N'_{i,\pi_i,j} + N_{i,\pi_i,j} )} \prod_{j}\frac{\Gamma(N'_{i,\pi_i,j} + N_{i,\pi_i,j})}{\Gamma(N'_{i,\pi_i,j})}\right].
\]
Here $\pi_i$ refers to the parent configuration of the variable $i$ and $N_{i,\pi_i,j}$ is the count of variable $i$ taking value $j$ with parent configuration $\pi_i$. $N'$ represents the counts in the prior.

With a prior over the model parameters  $p(\Theta_{\mathcal{G}})$ (say, a uniform one), the BD score is defined as
\[
\log p(\mathcal{D} \mid \Theta_{\mathcal{G}}) + \log p(\Theta_{\mathcal{G}}).
\] 

Notice that there is no penalty term appended to the BD score, as it will penalize overfitting implicitly via the integral over the parameter space.

While score-based methods do not suffer from the multiple testing problems that are innate to constraint-based structure learning, most score metrics impose parametric assumptions on the distributional forms of random variables and/or the functional forms of parent-child relationships. Such parametric assumptions can be difficult to justify in real-world settings, and violations of these assumptions can result in inaccurate structure learning. These limitations have given rise to various \textit{nonparametric score metrics}, including those that accommodate mixed data types. These often employ regression in Reproducing Kernel
Hilbert Space, though kernel-based methods can be slow to run in practice \citep{huang2018generalized}.

\subsubsection{Search Algorithms}

Search procedures commonly take a \textit{local search} or \textit{greedy search} approach. Both approaches are computationally tractable and sometimes offer guarantees on correctness under infinite data and other sufficient conditions. However, neither type of approach can guarantee the quality of the learned graph in finite sample regimes (like those we encounter in real data). Our search space is highly non-convex, and search algorithms can get stuck in sub-optimal regions (e.g., local maxima).

A popular set of search frameworks is owed to \textit{Greedy Equivalence Search} (GES) and its variants \citep{chickering2002optimal}. To reasonably constrain our search space, GES searches over Markov equivalence classes (MECs; Definition \ref{def:markov_equivalence_class}) instead of the space of unique DAGs. First, we start with the empty graph and iteratively add edges until goodness-of-fit reaches a local maximum. Next, we iteratively eliminate edges until the score again reaches a local maximum. %Guarantees under infinite data: global minimizer when assumptions are met.
GES is proven to correctly identify the optimal solution in the large sample limit when applied to the sparsely-connected search space of MECs.

More concretely, the two phases of GES are as follows:

\begin{enumerate}
    \item \textit{Forward Equivalence Search (FES).} Starting with the equivalence class of no dependencies, greedily make single-edge additions until a local maximum is reached. When all assumptions are met, the local maximum reached after FES contains the generative distribution.
    \item \textit{Backward Equivalence Search (BES).} Consider all single-edge deletions within the current equivalence class until a local maximum is reached. When all assumptions are met, the equivalence class reached after BES must be a perfect map of the generative distribution.
\end{enumerate}

Though GES has spawned many influential extensions, it presents with several limitations. While GES is guaranteed to return a perfect map of the generative distribution in the large sample limit  when all assumptions are met, these conditions are generally unrealistic and unverifiable. Though computationally convenient, returning the MEC provides less graphical information than returning the unique DAG. While the original formulation of GES assumes no latent confounding, recent extensions address this limitation \citep{claassen2022greedy}.

%For local search, we begin with an empty graph or a complete graph. At each step, we update the graph structure by a single operation of adding an edge, removing an edge, or reversing an edge. (Of course, each operation should preserve the acyclicity constraint.) If the score increases, then we keep the current update; otherwise, the algorithm attempts a different update and recomputes the score.

%For greedy search with K3, we first assume a topological order of the graph. For each variable, we restrict its parent set to the variables with a higher order. While searching for parent set for each variable, it takes a greedy approach by adding the parent that increases the score most until no improvement can be made.

%This search algorithm often takes a greedy approach, as in \textit{Greedy Equivalence Search} and its variants \citep{chickering2002optimal}.

\subsubsection{Chow-Liu Algorithm}

The Chow-Liu Algorithm is a classical score-based approach for identifying the maximum likelihood tree-structured graph \citep{chow1968approximating}. We restrict our attention to trees where each node has exactly one parent, except for a parentless root node (Definition \ref{def:directed_tree}). The score that we employ here is simply the log-likelihood. There is no penalty term for graph structure complexity, since the algorithm only considers tree structures.

\begin{figure}[!t]
    \centering

\begin{tikzpicture}[scale=0.7,every edge quotes/.style = {font=\footnotesize,fill=white,sloped}]
    % Nodes
    \node[draw,scale=0.8,circle,thick,black,fill=Lavender!30] (A) at (0, 2) {$A$};
    \node[draw,scale=0.8,circle,thick,black,fill=Lavender!30]  (B) at (-2, 0) {$B$};
    \node[draw,scale=0.8,circle,thick,black,fill=Lavender!30]  (C) at (2, 0) {$C$};
    \node[draw,scale=0.8,circle,thick,black,fill=Lavender!30]  (D) at (0, -2) {$D$};

    % Edges with weights
    \draw[thick] (A) edge["\color{RubineRed}$0.07$"] (B);
    \draw[thick] (A) edge["\color{RubineRed}$0.32$"] (C);
    \draw[thick] (B) edge["\color{RubineRed}$0.32$"] (0,0);
    \draw[thick] (0,0) edge[] (C);
    \draw[thick] (B) edge["\color{RubineRed}$0.32$"] (D);
    \draw[thick] (C) edge["\color{RubineRed}$0.02$"] (D);
    %\draw[thick] (A) edge["\color{RubineRed}$0.17$" left] (center);
    \draw[thick] (A) edge["\color{RubineRed}$0.17$"] (0,0);
    \draw[thick] (0,0) edge[] (D);

    % Label
    \node[] (label) at (0,-3) {\small\textbf{(a)} \textsc{weighted skeleton}};
\end{tikzpicture}
%%%%%%%%%%%%
\hspace{10mm}
%%%%%%%%%%%%
\begin{tikzpicture}[scale=0.7,every edge quotes/.style = {font=\footnotesize,fill=white,sloped}]
    % Nodes
    \node[draw,scale=0.8,circle,thick,black,fill=Lavender!30] (A) at (0, 2) {$A$};
    \node[draw,scale=0.8,circle,thick,black,fill=Lavender!30]  (B) at (-2, 0) {$B$};
    \node[draw,scale=0.8,circle,thick,black,fill=Lavender!30]  (C) at (2, 0) {$C$};
    \node[draw,scale=0.8,circle,thick,black,fill=Lavender!30]  (D) at (0, -2) {$D$};

    % Edges with weights
    \draw[thick] (A) edge["\color{RubineRed}$0.32$"] (C);
    \draw[thick] (B) edge["\color{RubineRed}$0.32$"] (C);
    \draw[thick] (B) edge["\color{RubineRed}$0.32$"] (D);
    
    % Label
    \node[] (label) at (0,-3) {\small\textbf{(b)} \textsc{spanning tree}};
\end{tikzpicture} \\
%%%%%%%%%%%%
\vspace{5mm}
%%%%%%%%%%%%
\begin{tikzpicture}[scale=0.7,every edge quotes/.style = {font=\tiny,fill=white,sloped}]
    % Nodes
    \node[draw,scale=0.8,circle,thick,ultra thick,RubineRed,fill=Lavender!30] (A) at (0, 2) {\color{black}$A$};
    \node[draw,scale=0.8,circle,black,thick,fill=Lavender!30]  (B) at (-2, 0) {$B$};
    \node[draw,scale=0.8,circle,thick,black,fill=Lavender!30]  (C) at (2, 0) {$C$};
    \node[draw,scale=0.8,circle,thick,black,fill=Lavender!30]  (D) at (0, -2) {$D$};

    % Edges with weights
    \draw[thick,-{Stealth[width=5pt,length=5pt]}] (A) -- (C);
    \draw[thick,-{Stealth[width=5pt,length=5pt]}] (C) -- (B);
    \draw[thick,-{Stealth[width=5pt,length=5pt]}] (B) -- (D);
    
    % Label
    \node[] (label) at (0,-3) {\small\textbf{(c)} \textsc{dag with root a}};
\end{tikzpicture}
%%%%%%%%%%%%
\hspace{10mm}
%%%%%%%%%%%%
\begin{tikzpicture}[scale=0.7,every edge quotes/.style = {font=\footnotesize,fill=white,sloped}]
    % Nodes
    \node[draw,scale=0.8,circle,thick,black,fill=Lavender!30] (A) at (0, 2) {$A$};
    \node[draw,scale=0.8,circle,ultra thick,RubineRed,fill=Lavender!30]  (B) at (-2, 0) {\color{black}$B$};
    \node[draw,scale=0.8,circle,thick,black,fill=Lavender!30]  (C) at (2, 0) {$C$};
    \node[draw,scale=0.8,circle,thick,black,fill=Lavender!30]  (D) at (0, -2) {$D$};

    % Edges with weights
    \draw[thick,-{Stealth[width=5pt,length=5pt]}] (C) -- (A);
    \draw[thick,-{Stealth[width=5pt,length=5pt]}] (B) -- (C);
    \draw[thick,-{Stealth[width=5pt,length=5pt]}] (B) -- (D);
    
    % Label
    \node[] (label) at (0,-3) {\small\textbf{(d)} \textsc{dag with root b}};
\end{tikzpicture}

    \caption{Graph learning with Chow-Liu (Algorithm \ref{alg:chow_liu}). Step 1 (\textbf{a}) yields an undirected skeleton where each edge is weighted by the mutual information for the incident pair of random variables. Step 2 (\textbf{b}) identifies the maximum weight spanning tree. Step 3 (\textbf{c} and \textbf{d}) yields the DAG corresponding to the selected root.}
    \label{fig:chow_liu}
\end{figure}
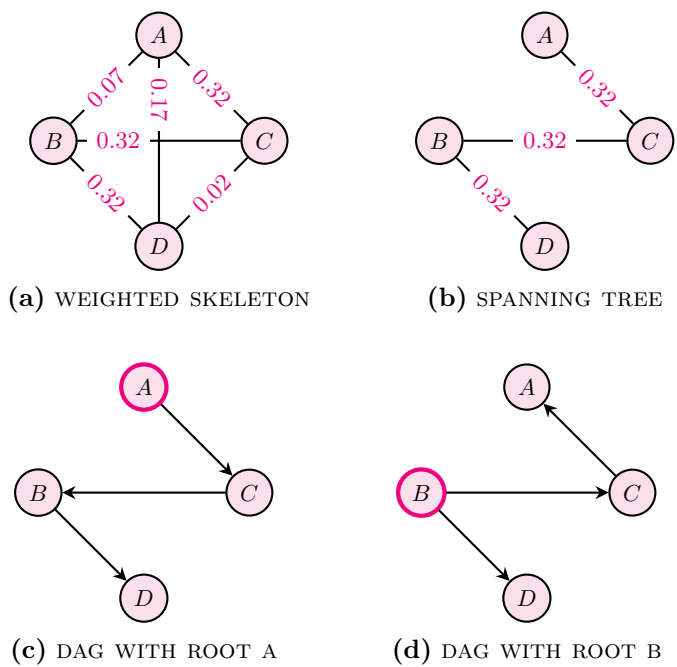

The algorithm has three steps, which we outline in Algorithm \ref{alg:chow_liu}. We visualize possible inputs and outputs for Algorithm \ref{alg:chow_liu} in Figure \ref{fig:chow_liu}.

\paragraph{Time Complexity} Let $n$ be the number of nodes in our graph. The time complexity of the Chow-Liu Algorithm is quadratic in $n$, as it takes $O(n^2)$ time to compute the mutual information for all pairs of variables and $O(n^2)$ time to compute the maximum spanning tree.

\begin{algorithm}[!t]
\caption{\textit{Chow-Liu Algorithm}} \label{alg:chow_liu} 
\vspace{3mm}

\begin{enumerate}
    \item Compute the mutual information for all pairs of variables $\{X,U\}$, and form a complete graph from the variables where the edge between variables $\{X,U\}$ has weight equivalent to the mutual information term $MI(X,U)$:
    \[
    \text{MI}(X,U) =\sum_{x,u} \hat p(x,u)\log\left[\frac{\hat p(x,u)}{\hat p(x) \hat p(u)}\right].
    \]
    This function measures how much information $U$ provides about $X$. %The graph with computed MI edge weights might resemble:
    %{% include maincolumn_img.html src='assets/img/mi-graph.png' %}
    Recall that from our empirical distribution
    \[
    \hat p(x,u) = \frac{Count(x,u)}{\# \text{ data points}}.
    \]
    \item Find the \textit{maximum weight spanning tree}, which can be done programmatically by using Kruskal's Algorithm \citep{kruskal1956shortest} or Prim's Algorithm \citep{prim1957shortest}.
    %{% include maincolumn_img.html src='assets/img/max-spanning-tree.png' %}
    \item Pick any node to be the \textit{root variable}, and assign directions radiating outward from this node (i.e., all arrows point away from the root). This step transforms the resulting undirected tree into a directed one. We then return the directed tree.
    %{% include maincolumn_img.html src='assets/img/chow-liu-tree.png' %}
\end{enumerate}

\end{algorithm}

\paragraph{Why it Works} Having described the algorithm, let's explain why this works. It turns out that the likelihood score decomposes into mutual information and entropy terms:
\[
\log p(\mathcal D \,;\, \theta, \mathcal{G}) = |\mathcal D| \sum_i \underbrace{\text{MI}(X_i, \textbf{pa}(X_i))}_{\text{mutual information}} - |\mathcal D| \sum_i \underbrace{\mystrut{2ex} H(X_i)}_{\text{entropy}}.
\]
We would like to find a graph $\mathcal{G}$ that maximizes this log-likelihood. Since the entropies are independent of the dependency ordering in the tree, the only terms that change with the choice of $\mathcal{G}$ are the mutual information terms. So we want
\[
\argmax_\mathcal{G} \log p(\mathcal D \,;\, \theta, \mathcal{G}) = \argmax_\mathcal{G} \sum_i \text{MI}(X_i, \textbf{pa}(X_i)).
\]
Now if we assume $\mathcal{G} = (\mathbf{V},\mathbf{E})$ is a tree (where each node has at most one parent), then
\[
\argmax_{\mathcal{G} \in \text{trees}} \log p(\mathcal D \,;\, \theta, \mathcal{G}) = \argmax_{\mathcal{G} \in \text{trees}} \sum_{(i,j)\in \mathbf{E}} \text{MI}(X_i,X_j).
\]
The orientation of edges does not matter here, as mutual information is symmetric. Thus, we can see why the Chow-Liu algorithm finds a tree structure that maximizes the log-likelihood of the data. As illustrated in Figure \ref{fig:chow_liu}, this can result in different DAGs depending on which root is selected.

\begin{figure}[!t]
    \centering

\begin{tikzpicture}[scale=0.8,->,>={Stealth[length=5,width=5]}]
    % Nodes.
    \node[draw,thick,circle,scale=0.7] (A) at (-1,2.5) {$A$};
    \node[draw,thick,circle,scale=0.7] (B) at (1,2.5) {$B$};
    \node[draw,thick,circle,scale=0.7] (C) at (0,1.5) {$C$};
    \node[draw,thick,circle,scale=0.7] (D) at (0,0) {$D$};
    % Edges.
    \draw[thick] (A) -- (C);
    \draw[thick] (B) -- (C);
    \draw[thick] (C) -- (D);
    % Labels.
    \node[] (label) at (0,-1) {\footnotesize\textbf{(a)} \textsc{true dag}};
\end{tikzpicture}
%%%%%%%%%%%%%%%%%%%%%%%%%%%%%%%%
\hspace{10mm}
%%%%%%%%%%%%%%%%%%%%%%%%%%%%%%%%
\begin{tikzpicture}[scale=0.8]
    % Nodes.
    \node[draw,thick,circle,scale=0.7] (A) at (-1,2.5) {$A$};
    \node[draw,thick,circle,scale=0.7] (B) at (1,2.5) {$B$};
    \node[draw,thick,circle,scale=0.7] (C) at (0,1.5) {$C$};
    \node[draw,thick,circle,scale=0.7] (D) at (0,0) {$D$};
    % Edges.
    \draw[thick] (A) -- (C);
    \draw[thick] (A) -- (B);
    \draw[thick] (A) edge[bend right=20] (D);
    \draw[thick] (B) -- (C);
    \draw[thick] (B) edge[bend left=20] (D);
    \draw[thick] (C) -- (D);
    % Labels.
    \node[] (label) at (0,-1) {\footnotesize\textbf{(b)} \textsc{begin}};
\end{tikzpicture}
%%%%%%%%%%%%%%%%%%%%%%%%%%%%%%%%
\hspace{10mm}
%%%%%%%%%%%%%%%%%%%%%%%%%%%%%%%%
\begin{tikzpicture}[scale=0.8]
    % Nodes.
    \node[draw,thick,circle,scale=0.7] (A) at (-1,2.5) {$A$};
    \node[draw,thick,circle,scale=0.7] (B) at (1,2.5) {$B$};
    \node[draw,thick,circle,scale=0.7] (C) at (0,1.5) {$C$};
    \node[draw,thick,circle,scale=0.7] (D) at (0,0) {$D$};
    % Edges.
    \draw[thick] (A) -- (C);
    \draw[thick] (A) edge[bend right=20] (D);
    \draw[thick] (B) -- (C);
    \draw[thick] (B) edge[bend left=20] (D);
    \draw[thick] (C) -- (D);
    % Labels.
    \node[] (label) at (0,-1) {\footnotesize\textbf{(c)} $A \ind B$};
\end{tikzpicture} \\
%%%%%%%%%%%%%%%%%%%%%%%%%%%%%%%%
\vspace{5mm}
%%%%%%%%%%%%%%%%%%%%%%%%%%%%%%%%
\begin{tikzpicture}[scale=0.8]
    % Nodes.
    \node[draw,thick,circle,scale=0.7] (A) at (-1,2.5) {$A$};
    \node[draw,thick,circle,scale=0.7] (B) at (1,2.5) {$B$};
    \node[draw,thick,circle,scale=0.7] (C) at (0,1.5) {$C$};
    \node[draw,thick,circle,scale=0.7] (D) at (0,0) {$D$};
    % Edges.
    \draw[thick] (A) -- (C);
    \draw[thick] (B) -- (C);
    \draw[thick] (C) -- (D);
    % Labels.
    \node[] (label) at (0,-1) {\footnotesize\textbf{(d)} $A \ind D \mid C, B \ind D \mid C$};
\end{tikzpicture}
%%%%%%%%%%%%%%%%%%%%%%%%%%%%%%%%
\hspace{10mm}
%%%%%%%%%%%%%%%%%%%%%%%%%%%%%%%%
\begin{tikzpicture}[scale=0.8]
    % Nodes.
    \node[draw,thick,circle,scale=0.7] (A) at (-1,2.5) {$A$};
    \node[draw,thick,circle,scale=0.7] (B) at (1,2.5) {$B$};
    \node[draw,thick,circle,scale=0.7] (C) at (0,1.5) {$C$};
    \node[draw,thick,circle,scale=0.7] (D) at (0,0) {$D$};
    % Edges.
    \draw[thick,-{Stealth[width=5pt,length=5pt]}] (A) -- (C);
    \draw[thick,-{Stealth[width=5pt,length=5pt]}] (B) -- (C);
    \draw[thick] (C) -- (D);
    % Labels.
    \node[] (label) at (0,-1) {\footnotesize\textbf{(e)} \textsc{$v$-structure}};
\end{tikzpicture}
%%%%%%%%%%%%%%%%%%%%%%%%%%%%%%%%
\hspace{10mm}
%%%%%%%%%%%%%%%%%%%%%%%%%%%%%%%%
\begin{tikzpicture}[scale=0.8]
    % Nodes.
    \node[draw,thick,circle,scale=0.7] (A) at (-1,2.5) {$A$};
    \node[draw,thick,circle,scale=0.7] (B) at (1,2.5) {$B$};
    \node[draw,thick,circle,scale=0.7] (C) at (0,1.5) {$C$};
    \node[draw,thick,circle,scale=0.7] (D) at (0,0) {$D$};
    % Edges.
    \draw[thick,-{Stealth[width=5pt,length=5pt]}] (A) -- (C);
    \draw[thick,-{Stealth[width=5pt,length=5pt]}] (B) -- (C);
    \draw[thick,-{Stealth[width=5pt,length=5pt]}] (C) -- (D);
    % Labels.
    \node[] (label) at (0,-1) {\footnotesize\textbf{(f)} \textsc{output}};
\end{tikzpicture}
%%%%%%%%%%%%%%%%%%%%%%%%%%%%%%%%

    \caption{A demonstration of the classic constraint-based PC algorithm for causal discovery \citep{spirtes2001causation}. In this case, the true DAG is a $y$-structure (\textbf{a}). To begin, we consider the complete undirected graph induced by $\{A,B,C,D\}$ (\textbf{b}). A test of marginal independence indicates that $A \ind B$, and so the edge $A - B$ is removed (\textbf{c}). Next, tests of conditional independence indicate that $A \ind D \mid C$ and $B \ind D \mid C$, and so $A - D$ and $B - D$ are removed as well (\textbf{d}). As $A \ind B$ and $A \nind B \mid C$, we can orient the $v$-structure $A \rightarrow C \leftarrow B$ (\textbf{e}). Finally, we can apply orientation propagation rules to infer that $C \to D$ (\textbf{f}). This yields the correct output: a $y$-structure. Figure and caption adapted from \citet{glymour2019review}.}
    \label{fig:pc_y_structure}
\end{figure}
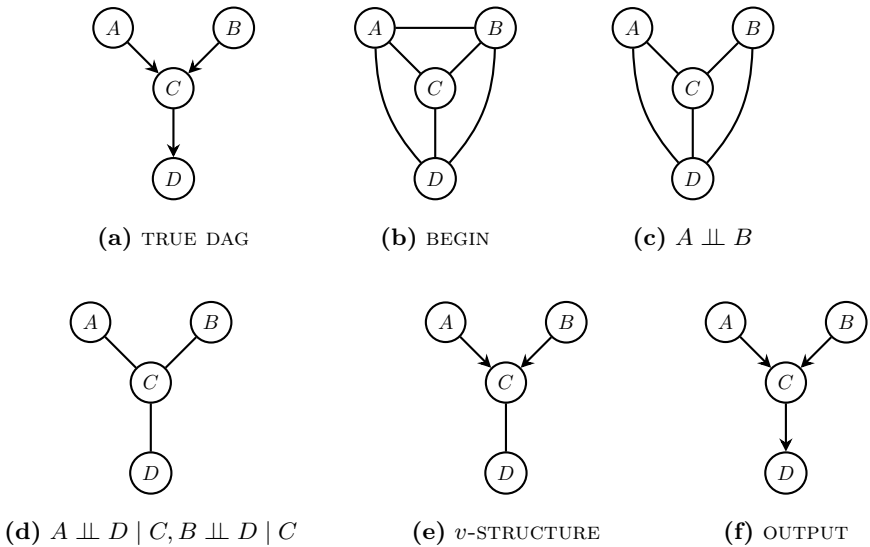

\subsection{Constraint-Based Methods}
\label{sec:constraint_based}

%\vk{these sections are well written, but they read a bit like a survey paper---your call about whether you want to do anything here (i think earlier sections are higher priority)}

Constraint-based structure learning employs sequential conditional independence testing to identify a set of edge constraints for the graph. Then, edge orientation rules are applied to the undirected skeleton to obtain the final output. Often, the conditional independence test is chosen by the user and simply plugged into the general algorithm, enabling significant flexibility. This family of algorithms generally benefits from a well-established body of theory, asymptotic guarantees on correctness, human interpretability, and no innate parametric assumptions (though the user-selected independence test or specific algorithmic design choices might impose their own assumptions).

The most famous example of constraint-based structure learning is the causal discovery algorithm \textit{PC}, which takes its name from the names of its authors \citep{spirtes2001causation}. Constraint-based methods like PC exploit the conditional independence properties of $v$-structures, which are distinct from the conditional independencies present in forks and chains (Table \ref{tab:primitives_independencies}, Figure \ref{fig:triples}). Figure \ref{fig:pc_y_structure} provides a demonstration of PC for learning a $y$-structured DAG (itself a canonical structure that can be exploited for graph learning; see \citealt{mani2006theoretical}). Note that in this case, PC is able to identify the full unique graph (i.e., all edges are oriented). In general, PC can only identify up to the MEC and outputs often contain at least some unoriented edges.

While the PC algorithm assumes that all confounders in the true graph are observed (i.e., \textit{causal sufficiency}), the popular \textit{Fast Causal Inference} (FCI) algorithm and its variants extend PC to allow for latent confounding (i.e., \textit{causal insufficiency}). See \citet{spirtes2013causal} for a deep discussion of related methods. Other prevalent assumptions in constraint-based learning involve the equivalence between properties of the data and properties of the causal graph:
\begin{enumerate}
    \item \textit{Causal Markov condition:} Any $d$-separation in graph $\mathcal{G}$ ($\ind_{\mathcal{G}}$) implies conditional independence in data distribution $p$ ($\ind_{{p}}$),
    \begin{align*}
        X \ind_{\mathcal{G}} Y | Z \Longrightarrow X \ind_{{p}} Y | Z .
    \end{align*}
    \item \textit{Faithfulness:} Conditional independence implies $d$-separation,
    \begin{align*}
         X \ind_{{p}} Y | Z \Longrightarrow  X \ind_{\mathcal{G}} Y | Z .
    \end{align*}
\end{enumerate}

Despite their popularity, constraint-based methods also face several drawbacks. Many recent advances in constraint-based discovery aim to alleviate the following challenges:
\begin{itemize}
    \item \textit{Statistical challenges.} Conditional independence testing is a hard problem in itself \citep{shah2020hardness}, and sequential testing raises its own set of statistical problems. Thus, the performance of constraint-based methods will be bottlenecked by the performance of the chosen independence test.
    \item \textit{Sample complexity.} The sample complexity of conditional independence testing is exponential in the cardinality of the conditioning set, which can result in poor finite sample performance. 
    \item \textit{Time complexity.} Unless assumptions are imposed, the total number of conditional independence tests will be worst-case exponential in the total number of nodes. Several approaches can be taken to improve runtimes. For example, many constraint-based procedures are \textit{embarrassingly parallelizable}: at least some speedups can be gained simply by paralellizing the sequential tests whose outputs do not rely on each other.
    \item \textit{Informativeness.} As with GES, purely constraint-based methods can only identify the true graph up to its MEC. Additionally, methods that rely solely on conditional independence testing require at least three variables. Therefore, these cannot handle the case of bivariate direction inference, where only two variables are considered \citep{mooij2016distinguishing}.
\end{itemize}

\subsection{Functional Causal Models}
\label{sec:functional_causal_models}

%Nonlinear additive noise model \citep{hoyer2008nonlinear}, post-nonlinear additive noise model \citep{zhang2009identifiability}, linear non-Gaussian acyclic models \citep{shimizu2011directlingam}.

\begin{figure}[!t]
    \centering
    \includegraphics[height=0.15\textheight]{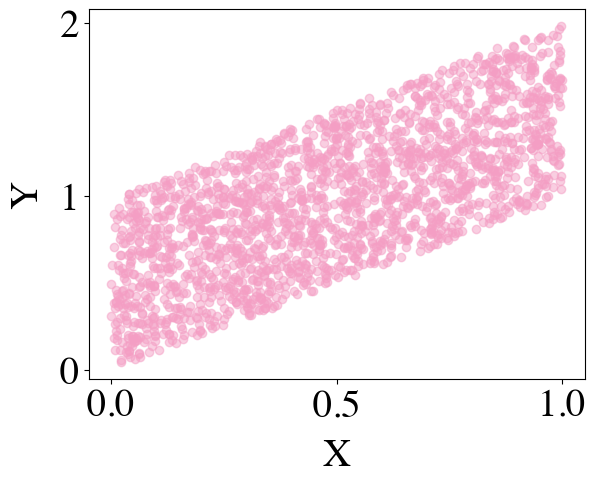}
    \hspace{2mm}
    \includegraphics[height=0.15\textheight]{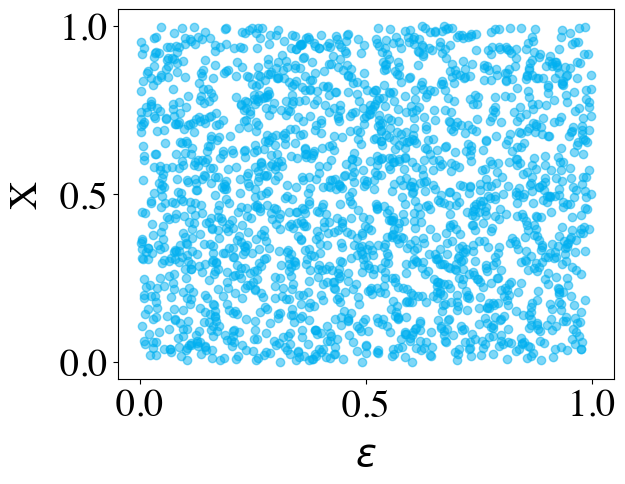}
    \hspace{2mm}
    \includegraphics[height=0.15\textheight]{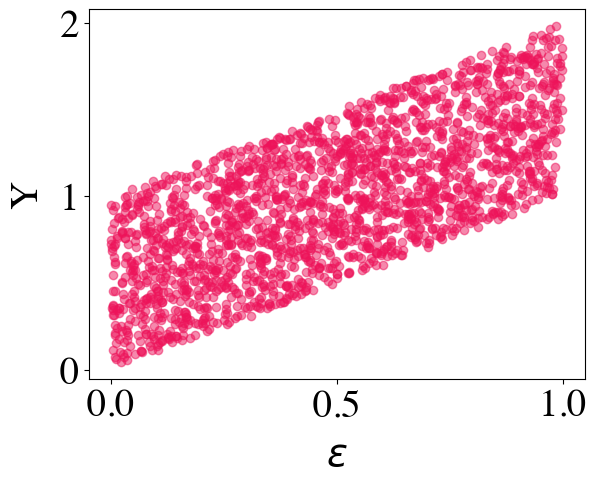}
    \caption{A bivariate ANM with linear causal functions and uniform noise distributions. In the true structural causal model, the variable $X$ and exogenous random noise $\epsilon$ are the causes of variable $Y$. As expected under the assumptions of the ANM, when we plot $X$ against $\epsilon$, we observe independence (center). When we plot $Y$ against $\epsilon$, we observe dependence (right).}
    \label{fig:anm}
\end{figure}

If we are willing to impose parametric assumptions on our causal model, we can identify the unique DAG describing our data distribution instead of its MEC. A popular family of causal discovery algorithms takes this approach by specifying a constrained \textit{functional causal model} (FCM). The most common FCM is the \textit{additive noise model} (ANM), which represents effect $Y$ as a function of its direct causes $\mathbf{X}$ and some exogenous random noise $\epsilon$, where function $f$ can take arbitrary forms:
\[
y = f(\mathbf{x}) + \epsilon.
\]
Crucially, the ANM assumes that $\mathbf{X}$ and $\epsilon$ are \textit{independent} while $Y$ and $\epsilon$ are \textit{dependent}. This assumption allows us to detect  asymmetries in the observed data that can be used to infer edge directionality. Consider the bivariate case of $X$ and $Y$: if the hypothetical causal model $X \to Y$ displays independence between the putative parent and the noise term while the alternative model $Y \to X$ does not, the independent noise assumption dictates that we model the causal relationship as $X \to Y$ (Figure \ref{fig:anm}).

%Function $f$ can take arbitrary forms. When $f$ is nonlinear, we can assume a nonlinear AMN \citep{hoyer2008nonlinear}. When $f$ is linear, we can assume a linear non-Gaussian acyclic model (LiNGAM) \citep{shimizu2011directlingam}. Note that under the ANM, the linear-Gaussian case is not identifiable \citep{zhang2009identifiability}.

The ANM is not the only FCM, and we can instead define a more general class. The post-nonlinear causal model (PNL) is the most general of the well-defined FCMs. It is suited for bivariate inference and inference on larger structures. The PNL accounts for the nonlinear effect of the cause, the inner noise effect, as well as a measurement distortion effect in the observed variables. Under this model, each variable $X_i$ in graph $\mathcal{G}$ takes the form
\[
x_i = g_{i}\big(f_{i}(\mathbf{pa}(x_i)) + \epsilon_i\big),
\]
 where $\mathbf{pa}(x_i)$ are the parents of $X_i$, $\epsilon_i$ is the exogenous noise term, $f_{i}$ is a nonlinear causal function, and $g_{i}$ is an invertible post-nonlinear distortion.

The PNL comes with thorough identifiability results \citep{zhang2009identifiability}. Most notably, the PNL is not identifiable in the linear-Gaussian setting. The identifiability conditions of the PNL extend to its special cases, which include the following:
\begin{itemize}
    \item \textit{Linear additive models.} Given $x_i = g_i\big(f_i(\mathbf{pa}(x_i)) + \epsilon_i\big)$, $f_i$ is linear and $g_i$ is the identity function. Examples include the family of linear non-Gaussian acyclic models, also known as LiNGAM \citep{shimizu2011directlingam}. While assuming linear causal functions is generally unrealistic for real-world data, LiNGAM methods can be very time-efficient. 
    \item \textit{Nonlinear additive models.} Here, $f_i$ is nonlinear and $g_i$ is the identity function. Examples include the nonlinear ANM presented by \citet{hoyer2008nonlinear}. Methods for nonlinear ANM discovery have more realistic assumptions than LiNGAM, but can be slow to run in practice.
    \item \textit{Multiplicative noise models.} The multiplicative noise model is a special case of the PNL, as $x_i = \mathbf{pa}(x_i) \cdot \epsilon_i$ can be expressed as $\exp(\log \mathbf{pa}(x_i) + \log \epsilon_i)$ where $f_i(\mathbf{pa}(x_i)) = \log(\mathbf{pa}(x_i))$ and $g_i(\cdot) = \exp(\cdot)$.
\end{itemize}

At a high-level, a simple formulation for FCM discovery can take the following approach. Assume the bivariate case with variables $X$ and $Y$, where edge directionality is unknown.
\begin{enumerate}
    \item Assume that $\{X,Y\}$ have a direct causal relationship with no confounders and specify the desired parametric assumptions.
    \item Fit the FCM for both causal directions and test for independence between the estimated noise and the hypothetical cause. For example, we could perform nonlinear regression of $y$ on $x$ to obtain $\hat{f}$, where residuals $\hat{\epsilon} =  y - \hat{f}(x)$.
    \item The direction that finds the hypothetical cause and noise terms to be statistically independent is considered plausible.
\end{enumerate}

In practice, causal discovery for FCMs often disaggregates the overall problem into two subproblems: (1) learning the causal ordering, where a topological sort (Definition \ref{def:topological_sort}) of the random variables is obtained; and (2) spurious edge pruning, where the final set of directed edges is obtained from the set of all possible edges admitted by the causal ordering \citep{peters2014causal,buhlmann2014cam}.

While the parametric assumptions of FCMs enable a wide range of interesting algorithms, these assumptions are not generally verifiable. Thus, justifying the realism of these assumptions can be challenging. Violations of these unverifiable parametric assumptions can impact performance on real-world data, which is a primary drawback to using FCMs in practice.

%\subsection{Alternative Approaches}

%In this section, we will briefly introduce two recent algorithms for graph search: order-search (OS) approach and integer linear programming (ILP) approach.

%\jm{Write a better summary of topological ordering based methods.}

%The OS approach, as its name suggests, conducts a search over the topological orders and the graph space at the same time. The K3 algorithm assumes a topological order in advance and searches only over the graphs that obey the topological order. When the order specified is a poor one, it may end with a bad graph structure (with a low graph score). The OS algorithm resolves this problem by performing a search over orders at the same time. It swaps the order of two adjacent variables at each step and employs the K3 algorithm as a sub-routine.

%The ILP approach encodes the graph structure, scoring and the acyclic constraints into a linear programming problem. Thus it can utilize a state-of-art integer programming solver. That said, this approach requires a bound on the maximum number of parents any node in the graph can have (say to be 4 or 5). Otherwise, the number of constraints in the ILP will explode and the computation will become intractable.

%\jm{Orthoganality in information space \citep{janzing2012information,janzing2015justifying}.}

\begin{reading}
\vspace{1mm}
    \textit{Structure Learning in Bayesian Networks}
    \begin{itemize}[leftmargin=*]
        \item \fullcite{heckerman1995learning}.
        \item \fullcite{chickering_bayes_nets_1995}.
    \end{itemize}
    \textit{Causal Discovery}
    \begin{itemize}[leftmargin=*]
        %\item \fullcite{scheines1997introduction}.
        \item \fullcite{spirtes2001causation}.
        \item \fullcite{heckerman2006bayesian}.
        \item \fullcite{glymour2019review}.
        \item \fullcite{vowels2022d}.
    \end{itemize}
\end{reading}

%%%%%%%%%%%%%%%%%%%%%%%%%%%%%%
%% Learning in undirected models
%%%%%%%%%%%%%%%%%%%%%%%%%%%%%%

\section{Learning in Undirected Models}
\label{sec:learning_undirected_models}

We now turn our attention to parameter learning in undirected graphical models. Unfortunately, as in the case of inference, the higher expressivity of undirected models also makes them significantly more difficult to learn than directed models. Fortunately, maximum likelihood learning in these models can be reduced to repeatedly performing inference, which will allow us to apply all the approximate inference techniques that we have seen earlier.

\subsection{Learning in Markov Random Fields}

Let us begin with an MRF (Section \ref{sec:mrf}) of the form
\[
p(\mathbf{x} = x_1, \dotsc, x_n) = \frac{1}{Z(\varphi)} \prod_{c \in \mathbf{C}} \phi_c(\mathbf{x}_c; \varphi),
\]
where
\[
 Z(\varphi) = \sum_{x_1, \dotsc, x_n}\prod_{c \in \mathbf{C}} \phi_c(\mathbf{x}_c; \varphi)
\]
is the normalizing constant. We can reparametrize $p$ as follows:
\begin{align*}
p(\mathbf{x})
& = \frac{1}{Z(\varphi)} \exp\left( \sum_{c \in \mathbf{C}} \log\phi_c(\mathbf{x}_c; \varphi) \right) \\
& = \frac{1}{Z(\varphi)} \exp\left( \sum_{c \in \mathbf{C}} \sum_{\mathbf{x}'_c} 1\{\mathbf{x}'_c = \mathbf{x}_c\} \log\phi_c(\mathbf{x}'_c; \varphi) \right) \\
& = \frac{\exp(\theta^T f(\mathbf{x}))}{Z(\theta)},
\end{align*}
where $f(\mathbf{x})$ is a vector of indicator functions and $\theta$ is the set of all the model parameters, as defined by the $\log \phi_c(\mathbf{x}_c'; \varphi)$.

Note that $Z(\theta)$ is different for each set of parameters $\theta$. Therefore, we also refer to $Z(\theta)$ as the \textit{partition function}. Bayesian networks are constructed such that $Z(\theta) = 1$ for all $\theta$. MRFs do not make this modeling assumption, which makes them more flexible but also more difficult to learn.

\subsubsection{Exponential Families}

More generally, distributions of the above form are members of the \textit{exponential family} of distributions. Many other distributions are in the exponential family, including the Bernoulli, Multinomial, Normal, and Poisson distributions. See \citet{wainwright_graphical_2008} for an overview of exponential families in graphical modeling.

Exponential families are widely used in machine learning. Suppose that you have an exponential family distribution of the form
\[
p(\mathbf{x}; \theta) = \frac{\exp(\theta^T f(\mathbf{x}))}{Z(\theta)}. 
\] %Actually, exponential families have a slightly more general form, but this one will be enough for our purposes.
Here are few facts about these distributions that are useful to keep in mind:
\begin{itemize}
    \item Exponential families are log-concave in their \textit{natural parameters} $\theta$. The partition function $Z(\theta)$ is also log-convex in $\theta$.
    \item The vector $f(\mathbf{x})$ is called the vector of \textit{sufficient statistics}. These fully describe the distribution $p$. For example, if $p$ is Gaussian, then $f(\mathbf{x})$ contains (simple reparametrizations of) the mean and the variance of $p$.
    \item Exponential families make the fewest unnecessary assumptions about the data distribution. More formally, the distribution maximizing the entropy $H(p)$ under the constraint $\mathbb{E}_p[\phi(\mathbf{x})] = \alpha$ (i.e., the sufficient statistics equal some value $\alpha$) is in the exponential family.
\end{itemize}

Exponential families are also very convenient to work with computationally. Their sufficient statistics can summarize arbitrary amounts of iid variables from the same distribution, and they admit so-called conjugate priors which makes them easily applicable in variational inference.
%In short, it definitely pays off to learn more about exponential families!

\subsection{Maximum Likelihood Learning in MRFs}

Suppose now that we are given a dataset $\mathcal{D}$ and we want to estimate $\theta$ via maximum likelihood. Since we are working with an exponential family, the maximum likelihood will be concave. We can express the log-likelihood as
\[
\frac{1}{|\mathcal{D}|} \log p(\mathcal{D}; \theta) = \frac{1}{|\mathcal{D}|} \sum_{\mathbf{x} \in \mathcal{D}} \theta^T f(\mathbf{x}) - \log Z(\theta).
\]
The first term is linear in $\theta$ and is easy to handle. The second term is equal to
\[
\log Z(\theta) = \log \sum_\mathbf{x} \exp(\theta^T f(\mathbf{x})).
\]
Unlike the first term, this one does not decompose across $\mathbf{x}$. It is not only hard to optimize, but also to evaluate.

Now consider the gradient of the log-likelihood. Obtaining the gradient of the linear part is very easy. However, the gradient of $\log Z(\theta)$ takes a more complicated form:
\[
\nabla_{\theta} \log Z(\theta) = \mathbb{E}_{\mathbf{x} \sim p} [ f(\mathbf{x}) ]. 
\]
This expression follows from differentiating the summation and applying the chain rule.
% This expression can be derived using simple algebra. \vk{might want to clarify?}

Computing the expectation on the right hand side of the equation requires inference with respect to $p$. For example, we could sample from $p$ and construct a Monte Carlo estimate of the expectation.
However, as we have seen, inference in general is intractable, and therefore so is computing the gradient.

We can similarly derive an expression for the Hessian of $\log Z(\theta)$:
\[
\nabla_{\theta}^2 \log Z(\theta) = \mathbb{E}_{\mathbf{x} \sim p} [ f(\mathbf{x}) f(\mathbf{x})^T ] - \mathbb{E}_{\mathbf{x} \sim p} [ f(\mathbf{x}) ] \mathbb{E}_{\mathbf{x} \sim p} [ f(\mathbf{x}) ]^T = \text{Cov}[f(\mathbf{x})].
\]
Since covariance matrices are always positive semi-definite, this proves that $\log Z (\theta)$ is convex (and therefore that the log-likelihood is concave). Recall that this was one of the properties of exponential families that we stated earlier.

In summary, even though the log-likelihood objective is convex, optimizing it is hard. Usually non-convexity is what makes optimization intractable, but in this case, it is the computation of the gradient that poses challenges.

\subsubsection{Approximate Learning Techniques}

Interestingly, however, maximum likelihood learning reduces to inference in the sense that we may repeatedly use inference to compute the gradient and determine the model weights using gradient descent.

This observation lets us apply to learning many of the approximate inference methods that we introduced in Chapter \ref{sec:approximate_inference}, such as:
\begin{itemize}
    \item \textit{Gibbs sampling} (Algorithm \ref{alg:gibbs_sampling}) from the distribution at each step of gradient descent. We can then approximate the gradient using Monte Carlo.
    \item \textit{Persistent contrastive divergence}, a variant of Gibbs sampling which re-uses the same Markov Chain between iterations. After a step of gradient descent, our model has changed very little. Thus, we can essentially keep taking samples from the same Gibbs sampler instead of starting a new one from scratch.\footnote{As an aside, persistent contrastive divergence is one of the most popular methods for training Restricted Boltzmann Machines, an early deep learning model that is also an undirected graphical model \citep{tieleman2008training}.}
\end{itemize}
These approaches replace exact gradient computation with approximations based on samples, allowing learning to proceed even when exact inference is intractable.
While such methods may introduce bias and variance into the gradient estimates, they are sometimes the only practical option for learning in large-scale undirected models.

\subsubsection{Pseudo-Likelihood}

Another popular approach to learning $p$ is called the \textit{pseudo-likelihood}. The pseudo-likelihood replaces the likelihood
\[
\frac{1}{|\mathcal{D}|} \log p(\mathcal{D}; \theta) = \frac{1}{|\mathcal{D}|} \sum_{x\in \mathcal{D}} \log p(\mathbf{x}; \theta)
\]
with the following approximation:
\[
\ell_\text{PL}(\theta ; \mathcal{D}) = \frac{1}{|\mathcal{D}|} \sum_{\mathbf{x} \in \mathcal{D}} \sum_{i} \log p(x_i \mid \textbf{ne}(x_i) ; \theta),
\]
where $x_i$ is the $i$-th variable in $\mathbf{x}$ and $\textbf{ne}(x_i)$ is the set of neighbors of $x_i$ (i.e., the Markov blanket of $x_i$).

Note that each term $\log p(x_i \mid \textbf{ne}(x_i); \theta)$ only involves one variable $x_i$ and hence its partition function is going to be tractable (we only need to sum over the values of one variable).

However, this approximation is not equal to the likelihood. Note that the correct way to expand the likelihood would involve the chain rule. That is, the terms would be the $\log p(x_i \mid x_{-i}; \theta)$ objective, where $x_{-i}$ are variables preceding $i$ in some ordering.

Intuitively, the pseudo-likelihood objective assumes that $x_i$ depends mainly on its neighbors in the graph and ignores the dependencies on other more distant variables. Observe also that if the pseudo-likelihood succeeds in matching all the conditional distributions to the data, a Gibbs sampler run on the model distribution will have the same invariant distribution as a Gibbs sampler run on the true data distribution, ensuring that they are the same.

More formally, we can show that the pseudo-likelihood objective is concave. Assuming the data are drawn from an MRF with parameters $\theta^*$, we can show that as the number of data points gets large, $\theta_\text{PL} \to \theta^*$. The pseudo-likelihood often works well in practice, although there are exceptions.

\subsubsection{Moment Matching}

We will end with an interesting observation about the maximum likelihood estimate $\hat{\theta}$ of the MRF parameters. Recall that the log-likelihood of an MRF is
\[
\frac{1}{|\mathcal{D}|} \log p(\mathcal{D}; \theta) = \frac{1}{|\mathcal{D}|} \sum_{\mathbf{x} \in \mathcal{D}} \theta^T f(\mathbf{x}) - \log Z(\theta).
\]
Taking the gradient using our expression for the gradient of the partition function, we obtain the expression
\[
\frac{1}{|\mathcal{D}|} \sum_{\mathbf{x} \in \mathcal{D}} f(\mathbf{x}) - \mathbb{E}_{\mathbf{x} \sim p} [ f(\mathbf{x}) ]. 
\]
Note that this is precisely the difference between the expectations of the natural parameters under the empirical and the model distributions. Let's now look at one component of $f(\mathbf{x})$. Recall that we have defined $f$ in the context of MRFs to be the vector of indicator functions for the variables of a clique: one entry of $f$ equals $\mathbb{I}[\mathbf{x}_c = \bar{\mathbf{x}}_c]$ 
for some $\mathbf{x}_c, \bar{\mathbf{x}}_c$. The gradient over that component equals
\[
\frac{1}{|\mathcal{D}|} \sum_{\mathbf{x} \in \mathcal{D}} \mathbb{I}[\mathbf{x}_c = \bar{\mathbf{x}}_c] - \mathbb{E}_{\mathbf{x} \sim p} [ \mathbb{I}[\mathbf{x}_c = \bar{\mathbf{x}}_c] ].
\]
This gives us an insight into how MRFs are trained. The log-likelihood objective forces the model marginals to match the empirical marginals.

We refer to the above property as \textit{moment matching}. This property of maximum likelihood learning is very general: whenever we choose distribution $q$ to minimize the inclusive KL divergence  $\text{KL}(p\, \| \, q)$ across $q$ in an exponential family, the minimizer will match the moments of the sufficient statistics to the corresponding moments of $p$. Recall that the MLE estimate is the minimizer over $q$ of $\text{KL}(\tilde{p} \; || \; q)$, where $\tilde{p}$ is the empirical distribution of the data (Section~\ref{sec:learning_theory_basics_the_learning_task}), so MLE estimation is a special case of this minimization. This property has connections to variational inference, where this minimization over $q$ in a smaller set of distributions $\mathcal{Q}$ is known as M-projection (``moment projection'').

\subsection{Learning in Conditional Random Fields}

Finally, we will consider how maximum likelihood learning extends to CRFs (Section \ref{sec:crf}). Recall that a CRF is a probability distribution of the form
\[
p(\mathbf{y} \mid \mathbf{x}) = \frac{1}{Z(\mathbf{x}, \varphi)} \prod_{c \in \mathbf{C}} \phi_c(\mathbf{y}_c, \mathbf{x}; \varphi),
\]
where
\[
Z(\mathbf{x}, \varphi) = \sum_{y_1, \dotsc, y_n}\prod_{c \in \mathbf{C}} \phi_c(\mathbf{y}_c, \mathbf{x}; \varphi)
\]
is the partition function. The feature functions now depend on $\mathbf{x}$ in addition to $\mathbf{y}$. The $\mathbf{x}$ variables are fixed and the distribution is only over $\mathbf{y}$. The partition function is thus a function of both $\mathbf{x}$ and $\varphi$. We can reparametrize $p$ as we did for MRFs:
\[
p(\mathbf{y} \mid \mathbf{x}) = \frac{\exp(\theta^T f(\mathbf{x}, \mathbf{y}))}{Z(\mathbf{x}, \theta)}, 
\]
where $f(\mathbf{x}, \mathbf{y})$ is again a vector of indicator functions and $\theta$ is a reparametrization of the model parameters. The log-likelihood for this model given a dataset $\mathcal{D}$ is
\[
\frac{1}{|\mathcal{D}|} \log p(\mathcal{D}; \theta) = \frac{1}{|\mathcal{D}|} \sum_{\mathbf{x},\mathbf{y} \in \mathcal{D}} \theta^T f(\mathbf{x}, \mathbf{y}) - \frac{1}{|\mathcal{D}|} \sum_{\mathbf{x} \in \mathcal{D}} \log Z(\mathbf{x}, \theta).
\]
Note that this is almost the same form as we had for MRFs, except that now there is a different partition function $\log Z(\mathbf{x}, \theta)$ for each data point $\mathbf{x},\mathbf{y}$. The gradient is now
\[
\frac{1}{|\mathcal{D}|} \sum_{\mathbf{x}, \mathbf{y} \in \mathcal{D}} f(\mathbf{x}, \mathbf{y}) - \frac{1}{|\mathcal{D}|} \sum_{\mathbf{\mathbf{x}} \in \mathcal{D}} \mathbb{E}_{\mathbf{y} \sim p(\mathbf{y} \mid \mathbf{x})} [ f(\mathbf{x},\mathbf{y}) ]
\]
Similarly, the Hessian is going to be the covariance matrix
\[
\text{Cov}_{\mathbf{y} \sim p(\mathbf{y} \mid \mathbf{x})} [ f(\mathbf{x},\mathbf{y}) ].
\]

The good news is that the conditional log-likelihood is still a concave function. We can optimize it using gradient ascent as before. The bad news is that computing the gradient now requires one inference per training data point $\mathbf{x}, \mathbf{y}$ in order to compute the term
\[
\sum_{\mathbf{x}, \mathbf{y} \in \mathcal{D}} \log Z(\mathbf{x}, \theta). 
\]
This makes learning CRFs more expensive than learning in MRFs. In practice, however, CRFs are more widely used than MRFs:  supervised learning is a widely used learning paradigm, and discriminative models (like CRFs) often learn a better classifier than their generative counterparts (which model $p(\mathbf{x},\mathbf{y})$).

To deal with the computational difficulties introduced by the partition function, we can use simpler models in which exact inference is tractable. This was the approach taken in the optical character recognition example that we introduced in Section \ref{sec:crf} (Figure \ref{fig:optical_character_recognition}). More generally, one should try to limit the number of variables or make sure that the model's graph is not too densely connected.

Finally, we would like to add that there exists another popular objective for training CRFs called the \textit{max-margin loss}, a generalization of the objective for training support vector machines. Models trained using this loss are called structured support vector machines or max-margin networks \citep{tsochantaridis2004support,tsochantaridis2005large}. This loss is more widely used in practice because it often leads to better generalization. Additionally, it requires only MAP inference to compute the gradient rather than general (e.g., marginal) inference, which is often more expensive to perform.

%%%%%%%%%%%%%%%%%%%%%%%%%%%%%%
%% Learning in latent variable models
%%%%%%%%%%%%%%%%%%%%%%%%%%%%%%

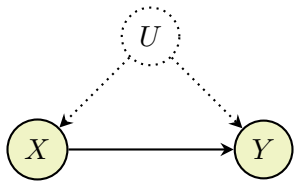
\begin{figure}[!t]
    \centering

\begin{tikzpicture}

% Nodes
\node[draw,circle,thick,black,dotted,scale=1] (U) at (0,0) {$U$};
\node[draw,circle,thick,black,scale=1,fill=GreenYellow!40]  (X) at (-1.5,-1.5) {$X$};
\node[draw,circle,thick,black,scale=1,fill=GreenYellow!40]  (Y) at (1.5,-1.5) {$Y$};

% Edges
\draw[thick,-{Stealth[width=5pt,length=5pt]}, dotted] (U) -- (X);
\draw[thick,-{Stealth[width=5pt,length=5pt]}, dotted] (U) -- (Y);
\draw[thick,-{Stealth[width=5pt,length=5pt]}] (X) -- (Y);
    
\end{tikzpicture}

    \caption{A DAG with latent variable $U$ and observed variables $X$ and $Y$.}
    \label{fig:latent_triangle}
\end{figure}

\section{Learning in Latent Variable Models}
\label{sec:learning_latent_models}

So far, we have assumed that learning takes place in settings where all variables of interest are \textit{observed} or \textit{measured}. However, this is not always the case in practice.

Consider a probabilistic language model of news articles. Each article $\mathbf{x}$ focuses on a specific topic $\mathbf{t}$ (e.g., finance, sports, politics, etc.). Using this prior knowledge, we can build a more accurate model $p(\mathbf{x} \mid \mathbf{t})p(\mathbf{t})$, in which we introduce an additional variable $\mathbf{t}$ that we have not directly observed. In this case, we say that $\mathbf{t}$ is \textit{unobserved}, \textit{unmeasured}, \textit{hidden}, or \textit{latent}. Our new model $p(\mathbf{x} \mid \mathbf{t})p(\mathbf{t})$ can provide greater accuracy, as we can now learn a separate $p(\mathbf{x} \mid \mathbf{t})$ for each topic rather than trying to model everything with $p(\mathbf{x})$. However, since $\mathbf{t}$ is unobserved, we cannot directly use the learning methods that we have seen so far. In fact, unobserved variables make learning much more difficult.

In this section, we will look at how to use and how to learn models that involve latent variables. More formally, we define a latent variable model (LVM) $p$ as a probability distribution
\[
p(\mathbf{x}, \mathbf{z}; \theta)
\]
over two sets of variables $\mathbf{X}, \mathbf{Z}$, where variables in $\mathbf{X}$ are observed at learning time and variables in $\mathbf{Z}$ are never observed (and where $\theta$ are model parameters, as before). LVMs can be directed or undirected (Figure \ref{fig:latent_triangle}). There exist both discriminative and generative LVMs, although we will focus on the latter (though the key ideas will hold for discriminative models as well).

\subsubsection{Why are Latent Variable Models Useful?}

There are several notable reasons to use LVMs. The simplest reason is that some data are naturally unobserved. For example, clinical trials commonly see participants drop out before all measurements are taken. LVMs can be used to learn in the presence of such missing data.

Perhaps the most important reason for studying LVMs is that they enable us to leverage prior knowledge when defining a model. Our topic modeling example illustrates this. We know that our set of news articles is actually a mixture of $K$ distinct distributions (one for each topic). LVMs allow us to design models that capture this. 

LVMs can also be viewed as increasing the expressive power of our model. We illustrate this point by taking Gaussian Mixture Models (GMMs) as an example.

\begin{figure}
    \centering
    \includegraphics[width=0.45\linewidth]{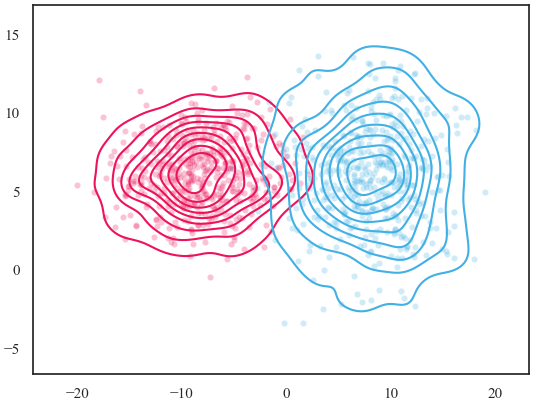}
    \includegraphics[width=0.45\linewidth]{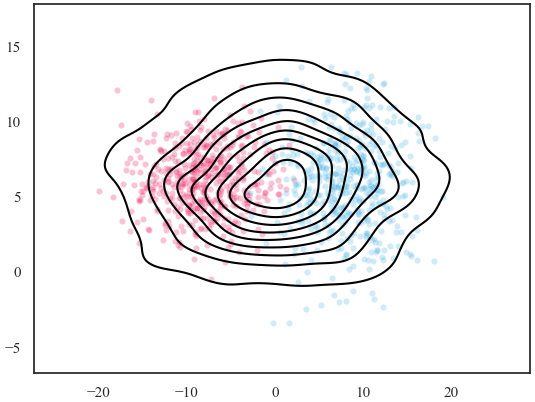}
    \caption{Example of a dataset that is better fit with a mixture of two Gaussians (left) than with only one (right). Mixture models allow us to model clusters in the dataset, as seen here in red and blue.}
    \label{fig:gmm}
\end{figure}

\paragraph{Gaussian Mixtures} GMMs are probabilistic LVMs that assume the data of interest are generated from a mixture of  Gaussian distributions --- that is, the data represent a finite number of normally distributed subpopulations. GMMs are one of the most widely used models in machine learning. The distribution that we can model using a mixture of Gaussian components is much more expressive than what we could have modeled using a single component (Figure \ref{fig:gmm}).

In a GMM, each data point is a tuple $(\mathbf{x}_i, z_i)$ with 
$d$-dimensional vector $\mathbf{x}_i \in \mathbb{R}^d$ and discrete scalar $z_i \in \{1, 2, \dotsc, K\}$. The joint $p$ is a directed model
\[
p(\mathbf{x}, z) = p(\mathbf{x} \mid z)p(z), 
\]
where $p(z = k) = \pi_k$ for vector of class probabilities $\pi \in \Delta_{K-1}$ and
\[
p(\mathbf{x} \mid z=k) = \mathcal{N}(\mathbf{x}; \mu_k, \Sigma_k) 
\]
is a multivariate Gaussian with mean and variance $\mu_k, \Sigma_k$. This model postulates that our observed data are comprised of $K$ clusters with proportions specified by $\pi_1, \dotsc, \pi_K$. The distribution within each cluster is a Gaussian. We can see that $p(\mathbf{x})$ is a mixture (i.e., a weighted sum) by explicitly writing out this probability:
\[
p(\mathbf{x}) = \sum_{k=1}^K p(\mathbf{x} \mid z=k)p(z=k) = \sum_{k=1}^K \pi_k \; \mathcal{N}(\mathbf{x}; \mu_k, \Sigma_k).
\]
To generate a new data point, we can sample a cluster $k$ and then draw a sample from the corresponding Gaussian $\mathcal{N}(\mathbf{x}; \mu_k, \Sigma_k)$. This way, we could sample from either the red or blue cluster in Figure \ref{fig:gmm}, which would not be possible if we had assumed that all data points were generated by a single Gaussian. We will revisit GMMs in Section \ref{sec:em_gmm} to illustrate our learning algorithms in action.

% include marginfigure.html id=``gmm1'' url=``assets/img/gmm1.png'' description=``Example of a dataset that is best fit with a mixture of two Gaussians. Mixture models allow us to model clusters in the dataset.'' %}

%{% include maincolumn_img.html src='assets/img/gmm2.png' caption='Example of a Gaussian mixture model, consisting of three components with different class proportions (a). The true class of each point is unobserved, so the distribution over $x$ looks like in (b); it is both multi-modal and non-Gaussian. Visualizing it in 3D shows the effects of class proportions on the magnitudes of the modes.' %}

\subsubsection{Marginal Likelihood Training}

How do we train an LVM? Our goal is still to fit the marginal distribution $p(\mathbf{x})$ over the set of variables $\mathbf{X}$ to our observed dataset $\mathcal{D}$. Hence our previous discussion about KL divergences applies here as well. By the same argument, we should be maximizing the \textit{marginal log-likelihood} of the data
\[
\log p(\mathcal{D}) = \sum_{\mathbf{x} \in \mathcal{D}} \log p(\mathbf{x}) = \sum_{\mathbf{x} \in \mathcal{D}} \log \left( \sum_\mathbf{z} p(\mathbf{x} \mid \mathbf{z}) p(\mathbf{z}) \right).
\]
This optimization objective is considerably more difficult than the regular log-likelihood, even for directed graphical models. For one, we can see that the summation inside the log makes it impossible to decompose $p(\mathbf{x})$ into a sum of log-factors. Even if the model is directed, we can no longer derive a simple closed-form expression for the parameters.

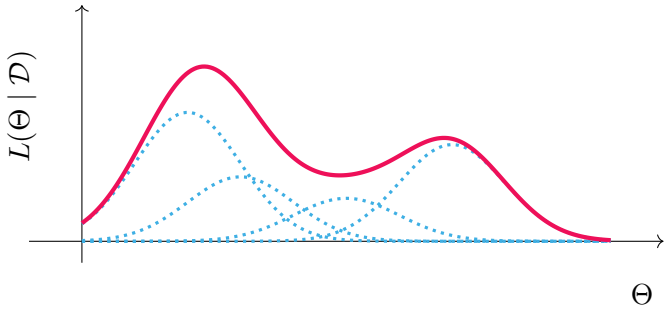
\begin{figure}[!t]
    \centering

\begin{tikzpicture}

    \begin{axis}[
        axis lines=middle,
        enlargelimits = true,
        xlabel={$\Theta$},
        ylabel={$L(\Theta \mid \mathcal{D})$},
        xlabel style={at={(1,-0.2)}},
        ylabel style={rotate=90, at={(-0.05,0.35)}},
        xmin=0, xmax=10,
        ymin=0, ymax=2,
        xtick=\empty,
        ytick=\empty,
        domain=0:10,
        samples=100,
        width=10cm, height=5cm,
        axis line style={->}
    ]

        % Individual components (gray curves)
        \addplot[smooth, very thick, dotted, CornflowerBlue] {1.2*exp(-0.5*(x-2)^2)};
        \addplot[smooth, very thick, dotted, CornflowerBlue] {0.4*exp(-0.5*(x-5)^2)};
        \addplot[smooth, very thick, dotted, CornflowerBlue] {0.9*exp(-0.5*(x-7)^2)};
        \addplot[smooth, very thick, dotted, CornflowerBlue] {0.6*exp(-0.5*(x-3)^2)};

        % Combined curve (black)
        \addplot[smooth, ultra thick, OrangeRed] {
            1.2*exp(-0.5*(x-2)^2) +
            0.4*exp(-0.5*(x-5)^2) +
            0.9*exp(-0.5*(x-7)^2) +
            0.6*exp(-0.5*(x-3)^2)
        };

    \end{axis}
\end{tikzpicture}

    \caption{Exponential family distributions (dotted blue lines) have concave log-likelihoods. However, a weighted mixture of such distributions is no longer concave (solid red line).}
    \label{fig:exponential_families}
\end{figure}

Looking closer at the distribution of a data point $\mathbf{x}$, we also see that it is actually a mixture
\[
p(\mathbf{x}) = \sum_\mathbf{z} p(\mathbf{x} \mid \mathbf{z}) p(\mathbf{z})
\]
of distributions $p(\mathbf{x} \mid \mathbf{z})$ with weights $p(\mathbf{z})$. Whereas a single exponential family distribution $p(\mathbf{x})$ has a concave log-likelihood (as we have seen in our discussion of undirected models), the log of a weighted mixture of such distributions is no longer concave or convex (Figure \ref{fig:exponential_families}). This non-convexity requires the development of specialized learning algorithms.

\subsection{Learning with Expectation-Maximization}
\label{sec:learning_with_expectation_maximization}

Since our objective is non-convex, we will resort to approximate learning algorithms. These methods are widely used and quite effective in practice. We will focus our discussion on \textit{Expectation-Maximization} (EM): a hugely important and widely used algorithm for learning directed latent variable graphical models \citep{dempster1977maximum}.

The EM algorithm learns a LVM $p(\mathbf{x},\mathbf{z}; \theta)$ with parameters $\theta$ and latent $\mathbf{z}$. This approach relies on two simple observations:
\begin{enumerate}
    \item If the latent $\mathbf{z}$ were fully observed, then we could optimize the log-likelihood exactly using the closed form solution for $p(\mathbf{x},\mathbf{z})$.
    \item Knowing the parameters $\theta$, we can often efficiently compute the posterior $p(\mathbf{z} \mid \mathbf{x}; \theta)$. Note that this is an assumption, and this does not hold for some models.
\end{enumerate}

EM follows a simple iterative two-step strategy (Algorithm \ref{alg:em}): given an estimate $\theta_t$ of the parameters at time-step $t$, compute $p(\mathbf{z} \mid \mathbf{x})$ and use it to ``hallucinate'' values for $\mathbf{z}$. Then, find a new $\theta_{t+1}$ by optimizing the resulting tractable objective. This process will eventually converge.

We haven't exactly defined what we mean by ``hallucinating'' values for $\mathbf{z}$. The full definition is a bit technical, but its instantiation is very intuitive in most models (e.g., GMMs). Here, ``hallucinating'' means computing the expected log-likelihood
\[
\mathbb{E}_{\mathbf{z} \sim p(\mathbf{z} \mid \mathbf{x})} \log p(\mathbf{x},\mathbf{z}; \theta). 
\]
This expectation is what gives the EM algorithm half of its name. If $\mathbf{z}$ is not too high-dimensional (e.g., in GMMs it is a one-dimensional categorical variable), then we can compute this expectation.

Since the summation is now outside the log, we can maximize the expected log-likelihood. In particular, when $p$ is a directed model, $\log p$ again decomposes into a sum of log-conditional probability distribution terms that can be optimized independently.

\begin{algorithm}[!t]
\caption{\textit{Expectation-Maximization}}
\label{alg:em}
\vspace{3mm}
\textbf{Input:} Dataset $\mathcal{D}$.
\begin{algorithmic}[1]
\vspace{3mm}
\STATE Initialize parameters at $\theta_0$.
\STATE Repeat until convergence for $t = 1, 2, \dotsc$:
\STATE \quad\quad - \textit{E-Step}: For each $\mathbf{x} \in \mathcal{D}$, compute the posterior $p(\mathbf{z} \mid \mathbf{x}; \theta_t)$.
\STATE \quad\quad - \textit{M-Step}: Compute new parameters via
    \[
    \theta_{t+1} = \argmax_\theta \sum_{\mathbf{x} \in \mathcal{D}} \mathbb{E}_{\mathbf{z} \sim p(\mathbf{z} \mid \mathbf{x}; \theta_t)} \log p(\mathbf{x},\mathbf{z}; \theta).
    \]
\end{algorithmic}
\end{algorithm}

\subsubsection{EM as Variational Inference}

Why exactly does EM converge? We can understand the behavior of EM by casting it in the framework of variational inference.

Consider the posterior inference problem for $p(\mathbf{z} \mid \mathbf{x})$, where the $\mathbf{x}$ variables are held fixed as evidence. We can apply our variational inference framework by taking $p(\mathbf{x},\mathbf{z})$ to be the unnormalized distribution. In that case, $p(\mathbf{x})$ will be the normalization constant. Recall that variational inference maximizes the evidence lower bound (ELBO)
\[
\mathcal{L}(p,q) = \mathbb{E}_{q(\mathbf{z})} \left[ \log p(\mathbf{x},\mathbf{z}; \theta) - \log q(\mathbf{z}) \right]
\]
over distributions $q$. The ELBO satisfies the equation
\[
\log p(\mathbf{x}; \theta) = \text{KL}(q(\mathbf{z}) \; \| \; p(\mathbf{z} \mid \mathbf{x} ; \theta)) + \mathcal{L}(p,q). 
\]
Hence, $\mathcal{L}(p,q)$ is maximized when $q=p(\mathbf{z} \mid \mathbf{x})$. In that case, the KL term becomes zero and the lower bound is tight: $\log p(\mathbf{x}; \theta) = \mathcal{L}(p,q).$

The EM algorithm can be seen as iteratively optimizing the ELBO over $q$ (at the $E$-step) and over $\theta$ (at the $M$-step). Starting at some $\theta_t$, we compute the posterior $p(\mathbf{z} \mid \mathbf{x}; \theta_t)$ at the $E$-step. We evaluate the ELBO for $q = p(\mathbf{z} \mid \mathbf{x}; \theta_t)$. This makes the ELBO tight:
\[
\log p(\mathbf{x}; \theta_t) = \mathbb{E}_{p(\mathbf{z} \mid \mathbf{x}; \theta_t)} \log p(\mathbf{x},\mathbf{z}; \theta_t) - \mathbb{E}_{p(\mathbf{z} \mid \mathbf{x}; \theta_t)} \log p(\mathbf{z} \mid \mathbf{x}; \theta_t)
\]
Next, we optimize the ELBO over $p$, holding $q$ fixed. We then solve the problem
\[
\theta_{t+1} = \argmax_\theta \mathbb{E}_{p(\mathbf{z} \mid \mathbf{x}; \theta_t)} \log p(\mathbf{x},\mathbf{z}; \theta) - \mathbb{E}_{p(\mathbf{z} \mid \mathbf{x}; \theta_t)} \log p(\mathbf{z} \mid \mathbf{x}; \theta_t).
\]
Note that this is precisely the optimization problem solved at the $M$-step of EM (in the above equation, there is an additive constant independent of $\theta$). Solving this problem increases the ELBO. However, since we fixed $q$ to $\log p(\mathbf{z} \mid \mathbf{x}; \theta_t)$, the ELBO evaluated at the new $\theta_{t+1}$ is no longer tight. But since the ELBO was equal to $\log p(\mathbf{x} ;\theta_t)$ before optimization, we know that the true log-likelihood $\log p(\mathbf{x};\theta_{t+1})$ must have increased.

We now repeat this procedure, computing $p(\mathbf{z} \mid \mathbf{x}; \theta_{t+1})$ (the $E$-step), plugging $p(\mathbf{z} \mid \mathbf{x}; \theta_{t+1})$ into the ELBO (which makes the ELBO tight), and maximizing the resulting expression over $\theta$. Every step increases the marginal likelihood $\log p(\mathbf{x}; \theta_t)$, which is what we wanted to show.

\subsubsection{Properties of EM}

Following from the above discussion, EM has the following properties:
\begin{enumerate}
    \item The marginal likelihood increases after each EM cycle.
    \item Since the marginal likelihood is upper-bounded by its true global maximum and it increases at every step, EM must eventually converge.
\end{enumerate}

However, since we are optimizing a non-convex objective, we have no guarantee to find the global optimum. In fact, EM in practice converges almost always to a local optimum. Moreover, that optimum heavily depends on the choice of initialization. Different initial $\theta_0$ can lead to very different solutions, and so it is very common to use multiple restarts of the algorithm and choose the best one in the end. In fact EM is so sensitive to the choice of initial parameters, that techniques for choosing these parameters are still an active area of research.

\subsection{Illustrative Example: EM in Gaussian Mixture Models}
\label{sec:em_gmm}

Consider the use of EM in the context of GMMs. Suppose we have a dataset $\mathcal{D}$. In the $E$-step, we can compute the posterior for each data point $\mathbf{x}$ as follows:
\[
p(z \mid \mathbf{x}; \theta_t) = \frac{p(z, \mathbf{x}; \theta_t)}{p(\mathbf{x}; \theta_t)} = \frac{p(\mathbf{x} \mid z; \theta_t) p(z; \theta_t)}{\sum_{k=1}^K p(\mathbf{x} \mid z_k; \theta_t) p(z_k; \theta_t)}.
\]
Note that each $ p(\mathbf{x} \mid z_k; \theta_t) p(z_k; \theta_t) $ is simply the probability that $\mathbf{x}$ originates from component $k$ given the current set of parameters $\theta$. After normalization, these form the $K$-dimensional vector of probabilities $p(z \mid \mathbf{x}; \theta_t)$.

Recall that in the original model, $z$ is an indicator variable that chooses a component for $\mathbf{x}$. We can view this as a ``hard'' assignment of $\mathbf{x}$ to one component. The result of the $E$ step is a $K$-dimensional vector (whose components sum to one) that specifies a ``soft'' assignment to components. In that sense, we have hallucinated a ``soft'' instantiation of $z$. This is what we meant earlier by an ``intuitive interpretation'' for $p(z \mid \mathbf{x}; \theta_t)$. 

At the $M$-step, we optimize the expected log-likelihood of our model.
\begin{align*}
\theta_{t+1}
& = \argmax_\theta \sum_{\mathbf{x} \in \mathcal{D}} \mathbb{E}_{z \sim p(z \mid \mathbf{x}; \theta_t)} \log p(\mathbf{x},z; \theta) \\
& = \argmax_\theta \sum_{k=1}^K \sum_{\mathbf{x} \in \mathcal{D}} p(z_k \mid \mathbf{x}; \theta_t) \log p(\mathbf{x} \mid z_k; \theta)  \\
& \hspace{38pt} + \sum_{k=1}^K \sum_{\mathbf{x} \in \mathcal{D}} p(z_k \mid x; \theta_t) \log p(z_k; \theta).
\end{align*}
We can optimize each of these terms separately. We will start with $p(\mathbf{x} \mid z_k; \theta) = \mathcal{N}(\mathbf{x}; \mu_k, \Sigma_k)$. We have to find the $\mu_k, \Sigma_k$ that maximize
\[
\sum_{\mathbf{x} \in \mathcal{D}} p(z_k \mid \mathbf{x}; \theta_t) \log p(\mathbf{x} \mid z_k; \theta)
= c_k \cdot \mathbb{E}_{\mathbf{x} \sim Q_k(\mathbf{x})} \log p(\mathbf{x} \mid z_k; \theta),
\]
where $c_k = \sum_{\mathbf{x} \in \mathcal{D}} p(z_k \mid \mathbf{x}; \theta_t)$ is a constant that does not depend on $\theta$ and $Q_k(\mathbf{x})$ is a probability distribution defined over $\mathcal{D}$ as
\[
Q_k(x) = \frac{p(z_k \mid \mathbf{x}; \theta_t)}{\sum_{\mathbf{x} \in \mathcal{D}} p(z_k \mid \mathbf{x}; \theta_t)}.
\]
We know that $\mathbb{E}_{\mathbf{x} \sim Q_k(x)} \log p(\mathbf{x} \mid z_k; \theta)$ is optimized when $p(\mathbf{x} \mid z_k; \theta)$ equals $Q_k(\mathbf{x})$ (as discussed in the section on learning directed models, this objective equals the KL divergence between $Q_k$ and $p$, plus a constant). Moreover, since $p(\mathbf{x} \mid z_k; \theta) = \mathcal{N}(\mathbf{x}; \mu_k, \Sigma_k)$ is in the exponential family, it is entirely described by its sufficient statistics. Thus, we can set the mean and variance $\mu_k, \Sigma_k$ to those of $Q_k(\mathbf{x})$, which are
\[
\mu_k = \mu_{Q_k} = \sum_{\mathbf{x} \in \mathcal{D}} \frac{ p(z_k \mid \mathbf{x}; \theta_t)}{\sum_{\mathbf{x} \in \mathcal{D}} p(z_k \mid \mathbf{x}; \theta_t)} \mathbf{x}
\]
and
\[
\Sigma_k = \Sigma_{Q_k} = \sum_{\mathbf{x} \in \mathcal{D}} \frac{ p(z_k \mid \mathbf{x}; \theta_t)}{\sum_{\mathbf{x} \in \mathcal{D}} p(z_k \mid \mathbf{x}; \theta_t)} (x-\mu_{Q_k}) (x-\mu_{Q_k})^T.
\]
Note how these are the just the mean and variance of the data, weighted by their cluster affinities! Similarly, we may find out that the class priors are
\[
\pi_k = \frac{1}{|\mathcal{D}|}\sum_{\mathbf{x} \in \mathcal{D}} p(z_k \mid \mathbf{x} ; \theta_t). 
\]
Although we have derived these results using general facts about exponential families, it is equally possible to derive them using standard calculus techniques.

\begin{reading}

\begin{itemize}[leftmargin=*]
    \item \fullcite{bishop1998latent}.
    \item Chapter 27 in \fullcite{murphy2012machine}.
\end{itemize}
    
\end{reading}

%%%%%%%%%%%%%%%%%%%%%%%%%%%%%%
%% Bayesian learning
%%%%%%%%%%%%%%%%%%%%%%%%%%%%%%

\section{Bayesian Learning}

\label{sec:learning_bayesian_models}

The learning approaches we have discussed so far are based on the principle of maximum likelihood estimation. While being extremely general, there are limitations of this approach as illustrated in the two examples below.

\paragraph{Example 1}

Suppose we are interested in modeling the outcome of a biased coin, $X \in \{\text{heads}, \text{tails}\}$. We toss the coin 10 times and observe 6 heads. If $\theta$ denotes the probability of observing heads, the maximum likelihood estimate (MLE) is given by
\[
\hat{\theta} = \frac{n_\text{heads}}{n_\text{heads} + n_\text{tails}} = 0.6.
\]
Now, suppose we continue tossing the coin such that after 100 total trials (including the 10 initial trials), we observe 60 heads. Again, we can compute the  maximum likelihood estimate as
\[
\hat{\theta} = \frac{n_\text{heads}}{n_\text{heads} + n_\text{tails}} = 0.6.
\]
In both of the above situations, the maximum likelihood estimate does not change as we observe more data. This seems counterintuitive --- our \textit{confidence} in predicting heads with probability 0.6 should be higher in the second setting where we have seen many more trials of the coin. The key problem is that we represent our belief about the probability of heads $\theta$ as a single number $\hat{\theta}$, so there is no way to represent whether we are more or less sure about $\theta$. 

\paragraph{Example 2}

Consider a language model for sentences based on the bag-of-words assumption. A bag-of-words model has a generative process where a sentence is formed from a sample of words which are metaphorically ``pulled out of a bag'' (i.e., sampled independently). In such a model, the probability of a sentence can be factored as the probability of the words appearing in the sentence. For a sentence $S$ consisting of words $w_1, \ldots, w_n$, we have 
\[
p(S) = \prod_{i=1}^n p(w_n).
\]
For simplicity, assume that our language corpus consists of a single sentence, ``Probabilistic graphical models are fun. They are also powerful.'' We can estimate the probability of each of the individual words based on the counts. Our corpus contains 10 words with each word appearing once. Thus, each word in the corpus is assigned a probability of 0.1. Now, while testing the generalization of our model to the English language, we observe another sentence, ``Probabilistic graphical models are hard.'' The probability of the sentence under our model is
$0.1 \times 0.1 \times 0.1 \times 0.1 \times 0 = 0$. We did not observe one of the words (``hard'') during training. This caused our language model to infer that the sentence is impossible, even though it is a perfectly plausible sentence.

Out-of-vocabulary words are a common phenomenon, even for language models trained on large corpus. One of the simplest ways to handle these words is to assign a prior probability of observing an out-of-vocabulary word, such that the model will assign a low but non-zero probability to test sentences containing such words. As an aside, modern systems commonly use \textit{tokenization}, where a set of fundamental tokens can be combined to form any word. Hence the word ``Hello'' as a single token and the word ``Bayesian'' is encoded as ``Bay'' + ``esian'' under the common Byte Pair Encoding. This can be viewed as putting a prior over all words, where longer words are less likely.

\subsubsection{Modeling Uncertainty in Bayesian Learning}

We can address some of the limitations of  maximum likelihood estimation by adopting a Bayesian framework. In contrast to maximum likelihood learning, Bayesian learning explicitly models uncertainty over both the observed variables $\mathbf{X}$ and the parameters $\theta$. In other words, the parameters $\theta$ are random variables as well. 

A prior distribution over the parameters $p(\theta)$ encodes our initial beliefs. These beliefs are subjective. For example, we can choose the prior over $\theta$ for a biased coin to be uniform between 0 and 1. If we expect the coin to be fair, the prior distribution can be peaked around $\theta = 0.5$. We will discuss commonly used priors later in this chapter.

Say we observe the dataset $\mathcal{D} = \lbrace x_1, \cdots, x_N \rbrace$ (e.g., in the coin toss example, each $X_i$ is the outcome of one toss of the coin). We can update our beliefs using Bayes' rule 
\[
p(\theta \mid \mathcal{D}) = \frac{p(\mathcal{D} \mid \theta) \, p(\theta)}{p(\mathcal{D})},
\]
where 
\[
\underbrace{\mystrut{1.5ex} p(\theta \mid \mathcal{D})}_{\text{posterior}} \propto \underbrace{\mystrut{1.5ex} p(\mathcal{D} \mid \theta)}_{\text{likelihood}} \, \underbrace{\mystrut{1.5ex} p(\theta)}_{\text{prior}}.
\]
%\[
%\text{posterior} \propto \text{likelihood} \times \text{prior}.
%\]
Thus, Bayesian learning provides a principled mechanism for incorporating prior knowledge into our model. Bayesian learning is useful in many situations, such as when want to provide uncertainty estimates about the model parameters (Example 1) or when the data available for learning a model is limited (Example 2).

\begin{figure}[!t]
    \centering
    %\fbox{
    %\includegraphics[width=\linewidth]{images/conjugate_priors/normal_likelihood.png}
    \includegraphics[width=0.8\linewidth]{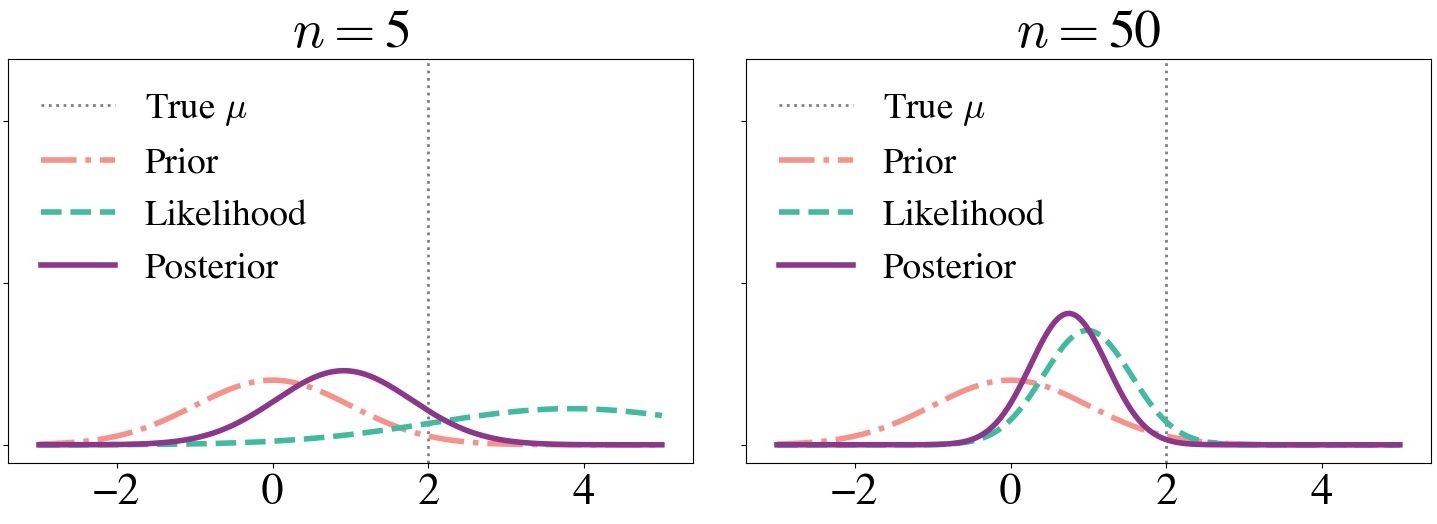} \\
    \vspace{5mm}
    \includegraphics[width=0.8\linewidth]{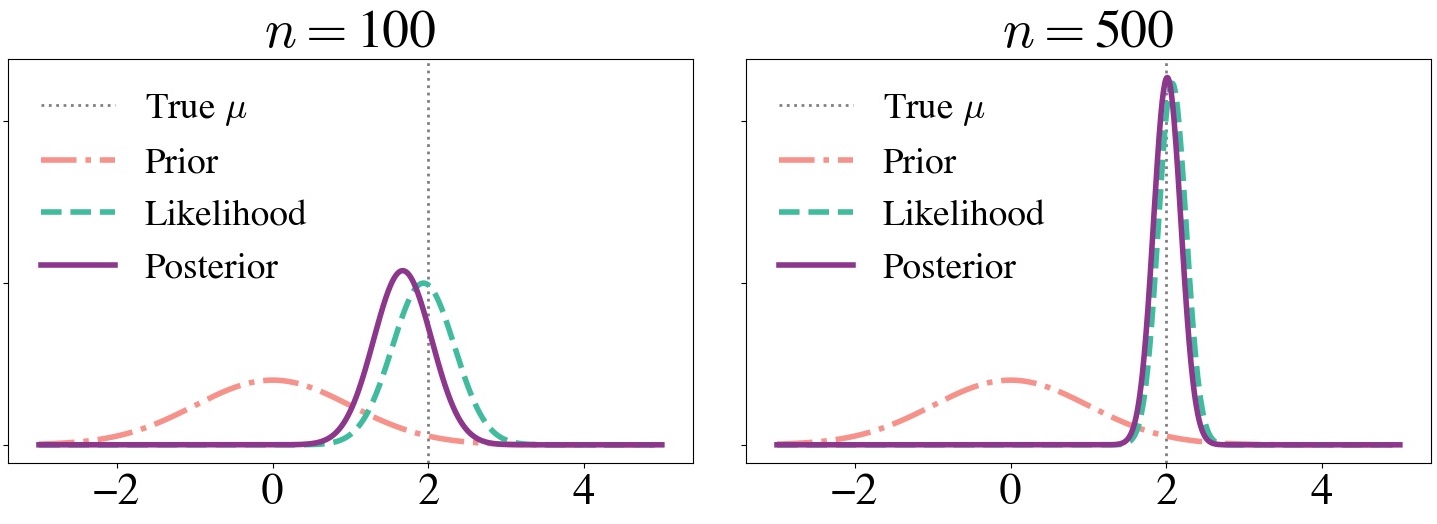}
    %} \\
    %\vspace{5mm}
    %\fbox{\includegraphics[width=\linewidth]{images/conjugate_priors/bimomial_likelihood.png}}
    \caption{As the number of data samples $n$ increases, the influence of the prior on our posterior diminishes. Meanwhile, the influence of the likelihood comes to dominate. In this example, our data are drawn from a Gaussian distribution with a true mean parameter ($\mu$) of 2.}
    \label{fig:likelihood_dominates}
\end{figure}

This framework is flexible in how much influence we place on prior knowledge versus the observed data. When we do not have much prior knowledge or certainty about $\theta$, we can choose a weak prior or uninformative prior. When we are very confident in our domain expertise, we can select a strong informative prior. Further, the prior $p(\theta)$ is independent of $\mathcal{D}$, while the likelihood changes as the sample size $n$ grows (Figure \ref{fig:likelihood_dominates}). When $n$ is small, the posterior is heavily influenced by the prior because the likelihood is relatively weak. As $n$ increases, the likelihood becomes sharper and more concentrated around the true parameter value. Consequently, the influence of the prior on the posterior diminishes as $n$ increases, as the likelihood function incorporates increasingly more information about the data.

\renewcommand\fbox{\fcolorbox{black}{white}}

\subsection{Conjugate Priors}

When calculating the posterior distribution using Bayes' rule, it should be pretty straightforward to calculate the numerator. However, calculating the denominator $p(\mathcal{D})$ requires that we compute the integral 
\[
p(\mathcal{D}) = \int_\theta p(\mathcal{D} \mid \theta)p(\theta)d\theta.
\]
This might cause us trouble, since integration is often difficult. Computing this integral might be feasible for very simple problems. However, computing integrals can be  challenging when $\theta$ is high-dimensional. 

To tackle this issue, people have observed that for some choices of prior $p(\theta)$, the posterior distribution $p(\theta \mid \mathcal{D})$ can be directly computed in closed form. This computational convenience is known as a \textit{conjugate prior}, where \textit{conjugacy} is formally defined as follows.

\begin{definition}[Conjugacy, \citealt{gelman1995bayesian}]
\label{def:conjugacy}
    Let $\mathcal{F}$ denote a class of sampling distributions $p(\mathcal{D} \mid \theta)$. Let $\mathcal{P}$ denote a class of prior distributions for $\theta$. We say that $\mathcal{P}$ is conjugate for $\mathcal{F}$ if
    \[
    p(\theta \mid \mathcal{D}) \in \mathcal{P} \; \text{for all} \; p(\cdot \mid \theta) \in \mathcal{F} \; \text{and} \; p(\cdot) \in \mathcal{P}.
    \]
\end{definition}

\begin{table}[!t]
    \centering
    \begin{tabular}{c c}
    \toprule
      \textsc{likelihood}   & \textsc{prior \& posterior}   \\
    \midrule
      Bernoulli & Beta   \\
      Binomial & Beta \\
      Geometric & Beta \\
      Poisson & Gamma \\
      Categorical & Dirichlet \\
      Multinomial & Dirichlet \\
    \bottomrule
    \end{tabular}
    \caption{Conjugate priors for common discrete likelihood distributions.}
    \label{tab:conjugate_priors}
\end{table}

When the prior and posterior are \textit{conjugate distributions} with respect to the likelihood, we can evade intractable numerical integrations. Table \ref{tab:conjugate_priors} lists conjugate priors for some common likelihood distributions.

We can return to the coin toss example where we are given a sequence of $N$ coin tosses
\[
\mathcal{D} = \{x_{1},\ldots,x_{N}\}
\]
and we want to infer the probability of getting heads ($\theta$) using Bayes' rule. Suppose we choose the prior $p(\theta)$ as the Beta distribution defined by
\[
p(\theta) = \textsf{Beta}(\theta \mid \alpha_H, \alpha_T) = \frac{\theta^{\alpha_H -1 }(1-\theta)^{\alpha_T -1 }}{B(\alpha_H,\alpha_T)}
\]
where $\alpha_H$ and $\alpha_T$ are the two parameters that determine the shape of the distribution (similar to how the mean and variance determine a Gaussian distribution), and $B(\alpha_H, \alpha_T)$ is some normalization constant that ensures $\int p(\theta)d\theta=1$. We will go into more details about the Beta distribution later. What matters here is that the Beta distribution has a very special property: the posterior $p(\theta \mid \mathcal{D})$ is always another Beta distribution (but with different parameters). More concretely, out of $N$ coin tosses, if the number of heads and the number of tails are $N_H$ and $N_T$ respectively, then it can be shown that the posterior is
\begin{align*}
    p(\theta \mid \mathcal{D}) &= \textsf{Beta}(\theta \mid \alpha_H+N_H,\alpha_T+N_T) \\
    &= \frac{\theta^{N_H+ \alpha_H -1 }(1-\theta)^{ N_T+ \alpha_T -1 }}{B(N_H+ \alpha_H,N_T+ \alpha_T)},
\end{align*}
which is another Beta distribution with parameters $(\alpha_H+N_H, \alpha_T+N_T)$. In other words, if the prior is a Beta distribution (i.e., we can represent it as two numbers $\alpha_H,\alpha_T$) then the posterior can be immediately computed by a simple addition $\alpha_H+N_H, \alpha_T+N_T$. Conveniently, there is no need to compute the complex integral $p(\mathcal{D})$. 

%{% include marginfigure.html id=``beta'' url=``assets/img/beta.png'' description=``The expectation of both $\textsf{Beta}(3,2)$ and $\textsf{Beta}(30,20)$ are $0.6$, but $\textsf{Beta}(30,20)$ is much more concentrated. This can be used to represent different levels of uncertainty in $\theta$'' %}

\subsubsection{The Beta Distribution}

\begin{figure}[!t]
    \centering
    \includegraphics[width=0.8\linewidth]{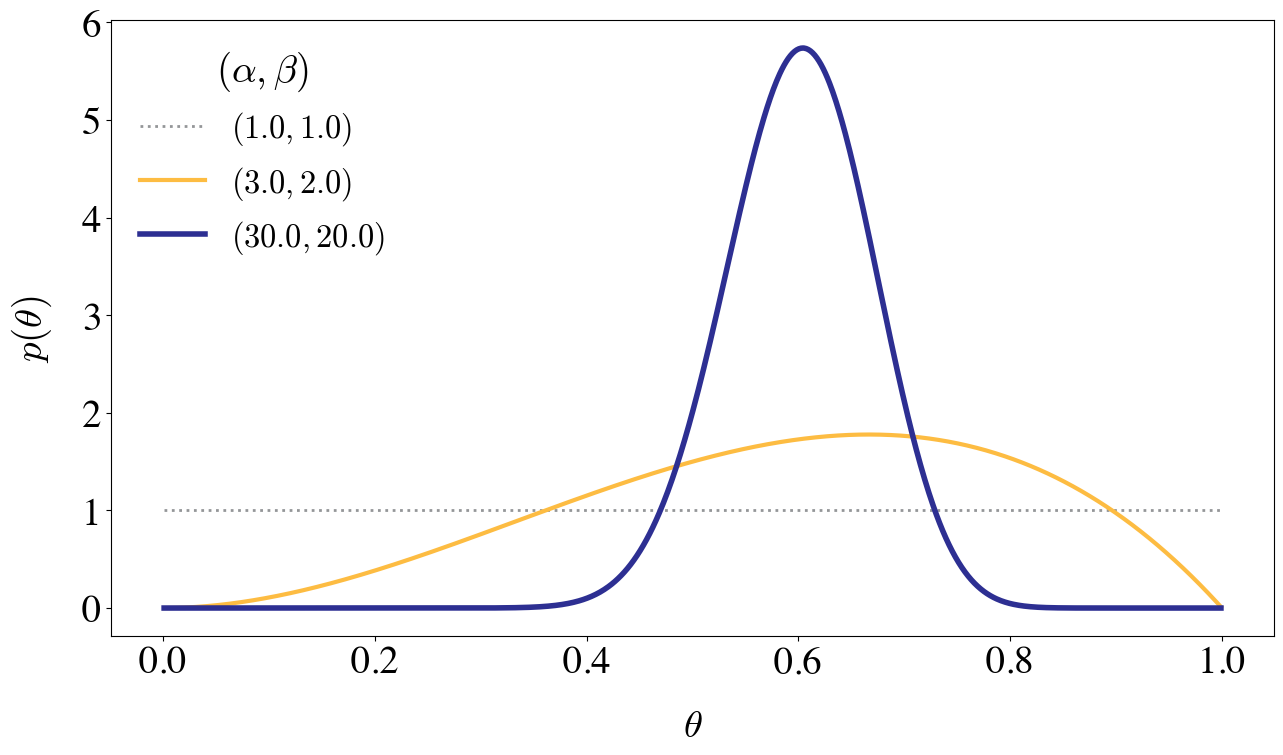}
    \caption{The expectation of both $\textsf{Beta}(3,2)$ and $\textsf{Beta}(30,20)$ are $0.6$, but $\textsf{Beta}(30,20)$ is much more concentrated. This can be used to represent different levels of uncertainty in $\theta$.}
    \label{fig:beta}
\end{figure}

Now we try to understand the Beta distribution better. If $\theta$ has distribution $\textsf{Beta}(\theta \mid \alpha_H, \alpha_T)$, then the expected value of $\theta$ is
\[
\frac{\alpha_H}{\alpha_H+\alpha_T}.
\]
Intuitively, $\alpha_H$ is larger than $\alpha_T$ if we believe that heads are more likely. The variance of the Beta distribution is the somewhat complex expression,
\[
\frac{\alpha_H\alpha_T}{(\alpha_H+\alpha_T)^2(\alpha_H+\alpha_T+1)},
\]
but we remark that (very roughly) the numerator is quadratic in $\alpha_H,\alpha_T$ while the denominator is cubic in $\alpha_H,\alpha_T$. Hence if $\alpha_H$ and $\alpha_T$ increases then the variance decreases, so we are more certain about the value of $\theta$ (Figure \ref{fig:beta}). We can use this observation to better understand the  posterior update rule: after observing more data $\mathcal{D}$, the prior parameters $\alpha_H$ and $\alpha_T$ increase by $N_H$ and $N_T$, respectively. Thus, the variance of $p(\theta \mid \mathcal{D})$ should be smaller than the variance of $p(\theta)$ (i.e., we are more certain about the value of $\theta$ after observing data $\mathcal{D}$). 

What we have shown here is that the Beta distribution family is a conjugate prior to the Bernoulli distribution family. Relating this back to the example above, if $p(\theta)$ is a Beta distribution and $p(\mathcal{D} \mid \theta)$ is a Bernoulli distribution, then $p(\theta \mid \mathcal{D})$ is still a Beta distribution. In general, we usually have a simple algebra expression to compute $p(\theta \mid \mathcal{D})$ (such as computing $\alpha_H+N_H, \alpha_T+N_H$ in the example above).

\subsubsection{The Categorical Distribution}

We now introduce another example of a conjugate prior, which generalizes the Bernoulli example above. Instead of being limited to binary outcomes, we can now consider the categorical distribution (think of a $K$-sided dice). Let $\mathcal{D} = \{ x_1, \ldots, x_N \}$ be $N$ rolls of the dice, where $x_j \in \{ 1, \ldots, K \}$ is the outcome of the $j$-th roll. The parameter of the categorical distribution is denoted by 
\[
\theta =(\theta_1, \cdots, \theta_K) := (p(X_j = 1), \ldots, p(X_j = K))
\]
where $\sum_{k = 1}^K \theta_k = 1$. 

We claim that the Dirichlet distribution is a conjugate prior for the categorical distribution. A Dirichlet distribution is defined by $K$ parameters $\mathbf{\alpha} = (\alpha_1, \ldots, \alpha_K)$, and its PDF is given by 
\[
p(\theta) = \textsf{Dirichlet}(\theta \mid \mathbf{\alpha}) = \frac{1}{B(\alpha)} \prod_{k=1}^K \theta_k^{\alpha_k - 1}
\]
where $B(\alpha)$ is still a normalization constant. 

To show that the Dirichlet distribution is a conjugate prior for the categorial distribution, we need to show that the posterior is also a Dirichlet distribution. To calculate the posterior $p(\theta \mid \mathcal{D})$ with Bayes' rule, we first calculate the likelihood $p(\mathcal{D} \mid \theta)$ as
\[
p(\mathcal{D} \mid \theta) = \prod_{k=1}^K \theta_k^{\sum_{j=1}^N 1\{ X_j = k \}}.
\]
To simplify the notation we denote $N_k = \sum_{j=1}^N 1\lbrace X_j=k\rbrace$ as the number of times we roll out $k$, so $p(\mathcal{D}\mid\theta)=\prod\theta_k^{N_k}$. Using this new notation, the posterior can be calculated as 
\begin{align*}
    p(\theta \mid \mathcal{D}) &\propto p(\mathcal{D} \mid \theta) p(\theta) \\
    &\propto \prod_{k=1}^K \theta_k^{N_k + \alpha_k - 1} \\
    &\coloneqq \textsf{Dirichlet}(\theta \mid \alpha_1+N_1,\cdots,\alpha_K+N_K).
\end{align*}

In other words, if the prior is a Dirichlet distribution with parameter $(\alpha_1, \cdots, \alpha_K)$, then the posterior $p(\theta \mid \mathcal{D})$ is a Dirichlet distribution with parameters $(\alpha_1+N_1, \cdots, \alpha_K+N_K)$. In example 2 above, we added a prior probability to observing an out-of-vocabulary word. We can see that this corresponds exactly to choosing a prior with nonzero prior $\alpha = \alpha_1 = \ldots = \alpha_K$.
% \williex{Not sure where this discussion of Laplace smoothing/heuristic comes from (not mentioned previously in book as far as I can tell)}
% This is also exactly the same as Laplace smoothing with parameter $\alpha$. We see that Laplace's heuristic for handling missing values has a rigorous justification when viewed with the Bayesian formalism.

\subsubsection{Some Concluding Remarks}

Many distributions have conjugate priors. In fact, any exponential family distribution has a conjugate prior. Although conjugacy seemingly solves the problem of computing Bayesian posteriors, there are two caveats:
\begin{enumerate}
    \item Usually practitioners will want to choose the prior $p(\theta)$ to best capture their knowledge about the problem, and using conjugate priors is a strong restriction.
    \item For more complex distributions, the posterior computation is not as easy as those in our examples. There are distributions for which the posterior computation is still NP-hard. 
\end{enumerate}

Conjugate priors are powerful tools used in many real-world applications and popular models, such as topic modeling (e.g., latent Dirichlet allocation~\citep{blei2003latent}) and generalized linear models.
However, their limitations should be considered when modeling tasks demand richer or more-nuanced prior structures.

\begin{reading}
    \begin{itemize}[leftmargin=*]
        \item \fullcite{gelman1995bayesian}.
        \item \fullcite{barber2012bayesian}.
        \item \fullcite{murphy2022probabilistic}.
    \end{itemize}
\end{reading}

%%%%%%%%%%%%%%%%%%%%%%%%
%% Discussion: VAE
%%%%%%%%%%%%%%%%%%%%%%%%

\chapter{Discussion: The Variational Autoencoder}
\label{sec:discussion_vae}

From the start of this tutorial, we have encouraged the reader to consider our three themes of representation, inference, and learning as interconnected endeavors. Now, we center our concluding discussion on an illustrative example of how these interconnections can play out to yield exceptionally powerful methods.

%%%%%%%%%%%%%%%%%%%%%%%%%%%%%%
%% Deep generative models
%%%%%%%%%%%%%%%%%%%%%%%%%%%%%%

% \williex{Possibly: remove 7.1 "Bringing it all Together" section, and move all underlying subsections up a hierarchical level.}

%\subsubsection{Bringing It All Together}

In this discussion, we present a highly influential deep probabilistic model: the \textit{variational autoencoder} (VAE; \citealt{kingma2014auto, kingma_introduction_2019}).  The VAE is a neural architecture for latent representation learning. VAEs have been famously used for image generation \citep{gregor2015draw}, language modeling \citep{bowman2016generating}, molecular design \citep{kusner2017grammar}, and semi-supervised learning tasks \citep{maaloe2016auxiliary}. As a testament to its impact, the VAE was recognized with the Test of Time Award at the 2024 International Conference on Learning Representations \citep{vae_iclr_tot_blog, vae_iclr_tot}. Using the VAE as a case study, we will draw connections among ideas from throughout this tutorial and demonstrate how these ideas are useful in machine learning research.

\section{Deep Generative Latent Variable Models}
\label{sec:vae}

Consider a directed LVM of the form
\[
p(\mathbf{x},\mathbf{z}) = p(\mathbf{x}|\mathbf{z})p(\mathbf{z})
\]
with observed variables $\mathbf{x} \in \mathcal{X}$ (where $\mathcal{X}$ can be continuous or discrete) and latent variables $\mathbf{z} \in \mathbb{R}^k$.

%{% include marginfigure.html id="faces" url="assets/img/faces.png" description="Variational autoencoder $p(x|z)p(z)$ applied to a face images (modeled by $x$). The learned latent space $\mathbf{z}$ can be used to interpolate between facial expressions." %}
To make things concrete, you can think of $\mathbf{x}$ as an image (e.g., a human face), and $\mathbf{z}$ as latent factors that explain the features of the image. For example, in the case of a face, one coordinate of $\mathbf{z}$ could encode whether the face is happy or sad, while another one encodes whether the face is male or female, etc.

We may also be interested in models with many layers of latent variables, e.g.,
\[
p(\mathbf{x} \mid \mathbf{z}_1)p(\mathbf{z}_1 \mid \mathbf{z}_2)p(\mathbf{z}_2 \mid \mathbf{z}_3)\cdots p(\mathbf{z}_{m-1}\mid \mathbf{z}_m)p(\mathbf{z}_m).
\]
These models are often referred to as \textit{deep latent variable models} and are capable of learning hierarchical latent representations. In this section, we will assume for simplicity that there is only one latent layer.

\subsection{Learning in Deep Generative Latent Variable Models}

Suppose that we are given a dataset $\mathcal{D} = \{\mathbf{x}_1, \mathbf{x}_2, \dotsc, \mathbf{x}_n\}$ and an LVM $p(\mathbf{x},\mathbf{z}; \theta)$ with observed variables $\mathbf{x}$, latent variables $\mathbf{z}$, and parameters $\theta$. We are interested in the following inference and learning tasks:
\begin{enumerate}
    \item \textit{Learning the parameters $\theta$ of $p$.}
    \item \textit{Approximate posterior inference over $\mathbf{z}$.} Given an image $\mathbf{x}_i$, what are its associated latent factors?
    \item \textit{Approximate marginal inference over $\mathbf{x}$.} Given an image $\mathbf{x}_i$ with missing parts, how do we fill-in these parts?
\end{enumerate}
We will also make the following assumptions:
\begin{enumerate}
    \item \textit{Intractability}. Computing the posterior probability $p(\mathbf{z} \mid \mathbf{x})$ is intractable.
    \item \textit{Big data}. The dataset $\mathcal{D}$ is too large to fit in memory, and so we can only work with small, subsampled batches of $\mathcal{D}$.
\end{enumerate}
Many interesting models fall in this class, including the VAE.

\subsection{The Standard Approaches}

We have learned several techniques that could be used to solve our three tasks. Let's try them out.

As we have discussed, EM (Algorithm \ref{alg:em}) can be used to learn LVMs. Recall, however, that performing the $E$-step requires computing the approximate posterior $p(\mathbf{z} \mid \mathbf{x})$, which we have assumed to be intractable. In the $M$-step, we learn parameters $\theta$ by looking at the entire dataset, which is going to be too large to hold in memory. 

%{% include sidenote.html id="note-onlineEM" note="Note, however, that there exists a generalization called online EM, which performs the $M$-step over mini-batches." %}

To perform approximate inference, we can use mean field variational inference (Section \ref{sec:variational_methods}). Recall, however, that one step of mean field requires us to compute an expectation whose time complexity scales exponentially with the size of the Markov blanket of the target variable (Definition \ref{def:markov_blanket_dag}, \ref{def:markov_blanket_mrf}).

What is the size of the Markov blanket for $\mathbf{z}$? If we assume that at least one component of $\mathbf{x}$ depends on each component of $\mathbf{z}$, then this introduces a $v$-structure (Table \ref{tab:primitives_independencies}) into the graph of our model (the $\mathbf{x}$, which are observed, are explaining away the differences among the $\mathbf{z}$). Thus, we know that all the $\mathbf{z}$ variables will depend on each other and the Markov blanket of some $\mathbf{z}_i$ will contain all the other $\mathbf{z}$ variables. This will make mean field intractable. An equivalent (and simpler) explanation is that $p$ will contain a factor $p(\mathbf{x}_i \mid \mathbf{z}_1, \dotsc, \mathbf{z}_k)$, in which all the $\mathbf{z}$ variables are tied.

%{% include sidenote.html id='note-meanfield' note='The authors refer to this when they say "the required integrals for any reasonable mean-field VB algorithm are also intractable." They also discuss the limitations of EM and sampling methods in the introduction and the methods section.' %}

Another approach would be to use sampling-based methods (Section \ref{sec:sampling_methods}). In the seminal  paper that first describes the VAE \citep{kingma2014auto}, the authors compare the VAE against these kinds of algorithms. However, they find that these sampling methods do not scale well to large datasets. In addition, techniques such as Metropolis-Hastings require a hand-crafted proposal distribution, which might be difficult to choose.

\section{Auto-Encoding Variational Bayes}

We are now going to learn about \textit{auto-encoding variational Bayes} (AEVB), an algorithm that can efficiently solve our three inference and learning tasks. The VAE is one instantiation of this algorithm.

In this section, following typical notation in the VAE literature, we will use $p_\theta(\mathbf{x},\mathbf{z})$ to denote a probabilistic model of observed variables $\mathbf{x}$ and latent variables $\mathbf{z}$ with parameter $\theta$ (and similarly for variational distributions $q_\phi$ with parameter $\phi$).

AEVB is based on ideas from variational inference (Section \ref{sec:variational_methods}). Recall that in variational inference, we are interested in maximizing the ELBO, written here as
\[ 
\mathcal{L}(p_\theta,q_\phi) = \mathbb{E}_{q_\phi(\mathbf{z} \mid \mathbf{x})} \left[ \log p_\theta(\mathbf{x},\mathbf{z}) - \log q_\phi(\mathbf{z} \mid \mathbf{x}) \right],
\]
over a family of distributions $q_\phi$. The ELBO satisfies the equation
\[
\log p_\theta(\mathbf{x}) = \text{KL}(q_\phi(\mathbf{z} \mid \mathbf{x})  \; || \;  p(\mathbf{z} \mid \mathbf{x})) + \mathcal{L}(p_\theta,q_\phi).
\]
Note that the variational approximation $q_\phi(\mathbf{z} \mid \mathbf{x})$ is conditioned on $\mathbf{x}$. This means that we are effectively choosing a different density $q(\mathbf{z})$ for every $\mathbf{x}$, which will be used to produce an improved posterior approximation, rather than always choosing a single static $q(\mathbf{z})$.

How exactly do we compute and optimize over $q_\phi(\mathbf{z} \mid \mathbf{x})$? As noted above, mean field could be applied here, but it tends to be both computationally challenging and inaccurate, since the fully factorized assumption on $q$ is too restrictive for our purposes.

\subsection{Black-Box Variational Inference}

The first important idea used in the AEVB algorithm is a general purpose approach for optimizing $q_\phi$ that works for a rich class of variational densities. Later, we will show special cases of this algorithm given specific choices of $q_\phi$.

This approach --- called \textit{black-box variational inference} --- consists of maximizing the ELBO using gradient descent over $\phi$ (instead of using, for example, a coordinate descent algorithm like in mean field). Hence, this procedure only assumes that $q_\phi$ is differentiable in its parameters $\phi$. The term \textit{black-box variational inference} was first coined by \citet{ranganath2014black}, while the core ideas were also inspired by earlier work, such as the Wake-Sleep algorithm \citep{hinton1995wake}. 

Additionally, rather than treating inference and learning as separate steps, the AEVB algorithm performs both inference and learning simultaneously by jointly optimizing $\phi$ and $\theta$ via gradient descent.
Optimization over $\phi$ will keep the ELBO tight around $\log p_\theta(\mathbf{x})$.
Optimization over $\theta$ will push the lower bound (and hence $\log p_\theta(\mathbf{x})$) up.
This is somewhat similar to how the EM algorithm optimizes a lower bound on the marginal likelihood (Section~\ref{sec:learning_with_expectation_maximization}).

\subsection{The Score Function Gradient Estimator}

To perform black-box variational inference, we need to compute the gradient
\[
\nabla_{\theta, \phi} \, \mathbb{E}_{q_\phi(\mathbf{z})} \left[ \log p_\theta(\mathbf{x},\mathbf{z}) - \log q_\phi(\mathbf{z}) \right],
\]
where we use $q_\phi(\mathbf{z})$ to denote an arbitrary variational distribution.
Taking the expectation with respect to $q_\phi$ in closed form will often not be possible. Instead, we may take Monte Carlo estimates of the gradient by sampling from $q_\phi$. This is easy to do for the gradient with respect to $\theta$. We can swap the gradient and the expectation and estimate the following expression via Monte Carlo:
\[
\mathbb{E}_{q_\phi(\mathbf{z})} \left[ \nabla_{\theta} \log p_\theta(\mathbf{x},\mathbf{z}) \right].
\]
However, taking the gradient with respect to $\phi$ is more difficult.
Since the expectation is taken with respect to $q_\phi$, which itself depends on $\phi$, we cannot simply move the gradient inside the expectation.

One way to estimate this gradient is via the so-called \textit{score function estimator}:
\begin{align*}
    &\nabla_\phi \mathbb{E}_{q_\phi(\mathbf{z})} \left[ \log p_\theta(\mathbf{x},\mathbf{z}) - \log q_\phi(\mathbf{z}) \right] \\
    &= \mathbb{E}_{q_\phi(\mathbf{z})} \left[ \left(\log p_\theta(\mathbf{x},\mathbf{z}) - \log q_\phi(\mathbf{z}) \right) \nabla_\phi \log q_\phi(\mathbf{z}) \right].
\end{align*}
This follows from some basic algebra and calculus and takes about half a page to derive. We leave it as an exercise to the reader, but for those interested, the full derivation may be found in Appendix B of \citet{mnih2014neural}.

The above identity places the gradient inside the expectation, which we may now evaluate using Monte Carlo. We refer to this as the score function estimator of the gradient.

Unfortunately, the score function estimator has an important shortcoming: it has a high variance. What does this mean? Suppose you are using Monte Carlo to estimate some quantity with expected value equal to 1. If your samples are all close to 1 (e.g., $\{0.9, 1.1, 0.96, 1.05, \dotsc\}$), then after a few samples you will get a good estimate of the true expectation.
On the other hand, suppose that 99\% of the time you sample 0, and 1\% of the time you sample 100.
% If, on the other hand, you sample zero 99\% of the time and you sample one hundred 1\% of the time, the
Then the expectation is still correct, but you will have to take a very large number of samples to figure out that the true expectation is actually 1.
We refer to the latter case as being \textit{high variance}.

This is the kind of problem we often run into with the score function estimator. In fact, its variance is so high that we cannot use it to learn many models.

The key contribution of the VAE paper is to propose an alternative estimator that is much better behaved. This is done in two steps: we first reformulate the ELBO so that parts of it can be computed in closed form (without Monte Carlo sampling), and then we use an alternative gradient estimator based on the so-called \textit{reparametrization trick}.

\subsection{The Stochastic Gradient Variational Bayes Estimator}

We will turn to a reformulation of the ELBO that takes the form
\[
\log p(\mathbf{x}) \geq \mathbb{E}_{q_\phi(\mathbf{z} \mid \mathbf{x})} \left[ \log p_\theta(\mathbf{x} \mid \mathbf{z}) \right] - \text{KL}(q_\phi(\mathbf{z} \mid \mathbf{x})  \; || \;  p(\mathbf{z})).
\]
This reparametrization is known as the \textit{Stochastic Gradient Variational Bayes} (SGVB) estimator, and it has a very interesting interpretation. First, think of $\mathbf{x}$ as an observed data point. The right-hand side consists of two terms. Both terms involve taking a sample $\mathbf{z} \sim q_\phi(\mathbf{z} \mid \mathbf{x})$, which we can interpret as a code describing $\mathbf{x}$. We are therefore going to call $q_\phi$ the \textit{encoder}.

In the first term, $\log p_\theta(\mathbf{x} \mid \mathbf{z})$ is the log-likelihood of the observed $\mathbf{x}$ given the code $\mathbf{z}$ that we have sampled. This term is maximized when $p_\theta(\mathbf{x} \mid \mathbf{z})$ assigns high probability to the original $\mathbf{x}$. It is thus trying to reconstruct $\mathbf{x}$ given the code $\mathbf{z}$. For that reason, we call $p_\theta(\mathbf{x} \mid \mathbf{z})$ the \textit{decoder} and call the first term the \textit{reconstruction error}.

The second term is the divergence between $q_\phi(\mathbf{z} \mid \mathbf{x})$ and the prior $p(\mathbf{z})$, which we will fix to be a unit Normal distribution. This encourages the codes $\mathbf{z}$ to look Gaussian. We call this the \textit{regularization} term. This prevents $q_\phi(\mathbf{z} \mid \mathbf{x})$ from simply encoding an identity mapping, and instead forces it to learn some more interesting representation (e.g., the facial features in our first example).

Thus, our optimization objective is trying to fit a $q_\phi(\mathbf{z} \mid \mathbf{x})$ that will map $\mathbf{x}$ into a useful latent space $\mathbf{z}$ from which we are able to reconstruct $\mathbf{x}$ via $p_\theta(\mathbf{x} \mid \mathbf{z})$. This type of objective is reminiscent of autoencoder neural networks. This is where the AEVB algorithm takes its name.

%{% include sidenote.html id="note-autoencoder" note="An autoencoder is a pair of neural networks $f, g$ that are composed as $\bar x=f(g(x))$. They are trained to minimize the reconstruction error $\|\bar x - x\|$. In practice, $g(x)$ learns to embed $x$ in a latent space that often has an intuitive interpretation." %}

\subsection{The Reparametrization Trick}

As we have seen earlier, optimizing our objective requires a good estimate of the gradient. The main technical contribution of the VAE paper is a low-variance gradient estimator based on the \textit{reparametrization trick}.

Under certain mild conditions, we may express the distribution $q_\phi(\mathbf{z} \mid \mathbf{x})$ as the following two-step generative process.
\begin{enumerate}
    \item First, we sample a noise variable $\epsilon$ from a simple distribution $p(\epsilon)$ (e.g., the standard Normal $\mathcal{N}(0,1)$):
    \[
    \epsilon \sim p(\epsilon).
    \]
    \item Then, we apply a deterministic transformation $g_\phi(\epsilon, \mathbf{x})$ that maps the random noise to a more complex distribution:
    \[
    \mathbf{z} = g_\phi(\epsilon, \mathbf{x}).
    \]
\end{enumerate}
For many interesting classes of $q_\phi$, it is possible to choose a $g_\phi(\epsilon, \mathbf{x})$, such that $\mathbf{z} = g_\phi(\epsilon, \mathbf{x})$ will be distributed according to $q_\phi(\mathbf{z} \mid \mathbf{x})$.

Gaussian variables provide the simplest example of the reparametrization trick. Instead of writing $\mathbf{z} \sim q_{\mu, \sigma}(\mathbf{z}) = \mathcal{N}(\mu, \sigma)$, we can write
\[
\mathbf{z} = g_{\mu, \sigma}(\epsilon) = \mu + \epsilon \cdot \sigma,
\]
where $\epsilon \sim \mathcal{N}(0,1)$. It is easy to check that the two ways of expressing the random variable $\mathbf{z}$ lead to the same distribution.

The biggest advantage of this approach is that we can now write the gradient of an expectation with respect to $q_\phi(\mathbf{z} \mid \mathbf{x})$ (for any $f$) as
\begin{align*}
    \nabla_\phi \mathbb{E}_{z \sim q_\phi(\mathbf{z} \mid \mathbf{x})}\left[ f(\mathbf{x},\mathbf{z}) \right] &= \nabla_\phi \mathbb{E}_{\epsilon \sim p(\epsilon)}\left[ f(\mathbf{x}, g_\phi(\epsilon, \mathbf{x})) \right] \\
    &= \mathbb{E}_{\epsilon \sim p(\epsilon)}\left[ \nabla_\phi f(\mathbf{x}, g_\phi(\epsilon, \mathbf{x})) \right].
\end{align*}
Since $p(\epsilon)$ does depend on $\phi$, we can swap the gradient $\nabla_\phi$ and the expectation $\mathbb{E}_{\epsilon \sim p(\epsilon)}$. As the gradient is now inside the expectation, we can take Monte Carlo samples to estimate the right-hand term. This approach has much lower variance than the score function estimator, and will enable us to learn models that we couldn't otherwise learn.

%{% include sidenote.html id="note-reparam" note="For more details as to why, have a look at the appendix of the paper by [Rezende et al.](https://arxiv.org/pdf/1401.4082.pdf)" %} 

\subsection{Choosing $q$ and $p$}

Until now, we did not specify the exact form of $p_\theta$ or $q_\phi$, besides saying that these could be arbitrary functions. How should one parametrize these distributions? The best $q_\phi(\mathbf{z} \mid \mathbf{x})$ should be able to approximate the posterior $p_\theta(\mathbf{z} \mid \mathbf{x})$. Similarly, $p_\theta(\mathbf{x} \mid \mathbf{z})$ should be flexible enough to represent the richness of the data.

For these reasons, we are going to parametrize $q$ and $p$ by \textit{neural networks}. These are extremely expressive function approximators that can be efficiently optimized over large datasets. This choice also draws a fascinating bridge between classical machine learning methods (approximate Bayesian inference in this case) and modern deep learning.

But what does it mean to parametrize a distribution with a neural network? Let's assume again that $q(\mathbf{z} \mid \mathbf{x})$ and $p(\mathbf{x} \mid \mathbf{z})$ are Normal distributions. We can express them as
\[
q(\mathbf{z} \mid \mathbf{x}) = \mathcal{N}(\mathbf{z}; \boldsymbol\mu(\mathbf{x}), \text{diag}(\boldsymbol\sigma(\mathbf{x}))^2)
\]
where $\boldsymbol\mu(\mathbf{x}), \boldsymbol\sigma(\mathbf{x})$ are deterministic vector-valued functions of $\mathbf{x}$ parametrized by an arbitrary complex neural network.

More generally, the same technique can be applied to any exponential family distribution by parameterizing the sufficient statistics by a function of $\mathbf{x}$.

\section{The Variational Autoencoder}

We are now ready to define the AEVB algorithm and the VAE, its most popular instantiation. The AEVB algorithm  simply combines
\begin{enumerate}
    \item the auto-encoding ELBO reformulation;
    \item the black-box variational inference approach; and
    \item the reparametrization-based low-variance gradient estimator.
\end{enumerate}
It optimizes the auto-encoding ELBO using black-box variational inference with the reparametrized gradient estimator. This algorithm is applicable to any deep generative model $p_\theta$ with latent variables that is differentiable in $\theta$.

\begin{figure}
    \centering
    \includegraphics[height=0.3\textheight]{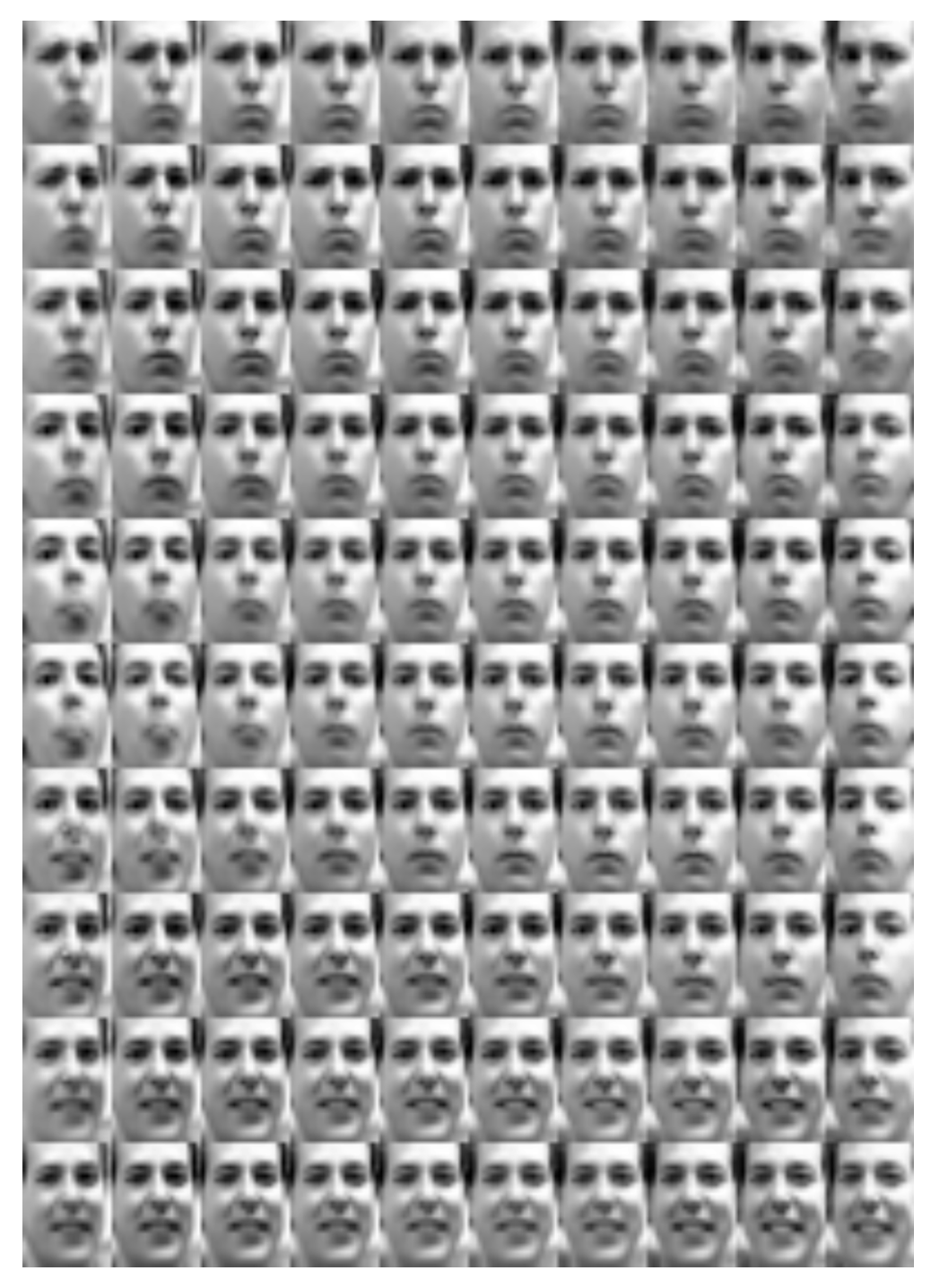}
    \includegraphics[height=0.3\textheight]{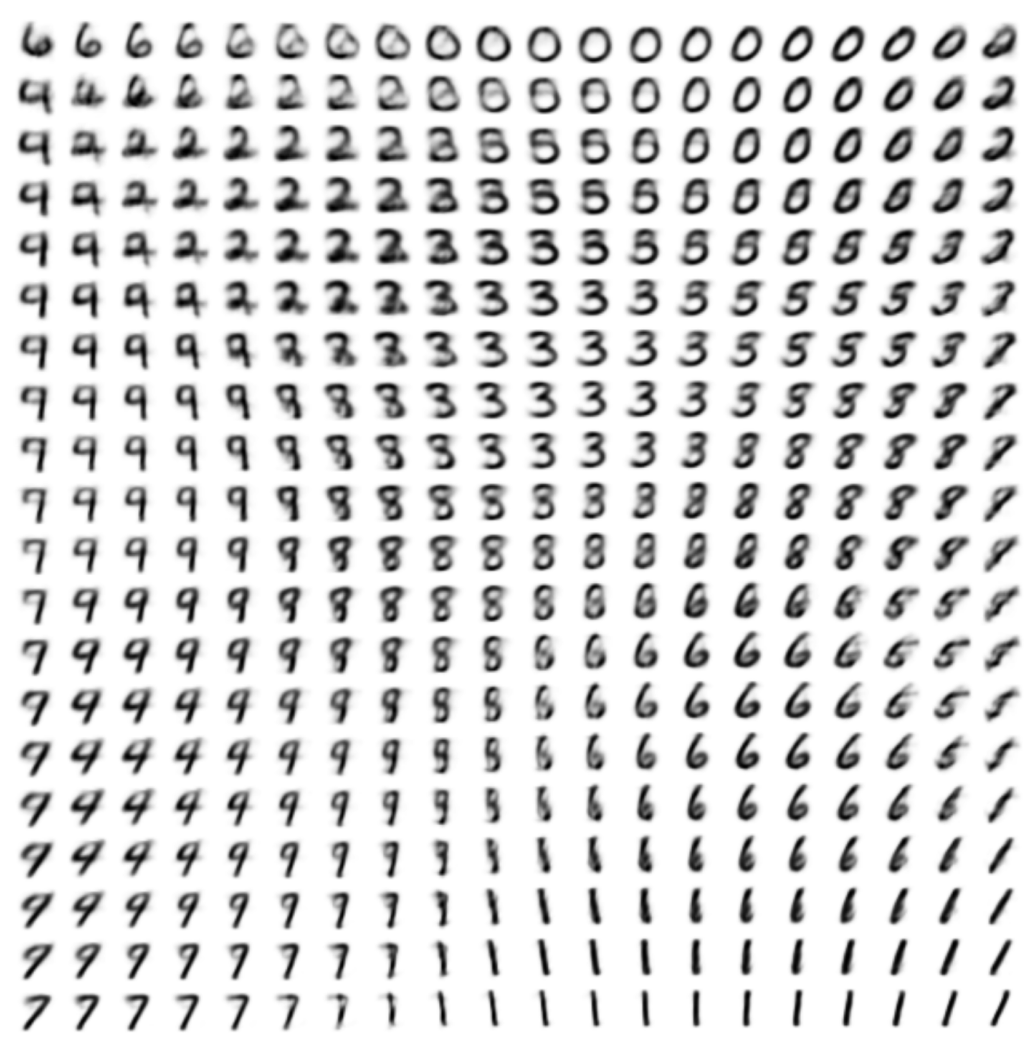}
    \caption{Data manifolds learned by AEVB for the Frey Face dataset (left) and MNIST (right). We can interpolate over human faces and written digits by interpolating over the latent variables. Figure  from \citet{kingma2014auto}.}
    \label{fig:frey_mnist}
\end{figure}

A VAE uses the AEVB algorithm to learn a specific model $p$ using a particular encoder $q$. The model $p$ is parametrized as
\begin{align*}
p(\mathbf{x} \mid \mathbf{z}) & = \mathcal{N}(\mathbf{x}; \boldsymbol\mu(\mathbf{z}), \text{diag}(\boldsymbol\sigma(\mathbf{z}))^2) \\
p(\mathbf{z}) & = \mathcal{N}(\mathbf{z}; 0, I),
\end{align*}
where $\boldsymbol\mu(\mathbf{z}), \boldsymbol\sigma(\mathbf{z})$ are parametrized by a neural network (in the original formulation, two dense hidden layers of 500 units each). Model $q$ is similarly parametrized as
\[
q(\mathbf{z} \mid \mathbf{x}) = \mathcal{N}(z; \boldsymbol\mu(\mathbf{x}), \text{diag}(\boldsymbol\sigma(\mathbf{x}))^2).
\]
This choice of $p$ and $q$ allows us to further simplify the auto-encoding ELBO. In particular, we can use a closed form expression to compute the regularization term, and we only use Monte Carlo estimates for the reconstruction term. These expressions are given in the original paper.

We can interpret the VAE as a directed latent variable probabilistic graphical model. We can also view it as a particular objective for training an autoencoder neural network. Unlike previous approaches, this objective derives reconstruction and regularization terms from a more principled, Bayesian perspective.

\subsection{Applications of the VAE}

%{% include marginfigure.html id="mnist" url="assets/img/mnist.png" description="Interpolating over MNIST digits by interpolating over latent variables" %}

The VAE can be applied to images in order to learn interesting latent representations, as illustrated on the Frey Face dataset and the MNIST handwritten digits database\footnote{\href{https://yann.lecun.com/exdb/mnist/}{https://yann.lecun.com/exdb/mnist/}} (Figure \ref{fig:frey_mnist}). On the face dataset, we can interpolate between facial expressions by interpolating between latent variables (e.g., we can generate smooth transitions between ``angry'' and ``surprised''). On the MNIST dataset, we can similarly interpolate between numbers. The VAE has also been used as a design tool in the basic sciences and engineering.  For example, VAEs have been used in chemistry for the automated design of novel molecules \citep{gomez2018automatic} and in materials science to generate new stable materials \citep{xiecrystal2022}.

%The authors also compare their methods against three alternative approaches: the Wake-Sleep algorithm, Monte Carlo EM, and hybrid Monte Carlo. The latter two methods are sampling-based approaches that are quite accurate, but they do not scale well to large datasets. Wake-Sleep is a variational inference algorithm that scales much better. However, it does not use the exact gradient of the ELBO (it uses an approximation instead), and hence it is not as accurate as AEVB. \citet{kingma2014auto}  illustrate this by plotting learning curves.

\vspace{5mm}

\begin{reading}
    \begin{itemize}[leftmargin=*]
        \item \fullcite{kingma_introduction_2019}.
    \end{itemize}
\end{reading}

%%%%%%%%%%%%%%%%%%%%%%%%
%% Acknowledgements
%%%%%%%%%%%%%%%%%%%%%%%%

% end of main matter

\begin{acknowledgements}
Author J. Maasch acknowledges support from the National Science Foundation Graduate Research Fellowship under Grant No. DGE – 2139899.
\end{acknowledgements}

%%%%%%%%%%%%%%%%%%%%%%%%
%% Appendix
%%%%%%%%%%%%%%%%%%%%%%%%

%\appendix

%%%%%%%%%%%%%%%%%%%%%%%%
%% Back matter
%%%%%%%%%%%%%%%%%%%%%%%%

%BACKMATTER SEE DOCUMENTATION
\backmatter  % references, restarts sample

%%%%%%%%%%%%%%%%%%%%%%%%
%% Bibliography
%%%%%%%%%%%%%%%%%%%%%%%%

\printbibliography

\end{document}